\documentclass{gameai-survey}

\usepackage{varwidth}
\usepackage{fontawesome}
\usepackage{needspace}
\usepackage[normalem]{ulem}
\usetikzlibrary{decorations.pathreplacing}
\newcommand{\concept}[1]{\uline{\emph{#1}}}

\newif\ifgameaidraftnotes
\gameaidraftnotesfalse

\usetikzlibrary{arrows.meta,fit,calc}
\usepackage{adjustbox}
\tikzset{
  dm node/.style={draw=GameLine,line width=.4pt,inner sep=4pt,align=center,font=\footnotesize,text width=#1},
  dm bridge/.style={font=\scriptsize\itshape,text=GameSlate,align=center,anchor=south,text width=#1},
  dm q/.style={font=\footnotesize\itshape,anchor=west,text=GameInk},
  dm axis/.style={-{Latex[length=1.6mm]},GameSlate,line width=.4pt},
  dm arrow/.style={-{Latex[length=1.8mm]},GameInk,line width=.5pt},
  dm frame/.style={draw=GameSlate,dashed,line width=.3pt,inner sep=6pt},
}
\newcommand{\dmcolor}{Player}
\newcommand{\dmhead}[1]{\textbf{\color{\dmcolor}#1}}
\newcommand{\dmtag}{\node[anchor=north east,font=\tiny\scshape,text=GameSlate,inner sep=1pt] at ([xshift=-2pt,yshift=-2pt]F.north east) {draft map};}
\newcommand{\draftmap}[1]{\ifgameaidraftnotes\par\medskip\noindent\begin{minipage}{\linewidth}\centering\adjustbox{max width=\linewidth}{#1}\end{minipage}\par\medskip\fi}

\definecolor{TitleGameG}{HTML}{C75A11}
\definecolor{TitleGameA}{HTML}{B48801}
\definecolor{TitleGameM}{HTML}{70AE47}
\definecolor{TitleGameE}{HTML}{589BD4}
\definecolor{TitleGameS}{HTML}{5757AE}
\newcommand{\gameaiGamesWord}{\textcolor{TitleGameG}{G}\textcolor{TitleGameA}{a}\textcolor{TitleGameM}{m}\textcolor{TitleGameE}{e}\textcolor{TitleGameS}{s}}
\title{AI for \texorpdfstring{\gameaiGamesWord}{Games} in the Foundation Model Era}

\newcommand{\gameaiProjectURL}{https://eurekaleo.github.io/awesome-ai-for-games}
\newcommand{\gameaiGitHubURL}{https://github.com/Eurekaleo/awesome-ai-for-games}
\newcommand{\gameaiAuthorNames}{\begin{tabular}{@{}c@{}}
    Meng Luo\textsuperscript{1},\enspace
    Yanlin Li\textsuperscript{1},\enspace
    Hao Li\textsuperscript{1},\enspace
    Hongzhan Lin\textsuperscript{1,\dag},\enspace
    Pengfei Zhou\textsuperscript{1},\enspace
    Tianjie Ju\textsuperscript{1},\\
    Ran Zhang\textsuperscript{2},\enspace
    Yeying Jin\textsuperscript{1},\enspace
    Mong-Li Lee\textsuperscript{1},\enspace
    Wynne Hsu\textsuperscript{1}
  \end{tabular}}
\newcommand{\gameaiAffiliations}{\begin{tabular}{@{}c@{}}
    \textsuperscript{1}National University of Singapore \qquad
    \textsuperscript{2}Nanyang Technological University
  \end{tabular}}
\newcommand{\gameaiTitleMark}{\makebox[\linewidth][l]{\raisebox{2mm}{\includegraphics[height=11mm,keepaspectratio,trim=49.2bp 50.64bp 49.2bp 47.52bp,clip]{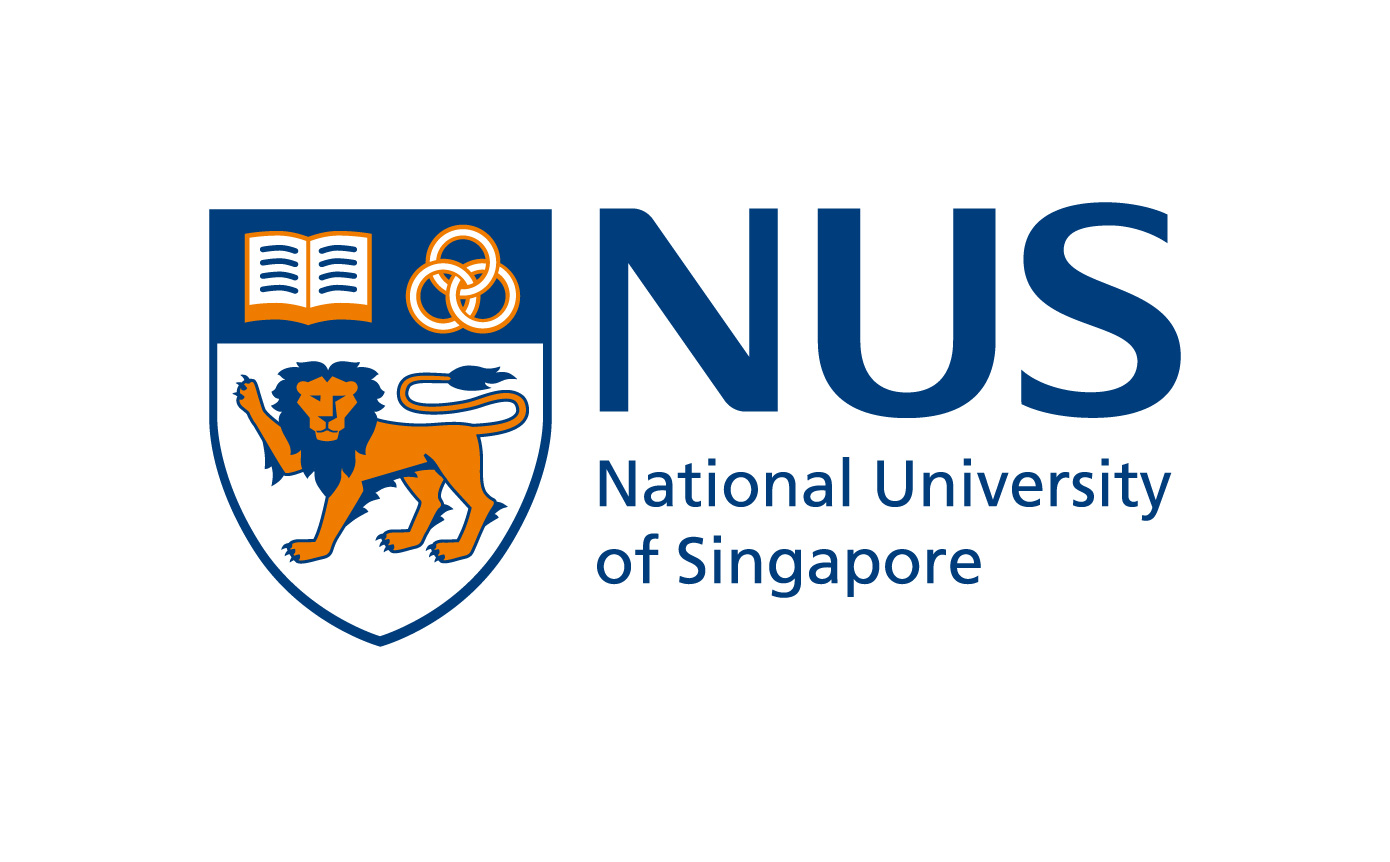}}\hspace{10mm}\raisebox{2.65mm}{\includegraphics[height=9.7mm,keepaspectratio]{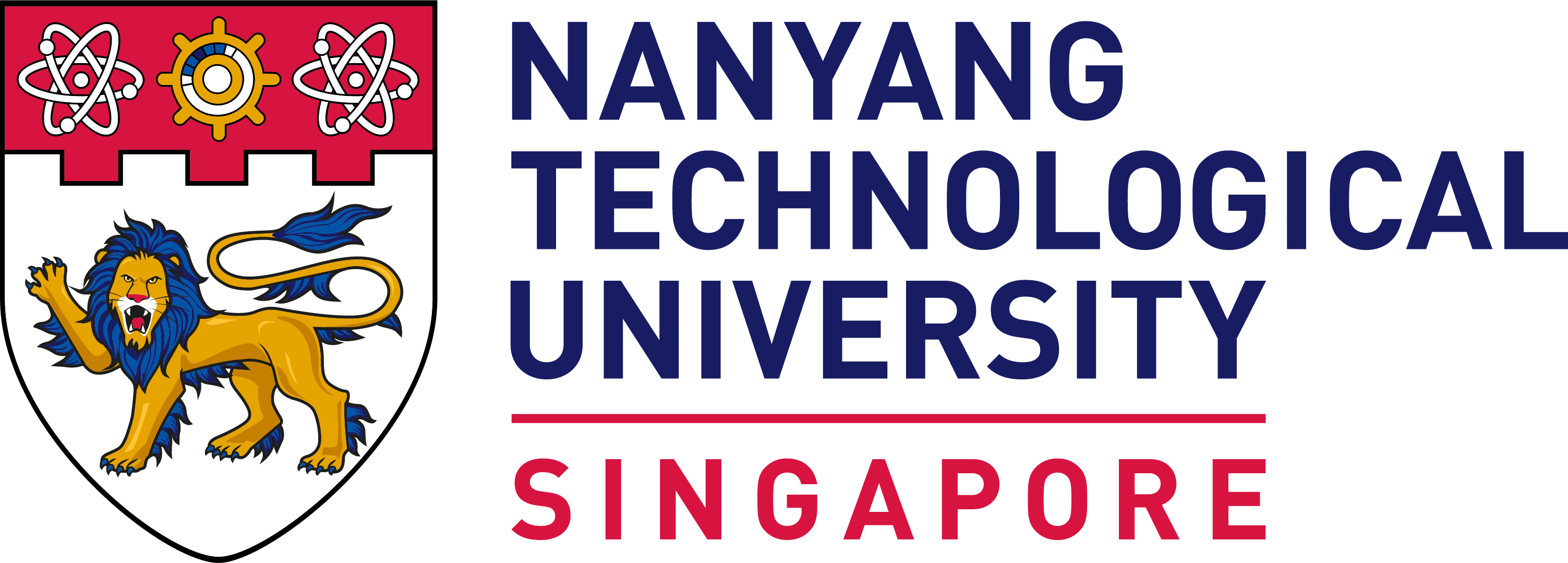}}\hfill
    \includegraphics[height=15mm,keepaspectratio]{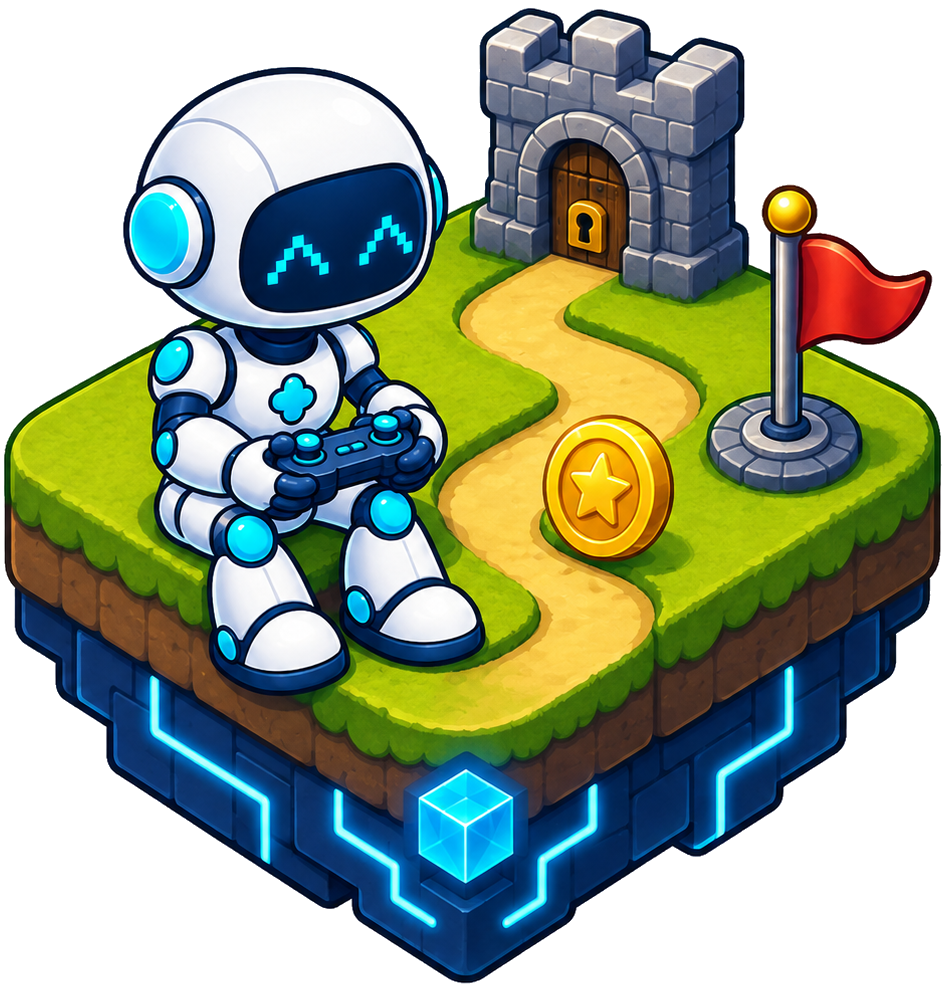}}}
\newcommand{\gameaiMetadataRow}[3]{\noindent
  \raisebox{-.35mm}{\makebox[3.4mm][c]{#1}}\hspace{1.2mm}{\rmfamily\small\bfseries\color{GameInk}#2:}\enspace
  {\footnotesize\rmfamily\color{GameLink}\href{#3}{#3}}\par}
\newcommand{\gameaiProjectGitHubRow}{\par\vspace{0.1mm}
  \begin{center}
    \begin{varwidth}{\linewidth}
      \raggedright

      \gameaiMetadataRow
        {\resizebox{!}{3.3mm}{\textcolor[HTML]{24292F}{\raisebox{\depth}{\faGlobe}}}}
        {Project}{\gameaiProjectURL}

      \vspace{.9mm}

      \gameaiMetadataRow
        {\includegraphics[height=3.3mm,keepaspectratio]{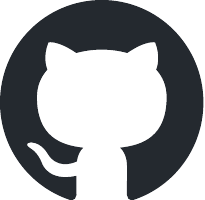}}
        {GitHub}{\gameaiGitHubURL}

      \vspace{.9mm}

      \noindent
      \raisebox{-.35mm}{\makebox[3.4mm][c]{\resizebox{!}{2.4mm}{\textcolor{GameInk}{\raisebox{\depth}{\faEnvelope}}}}}\hspace{1.2mm}{\rmfamily\small\bfseries\color{GameInk}Contact:}\enspace
      {\footnotesize\rmfamily\color{GameLink}\href{mailto:mluo@u.nus.edu}{mluo@u.nus.edu},
        \href{mailto:danielhzlin@nus.edu.sg}{danielhzlin@nus.edu.sg}}\par

    \end{varwidth}
  \end{center}
}
\newcommand{\gameaiTitleLens}{\par\vspace{2mm}
  \noindent\begin{minipage}{\linewidth}
    \centering
    \begin{tikzpicture}
      \clip[rounded corners=2.5mm] (0,0) rectangle (\linewidth,0.48485\linewidth);
      \node[anchor=south west,inner sep=0pt] at (0,0) {\includegraphics[width=\linewidth]{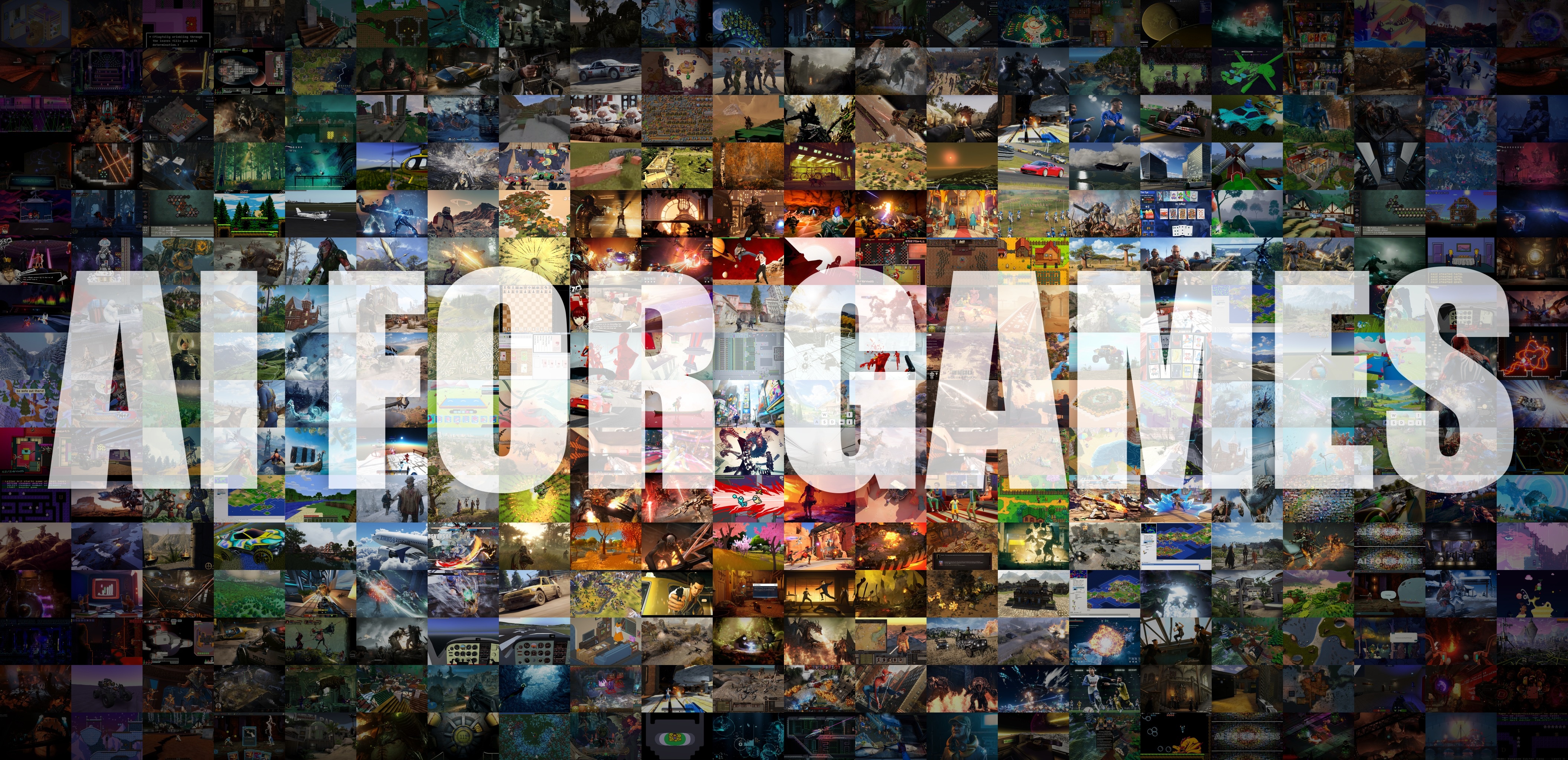}};
    \end{tikzpicture}
    \captionof{figure}{\textbf{AI for Games in the foundation model era.} The tiles are in-game frames from commercial and open-source games, research environments, and generative game systems reviewed in this survey.}
    \label{fig:title-teaser}
  \end{minipage}}
\renewenvironment{gameaifrontmatter}{\thispagestyle{empty}\noindent\begin{minipage}{\linewidth}\setlength{\parindent}{0pt}\setlength{\parskip}{0pt}\centering
  \gameaiTitleCore
  \vspace{3.0mm}}{\end{minipage}\par\vspace{1.5mm}}
\titlecontents{subsubsection}
  [4.4em]
  {\fontsize{8.4}{9.6}\selectfont\color{GameSlate}}
  {}
  {}
  {\titlerule*[0.55pc]{.}\contentspage}
\newcommand{\runin}[1]{\phantomsection\addcontentsline{toc}{subsubsection}{#1}\textbf{#1.}}
\makeatletter
\renewcommand{\gameaiContents}{\clearpage
  \pagestyle{gameaicontents}\phantomsection
  \pdfbookmark[1]{Contents}{contents}\noindent
  {\sffamily\bfseries\Large\color{GameInk}Contents}\par
  \vspace{1.5mm}{\color{GameLink}\rule{\linewidth}{1.0pt}}\par
  \vspace{3.5mm}\begingroup
    \hypersetup{linktoc=all}\setlength{\parskip}{0pt}\setlength{\parindent}{0pt}\@starttoc{toc}\endgroup
  \clearpage
  \pagestyle{gameai}}
\makeatother
\newcommand{\gameaiAuthorBlock}{\centering
  {\rmfamily\bfseries\fontsize{9.35}{11.1}\selectfont\color{GameInk}\gameaiAuthorNames}\par
  \vspace{1.0mm}{\rmfamily\bfseries\itshape\fontsize{8.2}{9.7}\selectfont\color[HTML]{475569}\gameaiAffiliations}\par
}
\author{\gameaiAuthorBlock}

\newcommand{\gameaiinsighticon}{\textmd{\faLightbulbO}}
\newlist{insightpoints}{itemize}{1}
\setlist[insightpoints]{label=\textbullet,leftmargin=1.25em,
  itemsep=4pt,parsep=0pt,topsep=1pt,partopsep=0pt}

\newcommand{\indexrole}[1]{\ifstrequal{#1}{P}{\textcolor{Player}{\textbf{P}}}{}\ifstrequal{#1}{M}{\textcolor{Simulator}{\textbf{M}}}{}\ifstrequal{#1}{D}{\textcolor{Creator}{\textbf{D}}}{}\ifstrequal{#1}{B}{\textcolor{Builder}{\textbf{B}}}{}\ifstrequal{#1}{R}{\textcolor{Resident}{\textbf{R}}}{}\ifstrequal{#1}{T}{\textcolor{Evaluator}{\textbf{T}}}{}}
\newcommand{\indexgroup}[3]{\addlinespace[3pt]\multicolumn{#2}{@{}l}{\textcolor{#1}{\bfseries #3}}\\*}
\tcbset{
  game insight common/.style={
    enhanced,
    boxrule=.8pt,arc=2.6mm,
    left=7pt,right=7pt,top=9pt,bottom=6pt,boxsep=3pt,
    fonttitle=\small\sffamily\bfseries,
    fontupper=\small,
    coltitle=white,
    attach boxed title to top left={xshift=7pt,yshift=-1.5mm},
    boxed title style={boxrule=0pt,arc=2pt,left=5pt,right=5pt,top=2.5pt,bottom=2.5pt},
    before={\par\Needspace{.38\textheight}},
    after={\par\medskip},
  }
}
\newtcolorbox{gameaiinsight}[2][Player]{
  game insight common,
  colback=#1!6!white,
  colframe=#1!48!GameLine,
  colbacktitle=#1!58!GameInk,
  title={{\color{white}\gameaiinsighticon}\enspace #2}
}

\newcommand{\gameaitablejust}{\justifying\parindent=0pt
  \hyphenpenalty=50\exhyphenpenalty=200
  \tolerance=1500\emergencystretch=2em\relax}
\newcolumntype{J}[1]{>{\gameaitablejust\arraybackslash}p{#1}}

\usepackage{tikz}
\newlength{\tilew}\newlength{\tilelabelh}
\newcommand{\figmark}[1]{\textsuperscript{\sffamily\scriptsize\color{GameSlate}[Fig.~#1]}}
\newcommand{\blocktile}[4]{\begin{minipage}[t]{\tilew}\centering
    \begin{tikzpicture}
      \clip[rounded corners=1.6mm] (0,0) rectangle (\tilew,0.625\tilew);
      \node[anchor=south west,inner sep=0] at (0,0) {\includegraphics[width=\tilew]{#1}};
    \end{tikzpicture}\par\vspace{2pt}
    \parbox[t][\tilelabelh][t]{\tilew}{\centering{\sffamily\bfseries\fontsize{7.8}{9}\selectfont\color{GameInk}#2}\\[1pt]{\fontsize{7.3}{8.6}\selectfont\itshape\color{GameInk}#4}}
  \end{minipage}}

\begin{document}

\newgeometry{top=13.5mm,bottom=15mm,left=23mm,right=23mm,headheight=16pt,headsep=8mm}
\begin{gameaifrontmatter}
  {\small\color{GameInk}\justifying\noindent Foundation models, alongside rapid advances in learned game-world models, are reshaping how AI is used across the game lifecycle. Beyond playing games, recent systems model players and game dynamics, support design and development, adapt player-facing experiences at runtime, and evaluate the resulting artifacts. Yet these directions have largely evolved as separate research threads, making it difficult to distinguish capabilities that transfer across settings from those that remain tied to particular games, engines, interfaces, or player populations. We organize the literature into six roles according to the immediate use of AI output: \emph{AI~that \textcolor[HTML]{599BD4}{\textbf{Plays and Acts}}}, \emph{AI~that \textcolor[HTML]{70AE47}{\textbf{Models Players and Games}}}, \emph{AI~that \textcolor[HTML]{5757AF}{\textbf{Designs Games}}}, \emph{AI~that \textcolor[HTML]{C75A11}{\textbf{Builds and Maintains Games}}}, \emph{AI~that \textcolor[HTML]{0ED1D2}{\textbf{Generates and Adapts at Runtime}}}, and \emph{AI~that \textcolor[HTML]{B58801}{\textbf{Tests and Evaluates Games}}}. For each role, we examine what structure is supplied by the game or workflow and what AI learns, generates, predicts, or revises; which capabilities transfer and which artifacts can be reused across settings and roles; and what claims are supported by the available evidence. We further identify concrete cross-role connections: trajectories can train world models, learned environments can provide experience for agents, design specifications can drive executable implementations, and feedback from play or testing can guide revision. Evaluation is most standardized and execution-grounded for bounded game playing and selected learned environments, while persistent state in learned worlds, repeated software revision, validated player modeling, sustained runtime adaptation, and representative automated testing remain less established. Together, these findings point to a central challenge for AI in games: enabling outputs and capabilities to be reused or transferred across roles while re-establishing evidence for their effectiveness in the game-specific structures, interfaces, and player contexts where they are ultimately used. \par}
  \gameaiProjectGitHubRow
\end{gameaifrontmatter}
\begingroup\renewcommand{\thefootnote}{\dag}\footnotetext{Corresponding author.}\endgroup
\gameaiTitleLens
\restoregeometry

\gameaiContents

\section{Introduction}
\label{sec:intro}

AI for games has long extended beyond playing. Recent uses of GPT-6 Astra and Claude Fable 5.2 make this breadth particularly visible. ARC-AGI-3 evaluates both models in unfamiliar interactive games without natural-language instructions, where they must discover how each game works, infer its goal, and determine how to act \citep{arcprize2026astra,arcprize2026benchmark,arcprize2026fable}. Playco reports using GPT-6 Astra within Playbot, an engine-connected development tool, to build three themed, playable prototypes from one grey-box foundation \citep{openai2026playbot}. An Anthropic engineer reports that Fable 5.2 generated the first version of Saltline, a browser-playable first-person shooter, from one prompt in about eight hours without human feedback \citep{caswell2026fable}. Together, these uses place each pretrained model in both player and builder roles across the game lifecycle, with different tasks to perform and different contributions to the game. \Cref{fig:title-teaser} illustrates the range of settings considered in this survey, from benchmark environments to commercial and generated games.

AI's participation across this lifecycle draws on several established research traditions. Game-playing research has developed methods for selecting actions under a game's rules, with advances through search, reinforcement learning, and self-play \citep{shannon1950chess,mnih2015dqn,silver2016alphago,vinyals2019alphastar,berner2019dota}. In parallel, researchers have used procedural generation to produce game content \citep{togelius2011searchpcg} and automated design to explore possible rules and mechanics \citep{browne2010evolutionary}. Interactive narrative systems shape how stories unfold in response to player actions \citep{mateas2005facade}, while mixed-initiative systems support authors during design \citep{smith2011tanagra}. Understanding the resulting experience has motivated work on player modeling, including the use of predicted preferences to guide content adaptation \citep{yannakakis2011edpcg,bakkes2012playermodeling}. Automated playtesting has also used simulated players to examine how different play styles expose different aspects of the same game \citep{holmgard2019playtesting,politowski2022testing}. Much of this work developed in separate research communities around particular AI roles.

\begin{figure}[H]
  \centering
  \includegraphics[width=\linewidth]{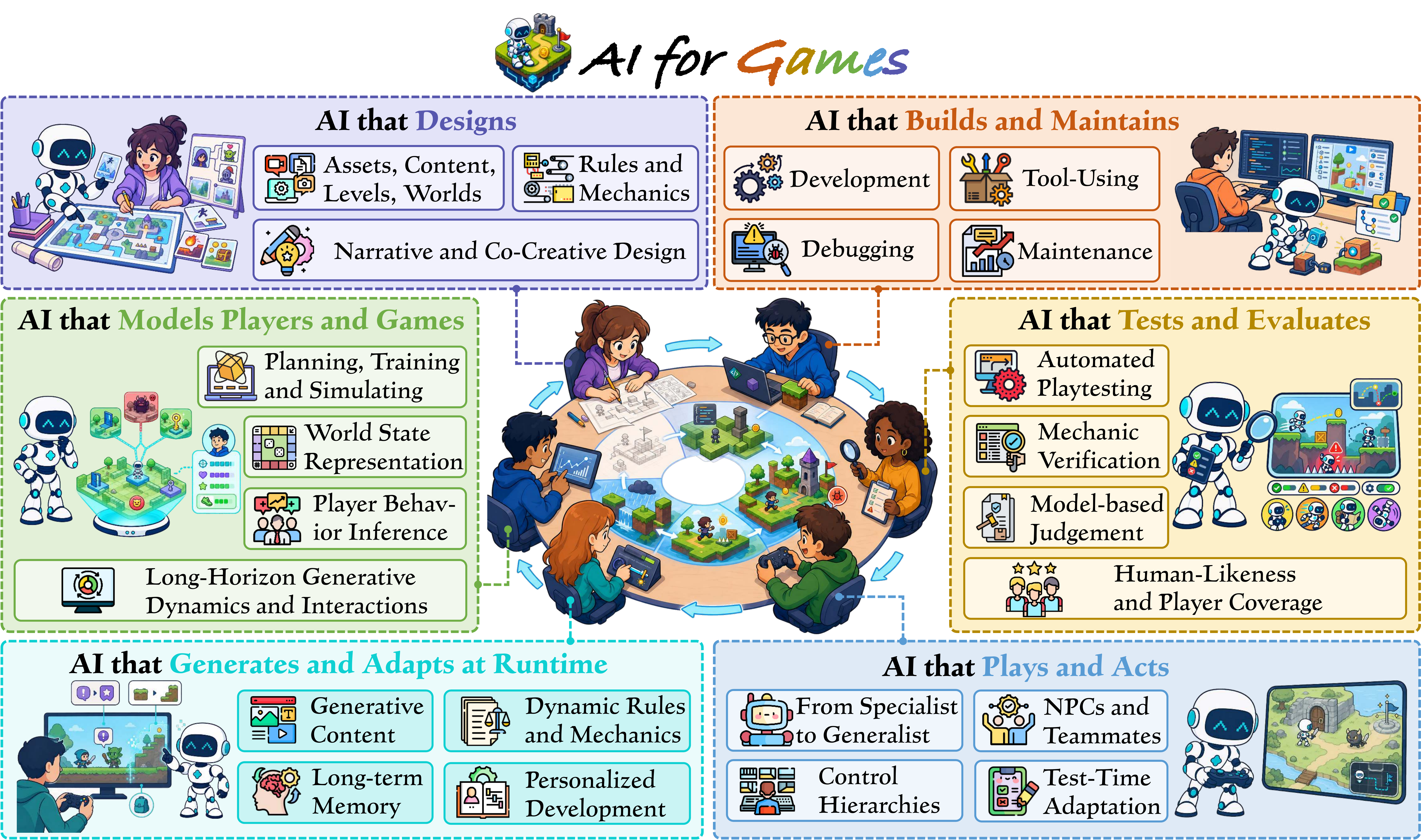}
  \caption{Six uses of AI output around an executable, player-facing game. The connectors indicate selective inputs and feedback rather than a mandatory pipeline.}
  \label{fig:role-lifecycle}
\end{figure}

Foundation models provide new ways to approach and connect these established tasks (\Cref{fig:role-lifecycle}). Developers can describe a design intention in natural language and connect pretrained models to tools for implementing it. DreamGarden, for example, develops a high-level idea into a hierarchical plan that designers can inspect and revise, while specialist modules generate assets and code \citep{earle2025dreamgarden}. Access to rendered gameplay also allows development systems to inspect the behavior of what they generate. Play2Code connects a coding agent to a browser-based playtester, which interacts with the running game and supplies observations for further revisions \citep{huang2026guigames}. Language, visual understanding, and tool use support both proposing changes and inspecting their effects in play.

Models can also participate directly in the interaction between a player and the game environment. In IF:CARGO, players express rules in natural language, and a language model translates them into constrained commands that the engine validates and executes \citep{hsu2026ifcargo}. The model's interpretation therefore becomes part of how the game responds to its players. GameNGen learns the environment's responses to actions from gameplay trajectories, producing an interactive simulator \citep{valevski2024gamengen}. Learned environments can also support the training of game-playing agents. World-model research has developed ways to learn transition dynamics from observations and supply imagined experience for policy learning \citep{ha2018worldmodels,hafner2020dreamer,luo2026physicalrsi}, and Dreamer~4 trains a policy inside a learned environment \citep{hafner2025dreamer4}. Across these applications, models contribute both to the environment in which interaction occurs and to the behavior of agents acting within it.

\textbf{Existing surveys establish the breadth of game AI and its major roles.} Broad syntheses cover game playing, content generation, and player modeling \citep{yannakakis2025book}, as well as LLM-centered applications \citep{gallotta2024,sweetser2024llmgames}. Gallotta et al. already organize LLM applications around roles within games \citep{gallotta2024}. Focused surveys examine game-playing agents \citep{hu2024gameagents,xu2024gameplaying}, machine-learned and LLM-assisted content generation \citep{summerville2018pcgml,maleki2024pcg}, player modeling \citep{bakkes2012playermodeling}, interactive world models \citep{liu2026interactivewm}, social agents \citep{feng2024socialagents}, LLM-based non-player characters \citep{mian2026llmnpc}, generative game development \citep{ternar2025gamedev}, and automated testing \citep{politowski2022testing}. Together, these accounts provide the foundations for comparing methods within individual areas.

Our survey connects these areas through the outputs that pass between them. It brings executable development and maintenance, runtime generation, and automated evaluation into the same analysis as acting, modeling, and design, while tracing their learned and symbolic predecessors. For each role, we examine the game structure that remains supplied, the transfer and reuse of capabilities and artifacts, and the evidence supporting an output at its point of use. This makes it possible to distinguish a proposed mechanic from its implementation, a player prediction from the adaptation it informs, and a test verdict from the repair it guides. The contribution is a synthesis of these relationships and of the empirical support available across the six roles.

\textbf{Concretely, we organize AI for games by the immediate use of the system's output.} Systems \concept{play and act} through decisions and communication, \concept{model players and games} through predictions and representations, and \concept{design} through content and rule proposals. Systems that \concept{build and maintain} implement and revise software; those that \concept{generate and adapt at runtime} change the live experience; and those that \concept{test and evaluate} produce evidence about behavior or quality. These uses distinguish contributions that can share an architecture: proposing an interesting mechanic, implementing it correctly, and checking how people encounter it are different achievements. Design assistance helps before implementation; testing guides successive development stages.

\textbf{Cross-role connections change what can be learned from gameplay.} A trajectory can become simulator training data, an experiment inside a learned environment, or diagnostic evidence for software repair. Its value depends on the receiving task. For example, a policy can exploit errors in a learned simulator \citep{ha2018worldmodels}, while automated and human playtesters can expose different behaviors and defects \citep{ariyurek2021testing}. Comparing these exchanges reveals both new opportunities for reuse and the game-specific information that must accompany an output, including action semantics, state, requirements, and the behavior of the intended players.

Taken together, the survey connects technical lineages across six roles, compares methods and reported results under their actual operating conditions, and synthesizes the exchanges demonstrated between roles. Three findings recur: broad pretraining expands available interfaces without removing game-specific structure; gameplay increasingly supplies data and feedback beyond an agent's final score; and progress remains task-dependent, with stronger shared benchmarks for bounded play than for sustained creation, adaptation, and human experience. Section~\ref{sec:taxonomy} introduces the organizing questions and role boundaries. Sections~\ref{sec:agents}--\ref{sec:testing} develop the technical review; Sections~\ref{sec:evaluation}--\ref{sec:challenges} compare evidence, connect the findings, and identify open problems.

\section{Roles Across the Game Lifecycle}
\label{sec:taxonomy}

We survey AI systems that directly participate in playing and acting in interactive games, modeling players or games, designing game content and mechanics, building or maintaining executable game artifacts, generating or adapting player-facing experiences at runtime, or testing and evaluating games and game artifacts. Particular attention is given to how broad pretraining and language, multimodal, code, and tool interfaces reshape these roles. The foundation-model era therefore serves as an analytical lens rather than an inclusion criterion. Earlier learned and symbolic systems show which task structures, constraints, and evaluation problems predate current models; recent specialist systems are included when they clarify how the same roles are being extended or connected. Player modeling is included when it informs behavior, adaptation, or evaluation, and media generation is included when it enters an evaluated game artifact.

Across the six roles, the analysis returns to three questions:

\begin{tcolorbox}[
  enhanced,
  colback=GameMist,
  colframe=GameLine,
  boxrule=.6pt,
  arc=2.2mm,
  left=3mm,right=3mm,top=2.2mm,bottom=2.2mm,
  before skip=3pt,after skip=5pt
]
\small
\begin{tabularx}{\linewidth}{@{}L{3.55cm}Y@{}}
\toprule
\textbf{Question} & \textbf{Recurring inquiry} \\
\midrule
\textbf{Boundary}
& What is supplied by the game or workflow, and what is assigned to AI? \\
\textbf{Transfer and Reuse}
& Which capabilities transfer, which artifacts can be reused, and what remains setting-specific? \\
\textbf{Evidence}
& What claims are supported by evaluation at the point of use? \\
\bottomrule
\end{tabularx}
\end{tcolorbox}

\emph{Boundary} distinguishes supplied structure from what AI learns, generates, predicts, or revises. \emph{Transfer} concerns competence under a changed game, engine, interface, player population, or task; \emph{reuse} concerns a representation, trace, specification, model, or feedback signal consumed elsewhere. Sharing a pretrained backbone or passing an artifact between components does not, by itself, demonstrate transfer. \emph{Evidence} asks what the resulting system has been shown to accomplish. Table~\ref{tab:survey-overview} lists outputs, applications, and empirical claims for each role.

\begin{table}[H]
\centering
\caption{Six roles: outputs, applications, and the principal empirical claim.}
\label{tab:survey-overview}
\footnotesize
\setlength{\tabcolsep}{3.5pt}
\renewcommand{\arraystretch}{1.16}
\begin{tabularx}{\textwidth}{@{}L{3.45cm}L{4.10cm}Y L{2.95cm}@{}}
\toprule
\textbf{Role} & \textbf{Output} & \textbf{Applications} & \textbf{Claim} \\
\midrule
\textcolor{Player}{\textbf{Play and Act}} & actions; plans; messages & players; teammates; NPCs & action quality \\
\textcolor{Simulator}{\textbf{Model Players and~Games}} & states; transitions; \mbox{player forecasts} & planning; simulation; \mbox{player models} & \mbox{predictive quality}; \mbox{policy transfer} \\
\textcolor{Creator}{\textbf{Design}} & levels; rules; story structures & PCG; automated design; co-creation & validity; control \\
\textcolor{Builder}{\textbf{Build and Maintain}} & code; scenes; project edits & engine agents; debugging; repair & working software \\
\textcolor{Resident}{\textbf{Generate and Adapt at~Runtime}} & dialogue; quests; live content & characters; \mbox{adaptive narrative} & \mbox{consistent content}; \mbox{player response} \\
\textcolor{Evaluator}{\textbf{Test and Evaluate}} & traces; verdicts; diagnoses & playtesting; verification; judging & \mbox{state coverage}; \mbox{verdict accuracy} \\
\bottomrule
\end{tabularx}
\end{table}

The taxonomy classifies outputs by role rather than model architecture, and roles are not mutually exclusive at the system level. It describes what an output is used for, whereas the lifecycle describes when it is used. A role may recur across phases, and a phase may involve several roles.

\subsection{Assigning Primary and Secondary Roles}

A system may serve several roles, so primary placement follows the immediate use of its output. Actions map to \emph{Play and Act}; predictions of player behavior or game dynamics map to \emph{Model Players and Games}; and content or rule proposals evaluated for design quality map to \emph{Design}. Executable artifacts evaluated for implementation quality map to \emph{Build and Maintain}; session-specific outputs evaluated through their effects on a live, player-facing experience map to \emph{Generate and Adapt at Runtime}; and test traces or judgments map to \emph{Test and Evaluate}. Predictions or simulations produced for planning or training remain in the modeling role even when they run online. When a system spans several roles, we assign its primary role according to the output on which its principal empirical claim rests, while cross-references record secondary roles.

For example, an NPC policy belongs to \emph{Play and Act} when the claim concerns action or coordination, whereas session-specific dialogue or behavior evaluated for its effect on the live player experience belongs to the runtime role. A playtester belongs to \emph{Test and Evaluate} when the claim concerns coverage or defects, even though it acts through a game interface. MarioGPT proposes level content and is therefore categorized as \emph{Design} \citep{sudhakaran2023mariogpt}; Play2Code implements and repairs software (\emph{Build and Maintain}) \citep{huang2026guigames}.

\begin{figure}[H]
  \centering
  \includegraphics[width=\linewidth]{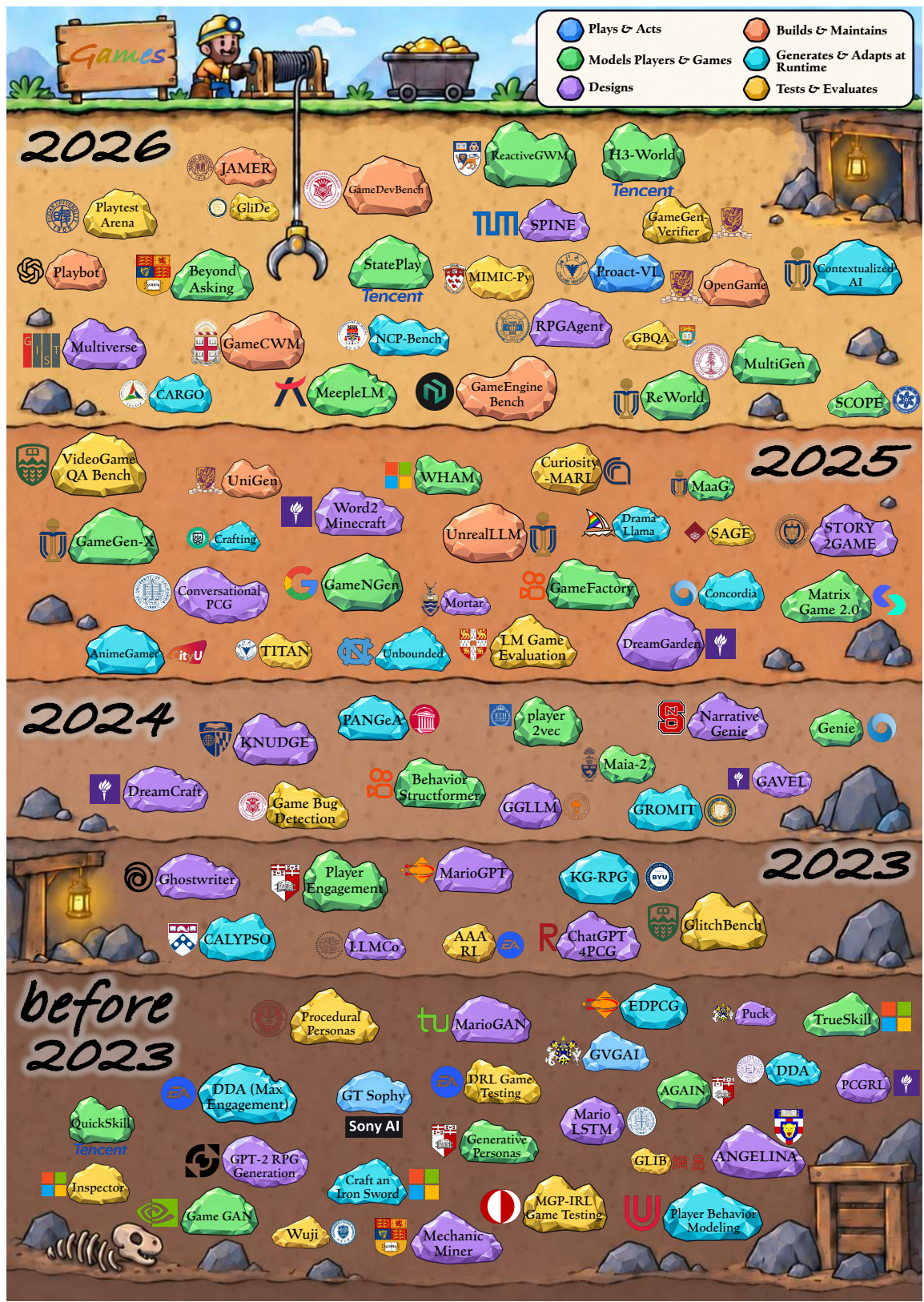}
  \caption{\textbf{Timeline of representative AI-for-games systems across the six roles.} Selected systems, benchmarks, and methods reviewed in this survey are arranged by year and colored according to their primary role in our taxonomy.}
  \label{fig:role-timeline}
\end{figure}

\Cref{fig:role-timeline} situates representative systems, benchmarks, and methods by publication year and primary role, while \Cref{fig:knowledge-map} provides a topic-based guide to the technical subareas and representative work covered in the following chapters. The system index in \Cref{app:resources} records primary and secondary roles for systems and named components, while the chapter illustrations summarize the research questions and workflows within each role.

\begingroup
\setcitestyle{numbers,square,comma}
\hypersetup{citecolor=GameCitation}
\setlength{\parskip}{0pt}
\tikzset{
  gameai coverage box/.style={rectangle,draw=GameInk!72,
    rounded corners=2pt,text=GameInk,line width=.65pt,
    inner xsep=4pt,inner ysep=3.6pt,align=center},
  gameai coverage category/.style={gameai coverage box,fill=white,
    font=\fontsize{10}{11.6}\selectfont\bfseries},
  gameai coverage function/.style={gameai coverage box,fill=white,
    font=\fontsize{10}{11.6}\selectfont\bfseries\itshape,
    text width=34mm,minimum height=9mm},
  gameai coverage panel/.style={gameai coverage box,align=left,
    font=\fontsize{10}{11.6}\selectfont,inner xsep=5pt},
  gameai play panel/.style={gameai coverage panel,fill=RolePanelPlayer},
  gameai model panel/.style={gameai coverage panel,fill=RolePanelSimulator},
  gameai design panel/.style={gameai coverage panel,fill=RolePanelCreator},
  gameai build panel/.style={gameai coverage panel,fill=RolePanelBuilder},
  gameai runtime panel/.style={gameai coverage panel,fill=RolePanelResident},
  gameai test panel/.style={gameai coverage panel,fill=RolePanelEvaluator}
}
\newcommand{\gameaiMapRole}[2]{\parbox[c][22pt][c]{98pt}{\centering\color{#1}\bfseries #2}}
\newcommand{\gameaiMapWorks}[1]{\parbox[t]{104mm}{\fontsize{10}{11.6}\selectfont\RaggedRight
    \hyphenpenalty=10000\exhyphenpenalty=10000 #1}}
\forestset{gameai knowledge tree/.style={
  forked edges,
  for tree={grow=east,reversed=true,anchor=west,parent anchor=east,
    child anchor=west,base=left,tier/.option=level,gameai coverage box,
    font=\fontsize{10}{11.6}\selectfont,
    edge+={GameInk!62,line width=.65pt},
    s sep=3.1pt,l sep=7pt,fork sep=3.5pt,
    ver/.style={rotate=90,child anchor=north,parent anchor=south,anchor=center}},
  where level=0{ver,fill=white,font=\fontsize{10}{11}\selectfont\bfseries,
    minimum width=82pt,minimum height=20pt,l sep=8pt}{},
  where level=1{ver,gameai coverage category,l sep=8pt}{},
  where level=2{gameai coverage function,l sep=7pt}{},
  where level=3{align=left}{}
}}

\begin{figure}[p]
\centering
\resizebox{.985\textwidth}{!}{\begin{forest} gameai knowledge tree
[AI for Games
  [{\gameaiMapRole{Player}{Play and Act}},draw=Player,text=Player
    [{Player and\\generalist agents}
      [{\gameaiMapWorks{DQN~\citep{mnih2015dqn}; AlphaZero~\citep{silver2018alphazero}; AlphaStar~\citep{vinyals2019alphastar}; GT Sophy~\citep{wurman2022sophy}; GVGAI~\citep{perezliebana2018gvgai}; Procgen~\citep{cobbe2020procgen}; DreamerV3~\citep{hafner2025dreamerv3}; Decision Transformer~\citep{chen2021decisiontransformer}; Gato~\citep{reed2022gato}; SIMA 2~\citep{sima22025}; NitroGen~\citep{magne2026nitrogen}; Gamer~\citep{luo2026gamer}; Cradle~\citep{tan2025cradle}.}},gameai play panel]]
    [{Learning and\\control hierarchies}
      [{\gameaiMapWorks{SPRING~\citep{wu2023spring}; GITM~\citep{zhu2023gitm}; DEPS~\citep{wang2023deps}; JARVIS-1~\citep{wang2023jarvis1}; Plan4MC~\citep{yuan2023plan4mc}; Voyager~\citep{wang2023voyager}; STEVE-1~\citep{lifshitz2023steve1}; MineDreamer~\citep{zhou2025minedreamer}; VPT~\citep{baker2022vpt}; OmniJARVIS~\citep{wang2024omnijarvis}; CombatVLA~\citep{chen2025combatvla}; ROCKET-1~\citep{cai2024rocket1}; TaskFrame~\citep{wang2026taskframe}; HiSTEVE~\citep{tang2026histeve}; Player-1~\citep{ouyang2026player1}; Game2Player~\citep{jiang2026game2player}.}},gameai play panel]]
    [{Test-time adaptation\\and memory}
      [{\gameaiMapWorks{REGENT~\citep{sridhar2025regent}; S3Gym~\citep{shi2026s3gym}; Twin (test-time digital twin)~\citep{skoutnev2026twin}; Code World Models (GGP)~\citep{lehrach2025cwm}; Optimus-1~\citep{li2024optimus}; EMemBench~\citep{li2026emembench}; GameVerse~\citep{zhang2026gameverse}; FlashAdventure~\citep{ahn2025flashadventure}.}},gameai play panel]]
    [{Opponents\\and teammates}
      [{\gameaiMapWorks{PSRO~\citep{lanctot2017unified}; QMIX~\citep{rashid2018qmix}; ToMnet~\citep{rabinowitz2018machine}; Overcooked~\citep{carroll2019overcooked}; Other-Play~\citep{hu2020otherplay}; Fictitious Co-Play~\citep{strouse2021fcp}; ProAgent~\citep{zhang2024proagent}; CICERO~\citep{fair2022cicero}; Werewolf Arena~\citep{bailis2024werewolf}; Project Sid~\citep{al2024project}; Proact-VL~\citep{yan2026proact}.}},gameai play panel]]
  ]
  [{\gameaiMapRole{Simulator}{Model Players\\and Games}},draw=Simulator,text=Simulator
    [{Planning models\\and simulators}
      [{\gameaiMapWorks{Dyna~\citep{sutton1990dyna}; World Models~\citep{ha2018worldmodels}; I2A~\citep{racaniere2017i2a}; PlaNet~\citep{hafner2019planet}; MuZero~\citep{schrittwieser2020muzero}; SimPLe~\citep{kaiser2020simple}; Dreamer~\citep{hafner2020dreamer}; Dreamer 4~\citep{hafner2025dreamer4}; GameGAN~\citep{kim2020gamegan}; Genie~\citep{bruce2024genie}; GameNGen~\citep{valevski2024gamengen}; GameGen-X~\citep{che2024gamegenx}.}},gameai model panel]]
    [{Representation and\\persistent state}
      [{\gameaiMapWorks{IRIS~\citep{micheli2023iris}; DIAMOND~\citep{alonso2024diamond}; PETS~\citep{chua2018pets}; Plan2Explore~\citep{sekar2020planning}; MOPO~\citep{yu2020mopo}; WorldMem~\citep{xiao2025worldmem}; Context-as-Memory~\citep{yu2025context}; VMem~\citep{li2025vmem}; ReWorld~\citep{chen2026reworld}; WorldCoder~\citep{tang2024worldcoder}; Model as a Game~\citep{chen2025modelgame}; StatePlay~\citep{lin2026stateplaystateawaregameworld}; Fight World Model~\citep{li2026fightworld}; BoxWorld~\citep{wu2026boxworld}; WorldRuler~\citep{zhai2026worldruler}; PlayGen2~\citep{zhang2026playgen2}.}},gameai model panel]]
    [{Player models}
      [{\gameaiMapWorks{TrueSkill~\citep{herbrich2006trueskill}; QuickSkill~\citep{zhang2022quickskill}; player2vec~\citep{wang2024player2vec}; Behavior Structformer~\citep{smirnov2024behaviorstructformer}; Maia~\citep{mcilroyyoung2020aligning}; Maia-2~\citep{tang2024maia2}; Maia-3 / Chessformer~\citep{monroe2026chessformer}; multimodal engagement~\citep{pinitas2023predicting}; Generative Personas~\citep{barthet2022generative}; behavior-inferred profiles~\citep{lu2026personalized}; MeepleLM~\citep{li2026meeplelm}; MMO player simulation~\citep{zhang2025beyond}; LM evaluations of games~\citep{collins2025gameevaluations}.}},gameai model panel]]
    [{Action interfaces\\and generation}
      [{\gameaiMapWorks{GameFactory~\citep{yu2025gamefactory}; WHAM~\citep{kanervisto2025wham}; SCOPE~\citep{tong2026scope}; Incantation~\citep{zhu2026incantationnaturallanguageaction}; ReactiveGWM~\citep{wang2026reactivegwmsteeringnpcreactive}; H3-World~\citep{chen2026h3world}; Diffusion Forcing~\citep{chen2024diffusionforcing}; CausVid~\citep{yin2024slow}; Self Forcing~\citep{huang2025self}; FramePack~\citep{zhang2025frame}; Matrix-Game 2.0~\citep{he2025matrixgame2}; MultiGen~\citep{po2026multigen}.}},gameai model panel]]
  ]
  [{\gameaiMapRole{Creator}{Design}},draw=Creator,text=Creator
    [{Assets, levels,\\and worlds}
      [{\gameaiMapWorks{Super Mario as a String~\citep{summerville2016mario}; MarioGAN~\citep{volz2018mariogan}; PCGRL~\citep{khalifa2020pcgrl}; MarioGPT~\citep{sudhakaran2023mariogpt}; Word2Minecraft~\citep{huang2025word2minecraft}; Moonshine~\citep{nie2025moonshine}; MarioDiffusion~\citep{schrum2025mariodiffusion}; ChatGPT4PCG~\citep{taveekitworachai2023chatgpt4pcg}; Multiverse~\citep{baek2026multiverse}; conversational generator control~\citep{whitehead2025conversational}; DreamCraft~\citep{earle2024dreamcraft}; database-driven 3D levels~\citep{xu2025database3d}.}},gameai design panel]]
    [{Rules and\\mechanics}
      [{\gameaiMapWorks{Mechanic Miner~\citep{cook2013mechanicminer}; ANGELINA~\citep{cook2017angelina1}; Puck~\citep{cook2022puck}; VGDL generation~\citep{hu2024gamegeneration}; GAVEL~\citep{todd2024gavel}; Mortar~\citep{nasir2025mortar}; ScriptDoctor~\citep{earle2025scriptdoctor}; STORY2GAME~\citep{zhou2025story2game}.}},gameai design panel]]
    [{Narrative and\\co-creative design}
      [{\gameaiMapWorks{Tanagra~\citep{smith2011tanagra}; Sentient Sketchbook~\citep{liapis2013sketchbook}; RPG quest generation~\citep{vanstegeren2021quests}; KNUDGE~\citep{weir2024knudge}; SceneCraft~\citep{kumaran2023scenecraft}; NarrativeGenie~\citep{kumaran2024narrativegenie}; RPGAgent~\citep{zhang2026rpgagent}; LLM interactive evolution~\citep{lanzi2023llmco}; DreamGarden~\citep{earle2025dreamgarden}; Ghostwriter~\citep{barth2023ghostwriter}; SPINE~\citep{geheeb2026spine}.}},gameai design panel]]
  ]
]
\end{forest}}
\caption{Taxonomy of AI for games with representative work. \textbf{(a) Playing, modeling, and design.} Each branch lists the systems, benchmarks, and environments of one survey subsection.}
\label{fig:knowledge-map}
\end{figure}
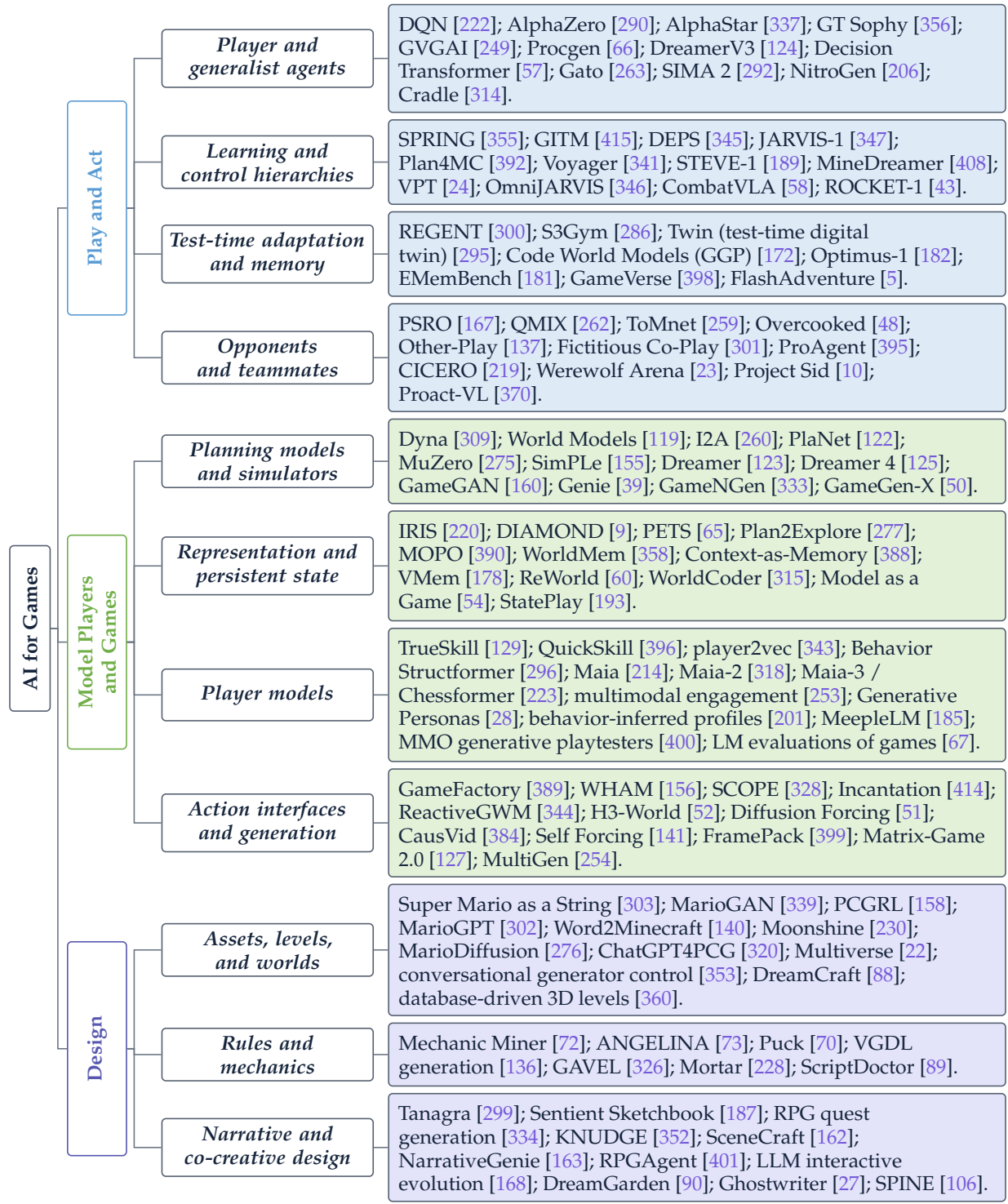

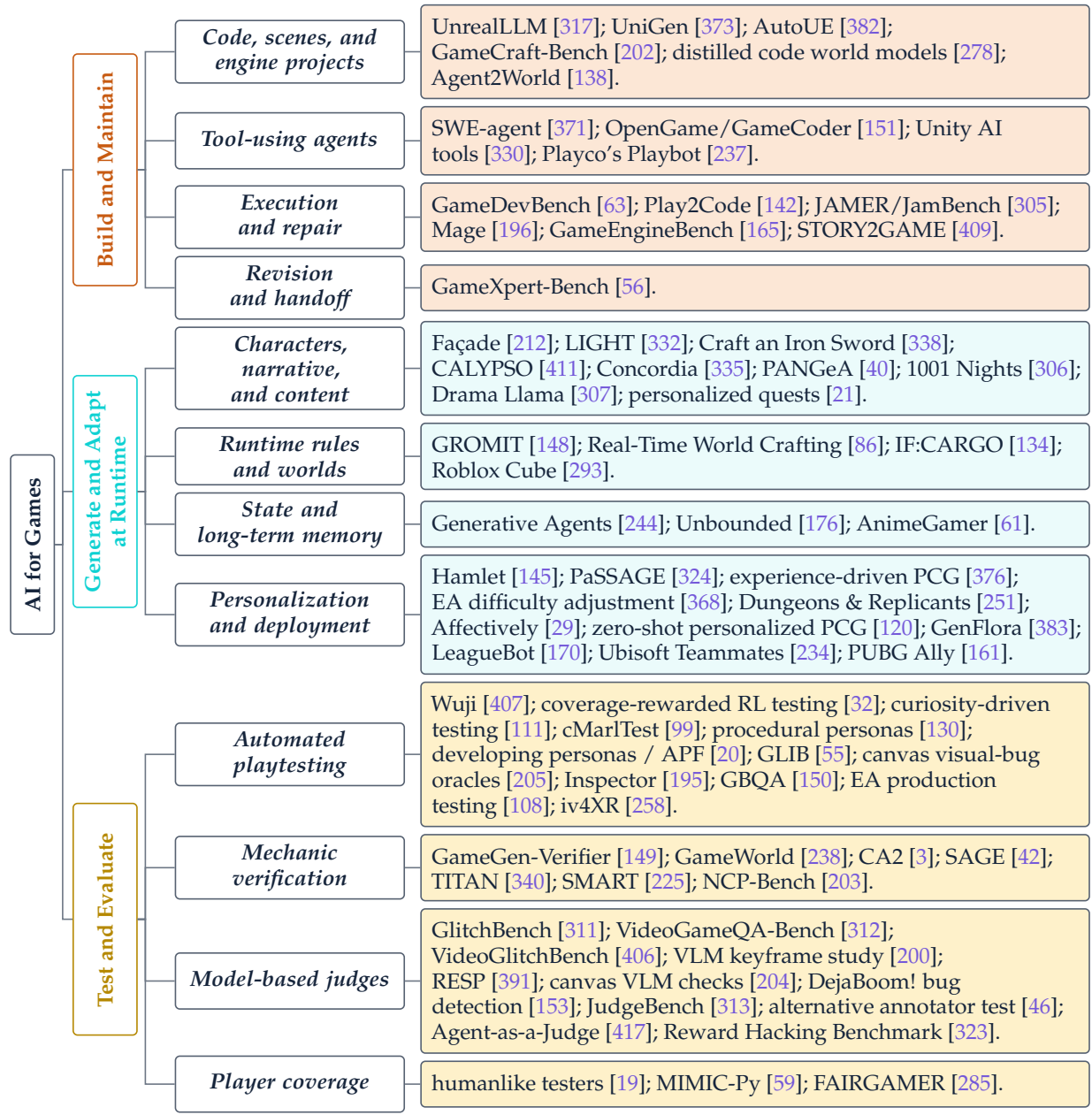
\begin{figure}[p]
\ContinuedFloat
\centering
\resizebox{.985\textwidth}{!}{\begin{forest} gameai knowledge tree
[AI for Games
  [{\gameaiMapRole{Builder}{Build and Maintain}},draw=Builder,text=Builder
    [{Code, scenes, and\\engine projects}
      [{\gameaiMapWorks{UnrealLLM~\citep{songtang-etal-2025-unrealllm}; UniGen~\citep{yang2025unigen}; AutoUE~\citep{yin2026autoue}; GameCraft-Bench~\citep{luo2026gamecraft}; distilled code world models~\citep{serapio2026gamecwm}; Agent2World~\citep{hu2025agent2world}.}},gameai build panel]]
    [{Tool-using agents}
      [{\gameaiMapWorks{SWE-agent~\citep{yang2024sweagent}; OpenGame/GameCoder~\citep{jiang2026opengame}; Unity AI tools~\citep{unity2026aitools}; Playco's Playbot~\citep{openai2026playbot}.}},gameai build panel]]
    [{Execution\\and repair}
      [{\gameaiMapWorks{GameDevBench~\citep{chi2026gamedevbench}; Play2Code~\citep{huang2026guigames}; JAMER/JamBench~\citep{sun2026jamer}; Mage~\citep{liu2026mage}; GameEngineBench~\citep{la2026gameenginebench}; UnityBench-Mini3D~\citep{zhou2026unitybench}; OpenGameEval~\citep{turkel2026opengameeval}; 3D GameDev Arena~\citep{bai2026gamedevarena}; OCTAVE~\citep{li2026octave}.}},gameai build panel]]
    [{Revision\\and handoff}
      [{\gameaiMapWorks{GameXpert-Bench~\citep{chen2026gamexpert}; IGMBench~\citep{ku2026worldediting}.}},gameai build panel]]
  ]
  [{\gameaiMapRole{Resident}{Generate and Adapt\\at Runtime}},draw=Resident,text=Resident
    [{Characters,\\narrative, and content}
      [{\gameaiMapWorks{Fa\c{c}ade~\citep{mateas2005facade}; LIGHT~\citep{urbanek2019light}; Craft an Iron Sword~\citep{volum2022craft}; CALYPSO~\citep{zhu2023calypso}; Concordia~\citep{vezhnevets2025concordia}; PANGeA~\citep{buongiorno2024pangea}; 1001 Nights~\citep{sun2023language}; Drama Llama~\citep{sun2025dramallama}; personalized quests~\citep{ashby2023quests}.}},gameai runtime panel]]
    [{Runtime rules\\and worlds}
      [{\gameaiMapWorks{GROMIT~\citep{jennings2024gromit}; Real-Time World Crafting~\citep{drake2025worldcrafting}; IF:CARGO~\citep{hsu2026ifcargo}; Roblox Cube~\citep{singh2026robloxcube}.}},gameai runtime panel]]
    [{State and\\long-term memory}
      [{\gameaiMapWorks{Generative Agents~\citep{park2023generativeagents}; Unbounded~\citep{li2025unbounded}; AnimeGamer~\citep{cheng2025animegamer}.}},gameai runtime panel]]
    [{Personalization\\and deployment}
      [{\gameaiMapWorks{Hamlet~\citep{hunicke2005dda}; PaSSAGE~\citep{thue2007passage}; experience-driven PCG~\citep{yannakakis2011edpcg}; EA difficulty adjustment~\citep{xue2017dynamic}; Dungeons \& Replicants~\citep{pfau2020dungeons}; Affectively~\citep{barthet2024affectively}; zero-shot personalized PCG~\citep{hafnar2025zeroshot}; GenFlora~\citep{yin2026contextualized}; LeagueBot~\citep{lee2026leaguebot}; Ubisoft Teammates~\citep{ubisoft2025teammates}; PUBG Ally~\citep{krafton2026allyduo}.}},gameai runtime panel]]
  ]
  [{\gameaiMapRole{Evaluator}{Test and Evaluate}},draw=Evaluator,text=Evaluator
    [{Automated playtesting}
      [{\gameaiMapWorks{Wuji~\citep{zheng2019wuji}; RL-augmented testing~\citep{bergdahl2021augmenting}; curiosity-driven testing~\citep{gordillo2021improving}; cMarlTest~\citep{ferdous2025curiosity}; procedural personas~\citep{holmgard2019playtesting}; developing personas / APF~\citep{ariyurek2021playtesting}; GLIB~\citep{chen2021glib}; canvas visual-bug oracles~\citep{macklon2022automatically}; Inspector~\citep{liu2022inspector}; GBQA~\citep{jiang2026gbqa}; EA production testing~\citep{gillberg2023productiontesting}; iv4XR~\citep{prasetya2022agent}.}},gameai test panel]]
    [{Mechanic verification}
      [{\gameaiMapWorks{GameGen-Verifier~\citep{jia2026gamegenverifier}; GameWorld~\citep{ouyang2026gameworld}; CA2~\citep{adaikkappan2026ca2}; SAGE~\citep{cai2025sage}; TITAN~\citep{wang2025titan}; SMART~\citep{mu2025smart}; NCP-Bench~\citep{ma2026ncpbench}.}},gameai test panel]]
    [{Model-based judges}
      [{\gameaiMapWorks{GlitchBench~\citep{taesiri2023glitchbench}; VideoGameQA-Bench~\citep{taesiri2025videogameqa}; VideoGlitchBench~\citep{zheng2026open}; VLM keyframe study~\citep{lu2026far}; RefGlitch-Bench~\citep{yu2026resp}; canvas VLM checks~\citep{macklon2025exploring}; DejaBoom! bug detection~\citep{jin2024bugdetection}.}},gameai test panel]]
    [{Player coverage}
      [{\gameaiMapWorks{humanlike testers~\citep{ariyurek2021testing}; MIMIC-Py~\citep{chen2026mimicpy}; FAIRGAMER~\citep{shi2026fairgamer}.}},gameai test panel]]
  ]
]
\end{forest}}
\caption[]{Taxonomy of AI for games with representative work (continued). \textbf{(b) Building, runtime generation and adaptation, and testing.}}
\end{figure}
\endgroup
\pretitlemark{section}{AI That Plays and Acts}
\clearpage

\gameaisectionaccent{Player}
\section{AI That Plays and Acts}
\label{sec:agents}

\begin{figure}[H]
  \centering
  \includegraphics[width=\linewidth]{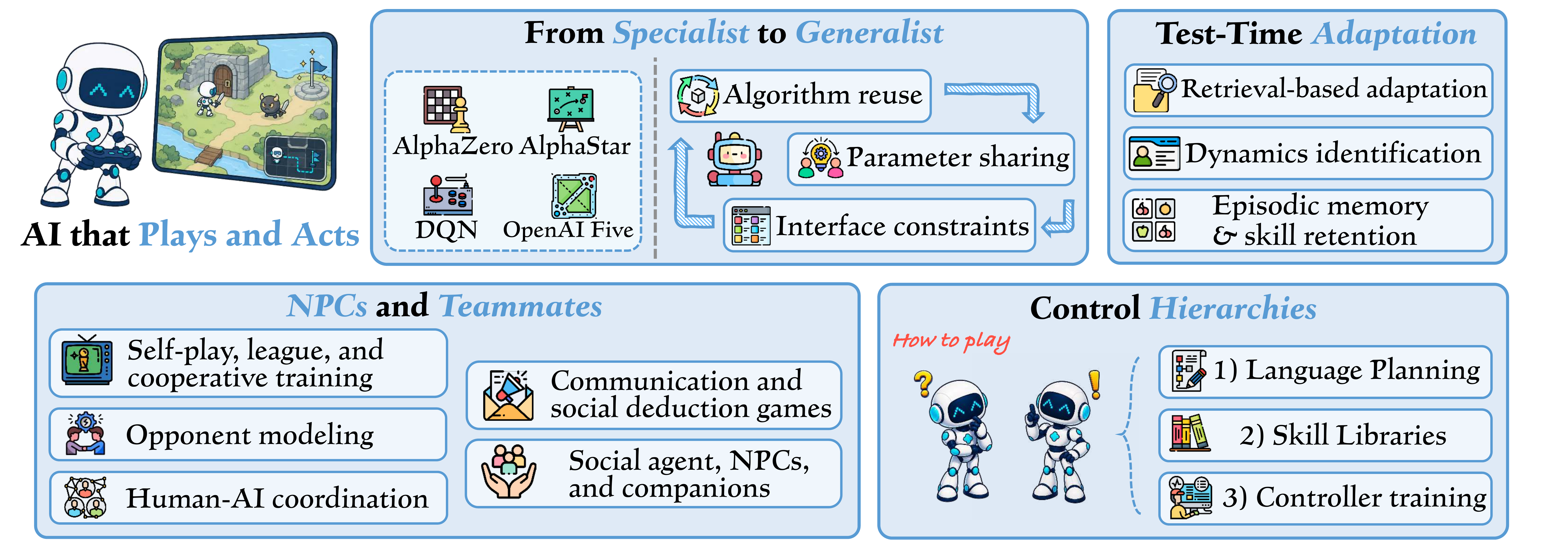}
  \caption{Research directions for AI that plays and acts: specialist-to-generalist policies, test-time adaptation, control hierarchies, and NPCs and teammates.}
  \label{fig:agent-control}
\end{figure}

A policy that masters one game has learned a particular combination of observations, controls, objectives, and interaction patterns. Foundation-model agents seek to reuse more of that competence: visual representations help interpret unfamiliar scenes, language supports planning from instructions, and learned or executable skills carry procedures into new tasks. A central design choice is how these resources reach the controls. A language planner using a semantic API receives different support \citep{wang2023voyager} from a policy acting through native keyboard and mouse input \citep{sima2024}. Comparing such agents requires following the division of work between perception, planning, memory, and action, including the adaptation needed when the game or its players change (\Cref{fig:agent-control}).

\draftmap{\begin{tikzpicture}[x=1cm,y=1cm]
\node[dm q] at (-1.65,2.35) {How the chapter moves: the scope widens from one policy, to one action, to many attempts, to many agents.};
\node[dm node=3.3cm] (a) at (0,0) {\dmhead{3.1 Specialist and generalist agents}\\[1pt]{\scriptsize algorithm, parameters, interface}};
\node[dm node=3.3cm] (b) at (4.3,0) {\dmhead{3.2 Control hierarchies}\\[1pt]{\scriptsize plan, skill, controller, timing}};
\node[dm node=3.3cm] (c) at (8.6,0) {\dmhead{3.3 Test-time adaptation}\\[1pt]{\scriptsize examples, mechanics, memory}};
\node[dm node=3.3cm] (d) at (12.9,0) {\dmhead{3.4 Multi-agent and social play}\\[1pt]{\scriptsize opponents, partners, people}};
\draw[dm arrow] (a) -- (b); \draw[dm arrow] (b) -- (c); \draw[dm arrow] (c) -- (d);
\node[dm bridge=2.7cm] at (2.15,0.8) {carried-in knowledge still has to be executed};
\node[dm bridge=2.7cm] at (6.45,0.8) {execution meets rules the agent does not know};
\node[dm bridge=2.7cm] at (10.75,0.8) {so far, one agent alone};
\node[dm frame,fit=(current bounding box)] (F) {};\dmtag
\end{tikzpicture}}

\subsection{Player and Generalist Agents}
\label{sec:play-generalist}

\draftmap{\begin{tikzpicture}[x=1cm,y=1cm]
\node[dm q] at (-1.65,2.35) {What makes an agent good at a game, and what of that carries into a game it was not trained on?};
\node[dm node=3.3cm] (a) at (0,0) {\dmhead{Specialist agents}\\[1pt]{\scriptsize AlphaZero, AlphaStar, OpenAI Five}};
\node[dm node=3.3cm] (b) at (4.3,0) {\dmhead{Algorithm reuse}\\[1pt]{\scriptsize GGP, GVGAI, DreamerV3, Procgen}};
\node[dm node=3.3cm] (c) at (8.6,0) {\dmhead{Parameter sharing}\\[1pt]{\scriptsize Gato, Multi-Game DT, SIMA, NitroGen}};
\node[dm node=3.3cm] (d) at (12.9,0) {\dmhead{Interface constraints}\\[1pt]{\scriptsize language goal, semantic API, native control; Cradle, Orak}};
\draw[dm arrow] (a) -- (b); \draw[dm arrow] (b) -- (c); \draw[dm arrow] (c) -- (d);
\node[dm bridge=2.7cm] at (2.15,0.8) {strength came from a fixed game; only the learning procedure transfers};
\node[dm bridge=2.7cm] at (6.45,0.8) {the procedure relearns action meanings; shared parameters carry them};
\node[dm bridge=2.7cm] at (10.75,0.8) {a shared policy reuses only what its interface exposes};
\draw[dm axis] (-1.65,-1.15) -- (14.55,-1.15) node[midway,below=1pt,font=\scriptsize,text=GameSlate] {from specialist competence to transferable competence};
\node[dm frame,fit=(current bounding box)] (F) {};\dmtag
\end{tikzpicture}}

Strong performance within one game and transfer to another are separate achievements. The literature reuses learning algorithms, trained parameters, and interaction interfaces in different combinations. Distinguishing them explains why a broadly applicable training procedure and a shared policy support different generalization claims.

\begin{figure}[!htb]
\centering
\setlength{\tilew}{0.32\linewidth}
\setlength{\tilelabelh}{10.5mm}
\blocktile{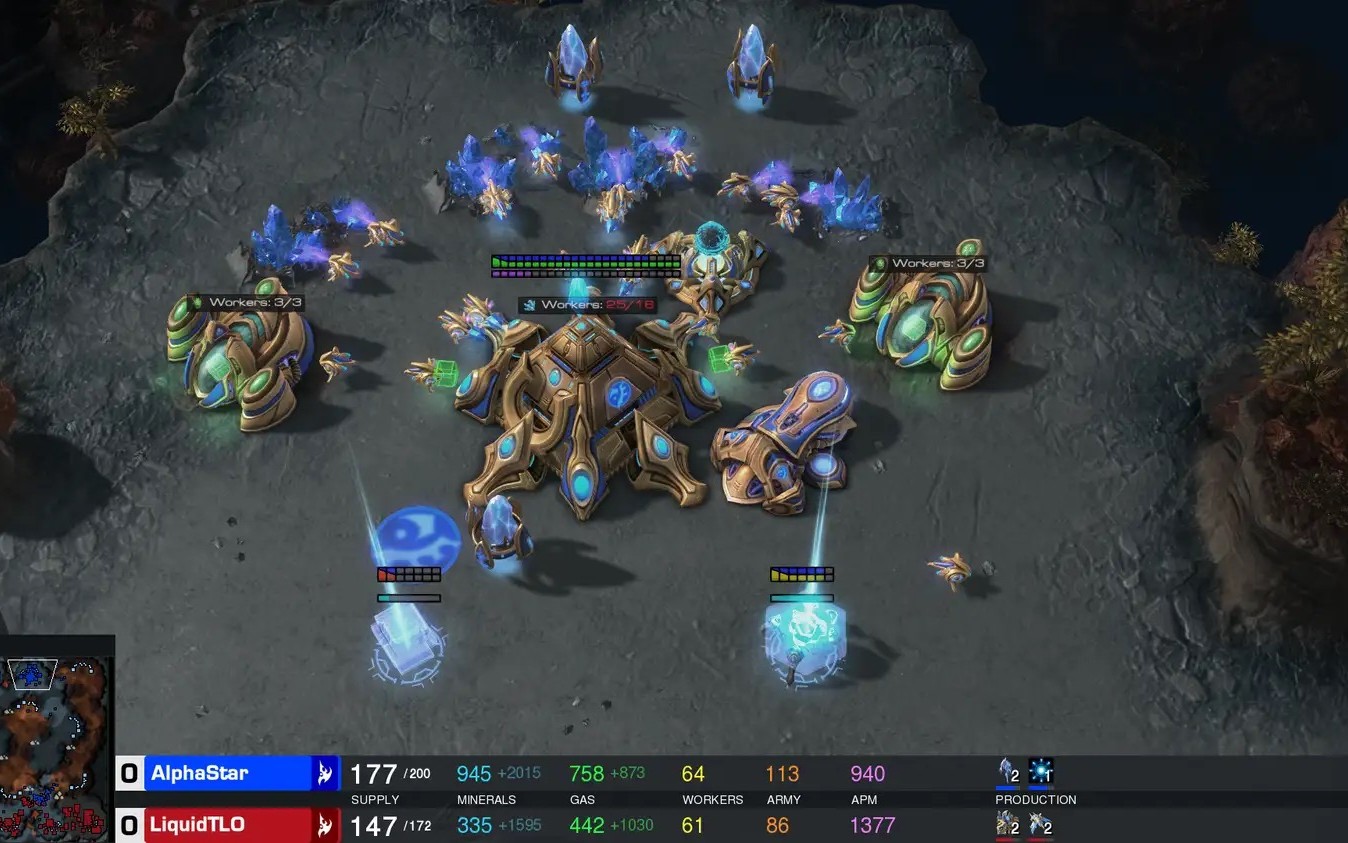}{(a) Specialist agents}{AlphaStar, StarCraft II}{Protoss army against a professional player}\hfill
\blocktile{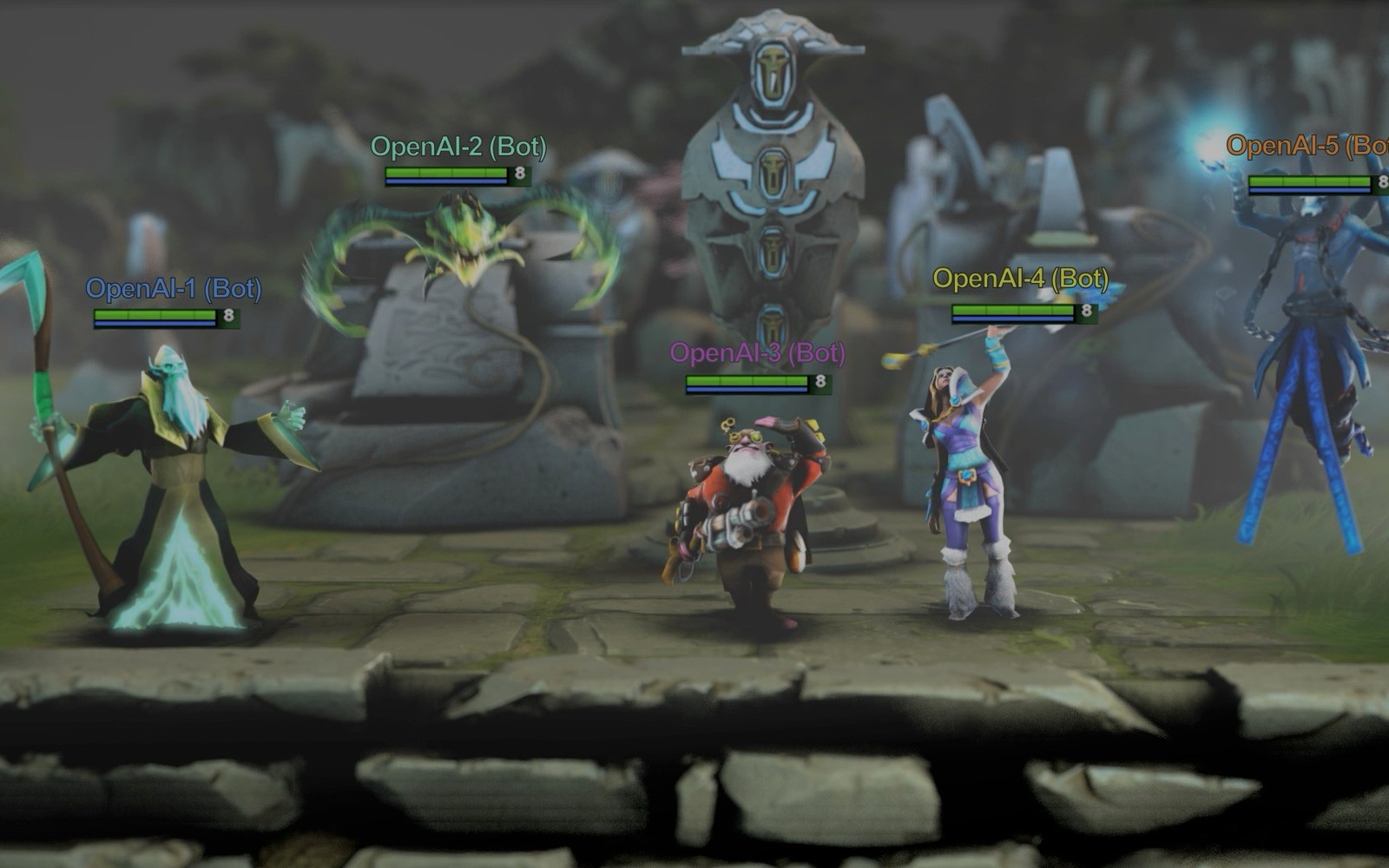}{(b) Specialist agents}{OpenAI Five, Dota 2}{All five heroes controlled by one policy}\hfill
\blocktile{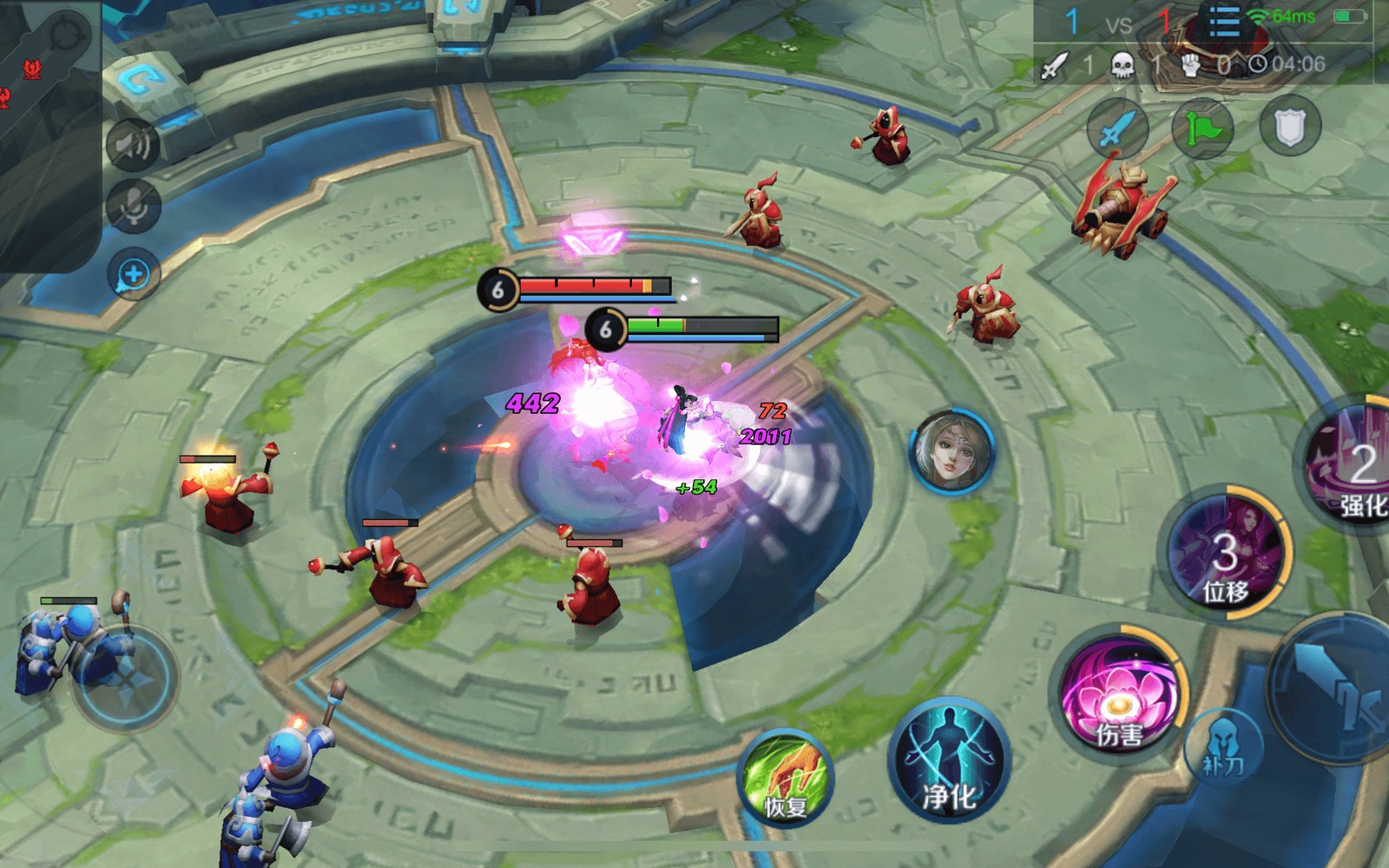}{(c) Specialist agents}{Honor of Kings AI}{Structured-state RL agent\\(illustrative game view)}\\[3pt]
\blocktile{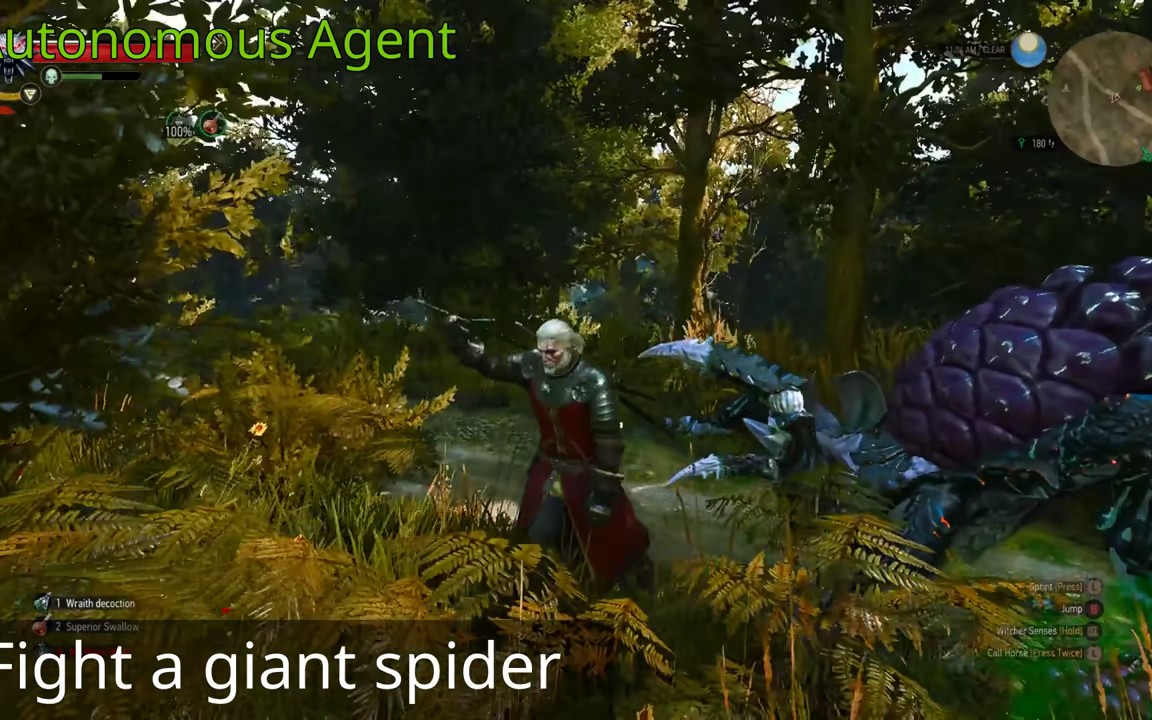}{(d) Parameter sharing}{NitroGen, The Witcher 3}{One shared policy fighting a giant spider}\hfill
\blocktile{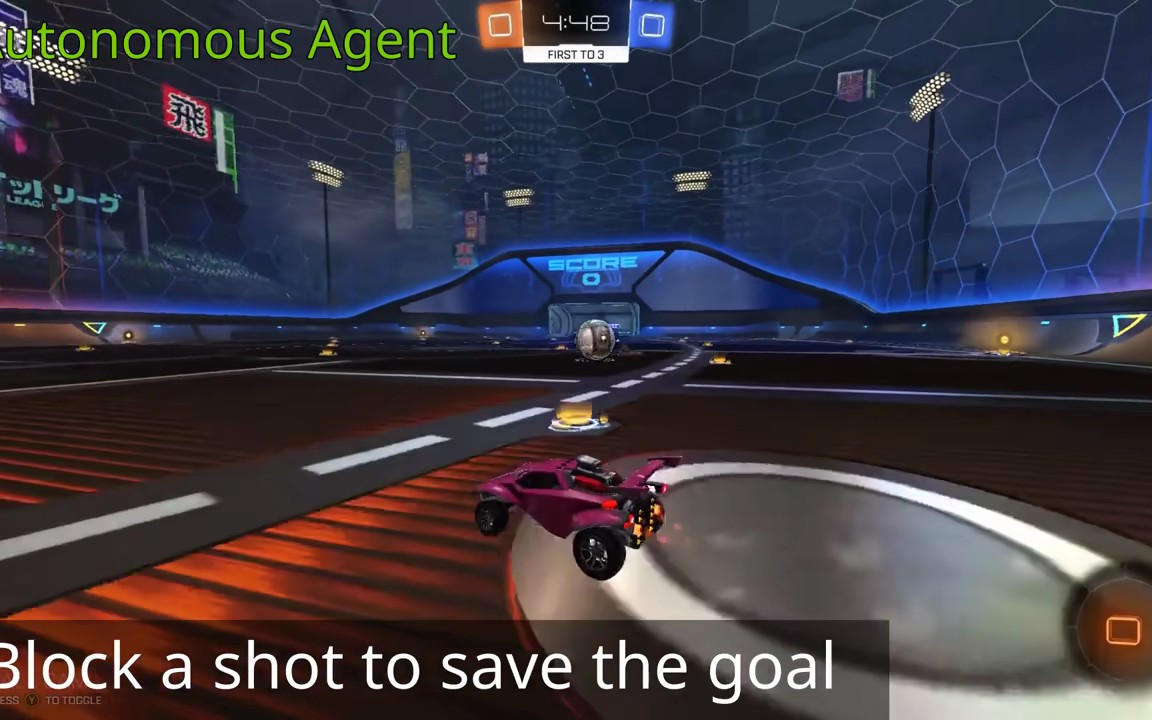}{(e) Parameter sharing}{NitroGen, Rocket League}{The same policy blocking a shot}\hfill
\blocktile{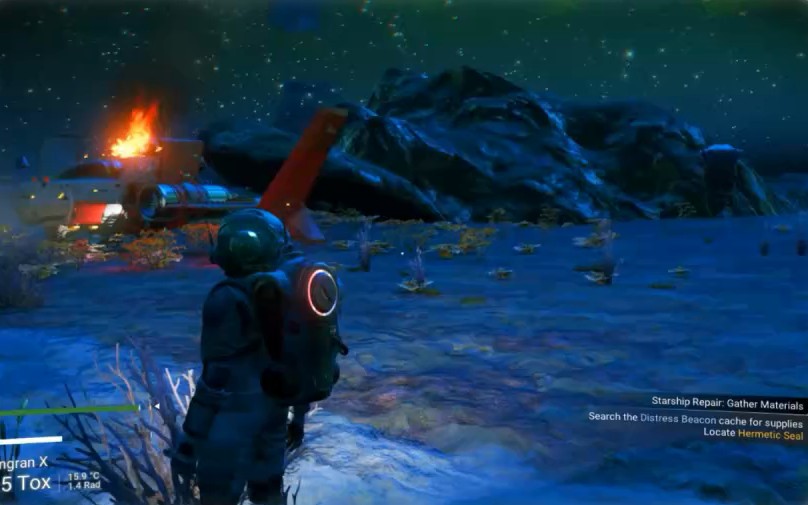}{(f) Parameter sharing}{SIMA 2, No Man's Sky}{One policy operating in a commercial title}
\caption{Specialist agents built for a single title (a--c) and policies whose parameters transfer across commercial games (d--f): (a) AlphaStar, StarCraft II \citep{vinyals2019alphastar}; (b) OpenAI Five, Dota 2 \citep{berner2019dota}; (c) Honor of Kings AI \citep{ye2020moba}; (d) NitroGen, The Witcher 3 \citep{magne2026nitrogen}; (e) NitroGen, Rocket League \citep{magne2026nitrogen}; (f) SIMA 2, No Man's Sky \citep{sima22025}.}
\label{fig:strip-3-1}
\end{figure}

\runin{Specialist agents}\figmark{\ref{fig:strip-3-1}a--c} Specialist results establish how effectively an agent can exploit a well-specified game and interaction protocol. Early chess and checkers programs used explicit rules and compact states \citep{shannon1950chess,samuel1959checkers}. Deep reinforcement learning connected pixels to actions in Atari \citep{mnih2015dqn}, while AlphaGo and AlphaZero combined learned policy and value functions with search and self-play \citep{silver2016alphago,silver2017alphagozero,silver2018alphazero}. AlphaStar \citep{vinyals2019alphastar} and OpenAI Five \citep{berner2019dota} extended large-scale training to real-time competition. Honor of Kings further addressed team-composition diversity through curriculum self-play and policy distillation \citep{ye2020moba}, while Gran Turismo Sophy combined continuous racing control with tactical interaction and racing etiquette \citep{wurman2022sophy}. These results broaden the kinds of expertise learned within a game. Their game-specific observations, rewards, and training protocols remain distinct from transfer of a trained policy to unfamiliar games.

\runin{Algorithm reuse} One candidate for reuse is the learning procedure. General Game Playing \citep{genesereth2005ggp} and GVGAI \citep{perezliebana2018gvgai} made the procedure the object of reuse: a solver keeps its reasoning machinery when the formal game changes. DreamerV3 provides a learning-era example, using one training configuration across more than 150 tasks while learning separate models and policies for those tasks \citep{hafner2025dreamerv3}. Procgen measures what algorithmic reuse leaves open by varying levels procedurally across 16 game-like environments and exposing the gap between memorizing a training distribution and generalizing to held-out levels \citep{cobbe2020procgen}. Different games can change action meanings and objectives as well as appearance, so neither algorithm reuse nor held-out-level performance alone establishes cross-game policy transfer.

\runin{Parameter sharing}\figmark{\ref{fig:strip-3-1}d--f} A shared policy retains trained weights across multiple tasks or games; transfer to an unseen game is an additional test. Decision Transformer casts offline control as sequence prediction, conditioning actions on past states, actions, and a desired return \citep{chen2021decisiontransformer}. Return conditioning selects behavior represented in the data but cannot supply missing exploration. Gato demonstrated one multimodal policy spanning Atari, robotics, and language \citep{reed2022gato}, while Multi-Game Decision Transformers trained on 41 Atari games and evaluated fine-tuning on five held-out games \citep{lee2022multigamedt}. MineDojo connected Minecraft control to large collections of video, language, and web knowledge \citep{fan2022minedojo}. SIMA \citep{sima2024}, SIMA~2 \citep{sima22025}, Game-TARS \citep{wang2025gametars}, and NitroGen \citep{magne2026nitrogen} extend shared-policy learning across diverse game collections. Their inputs also differ: SIMA agents follow instructions, whereas NitroGen learns short-context visual--motor behavior without language conditioning.

A controlled study of continued pretraining varies data volume, world diversity, and model size across 120 configurations: more data improves in-corpus performance, diversity improves transfer to related worlds with limited gains in distant ones, and an early interface phase with broad cross-world benefits gives way to a world phase whose gains are confined to in-corpus and related worlds \citep{zhang2026gamescaling}. Shared parameters thus appear to carry interface competence more broadly than world-specific competence. Gamer builds on this study. Continued pretraining on about 2,400 hours of gameplay from public game corpora and Honkai: Star Rail is followed by instruction tuning on about 1,000 hours of annotated play in that game and by online reinforcement learning with environment-verified rewards, all through one keyboard-and-mouse interface. On two side quests held out from training, verified milestone success rises from 34.4\% and 47.9\% after continued pretraining to 80.0\% and 83.6\% after reinforcement learning, which measures generalization to new tasks in a familiar game rather than transfer to an unseen one \citep{luo2026gamer}. A multi-teacher distillation study takes up that unseen case in browser games: a 4B student distilled from rule-based teachers on two or three source games succeeds in 26 and 24 of 30 runs on the first task in each target game, against at most 19 and 12 for the baselines. On the second task, whose threshold lies beyond the 100-step evaluation horizon, every method scores 0/30. Peak progress there is 58.0\% on GeoDash against 59.6\% for the best baseline, and 39.4\% on Doodle Jump against 24.3\% \citep{yi2026camopd}. \Cref{tab:agent-foundations} separates parameter sharing from the adaptation required in each evaluation setting.

\begin{table}[H]
\centering
\caption{Recurring training and evaluation settings for game-agent generality.}
\label{tab:agent-foundations}
\begingroup
\setcitestyle{numbers,square,comma}
\footnotesize
\setlength{\tabcolsep}{3.5pt}
\begin{tabularx}{\textwidth}{@{}L{3.15cm}L{3.10cm}L{3.55cm}Y@{}}
\toprule
\textbf{Setting} & \textbf{Reused} & \textbf{Game-specific work} & \textbf{Examples} \\
\midrule
Per-game training & algorithm; settings & agent training in each game, with known rules for AlphaZero and a learned world model for DreamerV3 & DQN; AlphaZero; DreamerV3~\citep{mnih2015dqn,silver2018alphazero,hafner2025dreamerv3} \\
Formal specification & solver; formal~interface & rules; forward~model & GGP; GVGAI~\citep{genesereth2005ggp,perezliebana2018gvgai} \\
Parameter sharing & policy weights & held-out testing; fine-tuning & Gato; Multi-Game DT; SIMA~2; NitroGen \citep{reed2022gato,lee2022multigamedt,sima22025,magne2026nitrogen} \\
Test-time adaptation & base model; retrieval & demonstrations; rules;~exploration & REGENT; Code World Models; Twin~\citep{sridhar2025regent,lehrach2025cwm,skoutnev2026twin} \\
\bottomrule
\end{tabularx}
\endgroup
\end{table}

The held-out unit also matters within these settings. A new map usually changes layout while retaining a game's controls and rules; a new mode can change rewards, opponents, or transition rules within the same title. Atari mode-transfer experiments already showed that a policy can fail under such relatively small changes, and that zero-shot evaluation and target-task fine-tuning must be distinguished \citep{farebrother2018generalization}. Holding out an entire game tests a broader change, but success may still concern shared navigation or object-use skills rather than unfamiliar rule reasoning. The cross-setting analysis in \Cref{sec:evaluation-generalization} distinguishes these settings and their supporting evidence.

\runin{Interface constraints} The interface places a practical boundary on each form of transfer. A shared visual encoder may transfer across changes in appearance more readily than a controller transfers from discrete buttons to camera-relative mouse movement. A language goal makes a task description portable across games while leaving the action grammar game-specific: a goal such as gathering a resource can be reused semantically, but the agent still needs to recognize the resource, discover its affordances, and execute the correct controls. A semantic action API reduces motor uncertainty by exposing inventory, legal actions, or navigation routines directly, and in doing so embeds affordances that a native-control agent must learn and execute for itself. Cradle standardizes observation and control around screenshots plus keyboard and mouse across games and applications \citep{tan2025cradle}, whereas Orak uses a structured MCP interface to support plug-and-play evaluation across 12 games \citep{park2025orak}. Cradle and Orak simplify comparison but support different control claims. Generalist performance therefore depends on which of the changing factors (visual appearance, objectives, controls, and timing) change together, on how much interaction is available for adaptation, and on which intermediate services the system receives. Comparisons are most informative when they state those services, since treating every successful task completion as equivalent hides the difference. System-level interfaces, adaptation requirements, and evaluation settings are compared in Appendix~\ref{app:resources}, \cref{tab:agent-systems}.

\subsection{Learning and Control Hierarchies}
\label{sec:play-hierarchy}

\draftmap{\begin{tikzpicture}[x=1cm,y=1cm]
\node[dm q] at (-1.65,2.35) {How does an intention become an action that arrives on time?};
\node[dm node=3.3cm] (a) at (0,0) {\dmhead{Language-based planning}\\[1pt]{\scriptsize SPRING, DEPS, JARVIS-1}};
\node[dm node=3.3cm] (b) at (4.3,0) {\dmhead{Reusable skills and policies}\\[1pt]{\scriptsize Plan4MC, Voyager, STEVE-1}};
\node[dm node=3.3cm] (c) at (8.6,0) {\dmhead{Controller training}\\[1pt]{\scriptsize VPT, JARVIS-VLA, CombatVLA, GTrXL}};
\node[dm node=3.3cm] (d) at (12.9,0) {\dmhead{Planner--controller interface and latency}\\[1pt]{\scriptsize ROCKET-1 (subgoal interface), SwarmBrain (timing)}};
\draw[dm arrow] (a) -- (b); \draw[dm arrow] (b) -- (c); \draw[dm arrow] (c) -- (d);
\node[dm bridge=2.7cm] at (2.15,0.8) {a plan needs a callable unit};
\node[dm bridge=2.7cm] at (6.45,0.8) {a skill needs an executor};
\node[dm bridge=2.7cm] at (10.75,0.8) {the layers exchange at the game's pace};
\draw[dm axis] (-1.65,-1.15) -- (14.55,-1.15) node[midway,below=1pt,font=\scriptsize,text=GameSlate] {top-down through the hierarchy};
\node[dm frame,fit=(current bounding box)] (F) {};\dmtag
\end{tikzpicture}}

Planning and motor control operate at different timescales. A model may know that a resource must be collected while lacking the policy that recognizes it, navigates to it, and completes the interaction. Hierarchical agents connect these decisions through language goals, executable skills, visual targets, or learned action representations.

\runin{Language-based planning} A language planner can reason over descriptions of rules and tasks even when motor realization is supplied elsewhere. SPRING reads game context from the Crafter paper \citep{hafner2022crafter} and selects actions by answering a fixed graph of reasoning questions over textual state descriptions \citep{wu2023spring}. GITM similarly exposes Minecraft knowledge and actions in text, using hierarchical decomposition and memory to handle technology-tree dependencies \citep{zhu2023gitm}. DEPS instead emphasizes revising plans from execution feedback \citep{wang2023deps}, while JARVIS-1 combines multimodal planning and memory with goal-conditioned controllers \citep{wang2023jarvis1}. Supplied knowledge can correct a mistaken plan, feedback can reveal an unmet prerequisite, and learned controllers supply motor execution. Success at the first two does not make that last service unnecessary. A useful comparison separates plan validity from plan grounding. Obtaining wood before crafting a tool can be a valid dependency without identifying the visible tree, reaching it, or recognizing completion. Text-state adapters can supply part of that grounding; visual agents must infer more of it from observations. Long-task success depends on plan decomposition and the services that execute and verify subgoals.

\begin{figure}[!htb]
\centering
\setlength{\tilew}{0.47\linewidth}
\setlength{\tilelabelh}{14.0mm}
\blocktile{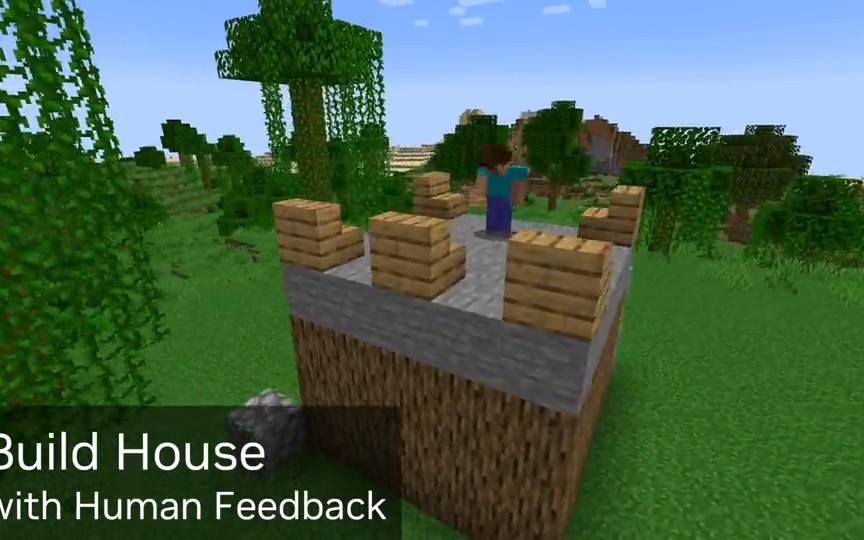}{(a) Reusable skills and policies}{Voyager, Minecraft}{A house built from a stored skill}\hfill
\blocktile{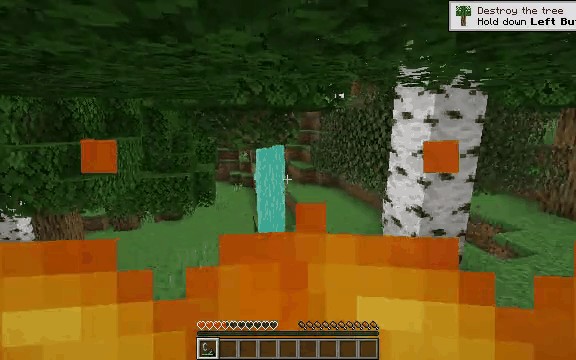}{(b) Planner--controller interface and latency}{ROCKET-1, Minecraft}{Planner-selected tree executed by the controller}
\caption{An intention passing down the control hierarchy, from a stored skill (a) to a planner-selected target executed by the controller (b): (a) Voyager, Minecraft \citep{wang2023voyager}; (b) ROCKET-1, Minecraft \citep{cai2024rocket1}.}
\label{fig:strip-3-2}
\end{figure}

\runin{Reusable skills and policies}\figmark{\ref{fig:strip-3-2}a} Skills package behavior that a planner can call repeatedly. Plan4MC combines a skill structure with learned policies \citep{yuan2023plan4mc}; Voyager retrieves and composes executable code \citep{wang2023voyager}; at larger scale, four Gemini~4 Argon agents controlling players through the Mineflayer library wrote 2,783 programs and defeated the Ender Dragon from a random world on Easy difficulty in about 27 hours, a single run that its authors describe as the first end-to-end autonomous completion of Minecraft, obtained through a semantic interface rather than native keyboard and mouse control \citep{faldor2026geminiminecraft}; STEVE-1 conditions a pretrained behavioral model on language-linked goals \citep{lifshitz2023steve1}. These are different representations of reusable behavior. A controlled NetHack study quantifies the trade-off for a supplied code-skill library: compared with primitive actions, skills nearly triple zero-shot game progression while reducing inference cost per episode by 86\%, combining skills with primitives retains much of this benefit while keeping a path back to low-level actions, and skill-based agents learn faster under reinforcement learning \citep{cupial2026codehack}. Code can expose API calls and checks for prerequisites, whereas a learned policy can accommodate variation in appearance and movement without enumerating every case. MineDreamer uses a visual intermediate: an image-editing diffusion model imagines a goal consistent with the instruction and current scene, and a goal-conditioned controller converts that target into keyboard and mouse actions \citep{zhou2025minedreamer}. The generated image helps specify what the controller should achieve, rather than predicting a validated sequence of transitions. Repair therefore depends on the representation: an erroneous code skill can be edited, a poor imagined target can be regenerated, and an unreliable motor policy may need additional training. A successful target image is not itself evidence of an accurate dynamics model.

\runin{Controller training} Reinforcement learning and self-play remain effective when rewards, legal actions, and repeated environment access are available. They optimize measured game outcomes directly and have produced precise specialist behavior, but reward design and fresh interaction are expensive across broad game collections. Behavioral cloning offers a complementary route: human trajectories can cover purposeful behavior, including regions of the state space that sparse-reward exploration reaches only slowly. Video PreTraining (VPT) learned an inverse-dynamics model from a smaller action-labeled Minecraft dataset and used it to recover controls for a much larger corpus of online video \citep{baker2022vpt}. JARVIS-VLA starts from the other side: it first adapts a vision--language model to Minecraft observations, then learns keyboard and mouse actions through imitation \citep{jarvisvla2025}. CombatVLA emphasizes efficient action generation for combat \citep{chen2025combatvla}. VPT and JARVIS-VLA differ in whether motor behavior or visual--language knowledge supplies the pretrained foundation.

Generated environments are a third source of experience. Player-1 follows supervised training on 6,044 gameplay trajectories with reinforcement learning in sixteen synthetic environments, and its 9B model reaches 38.33\% success on MineDojo-300 against 12.63\% for the strongest evaluated baseline \citep{ouyang2026player1}. In Game2Player, a builder agent adapts web games into 568 verified environments with 3,436 tasks while coding and vision players attempt them, and training on the resulting 307 hours of play raises a 9B model's progress on the independent GameWorld benchmark from 30.0\% to 53.1\% \citep{jiang2026game2player}.

OmniJARVIS learns an intermediate representation rather than using either free-form plans or raw controls as the sole bridge. A behavior encoder discretizes trajectories into tokens added to a multimodal language model's vocabulary, and an imitation-learning decoder executes the selected behaviors \citep{wang2024omnijarvis}. Instructions, observations, reasoning, and behavior tokens can then participate in a common sequence-prediction objective. Its Minecraft evaluation supports this connection between high-level instruction following and learned execution; the decoder and training trajectories still provide game-specific motor knowledge. Executable skills offer a useful contrast: code supplies explicit procedures over an API, while behavior tokens refer to procedures learned from interaction data.

Imitation and reinforcement learning also fail differently. Demonstrations inherit uneven skill and incomplete action labels, whereas environment optimization can specialize behavior to the reward and dynamics used in training. Temporal representation matters in both cases. Gated Transformer-XL stabilizes attention-based reinforcement learning and evaluates it on memory-demanding environments including DMLab-30 \citep{parisotto2020gtrxl}. A longer history helps only if relevant information is retrieved and acted on before it becomes stale.

\runin{Planner--controller interface and latency}\figmark{\ref{fig:strip-3-2}b} The layers meet at two points: how an intention is communicated downward, and how fast the exchange must run. ROCKET-1 addresses the first. For spatial targets that language cannot express precisely, its high-level model communicates tracked visual regions to the low-level policy \citep{cai2024rocket1}. Real-time games govern the second. A deliberative model may update goals every few seconds while a reactive controller manages movement, camera control, and combat. SwarmBrain makes this separation explicit in StarCraft~II \citep{shao2024swarmbrain}, while recent vision--language--action systems address the same latency pressure through faster action generation and execution \citep{chen2025combatvla}. TaskFrame pairs a 9B vision--language planner with a causal Transformer controller that emits keyboard, mouse, and camera actions at every game step. In Minecraft it reaches 51\% average kill success against 31\% for JARVIS-VLA-7B, while mining is close to saturated for both \citep{wang2026taskframe}. HiSTEVE has a GPT-4o planner select atomic subtasks for one native-interface policy and uses a separate detector to flag candidate subtask completion for the harness to verify, and reaches a MineStudio four-family average of 85.0\% against 65.0\% for the OpenHA Text/CoA baseline \citep{tang2026histeve}. In the deployed PUBG Ally teammate, a language-model agent inspects game information through a controlled interface and issues high-level choices that steer a faster control layer for movement, combat, and recovery under strict latency constraints \citep{pubg2026ally}.

Gamer measures the cost of delay on four public game benchmarks: when the game keeps running during inference instead of pausing, the average score of GPT-5.6 Sol falls by 34.5 points to 17.6, whereas that of Gamer's 8B continued-pretraining policy falls by 4.4 points to 29.5 \citep{luo2026gamer}. RealtimeBench makes the same comparison on 60 held-out test tasks in six game types with structured text state: GPT-6.1 Sol, which takes about 19 seconds per response, beats a faster decision model in five of the six game types when games are paused, but averages 1.51 against 13.51 on a 0--100 progress scale when they keep running \citep{pan2026realtimebench}. Decision quality, control rate, and action-to-effect delay jointly determine behavior: a strategically sound intention can still arrive after the state in which it was useful. Asynchronous planning, cached beliefs, interruptible skills, and safe defaults therefore shape system capability.

\subsection{Test-Time Adaptation and Memory}
\label{sec:play-adaptation}

\draftmap{\begin{tikzpicture}[x=1cm,y=1cm]
\node[dm q] at (-2.1,2.35) {What does the agent rebuild from an unfamiliar game while playing, and keep for the next attempt?};
\node[dm node=4.0cm] (a) at (0,0) {\dmhead{Retrieval-based adaptation}\\[1pt]{\scriptsize REGENT, S3Gym}};
\node[dm node=4.0cm] (b) at (5.5,0) {\dmhead{Dynamics identification}\\[1pt]{\scriptsize Twin, Code World Models}};
\node[dm node=4.0cm] (c) at (11.0,0) {\dmhead{Episodic memory and skill retention}\\[1pt]{\scriptsize EMemBench, GameVerse, Voyager}};
\draw[dm arrow] (a) -- (b); \draw[dm arrow] (b) -- (c);
\node[dm bridge=3.2cm] at (2.75,0.8) {precedent for action, not the mechanics behind it};
\node[dm bridge=3.2cm] at (8.25,0.8) {what is reconstructed must be retained across attempts};
\draw[dm axis] (-2.1,-1.15) -- (13.1,-1.15) node[midway,below=1pt,font=\scriptsize,text=GameSlate] {reconstruction grows more explicit};
\node[dm frame,fit=(current bounding box)] (F) {};\dmtag
\end{tikzpicture}}

Test-time adaptation changes what an agent can learn from a new game before or during evaluation. Demonstration retrieval supplies examples of successful behavior; interaction can reveal unknown mechanics; memory preserves experience for later attempts. Their costs and benefits depend on what information is supplied, what the agent must acquire, and which decisions the retained information improves.

\begin{figure}[!htb]
\centering
\setlength{\tilew}{0.47\linewidth}
\setlength{\tilelabelh}{14.0mm}
\blocktile{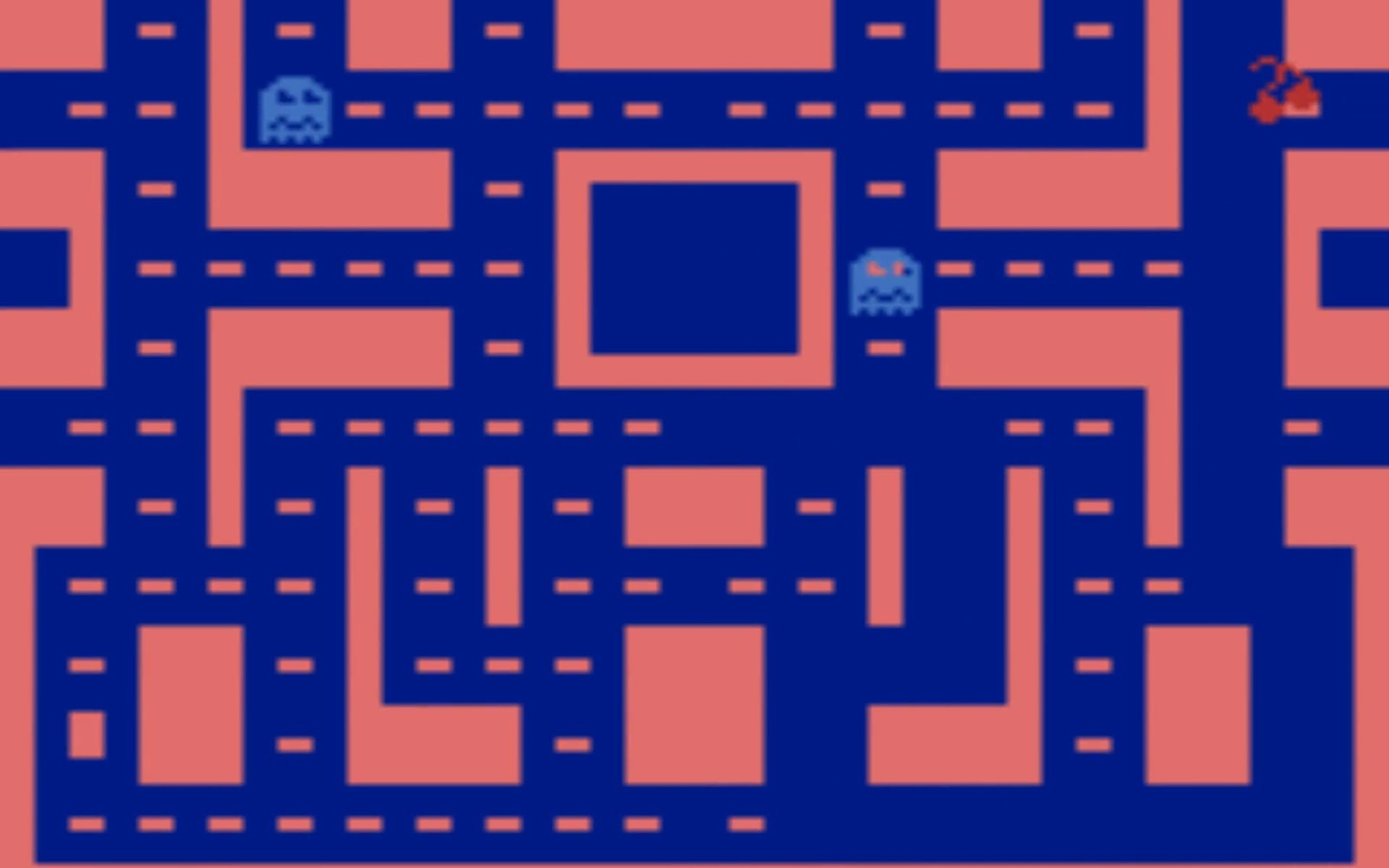}{(a) Retrieval-based adaptation}{REGENT, Ms. Pac-Man}{Unseen game played from retrieved demonstrations}\hfill
\blocktile{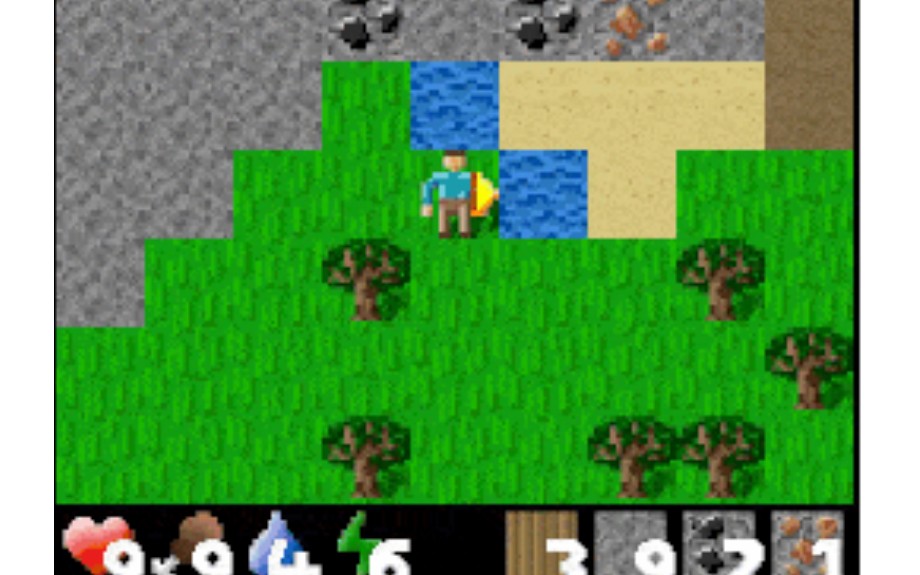}{(b) Episodic memory and skill retention}{EMemBench, Crafter}{Collected items and vitals shown in the HUD}
\caption{Test-time adaptation by retrieving demonstrations for an unseen game (a) and by carrying episodic state through an episode (b): (a) REGENT, Ms. Pac-Man \citep{sridhar2025regent}; (b) EMemBench, Crafter \citep{li2026emembench}.}
\label{fig:strip-3-3}
\end{figure}

\runin{Retrieval-based adaptation}\figmark{\ref{fig:strip-3-3}a} REGENT retrieves state--action examples from demonstrations of a held-out game and combines them with a pretrained policy, allowing in-context adaptation without fine-tuning. The demonstrations and known observation/action spaces remain part of the setup \citep{sridhar2025regent}. Retrieved examples provide local evidence about what to do without changing all of the policy's parameters. Their usefulness depends on whether similarity in the retrieval space corresponds to the same action problem: visually similar states may require different actions after a rule, objective, or inventory change. S3Gym examines reuse across attempts within a game. It separates permissive exploration from evaluation on held-out seeds or configurations in seven games and compares raw interaction histories, summarized experience, and supervised fine-tuning \citep{shi2026s3gym}. Its reported gains depend on the task: summaries help when experience compresses into reusable strategies, whereas precise state-dependent tasks can benefit from retaining more of the original trajectory. Its text interfaces and supplied game descriptions make this an evaluation of experience reuse, with native visual--motor transfer outside its scope.

\runin{Dynamics identification} Twin builds an executable hypothesis of unknown dynamics, checks it against replayed transitions, and searches within the resulting digital twin \citep{skoutnev2026twin}. Here interaction has two purposes: obtaining reward and distinguishing competing explanations of the mechanics. A useful exploratory action may expose a collision rule or terminal condition even when it does not immediately improve the score. Code World Models also synthesize executable transition, legality, and termination functions, but the evaluated protocol supplies natural-language rules and initial offline trajectories. The synthesized world model is refined before competitive play and held fixed during it, and inference functions handle hidden information in the partially observed games \citep{lehrach2025cwm}. The three systems therefore differ in what adaptation reconstructs: REGENT supplies relevant demonstrations to a shared policy, Twin identifies unknown mechanics through interaction, and Code World Models translate stated rules into a testable simulator. Active identification matters particularly in the unknown-rule setting, where an action can reveal a mechanic as well as advance the task.

\runin{Episodic memory and skill retention}\figmark{\ref{fig:strip-3-3}b} Memory supports this process at several levels. Recent observations maintain continuity of control, episodic records recover earlier attempts, summaries compress long trajectories, and skill libraries preserve procedures that have succeeded. EMemBench evaluates trajectory-grounded episodic memory through programmatically generated questions with verifiable answers \citep{li2026emembench}, while GameVerse tests whether reflection on gameplay video improves subsequent play \citep{zhang2026gameverse}. Episodic records and summaries help an agent act under partial observation, but stale summaries, conflicting episodes, and retrieval that misses a reset or rule change make them unreliable. Executable hypotheses and engine state provide stronger grounding when exact legality and termination matter.

Optimus-1 makes a concrete distinction between task knowledge and remembered experience. It retrieves crafting dependencies from a directed knowledge graph for planning, while a pool of compressed multimodal episodes supports a reflector that decides whether to continue, complete, or replan a subgoal. Successful and failed cases are retained, and a STEVE-1 controller executes the resulting goals \citep{li2024optimus}. The reported Minecraft experiments and memory ablations connect these components to task completion. Reuse here concerns explicit knowledge and experiences with a particular controller, rather than evidence that a larger context window alone produces general long-horizon competence.

AgenticSTS makes retrieval granularity explicit in \emph{Slay the Spire 2}: each decision receives a bounded selection of game knowledge, episodic summaries, and triggered strategic skills instead of an accumulating transcript. Its fixed-difficulty ablations compare memory components within one game; the small, ten-run conditions support exploratory comparisons rather than a general advantage for bounded memory \citep{cheng2026agenticsts}. The memory unit affects correction as well as capacity: an episode retains context, a summary may omit an exception, and a skill may depend on outdated prerequisites. FlashAdventure tests a different demand across 34 adventure games, where clues discovered in one scene must inform later object use and navigation \citep{ahn2025flashadventure}. Progression requires applying remembered facts, not merely recalling them.

\subsection{Opponents, Teammates, NPCs, and Companions}
\label{sec:play-others}

\draftmap{\begin{tikzpicture}[x=1cm,y=1cm]
\node[dm q] at (-1.5,2.35) {What must the agent learn about the others in the game?};
\node[dm node=2.8cm] (a) at (0,0) {\dmhead{Self-play, league, and cooperative training}[1pt]{\scriptsize PSRO, XLand, QMIX, MAPPO}};
\node[dm node=2.8cm] (b) at (3.6,0) {\dmhead{Opponent modeling and theory of mind}[1pt]{\scriptsize ToMnet, LOLA, DeepRole}};
\node[dm node=2.8cm] (c) at (7.2,0) {\dmhead{Zero-shot coordination and human--AI teamwork}[1pt]{\scriptsize Other-Play, FCP, MEP, HSP}};
\node[dm node=2.8cm] (d) at (10.8,0) {\dmhead{Communication and social deduction games}[1pt]{\scriptsize CICERO, Werewolf, Avalon}};
\node[dm node=2.8cm] (e) at (14.4,0) {\dmhead{Social agents, NPCs, and companions}[1pt]{\scriptsize Generative Agents, Project Sid}};
\draw[dm arrow] (a) -- (b); \draw[dm arrow] (b) -- (c); \draw[dm arrow] (c) -- (d); \draw[dm arrow] (d) -- (e);
\node[dm bridge=2.4cm] at (1.8,0.8) {a trained population is not the population met at test time};
\node[dm bridge=2.4cm] at (5.4,0.8) {a model of the other is what adaptation acts on};
\node[dm bridge=2.4cm] at (9.0,0.8) {with people, talk becomes part of the game};
\node[dm bridge=2.4cm] at (12.6,0.8) {a human partner also judges the character};
\draw[dm axis] (-1.5,-1.15) -- (15.9,-1.15) node[midway,below=1pt,font=\scriptsize,text=GameSlate] {the other party, from the most controlled population to the least};
\node[dm frame,fit=(current bounding box)] (F) {};\dmtag
\end{tikzpicture}}

Other agents make the interaction distribution depend on goals, conventions, and beliefs beyond the learner's own policy. Self-play, unfamiliar-team coordination, strategic communication, and character interaction therefore pose different learning problems \citep{albrecht2017autonomous,mirsky2022survey,feng2024socialagents}. Language models broaden how intentions and histories can be represented, while the task still determines whether success means winning, coordinating, or sustaining a believable interaction.

\begin{figure}[!htb]
\centering
\setlength{\tilew}{0.47\linewidth}
\setlength{\tilelabelh}{14.0mm}
\blocktile{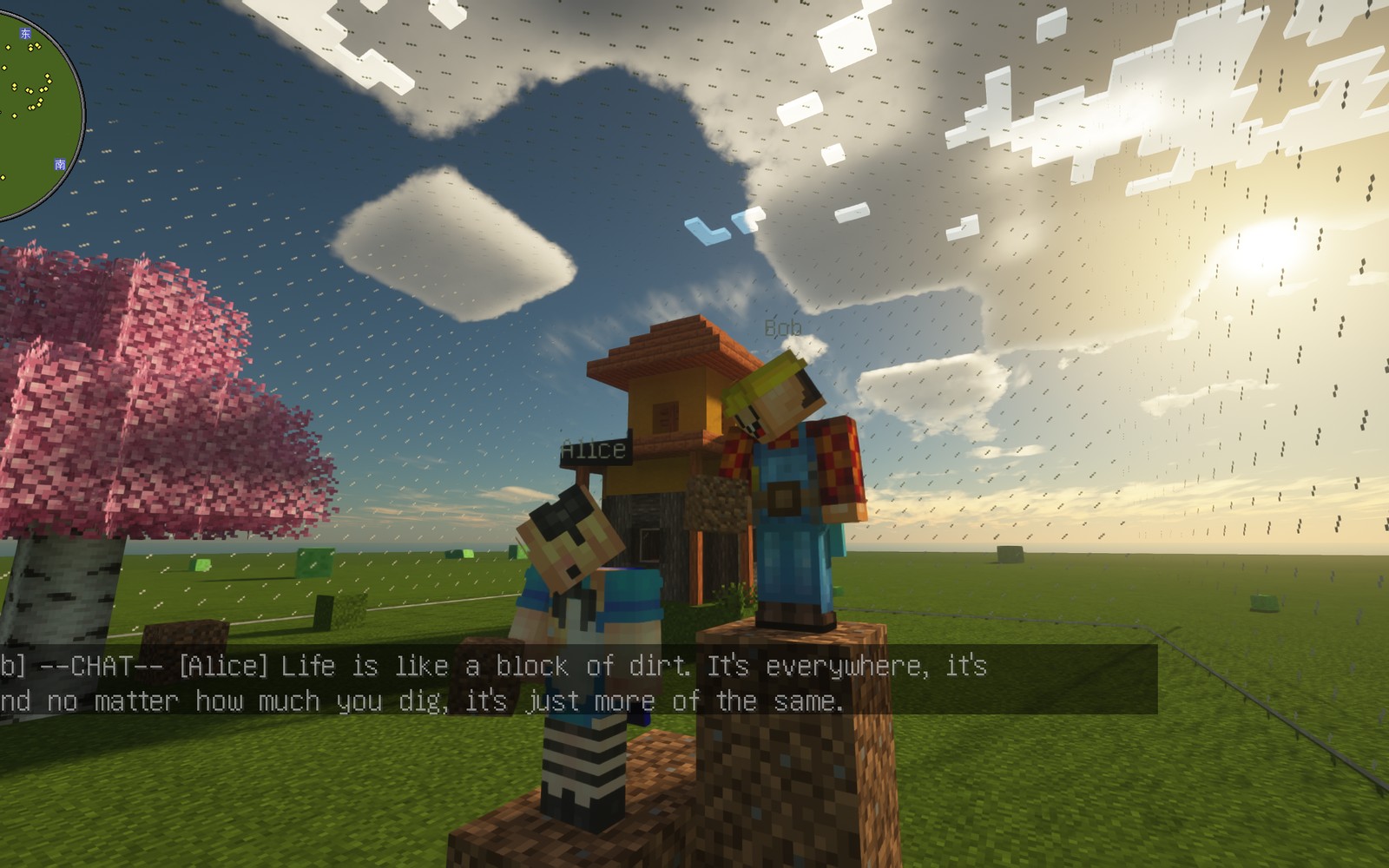}{(a) Task-dependency-based coordination}{VillagerAgent, Minecraft}{Agents coordinating dependency-linked tasks}\hfill
\blocktile{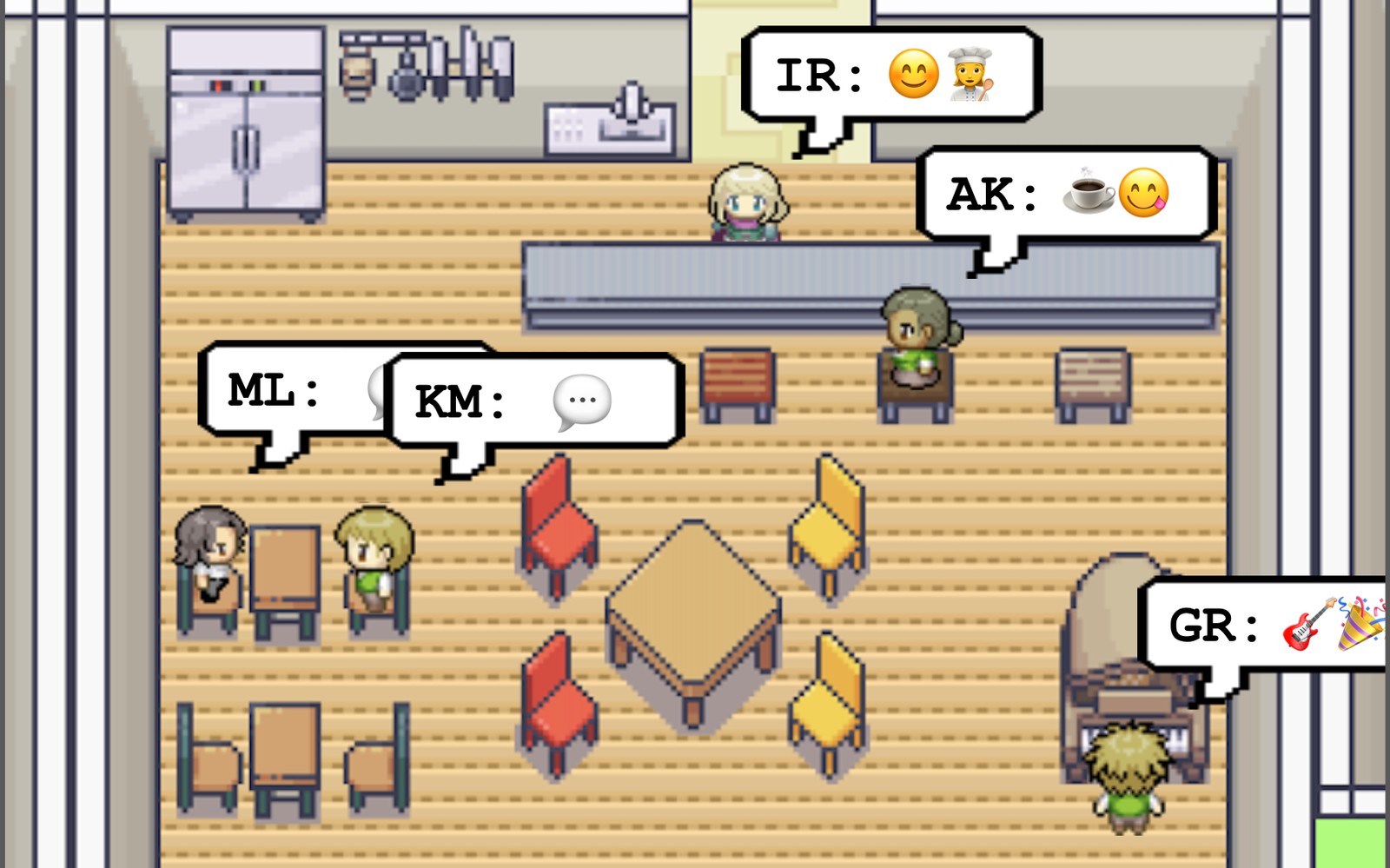}{(b) Social agents, NPCs, and companions}{Generative Agents, Smallville}{An unscripted party emerging from one intention}\\[3pt]
\blocktile{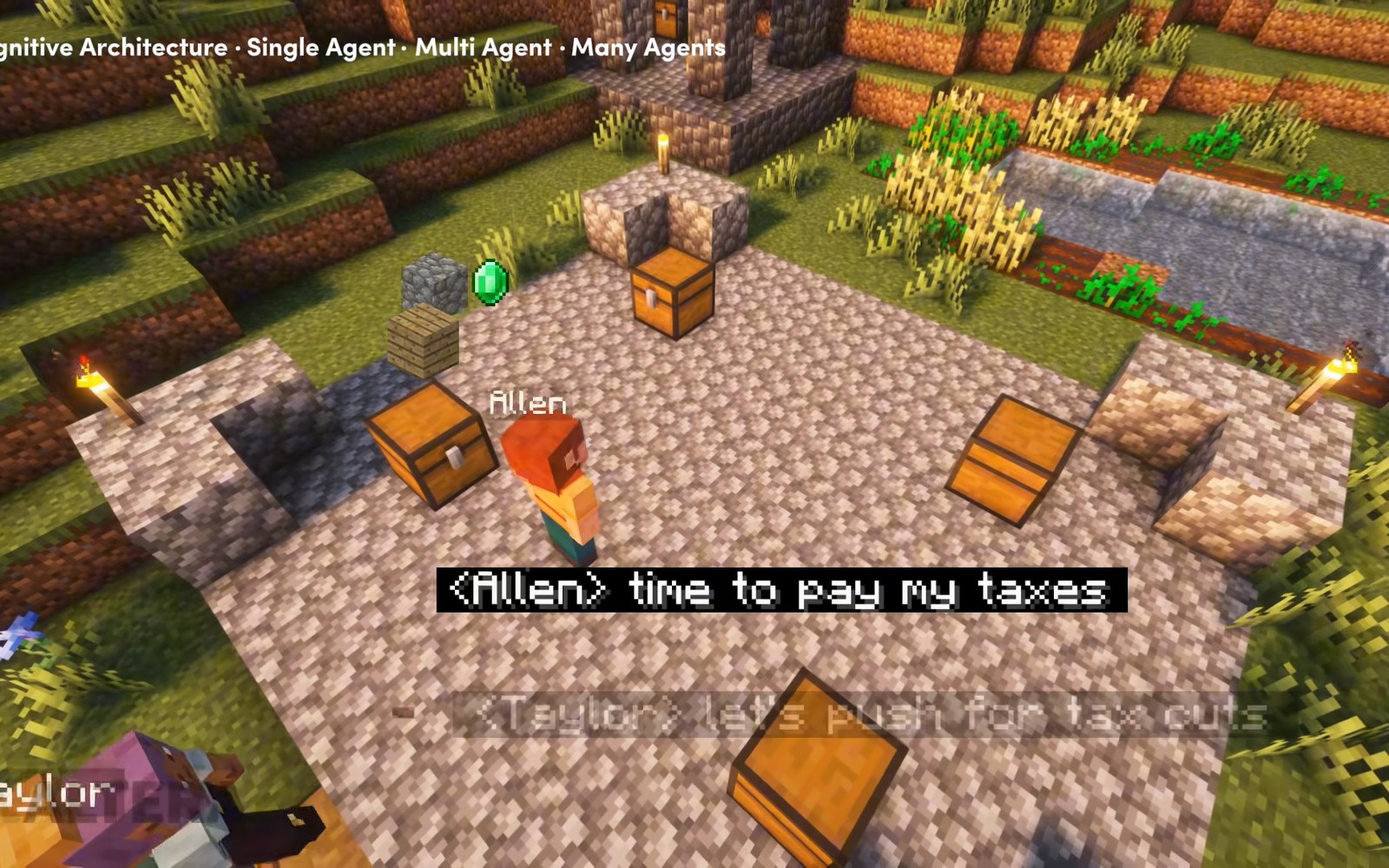}{(c) Social agents, NPCs, and companions}{Project Sid (PIANO), Minecraft}{Agents negotiating taxation in the village chat}\hfill
\blocktile{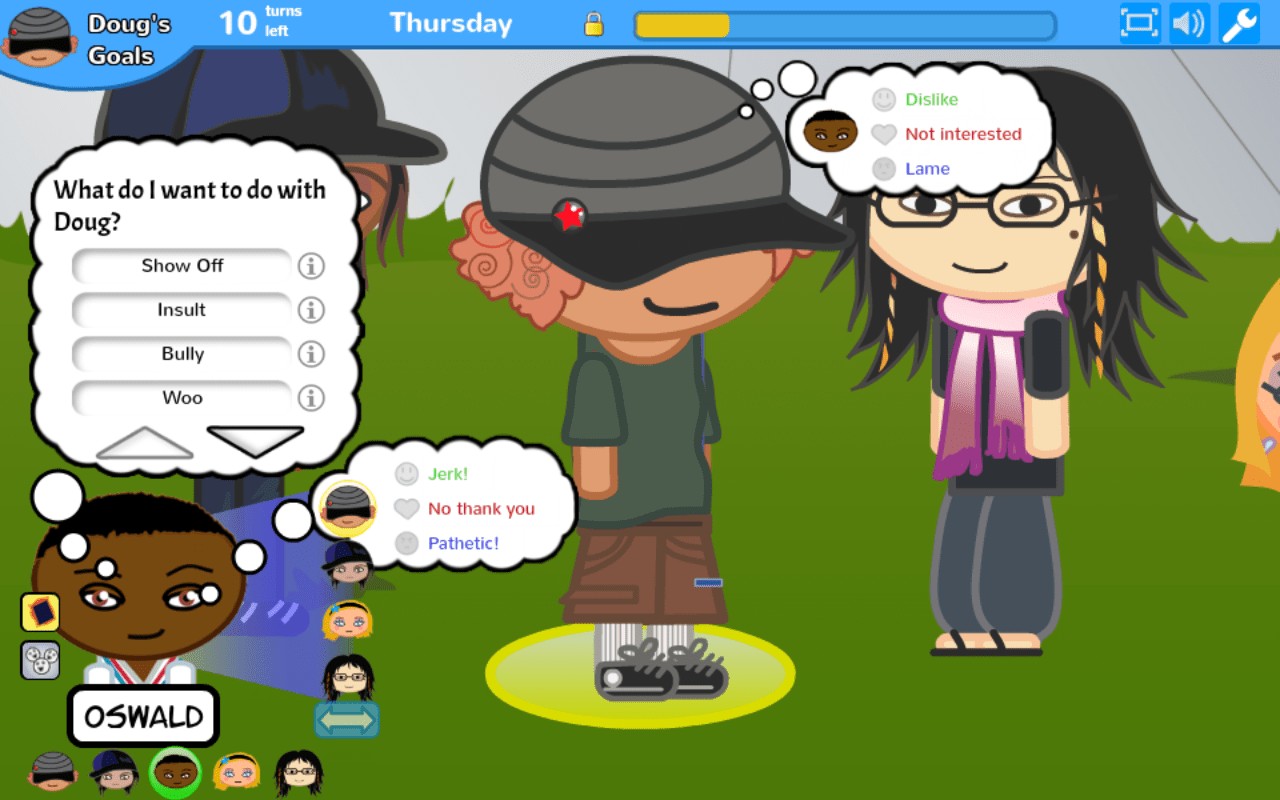}{(d) Social agents, NPCs, and companions}{Comme il Faut, Prom Week}{Social moves presented as a game menu}
\caption{Task-dependency-based multi-agent coordination (a) and social agents whose behaviour emerges from interaction (b--d): (a) VillagerAgent, Minecraft \citep{dong2024villageragent}; (b) Generative Agents, Smallville \citep{park2023generativeagents}; (c) Project Sid (PIANO), Minecraft \citep{al2024project}; (d) Comme il Faut, Prom Week \citep{mccoy2013promweek}.}
\label{fig:strip-3-4}
\end{figure}

\runin{Self-play, league, and cooperative training} Training populations determine which opponents an agent learns to handle. AlphaStar \citep{vinyals2019alphastar} and OpenAI Five \citep{berner2019dota} scale competitive training within particular games. Policy-space response oracles address overfitting to training opponents by constructing responses to mixtures of policies \citep{lanctot2017unified}. Population-based training in Quake III Capture the Flag \citep{jaderberg2019human} and the procedurally generated tasks of XLand \citep{oel2021xland} further vary the partners, opponents, and situations encountered during learning. ReBeL \citep{brown2020combining} and Pluribus \citep{brown2019superhuman} combine learning with game-theoretic reasoning in imperfect-information poker. These results establish strong ways to generate training experience; they do not require language interfaces or broad multimodal pretraining.

Cooperative learning additionally faces credit assignment under shared rewards and partial observation. Value-decomposition networks sum individual value functions \citep{sunehag2017value}, QMIX learns a monotonic mixing function compatible with decentralized action selection \citep{rashid2018qmix}, and MAPPO provides a policy-gradient alternative \citep{yu2021surprising}. Benchmarks such as SMAC \citep{samvelyan2019starcraft} and Hanabi \citep{bard2020hanabi} make these decisions reproducible within specified tasks. Hanabi also exposes convention dependence: an action can convey private information, and partners may interpret it differently. Melting Pot broadens evaluation across social scenarios and partner populations \citep{leibo2021meltingpot}. Population diversity and inference about a particular partner are complementary responses to this problem.

\runin{Opponent modeling and theory of mind} Models of other agents can predict actions, goals, or beliefs \citep{albrecht2017autonomous}. ToMnet learns a prior over agents and predicts a new agent's behavior from observations \citep{rabinowitz2018machine}; LOLA accounts for how one's own update affects another learner \citep{foerster2017learning}; DeepRole combines search, learned values, and deductive beliefs in The Resistance: Avalon \citep{serrino2019finding}. These approaches differ in what is inferred: a stable behavioral tendency, a learning response, or hidden information relevant to the current game. Language-based accounts of intentions offer another representation, but must still be checked against subsequent actions rather than judged by plausibility alone.

\runin{Zero-shot coordination and human--AI teamwork} Ad hoc teamwork is the problem of collaborating with new teammates without prior coordination, and zero-shot coordination is its no-adaptation special case \citep{mirsky2022survey}. Overcooked distinguishes coordinated task execution from compatibility with people \citep{carroll2019overcooked}. Two established approaches address partner variation without assuming a single shared convention. Other-Play trains with known game symmetries to discourage arbitrary coordination conventions \citep{hu2020otherplay}, and Fictitious Co-Play trains a best response to a population that includes agents at different stages of learning \citep{strouse2021fcp}. Later population methods vary the population deliberately: maximum-entropy population-based training adds a population entropy bonus and trains the final agent against a diversified pool with prioritized sampling, without human data \citep{zhao2021maximum}. Hidden-utility self-play addresses a different assumption, that every partner optimizes the environment reward, by modeling human biases as hidden reward functions inside the self-play objective \citep{yu2023learning}. ProAgent adds online interpretation, inferring and revising a partner's likely intentions from game state in Overcooked and comparing against trained-agent populations and human-proxy models \citep{zhang2024proagent}. Human-proxy performance, independent-agent coordination, and interaction with actual players provide distinct tests of the resulting ability.

MindAgent and VillagerAgent address coordination through task allocation rather than partner inference alone. MindAgent schedules work among agents in CuisineWorld and supports human--NPC collaboration \citep{gong2024mindagent}; VillagerAgent\figmark{\ref{fig:strip-3-4}a} uses task-dependency graphs to allocate and update Minecraft subtasks \citep{dong2024villageragent}. Compared with ProAgent's online interpretation of a partner, these systems make more of the joint plan explicit. Their gains concern coordination over available skills and interfaces, while the game or predefined APIs carry out low-level actions.

\runin{Communication and social deduction games} Language can be part of the action space, carrying promises, private information, or deception. Deal or No Deal learns negotiation from human dialogues and uses simulated dialogue continuations \citep{lewis2017deal}; CICERO couples dialogue to strategic plans in Diplomacy \citep{fair2022cicero}. In Werewolf, retrieval and reflection reuse prior communications without parameter updates \citep{xu2023exploring}, while a separate approach lets a reinforcement-learning policy select among language-model action proposals \citep{xu2023language}. Recursive Contemplation uses perspective-taking to reason about hidden roles in Avalon \citep{wang2023avalon}. These methods place different constraints on language: generating a credible utterance, selecting a useful action, and inferring concealed information are not the same objective. AvalonBench supplies role-specific agents and baseline opponents \citep{light2023avalonbench}; Werewolf Arena compares models in tournaments with bidding-based turn-taking \citep{bailis2024werewolf}. The interaction protocol affects the result alongside the model: who can speak, what history is visible, and how speech relates to legal actions shape the strategic problem. Matched-casting comparisons in Mafia measure this effect: in societies of language-model agents, simultaneous broadcast improves identification of hidden adversaries over silence in all nine compositions tested with 8--46 players, whereas turn-taking removes most of that advantage; at 70 players, agents reading eight statements per day identify adversaries worse than silent ones, and in controlled redeployments, societies carrying sixty generations of their own strategy notes score below societies with none \citep{huang2026mafiasocieties}. Cross-game comparisons therefore need both protocol and model details.

\runin{Social agents, NPCs, and companions}\figmark{\ref{fig:strip-3-4}b--d} NPC behavior can serve characterization and interaction, where competitive score may be beside the point. Comme il Faut in Prom Week models relationships and social affordances explicitly, making social state part of gameplay \citep{mccoy2013promweek}. Generative Agents instead retrieves memories and uses language-model reflection and planning to produce individual and emergent behavior in a sandbox \citep{park2023generativeagents}. Authored social rules constrain permissible interactions, whereas language-based reasoning broadens the behaviors a character can propose. Both approaches need to track who knows what and how a player's intervention changes the scene.

Larger populations add coordination across concurrent activities: Project Sid analyzes a 500-agent Minecraft simulation spanning multiple societies under the PIANO architecture, which keeps an agent coherent across several concurrent output streams while it interacts with humans and other agents, and reports groups of about 30 agents developing specialized roles and changing collective rules \citep{al2024project}. AgentSociety simulates more than 10,000 agents and five million interactions in a realistic societal environment, which places it beyond games but shows the scale such populations can reach in large-scale simulation \citep{piao2025agentsociety}.

Targeted studies examine the coordination, fairness, and timing of NPC behavior. CASCADE evaluates a layered NPC coordination architecture through micro-scenario prototypes and trace analysis \citep{xu2026cascade}, while FAIRGAMER tests social bias in NPC interactions \citep{shi2026fairgamer}. Proact-VL decides when to comment or assist during streaming play, making the timing of initiative part of companion behavior \citep{yan2026proact}. Coordination, fairness, and timing extend beyond selecting an effective movement or combat action. Runtime dialogue, memory, and their effects on players are developed in \Cref{sec:adapt}, and learned simulators with explicit NPC control are discussed in \Cref{sec:world-models}.

\begin{gameaiinsight}[Player]{Transfer happens at different levels of control}
\begin{insightpoints}
\item \textbf{A common interface changes what the agent must learn.} A language planner over executable skills can reuse task knowledge while a specialist controller supplies timing and movement. Native-control policies must learn more of that mapping themselves. Voyager \citep{wang2023voyager}, STEVE-1 \citep{lifshitz2023steve1}, and NitroGen \citep{magne2026nitrogen} therefore offer different kinds of reuse.
\item \textbf{Adaptation results depend on the information supplied.} REGENT \citep{sridhar2025regent} retrieves demonstrations; Twin \citep{skoutnev2026twin} experiments to infer rules. Both adapt without ordinary per-game policy training, but one receives examples and the other pays to acquire them. Comparing unfamiliar-game performance requires accounting for that information and interaction budget.
\item \textbf{Working with new partners needs its own evaluation.} Shared weights or strong self-play can preserve conventions that unfamiliar partners do not share. Other-Play \citep{hu2020otherplay} and Fictitious Co-Play \citep{strouse2021fcp} address partner variation explicitly; task success alone leaves this dimension of generalization unresolved.
\end{insightpoints}
\end{gameaiinsight}

\gameaisectionaccent{Simulator}
\renewcommand{\dmcolor}{Simulator}
\section[AI That Models Players and Games]{AI That Models Players and Games}
\label{sec:world-models}

The agents of \Cref{sec:agents} act inside an existing game, and much of what they and other systems decide depends on what will happen next: a planner evaluates possible actions, a policy learns from simulated experience, and an adaptive system anticipates a player's response. These uses place different demands on a model. Pretrained video and language representations broaden the observations and descriptions it can accept, but they do not specify which hidden variables, action effects, or behavioral regularities must be preserved. The comparison in this chapter centers on that choice: what is represented and learned, how predictions are used, and whether they remain useful through interaction. World simulation and player modeling share this dependence on use, but differ in data and validation.

\draftmap{\begin{tikzpicture}[x=1cm,y=1cm]
\node[dm q] at (-1.8,2.35) {How the chapter moves: the consumer, then what the model must hold, then how it runs live; the player is the parallel object.};
\node[dm node=3.6cm] (a) at (0,0) {\dmhead{4.1 Consumers of the model}\\[1pt]{\scriptsize planner, policy learner, player}};
\node[dm node=3.6cm] (b) at (4.8,0) {\dmhead{4.2 State and persistence}\\[1pt]{\scriptsize state, gaps, persistence}};
\node[dm node=3.6cm] (d) at (9.6,0) {\dmhead{4.4 Interactive generation}\\[1pt]{\scriptsize actions in, frames out, drift}};
\node[dm node=3.6cm] (c) at (0,-2.2) {\dmhead{4.3 Player models}\\[1pt]{\scriptsize logs, moves, profiles}};
\draw[dm arrow] (a) -- (b); \draw[dm arrow] (b) -- (d); \draw[dm arrow] (a) -- (c);
\node[dm bridge=2.9cm] at (2.4,0.8) {the consumer decides what must be preserved};
\node[dm bridge=2.9cm] at (7.2,0.8) {what is preserved must survive the loop that feeds on it};
\node[dm bridge=3.6cm,anchor=west] at (2.1,-1.1) {same question, other object: the player instead of the game (parallel branch)};
\node[dm frame,fit=(current bounding box)] (F) {};\dmtag
\end{tikzpicture}}

\subsection{Planning Models, Training Environments, and Interactive Simulators}
\label{sec:model-consumers}

\draftmap{\begin{tikzpicture}[x=1cm,y=1cm]
\node[dm q] at (-2.1,2.35) {Who uses the prediction, and what does each consumer need from it?};
\node[dm node=4.0cm] (a) at (0,0) {\dmhead{Planning models}\\[1pt]{\scriptsize Dyna, I2A, MuZero}};
\node[dm node=4.0cm] (b) at (5.5,0) {\dmhead{Learned training environments}\\[1pt]{\scriptsize SimPLe, Dreamer, Dreamer~4}};
\node[dm node=4.0cm] (c) at (11.0,0) {\dmhead{Interactive simulators}\\[1pt]{\scriptsize GameGAN, Genie, GameNGen, WHAM}};
\draw[dm arrow] (a) -- (b); \draw[dm arrow] (b) -- (c);
\node[dm bridge=3.2cm] at (2.75,0.8) {a learner drives the model into states it never saw};
\node[dm bridge=3.2cm] at (8.25,0.8) {for a person, the generated frame is the experience itself};
\draw[dm axis] (-2.1,-1.15) -- (13.1,-1.15) node[midway,below=1pt,font=\scriptsize,text=GameSlate] {from the most internal use to the most exposed};
\node[dm frame,fit=(current bounding box)] (F) {};\dmtag
\end{tikzpicture}}

\begin{figure}[H]
  \centering
  \includegraphics[width=\linewidth]{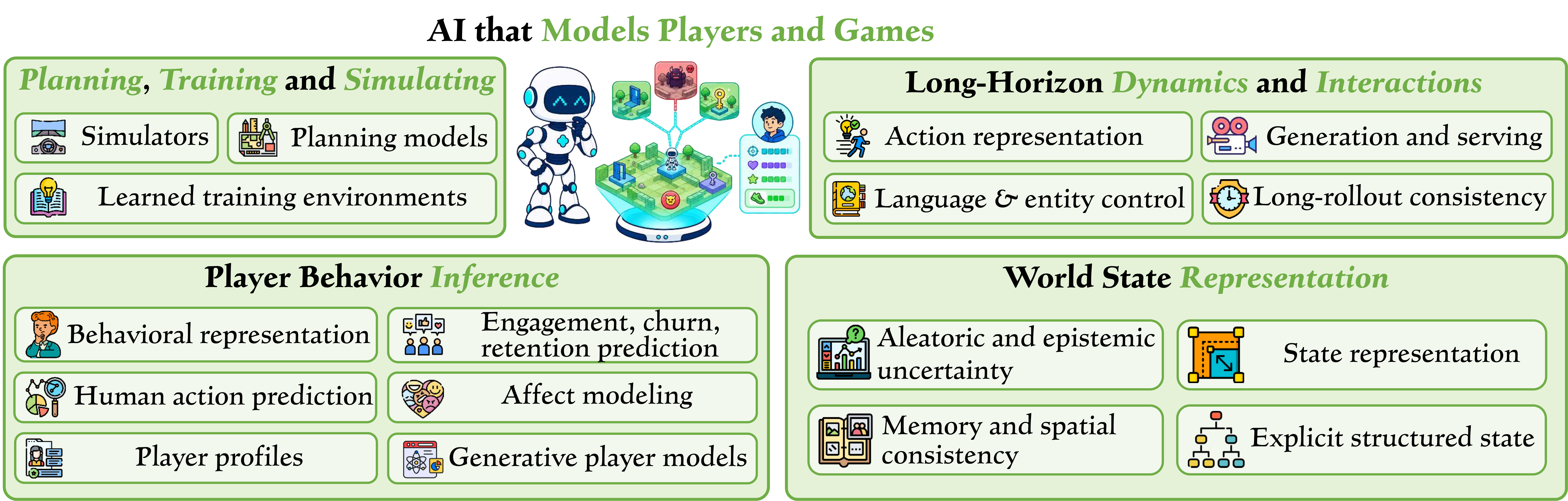}
  \caption{Research directions for modeling players and games: planning, training, and simulation; world-state representation; player behavior modeling; and long-horizon interaction.}
  \label{fig:wm-ladder}
\end{figure}

Learned dynamics may support planning, policy training, or direct interaction (\Cref{fig:wm-ladder}). A planner needs distinctions that affect action choice; a policy learner needs reliable rollouts under its evolving behavior; a player encounters the generated observations themselves. The same prediction loss can therefore conceal quite different errors in use.

\runin{Planning models} Dyna established the loop of learning transitions and using them for simulated planning \citep{sutton1990dyna}. Action-conditional video prediction for Atari \citep{oh2015actionconditional} and recurrent environment simulators \citep{chiappa2017res} then showed how pixels and actions could be rolled forward over many steps. World Models \citep{ha2018worldmodels}, PlaNet \citep{hafner2019planet}, and Dreamer \citep{hafner2020dreamer} brought this loop to visual control by learning compressed dynamics and optimizing behavior in imagined trajectories. Imagination-Augmented Agents made the consumer explicit: a policy learned how to interpret imperfect imagined rollouts, so model predictions never replaced the environment outright \citep{racaniere2017i2a}. MuZero instead learns latent dynamics together with the reward, value, and policy predictions used by search, without reconstructing future observations \citep{schrittwieser2020muzero}. For an agent-internal model, useful fidelity lies in preserving the distinctions that can change an action choice.

Predicting observations and predicting decision-relevant quantities produce different failure modes. A search model may ignore texture while preserving reward and legal strategic alternatives. A reconstruction model may reproduce texture while missing a rare transition that changes the optimal action. A transition-level diagnostic quantifies this mismatch: event transitions, where a reward is earned or an episode ends, are rendered about as well as other steps, yet replacing each one and its immediate neighbors with real observations removes a median 70\% of return error, whereas replacing equally many transitions with the largest MSE, SSIM, or LPIPS error removes less than a random choice. Event patches outperform random patches in 39 of 44 Atari-100k model--collection pairs \citep{liu2026eventsnotpixels}. I2A learns an interpretation of rollouts, whereas MuZero trains the quantities consumed by search directly. Planning quality therefore depends on the coupling between model objective and decision procedure, not simply on how much of the visible environment the model reconstructs. A planner can query a fixed model, whereas a policy learner may shift the distribution of states the model must predict as its behavior changes.

\runin{Learned training environments} A learned training environment is often evaluated under a changing policy. As the policy improves inside the model, it visits states and action sequences that may be rare in the training data. A small transition error can then look, to the improving policy, like an attractive strategy. SimPLe \citep{kaiser2020simple} and Dreamer \citep{hafner2020dreamer} connect model quality to the return achieved in the reference environment after learning from imagined trajectories. DreamerV3 improves the robustness of this online loop across diverse tasks using a fixed configuration \citep{hafner2025dreamerv3}. Dreamer~4 studies a different condition: learning Minecraft behavior from a fixed offline dataset. Using 2,500 hours of action-labeled contractor data, its reported policy obtains an iron pickaxe in 29\% of evaluation episodes and diamonds in 0.7\%. Episodes allow 60 minutes, native mouse/keyboard control, and a prescribed sequence of task prompts \citep{hafner2025dreamer4}. The Dreamer~4 results show useful policy improvement through imagination while also locating the remaining difficulty in the final stages of a long task. A ten-hero MOBA study instead judges a structured multi-agent world model from outside. The world model starts from 2,000 real games of the shipped policy, scripted bots, and random play. After that, the real game supplies no gradient, only the policy's own games as further training data and an online evaluation that selects and anchors the policy. In this continuous Dyna loop, a policy trained in 1,400-tick imagined episodes wins 70.2\% of real games as one side against the shipped opponent (95\% CI 66.4--73.7), up from 0\% for dream training alone, but none as the other side. The shipped opponent playing itself shows the same asymmetry, winning all 300 games on the first side and none on the other. Every unanchored run collapsed while no in-dream metric tracked the collapse; retraining the model on the policy's own games was worth 9.2 points of real win rate with the policy recipe fixed; and the model represented the macro game but not crowd control, cast timing, or lethality, while the policy won through map-wide pressure with almost no coordinated fighting \citep{kieto2026mobadyna}. Online model learning, offline policy learning, and successful deployment in the reference game should be distinguished when comparing simulators.

Online and offline training also differ in how errors are corrected. An online learner can collect reference-game transitions after its policy changes and fit the newly visited states. An offline learner must work with the coverage of a fixed dataset, even when optimization discovers actions outside that coverage. Short imagined rollouts, policy constraints, and conservative action selection address different parts of this problem: reducing compounding error, limiting distribution shift, and avoiding uncertain predictions. Their success is measured by the policy in the reference environment, not by its reward inside the learned simulator.

\begin{figure}[!htb]
\centering
\setlength{\tilew}{0.47\linewidth}
\setlength{\tilelabelh}{14.0mm}
\blocktile{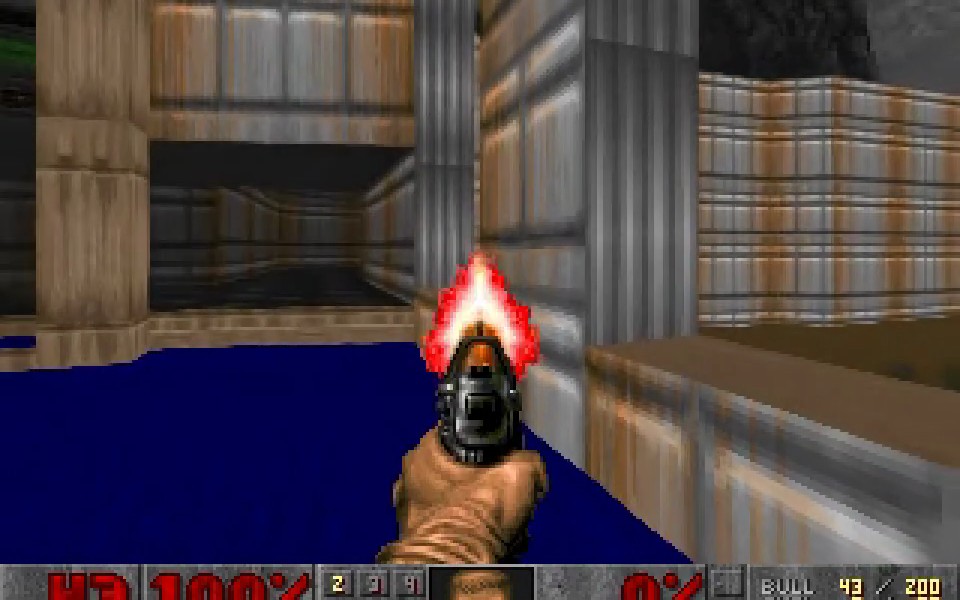}{(a) Interactive simulators}{GameNGen, DOOM}{A playable DOOM level generated frame by frame}\hfill
\blocktile{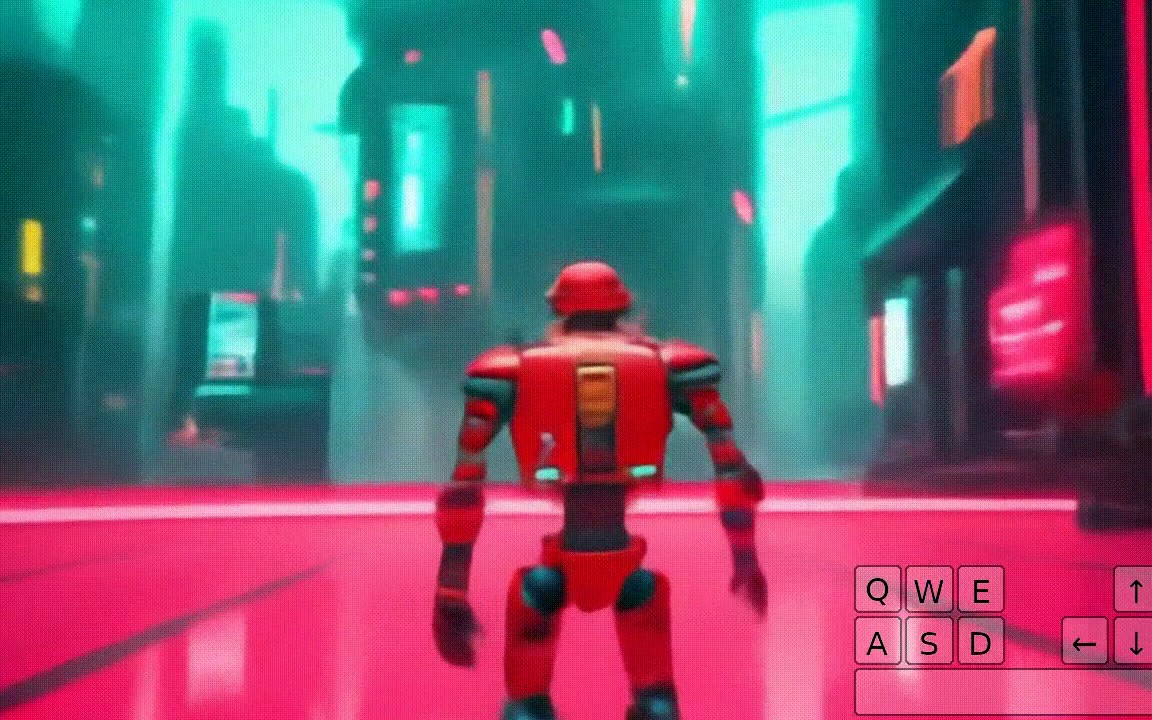}{(b) Interactive simulators}{Genie 2, generated street}{Generated street responding to the pressed key}\\[3pt]
\blocktile{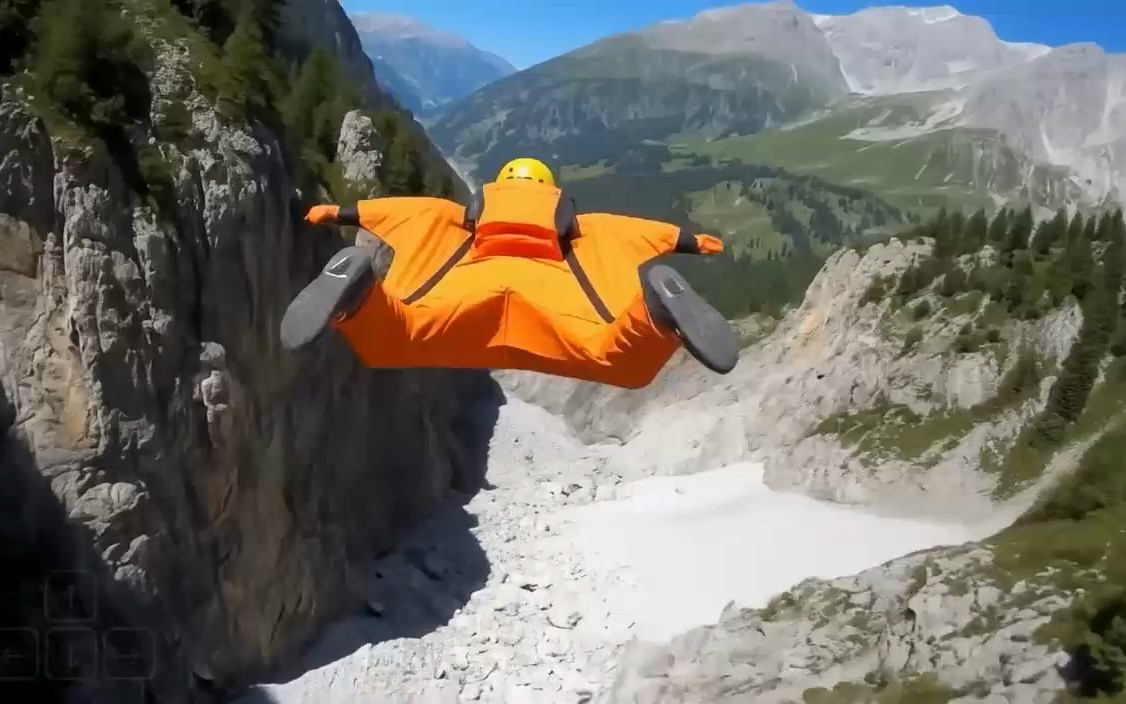}{(c) Interactive simulators}{Genie 3, generated world}{A flyable world generated from a prompt}\hfill
\blocktile{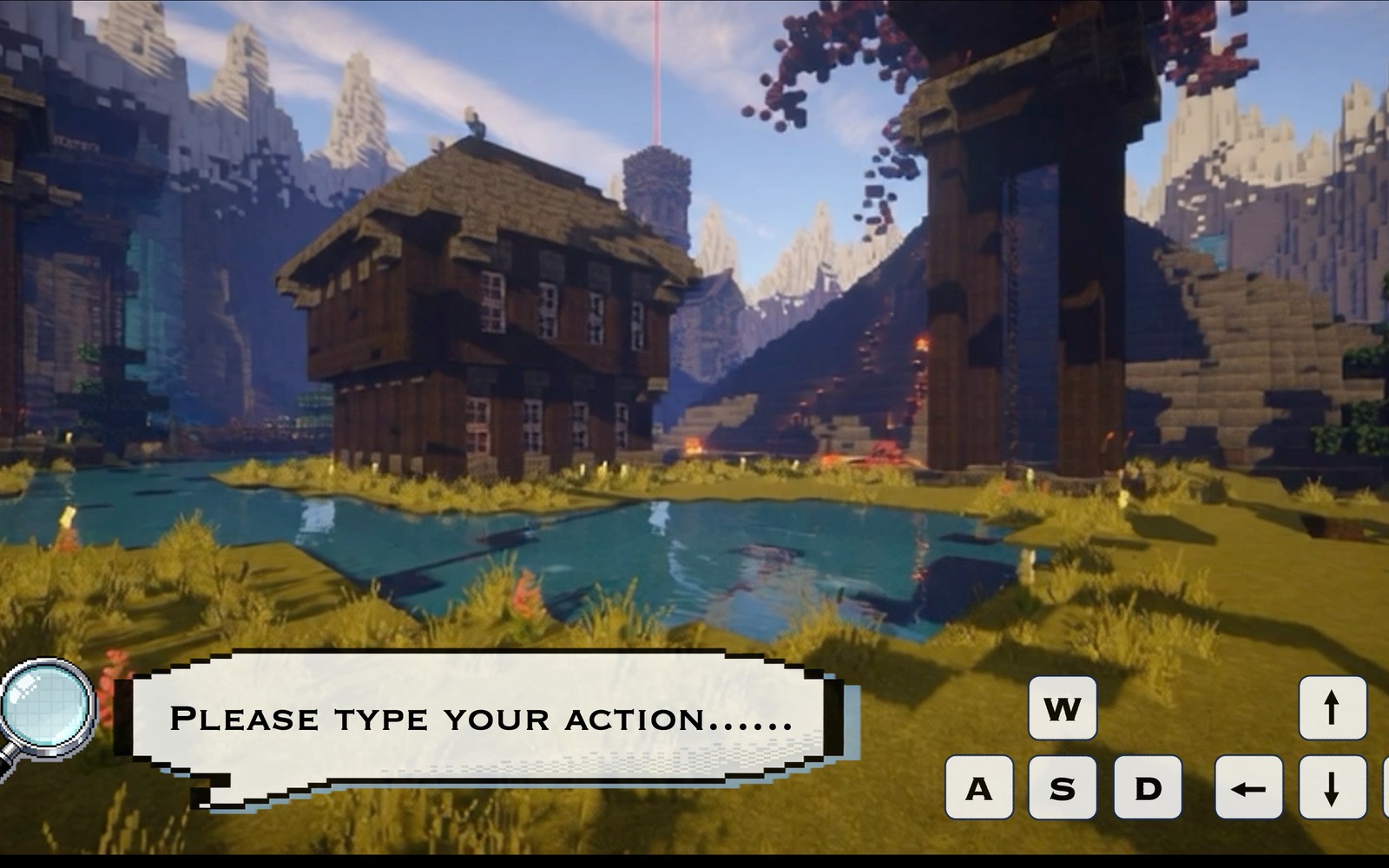}{(d) Interactive simulators}{Hunyuan-GameCraft-2, generated voxel world}{Generated village steered by keys and text}
\caption{Learned simulators that a player interacts with directly; each frame is produced in response to the input shown: (a) GameNGen, DOOM \citep{valevski2024gamengen}; (b) Genie 2, generated street \citep{parkerholder2024genie2}; (c) Genie 3, generated world \citep{parkerholder2025genie3}; (d) Hunyuan-GameCraft-2, generated voxel world \citep{tang2025hgc2}.}
\label{fig:strip-4-1}
\end{figure}

\runin{Interactive simulators}\figmark{\ref{fig:strip-4-1}a--d} Player-facing simulators make generated observations part of the experience itself. GameGAN demonstrated action-conditioned visual emulation of a particular game \citep{kim2020gamegan}, Genie learned interactive environments and latent controls from unlabeled video \citep{bruce2024genie}, and GameNGen produced real-time Doom interaction with an action-conditioned diffusion model \citep{valevski2024gamengen}. Recent systems such as GameGen-X \citep{che2024gamegenx} and Hunyuan-GameCraft-2 \citep{tang2025hgc2} broaden the visual and instruction-conditioned domain, although their evidence centers on video quality, controllability, and instruction or action response, and does not establish persistent game-state or mechanics fidelity. Google DeepMind's Genie~2 \citep{parkerholder2024genie2} and Genie~3 \citep{parkerholder2025genie3} announcements describe a related general-purpose direction, but they are technical announcements, not peer-reviewed papers. WHAM occupies a related position by generating gameplay continuations and controller actions for human ideation \citep{kanervisto2025wham}. Responsive control, visual continuity, and stable affordances all matter because players learn the world through the sequence of observations the model generates.

PlayGen combines game-specific data collection with an autoregressive diffusion Transformer and evaluates interactive mechanics as well as visual generation \citep{yang2024playgen}. Its selected 2D and 3D games show why interaction tests belong alongside video metrics, while leaving transfer to unfamiliar rules a separate question. MineWorld similarly studies real-time action-conditioned Minecraft simulation, with speed and control evaluated within that game's data and action space \citep{guo2025mineworld}. Both ground the generative model's controls in target-game interactions. Appendix~\ref{app:resources}, \Cref{tab:world-model-systems}, compares systems by representation, control inputs, and evaluation.

Players are not the only source of interaction a simulator meets. Agent trajectories can help train and test learned environments. GameNGen obtains simulator-training trajectories from an RL agent playing Doom \citep{valevski2024gamengen}, whereas Dreamer~4 uses a learned environment to train a policy \citep{hafner2025dreamer4}. PlayWorld uses agents to probe generated worlds, adapting action sequences to pursue the same objective across models \citep{ding2026playworld}. Such tests examine whether a model sustains controllable interaction in addition to producing visually plausible predictions (\Cref{sec:testing,sec:evaluation}). Supervision by interaction also appears where the model is symbolic and executable. Agent2World assigns separate agents to research a domain, implement a world model, and test its behavior through unit tests and simulation, and the resulting repair trajectories are then reused for fine-tuning \citep{hu2025agent2world}. Its benchmarks use PDDL and code representations in place of player-facing video, but they show that interaction can supervise the construction of a model as well as a policy.

\subsection{Representation, Uncertainty, and Persistent State}
\label{sec:model-state}

What a model retains helps determine which kinds of fidelity it can support. Perceptual quality concerns the rendered sequence; mechanics correctness concerns whether actions produce legal consequences; persistent state concerns whether those consequences survive later interaction and changes of view. These requirements overlap but are not interchangeable: a health bar can look convincing while displaying an incorrect value, and a correctly predicted pickup can be forgotten on a revisit. The evaluation analysis in \Cref{sec:evaluation-models} connects each requirement to a distinct test.

\draftmap{\begin{tikzpicture}[x=1cm,y=1cm]
\node[dm q] at (-1.65,2.35) {What does the model hold about the world, and does that account stay true as play continues?};
\node[dm node=3.3cm] (a) at (0,0) {\dmhead{State representation}\\[1pt]{\scriptsize PlaNet, IRIS, DIAMOND}};
\node[dm node=3.3cm] (b) at (4.3,0) {\dmhead{Aleatoric and epistemic uncertainty}\\[1pt]{\scriptsize PETS as precedent}};
\node[dm node=3.3cm] (c) at (8.6,0) {\dmhead{Memory and spatial consistency}\\[1pt]{\scriptsize ReWorld, WorldMem, VMem}};
\node[dm node=3.3cm] (d) at (12.9,0) {\dmhead{Explicit structured state}\\[1pt]{\scriptsize StatePlay, WorldMind, Marionette}};
\draw[dm arrow] (a) -- (b); \draw[dm arrow] (b) -- (c); \draw[dm arrow] (c) -- (d);
\node[dm bridge=2.7cm] at (2.15,0.8) {compression sets what is kept; data set what is known};
\node[dm bridge=2.7cm] at (6.45,0.8) {represented facts must persist across revisits};
\node[dm bridge=2.7cm] at (10.75,0.8) {spatial recall does not entail numerical state};
\draw[dm axis] (-1.65,-1.15) -- (14.55,-1.15) node[midway,below=1pt,font=\scriptsize,text=GameSlate] {from the state itself, to its gaps, to what must persist};
\node[dm frame,fit=(current bounding box)] (F) {};\dmtag
\end{tikzpicture}}

Compression, uncertainty estimation, retrieval, and explicit state each address part of what must be retained: omitting a visual detail is not the same as forgetting a collected item or predicting an unfamiliar transition with unjustified confidence about its outcome.

\runin{State representation} Representation determines which regularities are easy to preserve. Latent-state models move temporal prediction into a compressed representation, and their objectives determine how much reconstructive detail and control-relevant information that representation retains. PlaNet \citep{hafner2019planet} and Dreamer \citep{hafner2020dreamer} use stochastic recurrent state-space models, whereas MuZero learns a state shaped directly by search objectives \citep{schrittwieser2020muzero}. IRIS preserves a reconstructive visual target through a discrete autoencoder and models the resulting tokens autoregressively \citep{micheli2023iris}. DIAMOND provides an instructive counterpoint: its diffusion world model preserves visual details that improve Atari policy learning, suggesting that seemingly incidental pixels can carry action-relevant information \citep{alonso2024diamond}. Compression is useful only when it preserves the distinctions needed by the planner, learner, or player.

In a recurrent state-space model, the observed frame updates a posterior state during training or real interaction, while imagined rollouts use a predictive prior without the next observation. Reconstruction, reward prediction, and latent regularization shape what survives this transition \citep{hafner2019planet,hafner2020dreamer}. Token models place a discrete visual encoder before temporal prediction. Pixel diffusion retains a richer reconstruction target but allocates more computation to appearance. The relevant trade-off is thus not simply model size: representation, training losses, and rollout procedure jointly determine which errors the downstream controller can detect or exploit. Representation limits what is retained; training data limit what is learned.

\runin{Aleatoric and epistemic uncertainty} Random events and hidden state can make several futures plausible even when the dynamics are known. Epistemic uncertainty instead concerns what the model has not learned from its data. SimPLe \citep{kaiser2020simple} and DreamerV3 \citep{hafner2025dreamerv3} use discrete stochastic latent variables to represent alternative futures. Diverse samples, however, do not by themselves show that a model recognizes an unfamiliar mechanic or assigns calibrated probabilities to its outcomes. This distinction matters when a planner can exploit an incorrect transition that appears valuable. Ensembles offer a complementary way to estimate model uncertainty. PETS propagates stochastic-model and ensemble uncertainty through sampled trajectories \citep{chua2018pets}; Plan2Explore uses predictor disagreement to direct exploration \citep{sekar2020planning}. In offline learning, MOPO penalizes uncertain model rewards \citep{yu2020mopo}, while MOReL treats insufficiently supported transitions pessimistically \citep{kidambi2020morel}. These continuous-control studies supply technical precedents for game models, not game-simulator results. The practical distinction is between sampling a known random outcome and collecting evidence about an unknown rule. Visual diversity addresses neither rule uncertainty nor calibration automatically.

\begin{figure}[!htb]
\centering
\setlength{\tilew}{0.32\linewidth}
\setlength{\tilelabelh}{17.5mm}
\blocktile{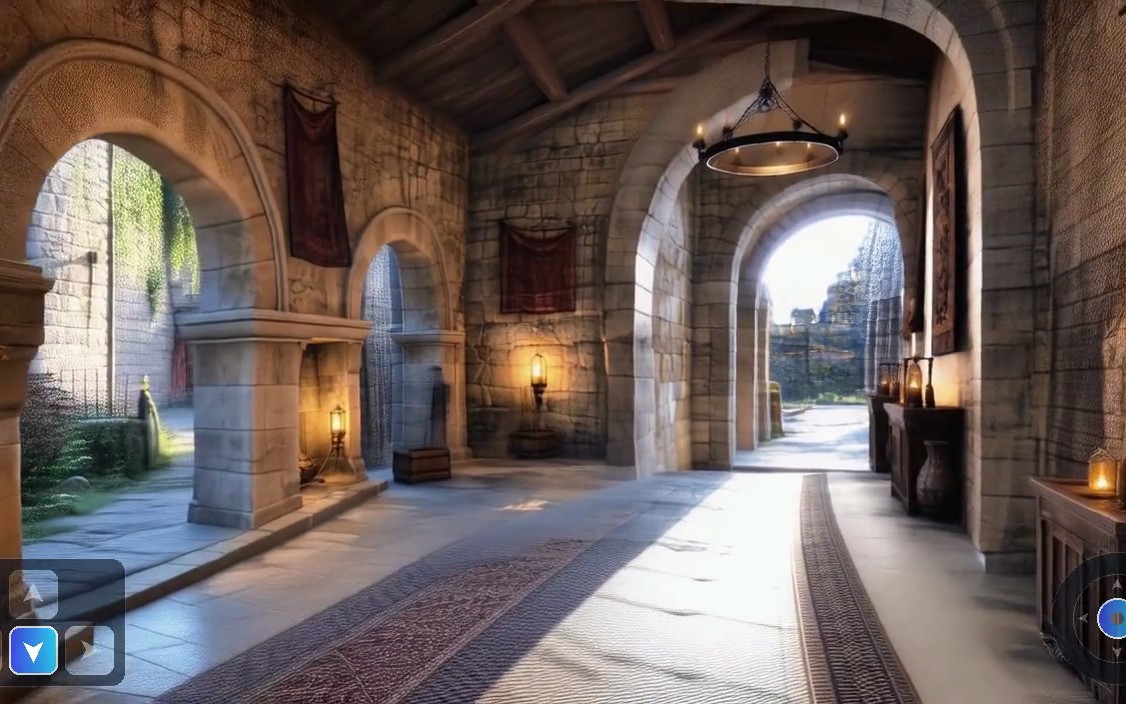}{(a) Memory and spatial consistency}{ReWorld, castle interior}{Hall unchanged after the camera returns}\hfill
\blocktile{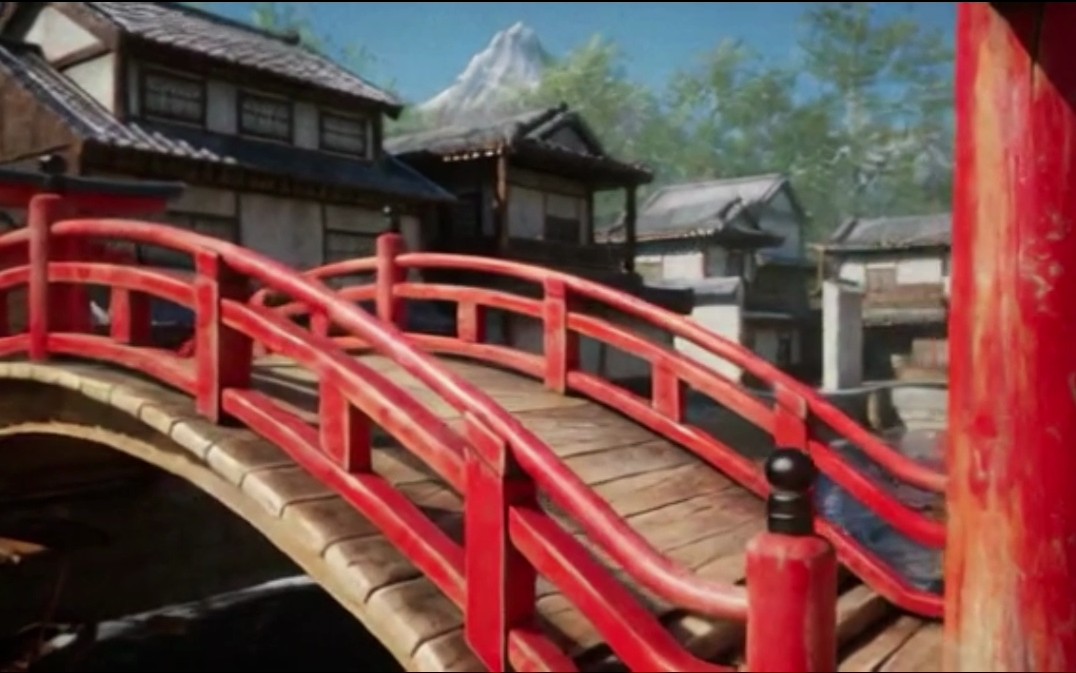}{(b) Memory and spatial consistency}{Context-as-Memory, Japanese townscape}{Bridge and roofs stable after a loop}\hfill
\blocktile{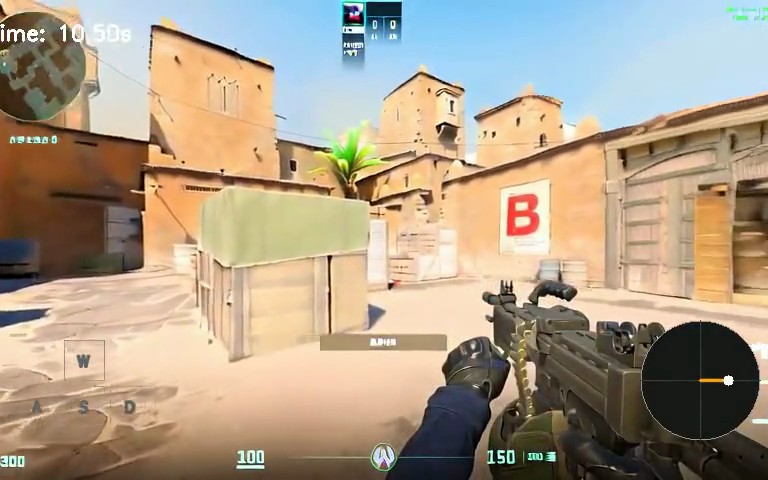}{(c) Memory and spatial consistency}{WorldCam, FPS map}{A courtyard revisited using camera-pose memory}\\[3pt]
\blocktile{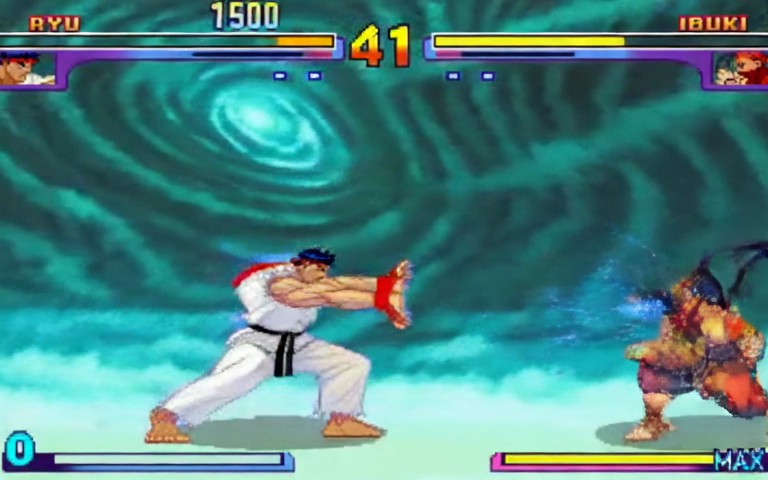}{(d) Explicit structured state}{StatePlay, Street Fighter III}{Health, timer and meters predicted as variables}\hfill
\blocktile{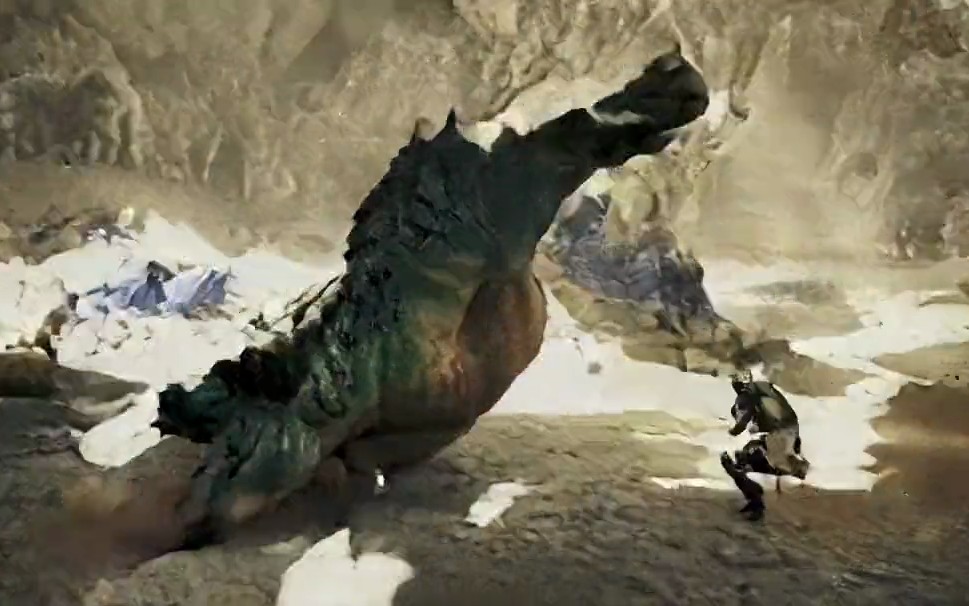}{(e) Explicit structured state}{Marionette, monster hunt}{Monster pose predicted as explicit state}\hfill
\blocktile{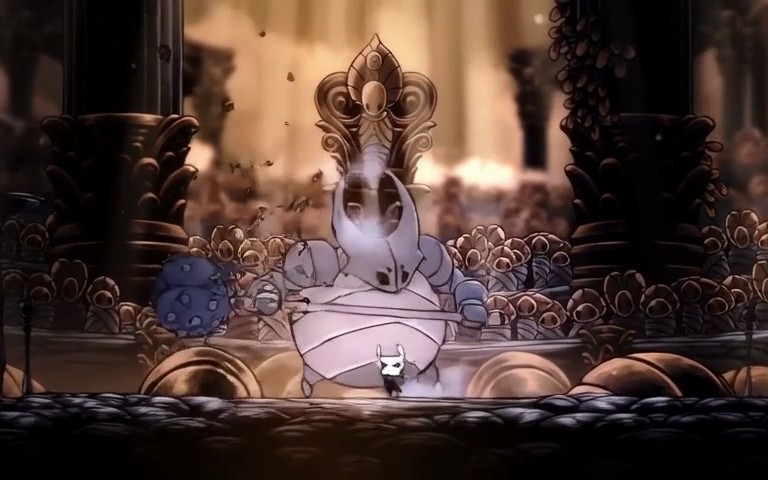}{(f) Explicit structured state}{WorldMind, Hollow Knight}{Boss behaviour driven by an explicit compact state}
\caption{Persistence in learned simulators: scenes that remain consistent when revisited (a--c) and game state predicted as explicit variables (d--f): (a) ReWorld, castle interior \citep{chen2026reworld}; (b) Context-as-Memory, Japanese townscape \citep{yu2025context}; (c) WorldCam, FPS map \citep{nam2026worldcam}; (d) StatePlay, Street Fighter III \citep{lin2026stateplaystateawaregameworld}; (e) Marionette, monster hunt \citep{meng2026marionettepredictingworldstates}; (f) WorldMind, Hollow Knight \citep{deng2026worldminddecoupledgameworld}.}
\label{fig:strip-4-2}
\end{figure}

\runin{Memory and spatial consistency}\figmark{\ref{fig:strip-4-2}a--c} Visual history often serves as the primary state store in video models. It naturally preserves recent motion and appearance, but off-screen objects, inventories, delayed events, and revisited geometry place increasing pressure on a finite context. Current systems combine three responses to this pressure.

The first extends what a fixed context can hold. ReWorld keeps a fixed-budget key--value cache behind mixed local and global attention \citep{chen2026reworld}. FramePack compresses input frames by importance so that more history fits within a constant context length \citep{zhang2025frame}, and Mixture of Contexts replaces dense attention with learned sparse routing, in which each query attends to a few informative chunks plus mandatory anchors \citep{cai2025mixture}. The second retrieves earlier views on demand. WorldMem indexes past frames by pose and time \citep{xiao2025worldmem}, Context-as-Memory selects context frames by field-of-view overlap between camera poses \citep{yu2025context}, and ReWorld adds pose-indexed landmark retrieval. The third anchors memory to geometry instead of frames. PERSIST evolves a latent 3D scene through separate environment, camera, and rendering components \citep{garcin2026persist}, WorldCam uses camera pose to organize an autoregressive 3D world \citep{nam2026worldcam}, and VMem indexes past views by the surface elements they observed so that generation retrieves only the views relevant to the current viewpoint \citep{li2025vmem}. AlayaWorld's revised renderer reprojects a streaming 3D point cache into the next viewpoint, while Alaya-EVOKE retrieves geometry-indexed history to reduce long-context conditioning costs~\citep{alayaworld2026full,yin2026evoke}. These extend the geometry-based route.

Evidence for all three routes is mostly visual. ReWorld reports visual recall over 64-second out-and-back trajectories and an overall rotation error of 11.95$^\circ$ \citep{chen2026reworld}, testing long-horizon view recovery and camera control, respectively. Numerical mechanics, spatial recall, and narrative or inventory persistence require different records and different tests, and success on one does not settle the others.

\runin{Explicit structured state}\figmark{\ref{fig:strip-4-2}d--f} Recent systems make selected missing variables explicit. Model as a Game retains numerical events in an external computation module and conditions visual generation on explicit spatial memory \citep{chen2025modelgame}. StatePlay jointly predicts frames and internal variables such as health, meters, and timers. In its \emph{Street Fighter III} testbed, the authors report normalized state error below $0.06$ and a gain of 18.6 percentage points over the strongest stateless baseline on mechanics-fidelity judgments. The model receives the initial numerical state. Subsequent state error is checked against recorded traces, whereas mechanics fidelity is scored by two vision--language judges on 100 test clips \citep{lin2026stateplaystateawaregameworld}. WorldMind reconstructs a compact state before selecting non-player-character behavior and rendering the next observation. On one-minute Game~A rollouts, two LLM judges prefer WorldMind to the baselines in approximately 70\% of pairwise comparisons for tactical appropriateness and coherence. Engine state is used for data collection but is not exposed at inference \citep{deng2026worldminddecoupledgameworld}. Marionette takes a related hybrid route by predicting an articulated state, applying a fixed geometry renderer, and then synthesizing appearance \citep{meng2026marionettepredictingworldstates}. StatePlay, WorldMind, and Marionette supply different kinds of structure: joint numerical prediction, explicit NPC decisions, and geometric rendering. Their reported improvements support those specific components, and none supports a claim of complete game-state correctness.

Explicit state is not a single representational choice. Numerical variables make updates such as damage or elapsed time measurable. Geometric state constrains where objects can reappear. Executable code can enforce a transition instead of predicting it: WorldCoder builds a Python program of its environment from interaction and edits that program to transfer across gridworlds \citep{tang2024worldcoder}, GIF-MCTS synthesizes code world models through a generate--improve--fix search evaluated on an 18-environment benchmark \citep{dainese2024generating}, and the code world models of \Cref{sec:play-adaptation} translate stated game rules into executable simulators \citep{lehrach2025cwm}. State World Models test this division with frozen video generators: on 50 counterfactual twin pairs that share an initial frame and action but differ in hidden state and required outcome, the best of 11 video backends produces both required endings in 32\% of pairs under rule-only prompting; for MiniMax-H3, specifying the required endings raises this to 64\%, and a harness that maintains world state under specified rules reaches 92\% with the same frozen model \citep{ki2026stateworldmodels}.

Later systems hand the transition to an engine and leave rendering to the video model. Fight World Model resolves Street Fighter~II combat in an executable engine and passes the outcomes to a video model through a compiled visual proxy, so rules such as damage scaling and lifesteal can be changed without retraining the renderer \citep{li2026fightworld}. In BoxWorld, a coding agent writes the engine from user rules and streams boxes, object attributes, and status variables to the renderer, which reaches a task score of 83.9 on its paired-state benchmark against at most 25.3 for the end-to-end video baselines \citep{wu2026boxworld}. WorldRuler keeps rule definitions and object states in an executable 3D rule graph, with perception feeding generated frames back into it, so that a saved and reloaded world follows the same rules \citep{zhai2026worldruler}. PROWBench then asks whether renderers show what the program logged: on 40 verified pairs, interaction success ranges from 0.381 to 0.727 across ten methods \citep{huang2026prowbench}. ContextRenderer works on the renderer side of this split. It trains on a 20K-clip dataset whose proxies span 11 formats so that one model can follow depth, semantic, low-poly, or primitive proxies, and its authors note that real-time use still needs faster inference \citep{wang2026contextrenderer}. PlayGen2 instead learns the state itself: an occupancy grid and identity-indexed 3D entity states evolve under player actions, and a separate renderer draws frames that never feed back into the state. After five minutes on its ViZDoom training map, PSNR is 19.4 against 16.1 for its frame-recurrent predecessor \citep{zhang2026playgen2}.

Numerical, geometric, and executable state incur different supervision costs. State traces require instrumentation, geometry requires spatial estimation or annotations, and executable rules require a specification and implementation. A model can combine them, but the resulting evaluation must still test whether the representations agree after action, especially when a visible animation and an internal variable suggest different outcomes.

\runin{State supervision from instrumented games} Games can expose synchronized observations that would be difficult to obtain from ordinary video. WildWorld records \emph{Monster Hunter Wilds} with aligned RGB, depth, camera poses, skeletons, actions, and state annotations. Its comparison of camera-, skeleton-, and state-conditioned generation separates sources of control that pixel-only datasets conflate \citep{li2026wildworld}. A separate data engine for \emph{Black Myth: Wukong} reports more than 90 hours of frame-aligned player inputs, engine states, and visual observations \citep{li2026pixelstates}. WorldRover instead renders prescribed explorations of artist-built Unreal environments, retaining geometry and trajectories while changing viewpoint or appearance. Its action signals are derived from trajectories, rather than recorded as human button presses \citep{xu2026worldrover}. This distinction matters when using the data to learn a controller.

Rendering buffers provide another kind of supervision. Generative World Renderer extracts RGB together with depth, normals, albedo, metallic, and roughness channels from two commercial games, and uses these buffers for generative appearance editing \citep{huang2026worldrenderer}. They describe visible geometry and materials, not health, inventory, or quest progress. Access to a game therefore creates an opportunity for precise supervision, not an automatic source of every relevant variable: collection still requires suitable instrumentation, permission, synchronized timestamps, and a known game version. Inferred poses or trajectories should remain distinguishable from quantities recorded directly by the engine.

\subsection{Player Models and Behavioral Inference}
\label{sec:model-players}

\draftmap{\begin{tikzpicture}[x=1cm,y=1cm]
\node[dm q] at (-1.4,2.35) {What is inferred about the player, from what, and what validates it?};
\node[dm node=2.6cm] (a) at (0,0) {\dmhead{Behavioral representation learning}\\[1pt]{\scriptsize player2vec, Structformer, bandits}};
\node[dm node=2.6cm] (b) at (3.3,0) {\dmhead{Human action prediction and human-likeness}\\[1pt]{\scriptsize Maia, Maia4All, LPLH}};
\node[dm node=2.6cm] (c) at (6.6,0) {\dmhead{Play-style, skill, and preference profiles}\\[1pt]{\scriptsize telemetry clustering, TrueSkill, Beyond Asking}};
\node[dm node=2.6cm] (d) at (9.9,0) {\dmhead{Affect and experience modeling}\\[1pt]{\scriptsize AGAIN, pixels-to-affect, VLM engagement}};
\node[dm node=2.6cm] (e) at (13.2,0) {\dmhead{Engagement, churn, and retention prediction}\\[1pt]{\scriptsize survival ensembles, DRL churn}};
\node[dm node=2.6cm] (f) at (16.5,0) {\dmhead{Generative player models and LLM player simulation}\\[1pt]{\scriptsize generative personas, MeepleLM}};
\draw[dm arrow] (a) -- (b); \draw[dm arrow] (b) -- (c); \draw[dm arrow] (c) -- (d); \draw[dm arrow] (d) -- (e); \draw[dm arrow] (e) -- (f);
\node[dm bridge=2.2cm] at (1.65,0.8) {an embedding is validated by what it predicts};
\node[dm bridge=2.2cm] at (4.95,0.8) {actions aggregate into a stable style or skill};
\node[dm bridge=2.2cm] at (8.25,0.8) {style is behavior; experience needs its own signal};
\node[dm bridge=2.2cm] at (11.55,0.8) {experience decides whether a player stays};
\node[dm bridge=2.2cm] at (14.85,0.8) {a model that predicts a player can also stand in for one};
\draw[dm axis] (-1.4,-1.15) -- (17.9,-1.15) node[midway,below=1pt,font=\scriptsize,text=GameSlate] {what is predicted, from events to actions, profiles, experience, retention, and finally whole players; validation grows stricter along the way};
\node[dm frame,fit=(current bounding box)] (F) {};\dmtag
\end{tikzpicture}}

Player models predict behavior or infer properties of people and populations from gameplay, questionnaires, or sensor data \citep{yannakakis2013playermodeling,smith2011inclusive,hooshyar2018data}. Experience-driven PCG \citep{yannakakis2011edpcg} and PaSSAGE \citep{thue2007passage} use such predictions to select content; procedural personas simulate styles for testing \citep{holmgard2019playtesting}. Broadly trained sequence and language models extend how behavior can be represented and generated. Their targets remain distinct: predicting a move, inferring a preference, estimating an experience, and simulating a player require different ground truth. Believable fictional behavior is not itself evidence of fidelity to a real person. The different prediction targets and their validation settings are indexed in Appendix~\ref{app:resources}, \Cref{tab:player-model-systems}.

\Needspace{4\baselineskip}
\runin{Behavioral representation learning} Representation learning offers one route from player logs to reusable models. player2vec serializes tracking events and trains a Longformer with masked-token prediction, and its evidence consists mainly of reconstruction metrics and qualitative embedding structure \citep{wang2024player2vec}. Behavior Structformer instead embeds structured event fields directly and evaluates supervised session-count targets constructed from behavioral logs \citep{smirnov2024behaviorstructformer}. The distinction matters: recovering a masked event, predicting an engagement proxy, and inferring a preference are different tasks. The two industrial studies demonstrate how game-specific event schemas can support sequence models, but do not independently validate a psychological interpretation of the learned embedding. Simpler representations can carry a surprising share of the signal. Behavioral features engineered from more than 75,000 battle-royale matches predicted player rank better than three mainstream rating systems, with some features informative for all players and others only for particular groups \citep{dehpanah2021player}. Modeling and adaptation can also be built as one process: a multi-armed-bandit formulation collects the data needed to model the current player while adapting the experience on the basis of that model, and can be tuned in simulation before a user study \citep{gray2021player}. Open player modeling asks whether and how players should see these models \citep{zhu2021open}.

Event encoding is a substantive part of this modeling problem. A sequence can represent a button press, a semantic action such as buying an item, or an aggregated session statistic. The encoding determines whether the model learns motor patterns, strategic routines, or population-level activity. Temporal order and time gaps can also carry different information: repeated actions in one encounter and the same actions across several days need not indicate the same preference. Reuse across games therefore requires an account of event correspondence, which a Transformer that accepts variable-length sequences does not by itself supply. An embedding is validated by what it predicts, and the most direct target is the player's next action.

\runin{Human action prediction and human-likeness} Predicting human actions, as opposed to optimal ones, is the aim of the Maia line in chess. Maia modeled the granular decisions that make up human play at each skill level, from twelve million online games per rating range, arguing that matching aggregate performance is not the same as matching behavior \citep{mcilroyyoung2020aligning}, and a follow-up learned models of individual players whose predictions are specific to one person \citep{mcilroyyoung2020learning}. Maia-2 conditions a shared chess model on player skill \citep{tang2024maia2}, while Maia-3 uses board-square tokens and geometric attention to improve human move matching \citep{monroe2026chessformer}, and both retain strong chess-specific representations. Maia4All targets the individual player beyond population-level skill: it first learns player prototypes, then adapts an embedding from as few as 800 positions, approximately 20 games. Its held-out move accuracy rises from about 51.4\% for Maia-2 to 53.2\% in that low-data condition \citep{tang2025maia4all}. This is evidence of more accurate individual action prediction within chess, without requiring the embedding to correspond to a named personality trait. Outside chess, the Learning to Play Like Humans framework guides a language model through interactive fiction with structured map building, action learning, and feedback-driven experience analysis, aligning the agent's behavior with narrative intent instead of task score alone \citep{zhang2025learning}.

Matching human actions and maximizing game strength are different objectives. A strong engine may systematically avoid mistakes that are characteristic of a particular skill group. Conditioning on skill models a population tendency, while adapting to one player's history can capture recurrent choices within that group. Held-out positions prevent memorizing specific examples. Held-out players test the population model. Later games by the same person test whether an individualized representation remains useful. The three splits answer different questions even when all are scored by move accuracy. Action histories can also support inference of broader properties such as style or skill, which are the next objects of analysis.

\runin{Play-style, skill, and preference profiles} Skill estimation and style inference use different observations. TrueSkill infers ability from match outcomes \citep{herbrich2006trueskill}, and QuickSkill estimates skill from early play \citep{zhang2022quickskill}; evaluations of rating systems examine accuracy and data efficiency in particular game populations \citep{boberirizar2024skill}. Telemetry clustering instead groups patterns of play, with results affected by the representation, clustering method, and observation period \citep{drachen2012guns,bauckhage2015clustering,drachen2014comparison,sifa2013behavior}. A useful caution comes from comparing computed and self-reported styles: behavioral features can predict agreement with procedural personas without predicting the style players attribute to themselves \citep{green2022predicting}. Foundation-model adaptation systems inherit this validation problem. Beyond Asking infers profiles from observed behavior, tests recovery of controlled synthetic traits, and includes an exploratory 12-participant pilot \citep{lu2026personalized}. A match-three deployment instead uses language-model reasoning to assign coarse player types and evaluates the resulting personalized levels \citep{hafnar2025zeroshot}. These test different parts of a pipeline. Synthetic trait recovery checks inference under the generator's assumptions; downstream player outcomes assess the intervention without necessarily identifying which inferred trait helped.

A player-modeling pipeline also needs to separate observations, inferred constructs, and intervention outcomes. An estimate of cautious play might predict avoidance behavior without corresponding to a stable personality trait. Conversely, a coarse profile may still be useful for selecting levels. For models inferred from interaction, calibration and temporal stability can be measured before deployment, while randomized content comparisons can assess downstream benefit. This prevents an improvement in level completion from being treated as proof that every intermediate profile is accurate. Style and skill can be inferred from behavior. Subjective experience requires a signal of its own.

\begin{figure}[!htb]
\centering
\setlength{\tilew}{0.32\linewidth}
\setlength{\tilelabelh}{21.0mm}
\blocktile{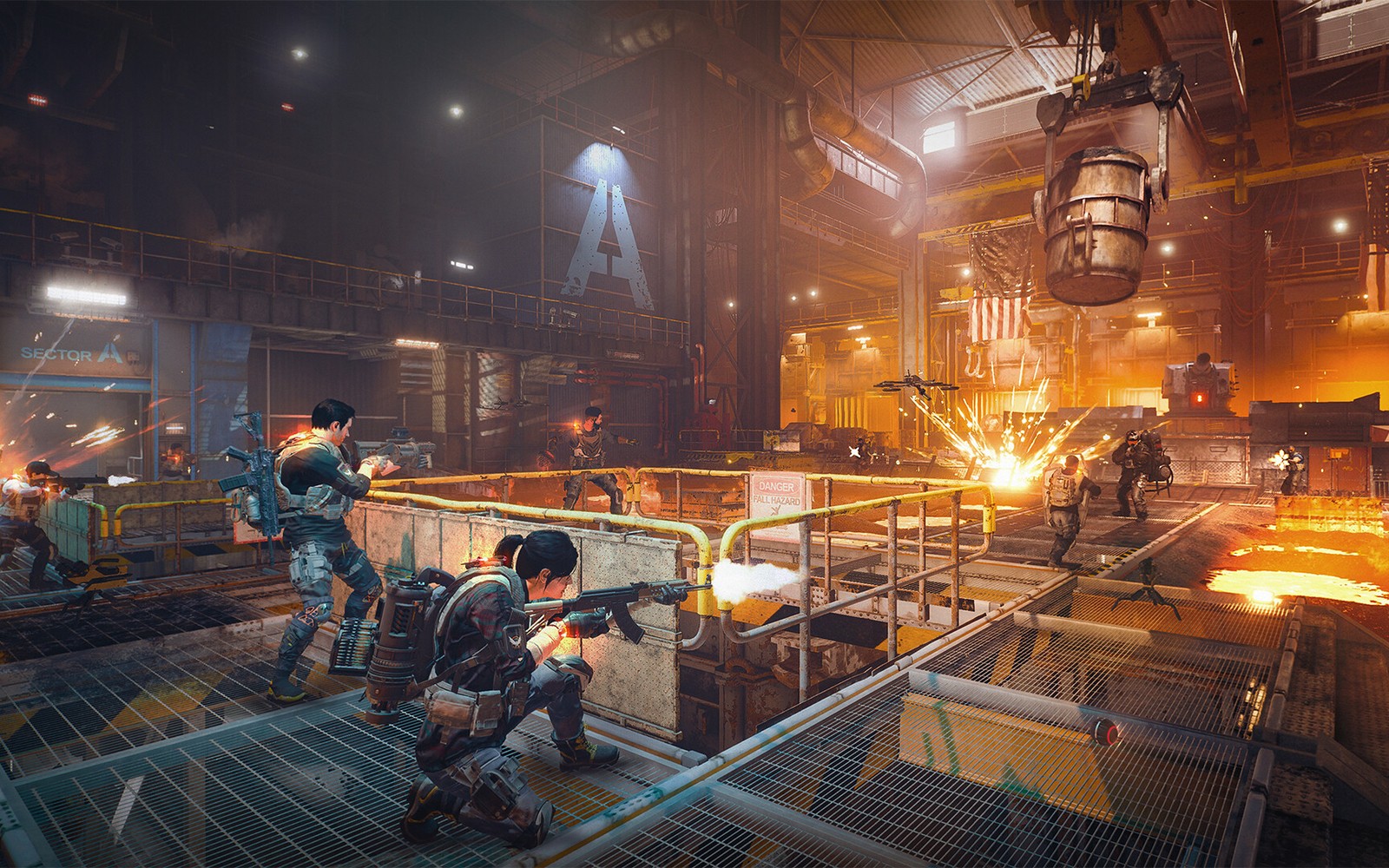}{(a) Affect and experience modeling}{Pinitas et al., The Division 2}{Commercial gameplay annotated\\for player engagement}\hfill
\blocktile{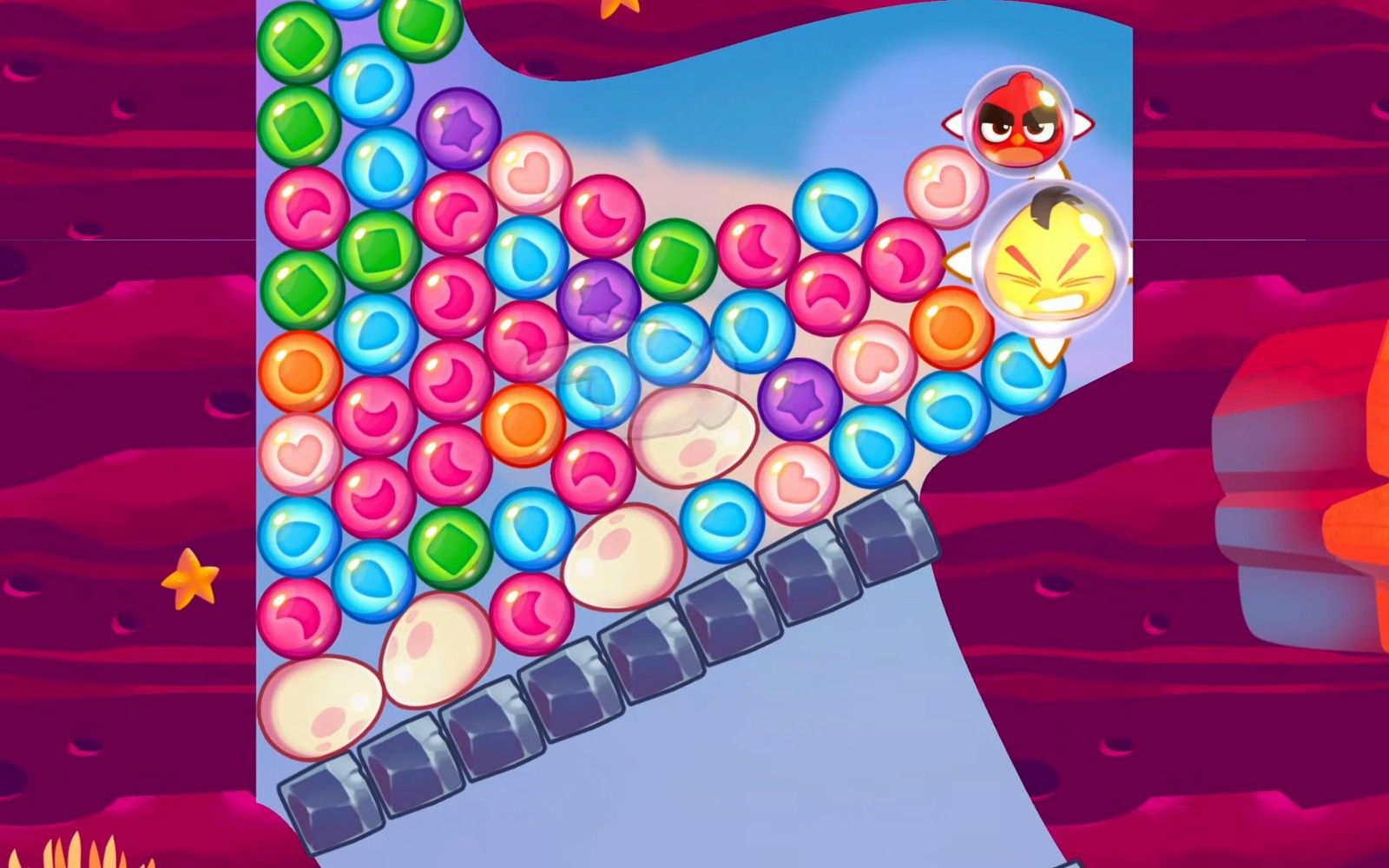}{(b) Engagement, churn,\\and retention prediction}{Roohi et al., Angry Birds Dream Blast}{Commercial title whose churn is predicted}\hfill
\blocktile{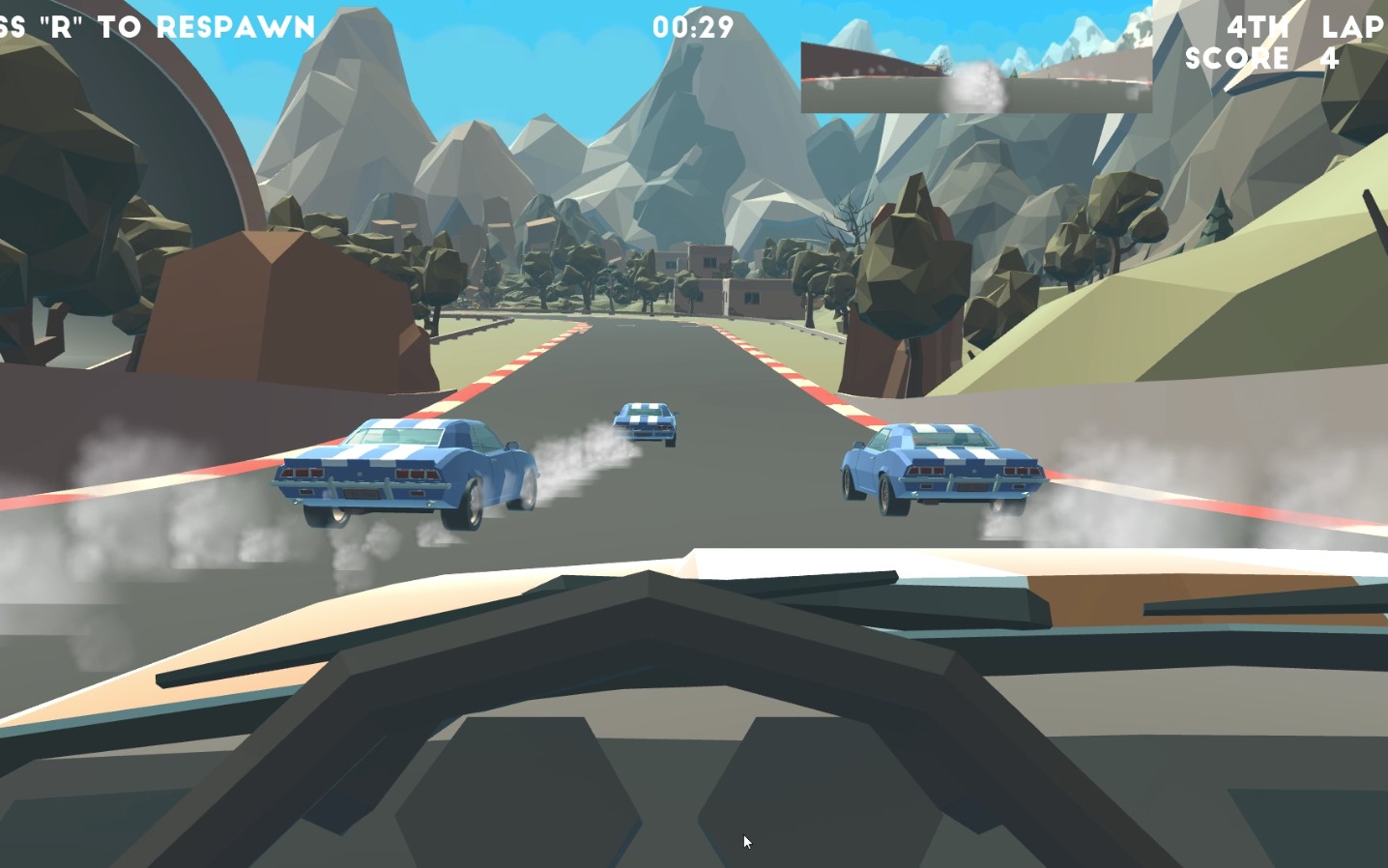}{(c) Generative player models\\and LLM player simulation}{Generative personas, Solid Rally}{Racing environment driven by generative personas}
\caption{Games in which player experience (a), retention (b), and simulated players (c) are modelled: (a) Pinitas et al., The Division 2 \citep{pinitas2023predicting}; (b) Roohi et al., Angry Birds Dream Blast \citep{roohi2020predicting}; (c) Generative personas, Solid Rally \citep{barthet2022generative}.}
\label{fig:strip-4-3}
\end{figure}

\runin{Affect and experience modeling}\figmark{\ref{fig:strip-4-3}a} Affective game computing treats player experience as the output of a loop of affect elicitation, sensing, detection, and adaptation, with its own annotation protocols and corpora \citep{yannakakis2023affective}. The AGAIN dataset illustrates this at scale: more than 1,100 in-game videos with gameplay data from nine games, annotated for arousal by 124 participants in a first-person, continuous fashion, totaling more than 37 hours of annotated gameplay \citep{melhart2021arousal}.

Gameplay footage also provides inputs for affect prediction. In leave-one-video-out evaluation on 45 retained videos of a survival shooter, convolutional networks classify high against low arousal from on-screen pixels. With annotations close to the mean arousal removed, 2DSeqCNN reaches the highest mean accuracy (78\%), and 2DFrameCNN reaches 98\% on its best held-out video \citep{makantasis2019pixels}. A general-purpose audiovisual approach, trained separately for each game, predicts arousal in four dissimilar games \citep{makantasis2021pixels}. In a commercial title, a study of 20 players' roughly one-hour sessions of Tom Clancy's The Division 2 (18.8 annotated hours) excludes 10-second windows within 0.05 of each player's normalized session mean and classifies the remaining windows as high or low engagement. On those windows, fusing footage with gamepad actions reaches 64.5\% average accuracy, and conditioning on the window's 20-minute segment of the session raises this to 72\% average and 87.7\% best-fold accuracy \citep{pinitas2023predicting}. A label-free approach instead trains on Let's Play videos, using the creator's voice amplitude as an affect proxy. In a 13-player study on an open-source Angry Birds clone, the model and players' self-reports both single out the second of three levels as the most intense. Correlations with electrodermal activity are mostly negative across players and levels, and the authors could not establish a moment-to-moment ground truth from the sensors \citep{goel2024label}.

Foundation-model studies examine whether broader visual representations support engagement prediction across games. A study of three vision--language models under six prompting strategies on nine first-person shooters finds zero-shot engagement predictions generally weak, often failing to beat per-game majority-class baselines, with retrieval-augmented prompting helping pointwise prediction in some settings while pairwise prediction of engagement change remains difficult \citep{wang2026vision}. The adaptation half of the loop is treated in \Cref{sec:runtime-players}.

\runin{Engagement, churn, and retention prediction}\figmark{\ref{fig:strip-4-3}b} Retention is an observable behavioral outcome, but not a direct measure of enjoyment. Survival models handle players whose eventual departure is not yet observed \citep{perianez2017churn}, while early-session heuristics can provide competitive retention baselines \citep{drachen2016rapid}. Usage-window modeling \citep{jang2021analyzing} and separate LSTM and survival-ensemble predictors \citep{guitart2019winning} further illustrate how the target depends on when observations are collected. These analytics studies provide context rather than the survey's foundation-model focus. More directly relevant to game design, simulated populations with different skill, persistence, and boredom can turn agent-estimated level difficulty into churn and pass-rate predictions \citep{roohi2020predicting}. Such estimates need validation against real populations and cannot be inferred from an agent's completion rate alone.

\runin{Generative player models and LLM player simulation}\figmark{\ref{fig:strip-4-3}c} Generative player models can predict actions, produce reactions to interventions, or estimate subjective judgments. Generative personas explicitly extend behavioral simulation to reported experience \citep{barthet2022generative}. Memory retrieval, reflection, and planning in Generative Agents provide mechanisms for individual and emergent social behavior in a sandbox \citep{park2023generativeagents}; fidelity to real players requires a human reference for the particular output being simulated.

Subjective judgment models aim to anticipate how different players will assess a game. MeepleLM is a virtual board-game playtester that aims to simulate different player groups' experiences and critiques. Its authors curated 1,727 structurally corrected rulebooks and 150,000 reviews and evaluated it on 207 games disjoint from the training corpus \citep{li2026meeplelm}. Collins et al. examine judgment prediction across 121 novel strategy games, comparing model evaluations with judgments of expected payoff and fun from over 450 participants. Agreement with game-theoretic estimates and agreement with people vary non-monotonically: stronger optimal-play reasoning alone does not establish a more faithful model of human judgments \citep{collins2025gameevaluations}. Subjective player models need validation against the intended group.

Behavioral simulation instead asks how players will act and respond to changes in the game. Beyond Playtesting adapts language models to a massively multiplayer game through supervised fine-tuning and reinforcement learning on large-scale real player behavior. Its offline simulations aim to reproduce player reasoning and reactions to interventions that would otherwise require live experiments \citep{zhang2025beyond}. Validation therefore needs held-out comparisons with the behaviors or responses being simulated. The representativeness of simulated testers is examined in \Cref{sec:test-representativeness}.

\subsection{Action Interfaces, Real-Time Generation, and Long-Horizon Consistency}
\label{sec:model-generative}

\draftmap{\begin{tikzpicture}[x=1cm,y=1cm]
\node[dm q] at (-1.65,2.35) {What goes in as an action, how does the next frame come out, and what breaks once the loop has run for long enough?};
\node[dm node=3.3cm] (a) at (0,0) {\dmhead{Action representation}\\[1pt]{\scriptsize GameFactory, SCOPE, Game2World, Genie}};
\node[dm node=3.3cm] (b) at (4.3,0) {\dmhead{Language and entity control}\\[1pt]{\scriptsize Incantation, ReactiveGWM}};
\node[dm node=3.3cm] (c) at (8.6,0) {\dmhead{Real-time generation and serving}\\[1pt]{\scriptsize Diffusion Forcing, Matrix-Game~2, FORGE}};
\node[dm node=3.3cm] (d) at (12.9,0) {\dmhead{Long-rollout consistency}\\[1pt]{\scriptsize Solaris, MultiGen}};
\draw[dm arrow] (a) -- (b); \draw[dm arrow] (b) -- (c); \draw[dm arrow] (c) -- (d);
\node[dm bridge=2.7cm] at (2.15,0.8) {language delegates part of the decision to the model};
\node[dm bridge=2.7cm] at (6.45,0.8) {any interface still needs a frame before the next input};
\node[dm bridge=2.7cm] at (10.75,0.8) {each generated frame becomes the next input};
\draw[dm axis] (-1.65,-1.15) -- (14.55,-1.15) node[midway,below=1pt,font=\scriptsize,text=GameSlate] {input $\to$ mechanism $\to$ horizon};
\node[dm frame,fit=(current bounding box)] (F) {};\dmtag
\end{tikzpicture}}

Interactive generation must reconcile action semantics, response time, and temporal consistency. A model trained on abundant unlabelled video may lack a precise control vocabulary; a detailed action-conditioned generator may respond too slowly. Longer history helps some forms of recall while increasing serving cost. These trade-offs depend on the training representation and rollout mechanism, not on frame quality alone.

\begin{figure}[!htb]
\centering
\setlength{\tilew}{0.32\linewidth}
\setlength{\tilelabelh}{17.5mm}
\blocktile{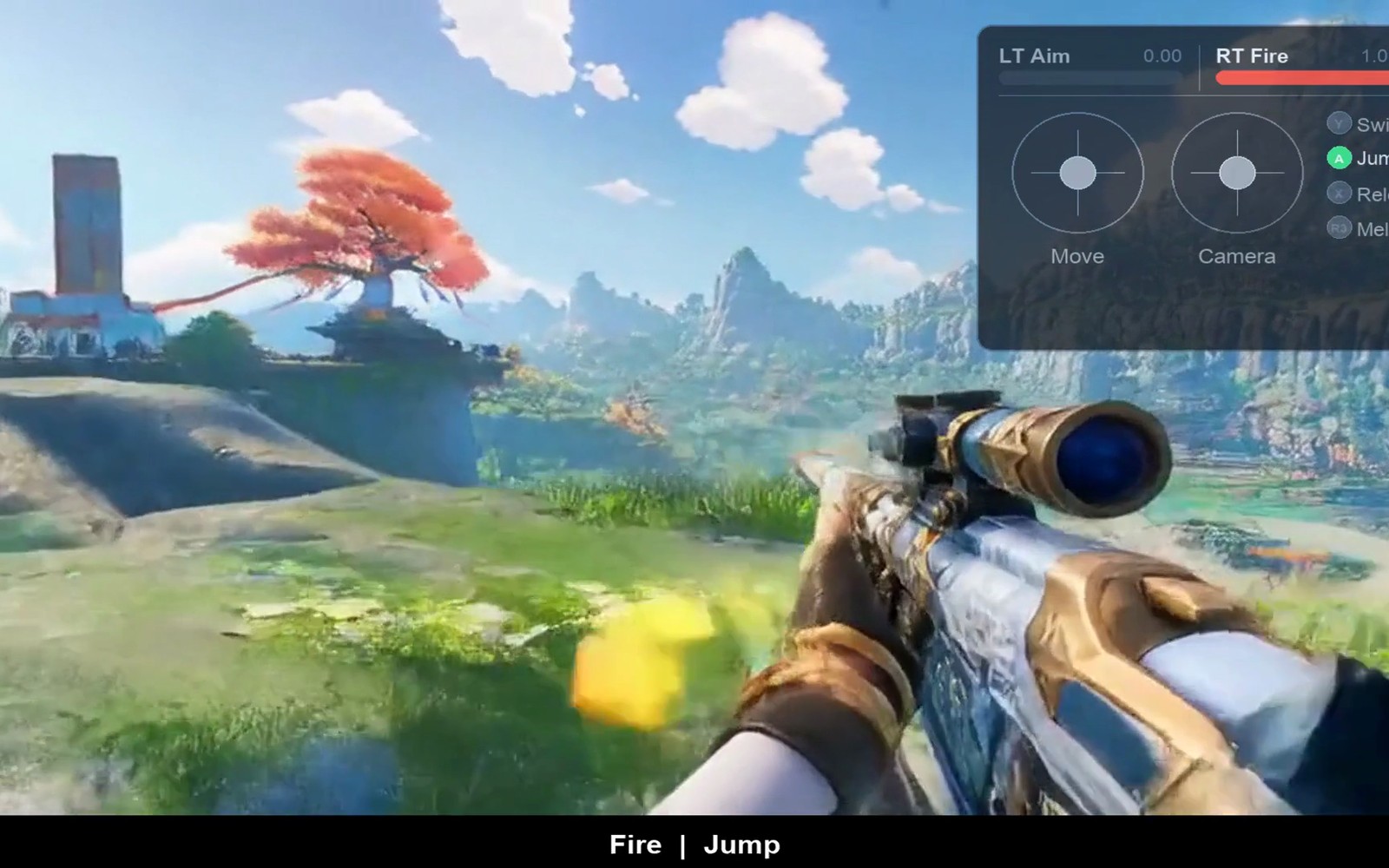}{(a) Action representation}{SCOPE, open-world FPS}{Gamepad overlay showing the driving action}\hfill
\blocktile{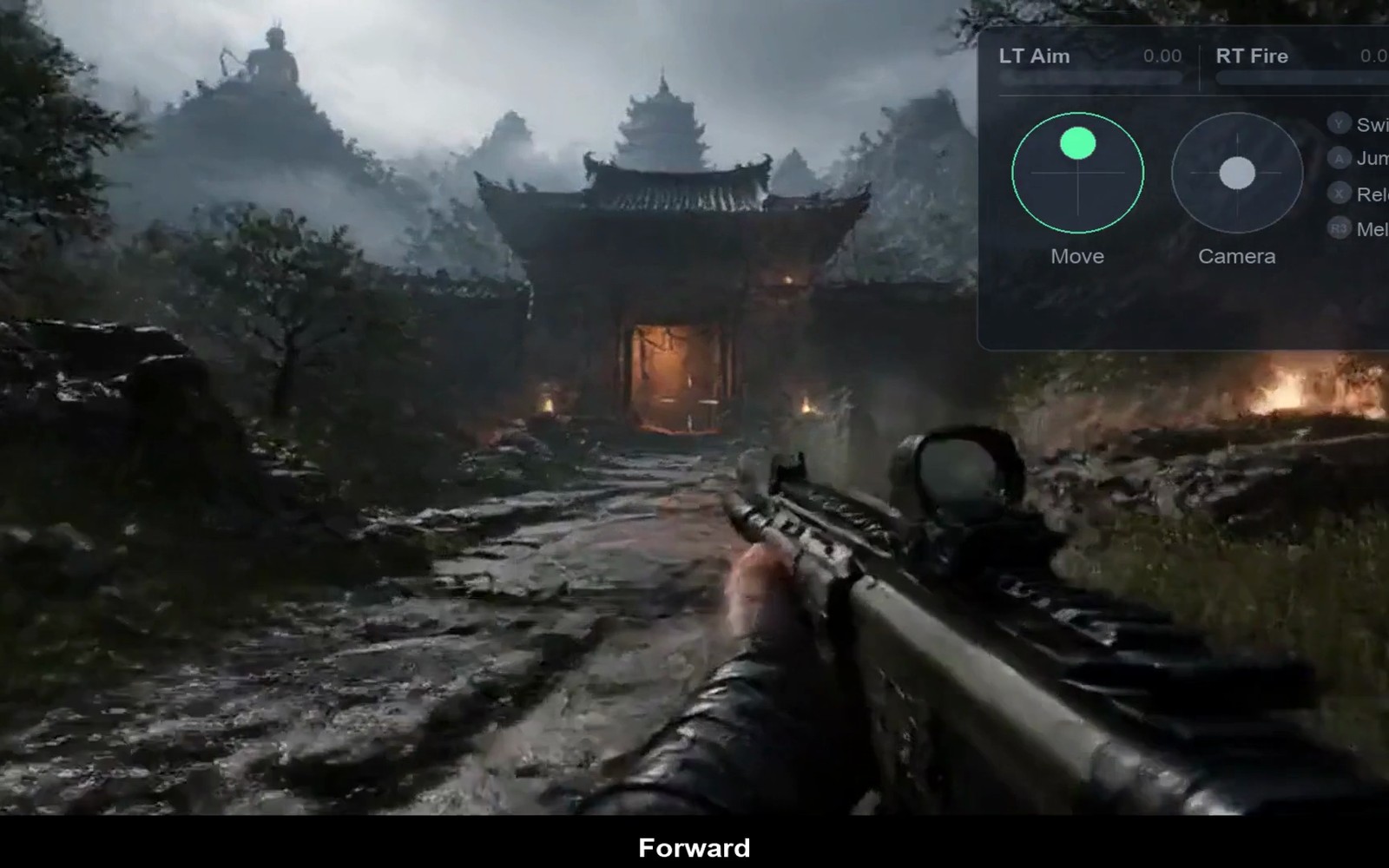}{(b) Action representation}{SCOPE, temple-gate scene}{The same controls in a different art style}\hfill
\blocktile{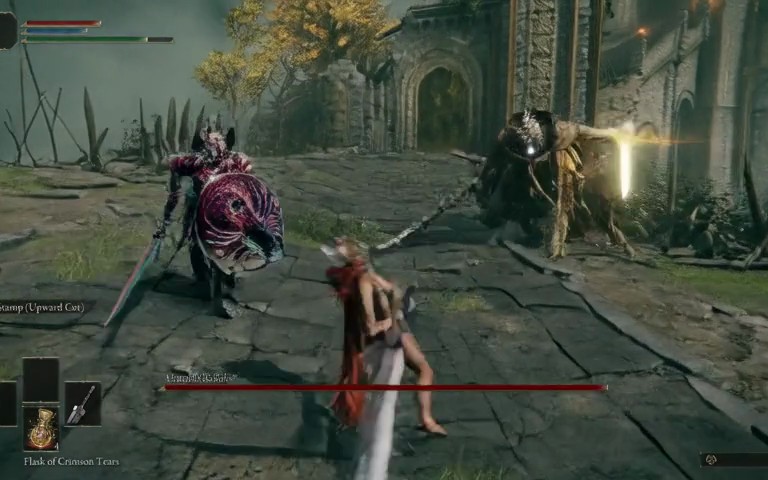}{(c) Language and entity control}{Incantation, Elden Ring}{Player and boss each given a text command}\\[3pt]
\blocktile{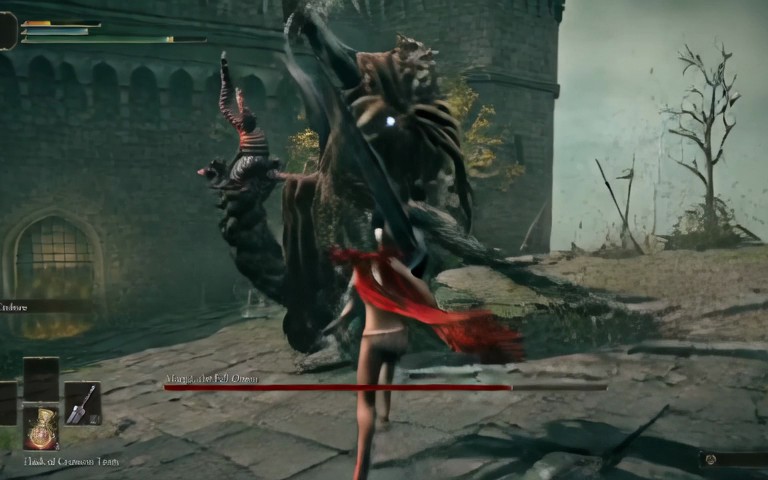}{(d) Language and entity control}{Incantation, Elden Ring}{Boss health bar visible during generated combat}\hfill
\blocktile{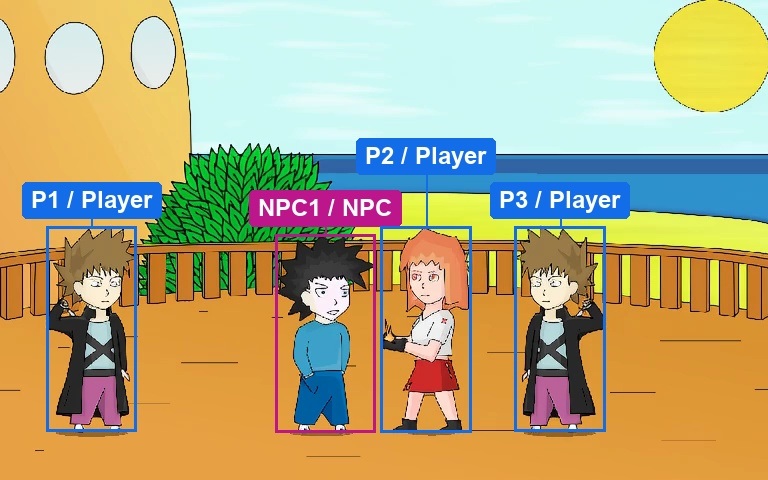}{(e) Player and NPC roles}{ReactiveGWM, Hikou no Mizu}{Three player-controlled characters\\and one rule-driven NPC}\hfill
\blocktile{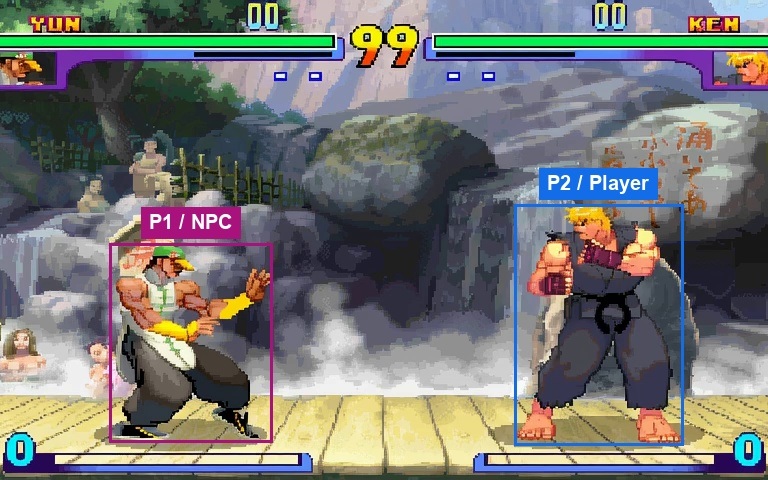}{(f) Player and NPC roles}{ReactiveGWM, Street Fighter III}{A player-controlled fighter and\\an NPC following if--then rules}
\caption{Action interfaces of generative simulators: device-action conditioning (a--b), natural-language entity commands (c--d), and per-character player and NPC roles (e--f): (a) SCOPE, open-world FPS \citep{tong2026scope}; (b) SCOPE, temple-gate scene \citep{tong2026scope}; (c) Incantation, Elden Ring \citep{zhu2026incantationnaturallanguageaction}; (d) Incantation, Elden Ring \citep{zhu2026incantationnaturallanguageaction}; (e) ReactiveGWM, Hikou no Mizu \citep{wang2026reactivegwmsteeringnpcreactive}; (f) ReactiveGWM, Street Fighter III \citep{wang2026reactivegwmsteeringnpcreactive}.}
\label{fig:strip-4-4}
\end{figure}

\runin{Action representation}\figmark{\ref{fig:strip-4-4}a--b} Action representation shapes both the scale of available data and the meaning of control. Native keyboard, mouse, gamepad, or telemetry action labels provide semantics that can be checked against the original game. GameFactory separates action control from visual appearance to transfer interactive dynamics across generated scenes \citep{yu2025gamefactory}, while WHAM interleaves visual and controller tokens to model both environmental change and human behavior \citep{kanervisto2025wham}. SCOPE trains on 65,557 of 69,000 clips from seven first-person shooters with aligned 10-DoF controls. It separates spatially local actions, such as firing, from global camera and movement signals, and its action control carries over to unseen visual scenes \citep{tong2026scope}. GameFactory, WHAM, and SCOPE depend on synchronized action data but preserve a direct mapping from user input to modeled transition.

Gameplay video is plentiful, but overlays can entangle scene dynamics with game-specific display conventions. Game2World Engine introduces a removal pipeline for heads-up-display elements, together with 96,000 paired synthetic videos and 1,079 in-the-wild clips from 303 games. In its controlled pilot, training on UI-free gameplay improves overall VideoReward by 6.83\% relative to UI-overlaid footage \citep{shen2026game2world}. That result supports scene-video preprocessing. Health, ammunition, and timers displayed in the interface may also be essential state observations, so a mechanics-oriented model may need to retain them as a separate channel.

Latent-action models draw on the much larger supply of ordinary video. Genie infers a discrete action between frames and trains an action-conditioned dynamics model without ground-truth controls \citep{bruce2024genie}. Repeated latent codes can reveal controllable structure within a domain. Their semantics arise from observed change, however, and a single code may combine camera movement, avatar motion, and scene dynamics. Latent actions are therefore effective for discovering interactive variation but less direct as a stable vocabulary for human control. ShadowDancer instead learns action latents from paired renderings that preserve motion while varying appearance. It uses demonstrated behavior to control a different rendering, including probes with previously unseen action assets \citep{cao2026shadowdancer}. This tests reuse of demonstrated dynamics, not discovery of an unfamiliar game's rules. A new scene, a new motion exemplar, and a new rule set expose different generalization problems.

\runin{Language and entity control}\figmark{\ref{fig:strip-4-4}c--f} Recent systems also use natural language as an action interface, beyond its use as a prompt for scene appearance. Incantation conditions latent frames on per-entity commands. On five held-out entity--action pairs, 89\% of its trials versus 43\% for an action-index baseline were rated as showing at least partial execution of the requested action, and 59\% versus 13\% as showing full execution. Its 90\% versus 0\% at-least-partial result on four out-of-vocabulary probes partly reflects interface expressiveness: the fixed-index baseline has no input slot for those prompts. Stable FVD over two-hour rollouts measures visual-distribution stability, not persistence of game state \citep{zhu2026incantationnaturallanguageaction}. ReactiveGWM instead lets users assign each character in a shared-view scene to player control or to an NPC driven by natural-language if--then rules, binding each role to a character through its first-frame instance mask, and trains a separate model for each of two fighting games \citep{wang2026reactivegwmsteeringnpcreactive}. Language composition and per-character role binding address different control bottlenecks within video generation.

H3-World examines how much language-conditioned control can be obtained by adapting a pretrained video model. It uses structured character and camera instructions, temporally aligned conditioning, and lightweight adaptation of a 33B backbone; attention routing separates control signals that might otherwise interfere \citep{chen2026h3world}. Its experiments use 7,872 training clips and 128 held-out clips, with action interventions that keep the initial observation and generation conditions fixed. This tests control rather than merely whether a prompt describes a plausible scene. The reported 124-frame generation with 50 denoising steps is a short-clip experiment, not evidence of real-time streaming or long-horizon mechanics.

Device, NPC-rule, and language interfaces distribute decisions differently. Device controls specify low-level input; an NPC rule delegates a sequence of actions; language can leave both realization and timing to the model. Entity-conditioned control adds an attribution problem: a command for one character should not alter another's motion unintentionally. ReactiveGWM, Incantation, and H3-World make different parts of this interface testable, complementing rather than replacing native-control evaluation.

\runin{Real-time generation and serving} Autoregression and diffusion operate at different levels in generative simulators. Genie \citep{bruce2024genie} and WHAM \citep{kanervisto2025wham} predict sequences of visual or action tokens autoregressively. GameNGen uses diffusion to generate the next visual observation and then feeds generated frames into subsequent predictions, so the rollout is autoregressive even though each frame is produced by diffusion \citep{valevski2024gamengen}. GameFactory similarly combines a video-diffusion prior with sequential continuation \citep{yu2025gamefactory}. Few-step generation, distillation, and streaming are used to reduce latency, while action semantics and persistent state remain open long-horizon problems \citep{he2025matrixgame2,forgewm2026}. Diffusion Forcing offers a general formulation for combining the two mechanisms: it trains a causal sequence model with independently chosen noise levels at different positions, supporting generation with partially noised history and variable prediction horizons \citep{chen2024diffusionforcing}. This illustrates why diffusion and autoregression are not mutually exclusive system categories. The former can specify how an observation or chunk is sampled, while the latter specifies how predictions are continued over time.

Yume-1.5 combines compressed history and linear attention with distilled streaming generation, adding textual event control to keyboard-based exploration \citep{mao2025yume15}. Its separation of navigation and event prompts illustrates why responsiveness involves more than camera tracking. Serving constraints then determine how much computation fits between input and response. Fewer denoising steps reduce work per frame. Chunked prediction amortizes computation across several frames but can delay response to a new command. Bounded caching reuses recent context while limiting the information retained for later revisits. A reported frame rate is therefore most informative alongside input latency, resolution, hardware, and the policy for interrupting or refreshing a generated chunk. Streaming-AAA shows how far serving speed can be pushed for one title: trained only on Genshin Impact, it generates 1080p video at 43 FPS with a computed latency of 198 ms that excludes network and codec overhead \citep{hu2026streamingaaa}.

AlayaRenderer-Flash uses a different real-time arrangement: a physics engine updates the game while a four-step streaming model renders RGB from synchronized geometry and material buffers. After target-game fine-tuning, the authors report 31.54~fps on one H200 and 30~fps for a live \emph{SuperTuxKart} integration \citep{lin2026rendererflash}. This demonstrates playable neural appearance over engine-executed rules; it does not show that the renderer learned those rules or transfers unchanged beyond the fine-tuned target game.

\runin{Long-rollout consistency} In autoregressive visual simulators, generated observations are usually fed into later predictions. Over time, a weakly represented fact can reappear as geometry drift, a duplicated object, a forgotten inventory item, or a changed rule. Multiplayer worlds add synchronization and action attribution across viewpoints, making shared-state failures visible from several perspectives \citep{savva2026solaris,hu2026multiplayerwm}. MultiGen produces synchronized generated viewpoints backed by an editable external map, but its multiplayer evaluation uses simulated Doom deathmatches on a single map, with no human use of the shared generative world \citep{po2026multigen}. For players, state failures can undermine navigation and trust in earlier choices. For agents, they can reverse the ranking of long-horizon actions. For policy training, they can teach behavior that fails when transferred back to the reference game.

Mitigation of this error accumulation has been studied mainly in video diffusion, where the train--test gap is explicit: a model trained to continue ground-truth frames must at inference continue its own imperfect outputs. CausVid distills a bidirectional diffusion transformer into a causal few-step generator, supervising the causal student with the bidirectional teacher, and reports reduced error accumulation in autoregressive generation \citep{yin2024slow}. Self Forcing trains on the model's own rollouts, conditioning each frame on previously self-generated output with key--value caching so that a video-level loss can score the whole sequence \citep{huang2025self}. FramePack adds drift-prevention methods, including early-established endpoints and adjusted sampling orders, alongside its context compression \citep{zhang2025frame}. None of the three reports results on game state. They address the visual drift that makes state failures visible.

A shared visual world requires more than each view being coherent separately. Two players may observe the same object at different times, act on it concurrently, or receive different observations of an event. Tests can therefore compare object identity, relative geometry, action attribution, and outcomes across synchronized views. An external map, as used by MultiGen, constrains one part of this agreement without specifying every hidden rule or score update. MASS separates a learned global-state transition model from per-camera rendering and evaluates both state accuracy and agreement between views in multiplayer Snake \citep{cai2026mass}. Unlike a conventional engine, its transition function is learned, so shared state prevents competing copies without guaranteeing correct rules.

MultiWorldBench measures the remaining gap against an engine reference in Minecraft: across 495 case configurations and ten capabilities, the best generated multiplayer world averages 21.39 against 91.69 for the engine, every generated system scores at most 8.00 on state persistence and 1.33 on structural consistency, and human pairwise judgments reproduce the automatic ranking \citep{xu2026multiworldbench}. Multics addresses the number of players on ten-player Counter-Strike~2 data. It exchanges information through player, group, and global tokens, and a model trained with four players and run with ten reaches cross-view consistency of 51.54\% against 47.82\% and 49.12\% for fully connected and centralized baselines \citep{huang2026multics}. The Esports World Model Dataset supplies data for this setting: about 4,880 player-view hours of human StarCraft~II, Dota~2, and Counter-Strike~2 play, part of it estimated for an ongoing extension, with each player's rendered view aligned to actions, engine state, and events \citep{ma2026esportswm}. Agreement across views still requires validation against reference transitions.

\begin{gameaiinsight}[Simulator]{What a model must capture depends on how it is used}
\begin{insightpoints}
\item \textbf{Prediction quality depends on the downstream task.} MuZero preserves information useful for planning without reconstructing pixels \citep{schrittwieser2020muzero}; interactive simulators must also produce observations \citep{valevski2024gamengen}. Dreamer tests the further question of whether imagined experience improves a policy in the reference game \citep{hafner2020dreamer}. A representation can succeed for one consumer and omit information another needs.
\item \textbf{Visual memory and game state record different things.} ReWorld retrieves spatial landmarks \citep{chen2026reworld}; StatePlay predicts selected numerical variables \citep{lin2026stateplaystateawaregameworld}. Recalling how a place looked cannot determine whether a reward was collected or a timer expired. Comparing these approaches points toward selective, updateable state alongside visual memory.
\item \textbf{Adaptation needs more than an accurate player predictor.} Maia4All validates individual move prediction \citep{tang2025maia4all}; personalized PCG evaluates the content players receive \citep{hafnar2025zeroshot}. Better prediction and better player experience are distinct outcomes, connected only when the inferred information changes a useful design decision.
\end{insightpoints}
\end{gameaiinsight}

\gameaisectionaccent{Creator}
\renewcommand{\dmcolor}{Creator}
\section{AI That Designs Games}
\label{sec:design}

\begin{figure}[H]
  \centering
  \includegraphics[width=\linewidth]{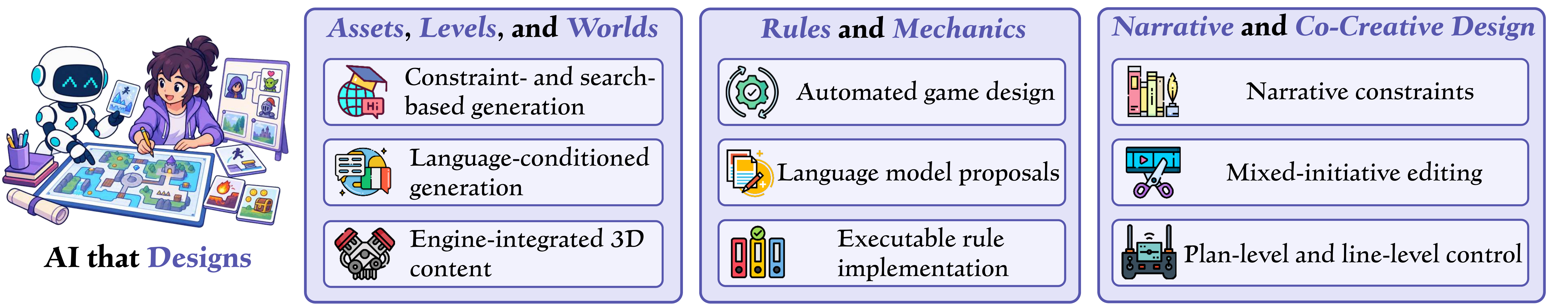}
  \caption{Research directions for AI that designs games: assets, levels, and worlds; rules and mechanics; and narrative and co-creative design.}
  \label{fig:design-directions}
\end{figure}

Playing and modeling start from a game that someone has already designed. Game design methods produce that game, generating and revising content, rules, and narrative structures under constraints set by designers. Earlier procedural systems already expressed intent through parameters, objectives, and direct editing \citep{togelius2011searchpcg,smith2011tanagra}. Foundation models add flexible language and multimodal interfaces for proposing and revising designs. Their contribution depends on how a request becomes a design representation and how candidate outputs are selected. A plausible level, a syntactically valid rule set, and a design that supports the intended play experience pose different generation and evaluation problems. The resulting methods combine learned proposals with search, explicit constraints, and human decisions (\Cref{fig:design-directions}).

\draftmap{\begin{tikzpicture}[x=1cm,y=1cm]
\node[dm q] at (-2.1,2.35) {How the chapter moves: the designed object becomes harder to validate by computation, so the human moves closer to the loop.};
\node[dm node=4.0cm] (a) at (0,0) {\dmhead{5.1 Content generation}\\[1pt]{\scriptsize validity computable; intent enters as prompt or constraint}};
\node[dm node=4.0cm] (b) at (5.5,0) {\dmhead{5.2 Rule and mechanic design}\\[1pt]{\scriptsize value is assessed through play}};
\node[dm node=4.0cm] (c) at (11.0,0) {\dmhead{5.3 Narrative and co-creative design}\\[1pt]{\scriptsize quality increasingly depends on human judgment}};
\draw[dm arrow] (a) -- (b); \draw[dm arrow] (b) -- (c);
\node[dm bridge=3.2cm] at (2.75,0.8) {a rule's effect depends on content and on play};
\node[dm bridge=3.2cm] at (8.25,0.8) {as computational checks become less complete, the human moves further into the loop};
\draw[dm axis] (-2.1,-1.15) -- (13.1,-1.15) node[midway,below=1pt,font=\scriptsize,text=GameSlate] {evaluation shifts from computable checks toward human judgment};
\node[dm frame,fit=(current bounding box)] (F) {};\dmtag
\end{tikzpicture}}

\subsection{Assets, Levels, and Worlds}
\label{sec:design-content}

\draftmap{\begin{tikzpicture}[x=1cm,y=1cm]
\node[dm q] at (-2.1,2.35) {Who guarantees that generated content is valid, and where does the designer's intent enter?};
\node[dm node=4.0cm] (a) at (0,0) {\dmhead{Constraint- and search-based generation}\\[1pt]{\scriptsize constraint solving, search-based PCG, PCGML, MarioGAN}};
\node[dm node=4.0cm] (b) at (5.5,0) {\dmhead{Language-conditioned generation}\\[1pt]{\scriptsize MarioGPT, Word2Minecraft, Moonshine}};
\node[dm node=4.0cm] (c) at (11.0,0) {\dmhead{Engine-compatible 3D generation}\\[1pt]{\scriptsize DreamCraft, database-driven 3D levels}};
\draw[dm arrow] (a) -- (b); \draw[dm arrow] (b) -- (c);
\node[dm bridge=3.2cm] at (2.75,0.8) {the designer's intent still needs a channel into the generator};
\node[dm bridge=3.2cm] at (8.25,0.8) {in 3D, valid also means the engine can use the asset};
\draw[dm axis] (-2.1,-1.15) -- (13.1,-1.15) node[midway,below=1pt,font=\scriptsize,text=GameSlate] {validity guaranteed by the generator, then by prompt plus checks, then by the engine};
\node[dm frame,fit=(current bounding box)] (F) {};\dmtag
\end{tikzpicture}}

Content generators must reconcile requested properties with the conditions for play. These conditions can be encoded in a construction procedure, checked during search, learned imperfectly from examples, or enforced through repair. Language conditioning changes how designers express requests; it can be combined with any of these mechanisms.

\begin{figure}[!htb]
\centering
\setlength{\tilew}{0.32\linewidth}
\setlength{\tilelabelh}{17.5mm}
\blocktile{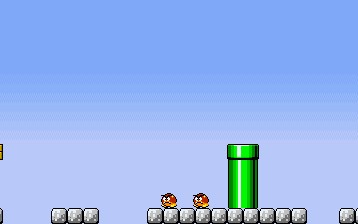}{(a) Constraint- and search-based generation}{MarioGAN, Super Mario Bros.}{A full level found by evolutionary search}\hfill
\blocktile{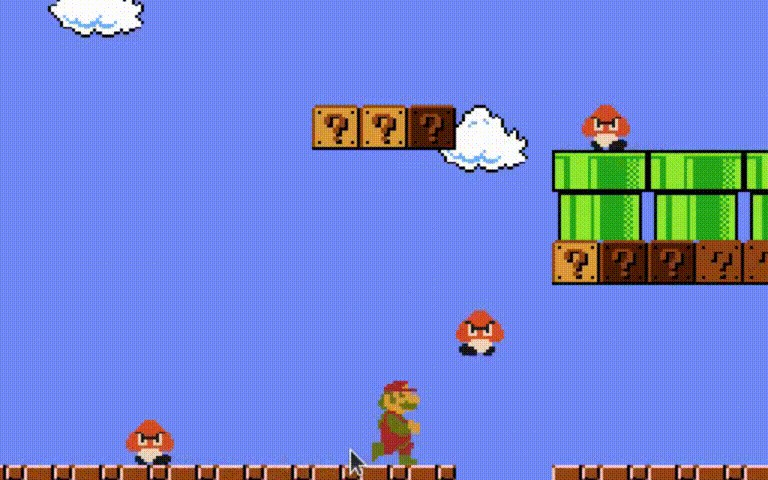}{(b) Language-conditioned generation}{MarioGPT, Super Mario Bros.}{A playable level written from a text prompt}\hfill
\blocktile{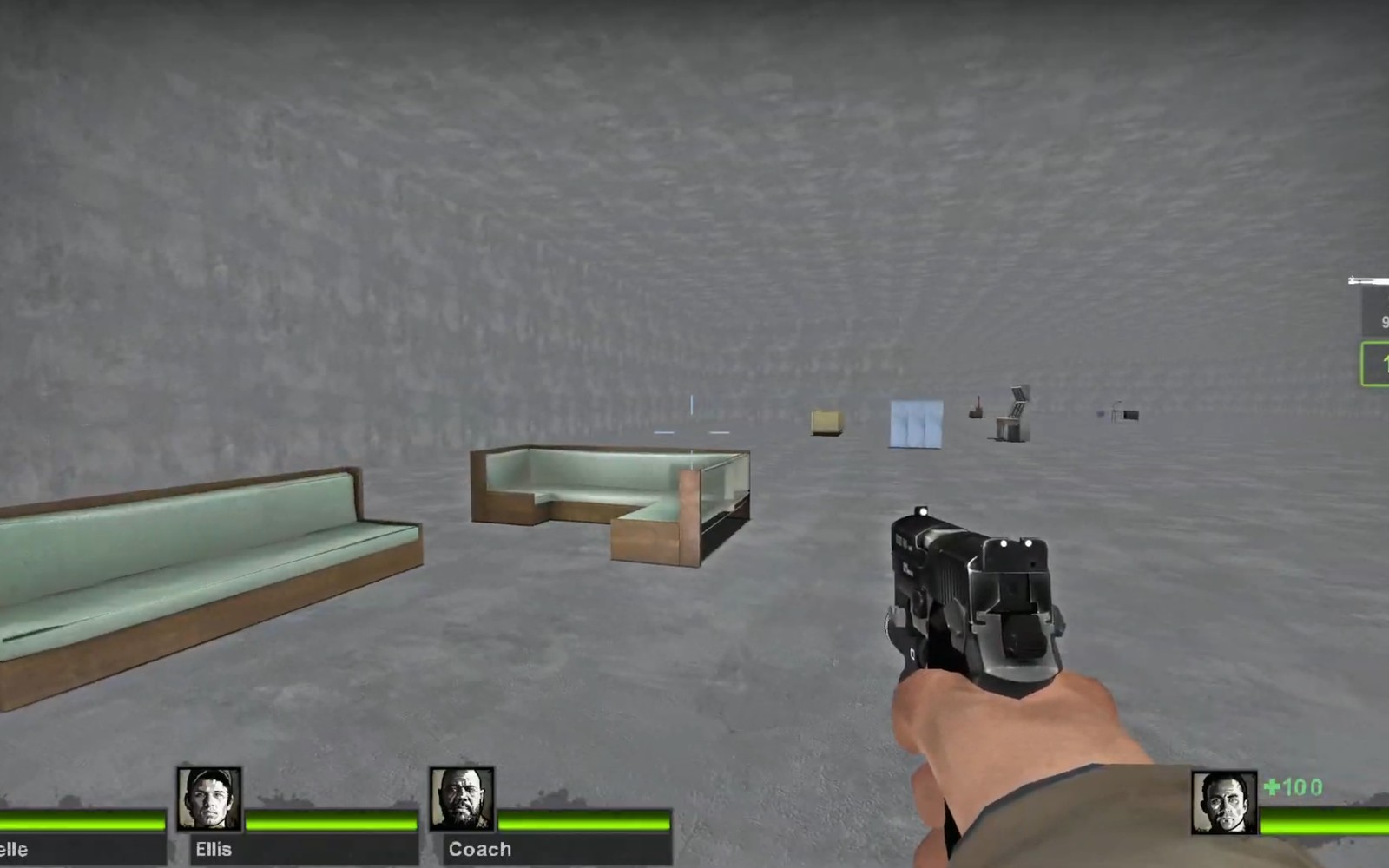}{(c) Engine-compatible 3D generation}{Xu et al., Left 4 Dead 2}{Generated room running\\inside a commercial engine}
\caption{Generated content that must remain playable: search-based (a), language-conditioned (b), and engine-compatible 3D generation (c): (a) MarioGAN, Super Mario Bros. \citep{volz2018mariogan}; (b) MarioGPT, Super Mario Bros. \citep{sudhakaran2023mariogpt}; (c) Xu et al., Left 4 Dead 2 \citep{xu2025database3d}.}
\label{fig:strip-5-1}
\end{figure}

\runin{Constraint- and search-based generation}\figmark{\ref{fig:strip-5-1}a} Much procedural content generation operates within an established game design, generating assets, levels, or worlds under fixed rules and goals. Constructive and constraint-based methods make validity constraints explicit \citep{compton2006procedural,smith2011asp}, search-based PCG explores candidates under quality functions \citep{togelius2011searchpcg}, and experience-driven PCG adapts content through player models \citep{yannakakis2011edpcg}. PCGML learns regularities from existing content \citep{summerville2018pcgml}, while PCGRL learns a sequential construction policy from environment feedback \citep{khalifa2020pcgrl}. Foundation models broaden the design interface: designers can express intent through language or multimodal examples and revise proposals iteratively.

The methods differ in where they obtain validity and variation. Constraint solving can exclude configurations that violate encoded rules, whereas search evaluates candidates and uses the result to propose replacements. Learned generators capture regularities in examples, which need not include every condition for successful play. Super Mario as a String represented levels as sequences for LSTM generation, establishing a route from sequence modeling to spatial content before pretrained language models \citep{summerville2016mario}. MarioGAN combines a learned design space with evolutionary search over structural properties and an A* agent \citep{volz2018mariogan}. Single-example diffusion instead learns reusable local patterns from one level through a restricted denoising receptive field \citep{dai2024singlediffusion}. Designer control comes from the representation and selection procedure as well as the model.

\runin{Language-conditioned generation}\figmark{\ref{fig:strip-5-1}b} MarioGPT fine-tunes a pretrained language model on 37 path-annotated Mario levels, with a separate text encoder conditioning generation on requested features. Its representation remains a game-specific tile vocabulary. Of 250 generated levels, 88.4\% were solved by an A* agent given up to five attempts per level. A separate 1,000-sample experiment measures adherence to prompts about pipes, enemies, blocks, and elevation \citep{sudhakaran2023mariogpt}. The two tests distinguish traversability from control over the design. Word2Minecraft instead converts structured stories into tile layouts and Minecraft block assignments, using A* and BFS-based checks and adjustments to preserve objective reachability \citep{huang2025word2minecraft}. Its 17-participant study finds higher story-coherence accuracy for objective-specific submaps than for main maps representing the whole story. An evolutionary baseline scores better on automated playability and path measures, but it directly optimizes objective distances and leaves story coherence unmeasured. These path-based measures do not evaluate narrative quality.

Sokoban exposes a different constraint: pushing a box can irreversibly destroy a solution. Todd et al. fine-tune GPT-2 on serialized levels and compare natural-language, code, and randomly initialized models. With abundant Boxoban data, pretraining brings little benefit; with small human-authored datasets, novelty is difficult to retain. Controlling empty-space proportion is also easier than controlling solution length, which requires solving the puzzle \citep{todd2023level}. Practical PCG studies the small-data problem in Metavoidal, using 60 authored rooms and human-in-the-loop fine-tuning to produce new rooms under project-specific constraints \citep{nasir2023practical}. These cases distinguish learning spatial regularities from learning what makes a level function.

Language models can also supervise a generator without producing its final output. Moonshine labels maps from a constructive generator and trains steerable generative models on the resulting pairs \citep{nie2025moonshine}. This separates semantic annotation from content synthesis. MarioDiffusion derives captions from tile properties and compares pretrained text encoders with a small Transformer trained for those captions. Its best diffusion configuration uses the small encoder, with prompt adherence, diversity, and agent-tested playability measured separately \citep{schrum2025mariodiffusion}. Together with the Sokoban comparison, this locates the benefit of pretraining in the information it contributes beyond the available game data, rather than in model size alone.

The ChatGPT4PCG competitions test language control under physics rather than traversal alone \citep{taveekitworachai2023chatgpt4pcg}. Prompts generate Science Birds structures resembling requested letters; simulation measures stability, while a classifier measures character similarity. The second edition adds diversity and permits programs that orchestrate prompting \citep{taveekitworachai2024chatgpt4pcg2}, and LLMs4PCG broadens the model choices \citep{llms4pcg2025}. A recognizable structure can collapse, just as a stable one can miss the requested shape. These are distinct checks, neither of which measures the enjoyment of a complete level.

Multiverse extends language-conditioned design across four games through aligned text--level embeddings and a conditional VQ-VAE. It abstracts game-specific entity names into shared categories and uses contrastive learning to relate descriptions across domains \citep{baek2026multiverse}. Reported blending gains are strongest within a genre; balanced cross-genre blends remain harder. Evaluation measures structural and semantic similarity rather than execution under a new combined rule set. Shared descriptions support content recombination before mechanics transfer.

\runin{Conversational editing and generator control} Another route keeps an existing generator or editable world behind a language interface. Whitehead et al.\ compare function calls with direct manipulation of a world representation: tool calls restrict changes to implemented operations, while direct editing permits broader changes but makes consistency harder to maintain \citep{whitehead2025conversational}. Their TinyTownQA experiments also show that the representation supplied to the model matters: structured world facts can support more accurate answers to questions about a scene than images or tile arrays alone. These are world-understanding tests, distinct from successful multistep editing. The practical comparison with direct generation is whether a request selects a known operation, revises a local region, or creates a new artifact. Preserving the rest of a design is part of the task, not an incidental consequence of generating a plausible replacement.

\runin{Engine-compatible 3D generation}\figmark{\ref{fig:strip-5-1}c} For 3D content, the choice of representation determines whether a visual prior can respect the game's construction rules. DreamCraft optimizes a quantized neural radiance field whose outputs map directly to Minecraft blocks. It combines a pretrained text-to-image prior with losses for block distributions and adjacency constraints, improving in-game text alignment over post-hoc conversion of unconstrained 3D output \citep{earle2024dreamcraft}. A database-driven approach places the language model earlier in the workflow: it constructs libraries of rooms, facilities, and mechanics that the authors review before topology optimization and repair assemble levels. Of 6,000 generation attempts, 95.47\% are validated as successfully repaired and free of subsequent anomalies in the Unity simulation, and the study also imports layouts into Left 4 Dead 2 \citep{xu2025database3d}. This result measures the combined curated-library and repair pipeline in one survival-horror setting, and says nothing about unassisted language-model generation. Both approaches make pretrained visual or semantic knowledge useful through an explicit, game-specific representation.

A central integration difficulty is that appearance alone does not determine an asset's role in play. A visually plausible wall must also have suitable collision, scale, and placement, and a room arrangement must preserve traversable connections and objectives. DreamCraft constrains the generated representation itself, while database-driven generation assembles reviewed components and then repairs their arrangement. The comparison suggests two practical routes for visual and semantic priors: generate directly within an engine-compatible vocabulary, or place a structured intermediate representation between generation and the engine.

\runin{Functional assets and compositional worlds} Asset generation also needs editable internal structure. CubePart conditions generation on a user-defined list of semantic parts and produces separate meshes that assemble into one object. Its engine examples attach animation and behavior scripts to those parts, such as vehicle components \citep{zhu2026cubepart}. The supplied schema and scripts explain how geometric generation becomes interactive; the model does not infer every physical parameter or behavior. At scene scale, WorldSculpt adapts a single-object prior to posed views and reconstructs individually addressable meshes, including from generated 3D Gaussian-splatting worlds. Its Unreal-derived benchmark evaluates reconstruction, not completed gameplay \citep{niu2026worldsculpt}. Object-level structure makes later editing possible, while collision, traversal, objectives, and inter-object behavior still require implementation and tests.

\Needspace{4\baselineskip}
Speech is another game asset whose content and delivery have different authoring requirements. Ghostwriter drafts dialogue text for writer selection \citep{barth2023ghostwriter}; a speech synthesizer produces its audible performance, and audio-driven animation can synchronize a character's face with an approved recording or synthetic track \citep{nvidia2025audio2faceplugin}. Producing these assets before release allows inspection and revision. Producing a response during play adds turn-taking and latency constraints, discussed in \Cref{sec:runtime-content}. In both cases, the relevant artifact is the integrated character performance rather than an isolated audio sample.

\Cref{tab:design-methods} compares these representation choices with the rule and co-creative methods discussed next. The useful contrast is where a proposal becomes constrained: in the output vocabulary, in executable checks, or through a designer's revision.
\begin{table}[htbp]
\centering
\caption{Design methods compared by output, feedback, and supplied game structure.}
\label{tab:design-methods}
\begingroup
\setcitestyle{numbers,square,comma}
\footnotesize
\setlength{\tabcolsep}{4pt}
\begin{tabularx}{\textwidth}{@{}L{3.30cm}L{3.30cm}Y L{3.85cm}@{}}
\toprule
\textbf{System} & \textbf{Generated object} & \textbf{Feedback / checks} & \textbf{Supplied structure} \\
\midrule
MarioGPT~\citep{sudhakaran2023mariogpt} & tile sequence & A*; prompt features & tiles; movement~rules \\
DreamCraft~\citep{earle2024dreamcraft} & block-aligned 3D scene & text alignment; adjacency~losses & Minecraft block~vocabulary \\
CubePart~\citep{zhu2026cubepart} & part-based 3D asset & part control; geometry & schemas; external~behavior \\
GAVEL~\citep{todd2024gavel} & Ludii rule program & compile; simulated play; diversity~search & game-description language \\
ScriptDoctor~\citep{earle2025scriptdoctor} & rules; tile levels & compiler; BFS; iterative~repair & DSL; engine; solver~budget \\
KNUDGE~\citep{weir2024knudge} & branching dialogue & lore and quest~consistency & lore; quest goals \\
DreamGarden~\citep{earle2025dreamgarden} & plans and scenes & compile; visual checks; user~edits & Unreal; designer~input \\
\bottomrule
\end{tabularx}
\endgroup
\end{table}

\subsection{Rules and Mechanics}
\label{sec:design-rules}

\draftmap{\begin{tikzpicture}[x=1cm,y=1cm]
\node[dm q] at (-2.1,2.35) {A rule's value shows only in play, so how is a proposed rule judged, and what makes it real?};
\node[dm node=4.0cm] (a) at (0,0) {\dmhead{Automated game design}\\[1pt]{\scriptsize ANGELINA, Mechanic Miner, Puck}};
\node[dm node=4.0cm] (b) at (5.5,0) {\dmhead{Language-model rule and mechanic proposals}\\[1pt]{\scriptsize VGDL generation, GAVEL, Mortar}};
\node[dm node=4.0cm] (c) at (11.0,0) {\dmhead{Executable rule implementation}\\[1pt]{\scriptsize STORY2GAME}};
\draw[dm arrow] (a) -- (b); \draw[dm arrow] (b) -- (c);
\node[dm bridge=3.2cm] at (2.75,0.8) {language widens the proposal space, not the judge};
\node[dm bridge=3.2cm] at (8.25,0.8) {a fluent rule still needs a state model and working code};
\draw[dm axis] (-2.1,-1.15) -- (13.1,-1.15) node[midway,below=1pt,font=\scriptsize,text=GameSlate] {judge $\to$ propose $\to$ implement};
\node[dm frame,fit=(current bounding box)] (F) {};\dmtag
\end{tikzpicture}}

Generating rules changes the space of possible actions and outcomes, so evaluating a proposal requires considering its interaction with goals, layouts, and player strategies. Current approaches differ in the rule language they expose, how they search for candidates, and how simulated play or human judgment selects among the resulting candidates.

\runin{Automated game design} Automated game design extends the generative boundary beyond content alone to rules, mechanics, and combinations of game components. Evolutionary search, constraint solving, common-sense reasoning, and planning have explored designs whose quality emerges through the behavior they permit \citep{nelson2007automated,browne2010evolutionary,zook2014mechanics}. Systems such as ANGELINA explored the coordinated generation of multiple game components and, eventually, complete small games \citep{cook2017angelina1}. Mechanic Miner searched for novel mechanics together with accompanying levels \citep{cook2013mechanicminer}, while Puck later emphasized continuous, user-oriented generation and testing \citep{cook2022puck}. Language generation can broaden the proposal space, while executable representations and simulated play provide evidence about how a proposal changes play.

Rule representations determine which designs a search can reach. A fixed grammar makes parsing and simulation tractable but excludes mechanics outside its vocabulary. More flexible code can express new behavior, at the cost of a harder validation problem. Joint mechanic--level search matters because a rule may have little effect in one layout and fundamentally change strategy in another. These earlier systems expose two choices that remain in foundation-model approaches: the unit of variation and the process used to judge it.

\runin{Language-model rule and mechanic proposals} Direct generation and search-based proposal use the model differently. In VGDL generation, prompts supply grammar, mappings, and examples, and the generated rule and level descriptions are checked for parsability, required interactions, termination conditions, and sprite mappings \citep{hu2024gamegeneration}. This tests whether language knowledge can be made compatible with an executable description language. The checks establish specific structural properties; they do not select for every quality of play. GAVEL instead fine-tunes a code model to mutate and recombine Ludii descriptions within quality-diversity search. Compilation and simulated play screen proposals, while rule-concept descriptors organize the archive \citep{todd2024gavel}. The model supplies variation; the interpreter and archive define what can execute and which differences are retained. This can preserve alternatives that a single-objective search would discard, although archive coverage is diversity under the chosen descriptors, not a general measure of design originality beyond those descriptors.

Mortar varies reusable Python mechanics and composes them into games through tree search \citep{nasir2025mortar}. A mechanic is evaluated through its contribution to games that distinguish a fixed ordering of stronger and weaker agents. Thus, the search unit moves from a complete rule description to a component whose value depends on its partners. Composition search and language-model variation play different roles: one selects combinations, while the other proposes new functions. This distinction matters when attributing improvements to the model rather than to the search procedure.

\runin{Checking designs through implementation and play} Validating a proposed mechanic requires both an adequate state representation and an implementation that realizes its effects. STORY2GAME derives a text game's state schema and action logic from story-event preconditions and effects, then generates executable action code \citep{zhou2025story2game}. The resulting code must express the intended state changes; compilation and semantic results are examined in \Cref{sec:build-repair}. An action that illuminates a forest, for example, requires illumination to be represented in the state. Compilation alone cannot reveal that omission. ScriptDoctor generates PuzzleScript games with human examples, grammar checks, compiler feedback, and breadth-first-search playtesting. It allows at most ten code-generation attempts in total, accepting a game when every level has a discovered solution longer than ten moves \citep{earle2025scriptdoctor}. Compilation and solver acceptance diverge in its experiments: additional examples can improve syntax without producing sufficiently nontrivial solvable levels. Unlike GAVEL's rule-concept archive or Mortar's skill ordering, this procedure selects by solution existence and length. Each objective favors different designs; an interesting short puzzle can be rejected by a length threshold.

Play can also drive revision directly. In PlayCraft, an agent that sees only rendered frames plays ten browser games, and a second agent revises the code from that play over three iterations. It aims for performance goals at each proficiency level and for meaningful chances to improve. In fresh evaluation runs, the player agent's learning gain between early and late episodes rises from 12.4 to 34.3 points at the highest player setting, and whether this gain tracks human engagement is untested \citep{cheng2026playcraft}. Recursive Game Creator adds a reviewer to the same loop: a designer, a builder, a player that writes executable policies, and a reviewer refine each game over three rounds, raising the GameCraft-Bench score on 45 tasks from 72.70 to 77.89 \citep{chen2026recursivegame}.

Selection introduces a different source of error. Mortar's 14-participant study of three game pairs agrees with its agent-based ordering in two pairs and reverses it in the third \citep{nasir2025mortar}. Simulated play is useful for screening, but its preference depends on the agents and objectives used. Reporting the proposal, repair, and simulation budgets helps distinguish a reliable generator from a search procedure that eventually finds an acceptable design.

\subsection{Narrative and Co-Creative Design}
\label{sec:design-narrative}

\draftmap{\begin{tikzpicture}[x=1cm,y=1cm]
\node[dm q] at (-2.1,2.35) {What must the generator keep intact, and at what granularity does the human take over?};
\node[dm node=4.0cm] (a) at (0,0) {\dmhead{Narrative constraints}\\[1pt]{\scriptsize KNUDGE, SceneCraft, NarrativeGenie}};
\node[dm node=4.0cm] (b) at (5.5,0) {\dmhead{Mixed-initiative editing}\\[1pt]{\scriptsize Tanagra, Sentient Sketchbook, LLM co-evolution}};
\node[dm node=4.0cm] (c) at (11.0,0) {\dmhead{Granularity of creator control}\\[1pt]{\scriptsize DreamGarden, Ghostwriter}};
\draw[dm arrow] (a) -- (b); \draw[dm arrow] (b) -- (c);
\node[dm bridge=3.2cm] at (2.75,0.8) {constraints fix what must hold; the designer still needs control over the result};
\node[dm bridge=3.2cm] at (8.25,0.8) {as more is generated at once, control moves to the plan or to the line};
\draw[dm axis] (-2.1,-1.15) -- (13.1,-1.15) node[midway,below=1pt,font=\scriptsize,text=GameSlate] {where the human intervenes: in the constraints, on the output, on the plan};
\node[dm frame,fit=(current bounding box)] (F) {};\dmtag
\end{tikzpicture}}

Narrative authoring combines structural requirements with judgments about pacing, characterization, and player choice. Co-creative tools address a related but broader question: how a designer can inspect and revise generated material, whether it is a dialogue tree, a level, or a mechanic. The important comparison is the control available over intermediate decisions, not simply the amount of content generated.

\begin{figure}[!htb]
\centering
\setlength{\tilew}{0.47\linewidth}
\setlength{\tilelabelh}{14.0mm}
\blocktile{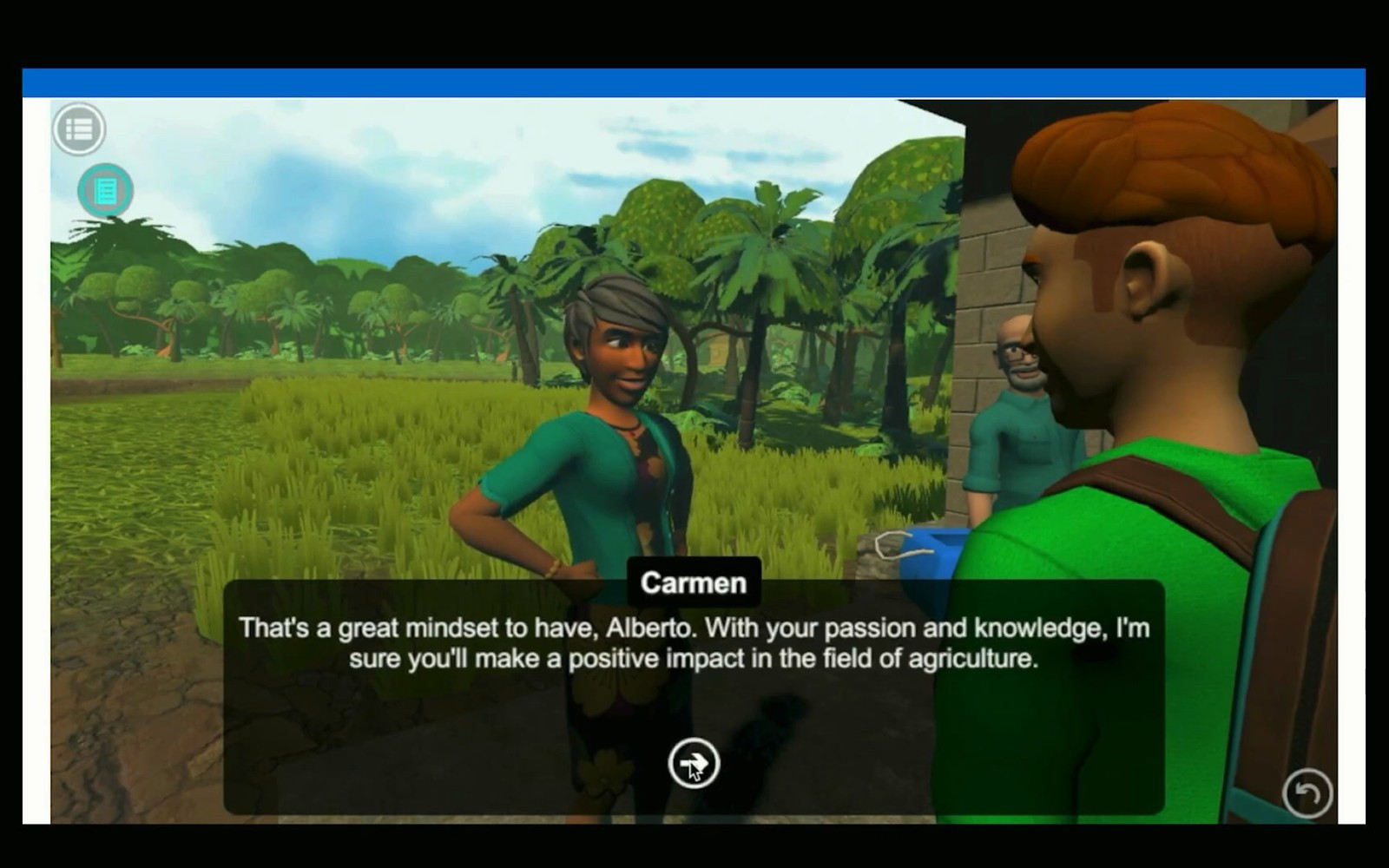}{(a) Narrative constraints}{NarrativeGenie, Unity episode}{NPC line generated within authored scene goals}\hfill
\blocktile{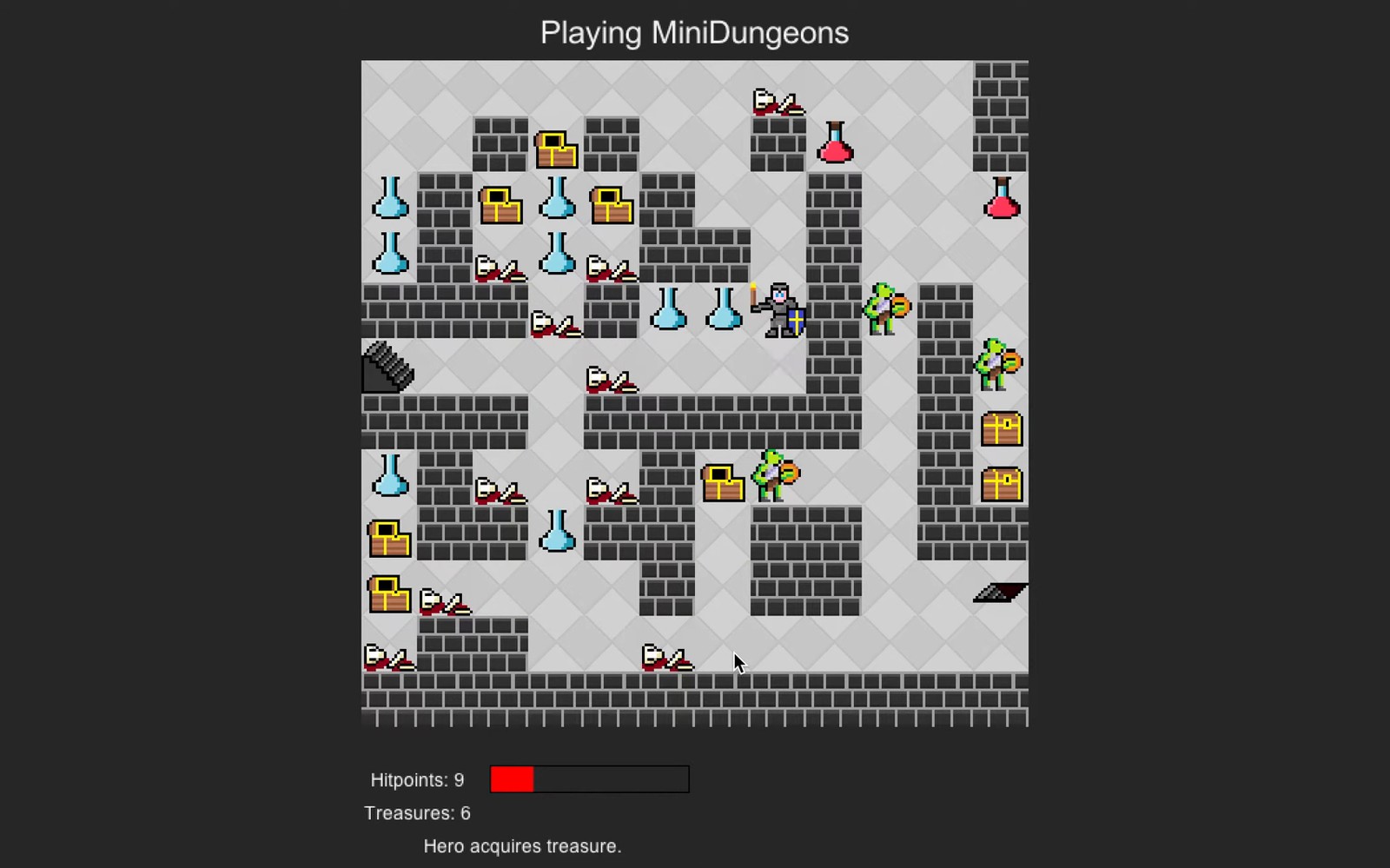}{(b) Mixed-initiative editing}{Sentient Sketchbook, MiniDungeons}{Persona-based playtesting in later work}
\caption{Narrative generated under authored constraints (a) and MiniDungeons, a testbed for persona-based playtesting (b): (a) NarrativeGenie, Unity episode \citep{kumaran2024narrativegenie}; (b) MiniDungeons, whose persona-training levels were designed in a mixed-initiative fashion \citep{liapis2013sketchbook,liapis2015personacritics}.}
\label{fig:strip-5-3}
\end{figure}

\runin{Narrative constraints}\figmark{\ref{fig:strip-5-3}a} Narrative authoring must coordinate what a character says with what a player can learn or do. Early GPT-2 fine-tuning on annotated RPG quests learned the format and language of quest-giving dialogue, evaluated as generated text for authoring \citep{vanstegeren2021quests}. KNUDGE adds branching quest dialogues from \emph{The Outer Worlds}: models generate trees under quest specifications, character facts, and lore, comparing supervised and in-context generation \citep{weir2024knudge}. This changes the target from a plausible passage to coordinated branches. A locally fluent reply can reveal a quest fact too early or conflict with information on another branch; neither task validates quest execution.

SceneCraft turns author-specified scene objectives, character traits, and variations into branching scripts for an existing game framework \citep{kumaran2023scenecraft}. NarrativeGenie constructs a partially ordered graph of story beats and a playable episode during authoring, before its runtime Adaptive Dialogue Manager takes over \citep{kumaran2024narrativegenie}. KNUDGE, SceneCraft, and NarrativeGenie differ in the structure supplied to generation: factual constraints, scene templates, and progression graphs. The comparison concerns how a branch preserves required information and remains compatible with other branches. Runtime selection and adaptation are discussed in \Cref{sec:adapt}.

\runin{Coordinating narrative, space, and mechanics} RPGAgent connects story outlines to a Unity prototype through agents for narrative, scenes, mechanics, and code. Structured intermediate outputs let designers review and edit decisions between stages. Its counterbalanced study with 18 participants reports higher overall user-experience and creativity-support scores than a GPT-assisted baseline, with both conditions using the same art assets and Unity Tilemap \citep{zhang2026rpgagent}. The study concerns a supported prototyping workflow, not fully autonomous project creation. Together with NarrativeGenie's beat graph \citep{kumaran2024narrativegenie} and DreamGarden's editable plan tree \citep{earle2025dreamgarden}, this illustrates a recurring design choice: intermediate structures let later generation use earlier decisions while giving the author a place to correct them. The important question is whether changes propagate coherently, rather than whether each component is plausible on its own.

\runin{Mixed-initiative editing}\figmark{\ref{fig:strip-5-3}b} Mixed-initiative tools preserve direct designer control while automating selected parts of construction or search. Tanagra responds to edits while maintaining structural constraints in platform-game levels \citep{smith2011tanagra}. Sentient Sketchbook couples map sketching with playability checks, gameplay-property evaluation, and alternative suggestions, and its initial evaluation includes a small study with industry experts \citep{liapis2013sketchbook}. In both cases, the designer can change the artifact directly, without repeatedly describing the desired result from scratch. Lanzi and Loiacono combine interactive evolution with large language models: designer feedback selects candidates while the model recombines and mutates proposals, evaluated across three design tasks \citep{lanzi2023llmco}. For 3D obstacle-course games, constrained solvability queries return sequences of player actions that satisfy designer-specified constraints, such as avoiding a region, passing a waypoint, or performing only two jumps; GPU-accelerated exploration guided by an offline-trained RL solver keeps the feedback fast enough for a design loop, and human playtesting confirms that players solve the resulting obstacles in the ways the designers intended \citep{majercik2026solvability}. Direct editing and candidate selection offer different forms of control. Editing can retain a preferred structure while changing a local detail; selection can reveal alternatives that the designer would not have specified. Neither interaction is captured by measuring the quality of a single generated artifact. Useful comparisons examine revision effort, inspectable decisions, and control over which parts remain unchanged.

CoBG extends co-creation to board-game rulebooks through separate ideation, realization, critique, and simulated-persona feedback modules. Evaluations on 207 held-out board games, a 30-participant creator study, and paper-prototype playtests assess rulebook quality and authoring support \citep{li2026autobg}. In contrast to executable rule search, revisions are screened by a learned critic and audience feedback is simulated. This can expose ambiguities and suggest alternatives. In the playtests, groups of three new players met fewer rule breakdowns with CoBG rulebooks than with GPT-5.4 rulebooks (1.2 vs.\ 3.0 per session), while balance and long-term enjoyment remain untested \citep{li2026autobg}. Its connection to MeepleLM's subjective playtester \citep{li2026meeplelm} is the use of player-oriented language feedback, rather than an engine supplying legal trajectories.

\runin{Granularity of creator control} DreamGarden connects co-creative design to implementation through an editable plan tree. Designers prune, expand, and annotate the hierarchy while specialist modules generate assets and C++ actors. In its 10-participant study, users valued seeing how the system interpreted prompts, but waiting for implementation and lacking intermediate inspection limited control \citep{earle2025dreamgarden}. The plan is therefore useful not simply as a decomposition for agents, but as a representation a designer can correct before expensive implementation completes.

Ubisoft's Ghostwriter illustrates a narrower industrial workflow: it drafts NPC barks from character and situation descriptions for writers to select and edit \citep{barth2023ghostwriter}. The official account documents authoring support rather than autonomous dialogue deployment. SPINE works at the level of design intent, using language models to help formulate design pillars and discuss whether proposed features fit them. Its evidence includes a game-jam application and interviews with four practitioners \citep{geheeb2026spine}. These systems place human decisions at different granularities: a principle, an overall plan, a candidate design, or an individual line. Their usefulness depends on whether those controls match the author's decisions and make correction manageable. The design entries in Appendix~\ref{app:resources}, \Cref{tab:development-systems}, compare these authoring mechanisms alongside artifact scope and feedback.

\begin{gameaiinsight}[Creator]{Richer proposals still need a way to judge quality}
\begin{insightpoints}
\item \textbf{Pretraining helps when it supplies knowledge the design data lack.} Todd et al.'s Sokoban experiments \citep{todd2023level} and MarioDiffusion \citep{schrum2025mariodiffusion} show that broad pretraining does not automatically improve performance in their restricted representations. The useful question is which new descriptions, constraints, or game vocabularies require that broader knowledge.
\item \textbf{Selection depends on what the evaluator rewards.} GAVEL \citep{todd2024gavel} and Mortar explore richer rule spaces through simulation-based selection. Mortar's small player study includes a preference reversal relative to its agent-based ordering \citep{nasir2025mortar}. Increasing proposal diversity therefore does not, by itself, broaden the kinds of play that the selection objective rewards.
\item \textbf{Control depends on what the designer can edit.} Sentient Sketchbook exposes map edits \citep{liapis2013sketchbook}; DreamGarden exposes a plan hierarchy \citep{earle2025dreamgarden}. Their comparison shifts attention from prompt expressiveness to the cost of inspecting, correcting, and retaining a design decision as generation proceeds.
\end{insightpoints}
\end{gameaiinsight}

\gameaisectionaccent{Builder}
\renewcommand{\dmcolor}{Builder}
\section{AI That Builds and Maintains Games}
\label{sec:build}

\begin{figure}[H]
  \centering
  \includegraphics[width=\linewidth]{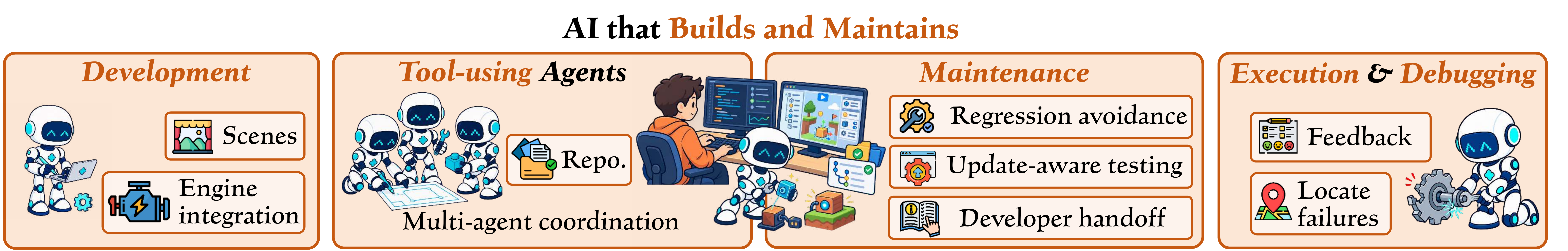}
  \caption{Research directions for AI that builds and maintains games: code, scenes, and engine projects; tool-using development agents; execution, debugging, and repair; and revision, maintenance, and handoff.}
  \label{fig:build-directions}
\end{figure}

A design becomes a game only once it runs. A game project can contain plausible code and attractive scenes yet fail when the pieces interact. Scripts depend on object bindings, engine callbacks, assets, and state that may be spread across files and editor settings. Foundation-model development agents bring code knowledge and language-based planning to this integration work, using repository tools, editor operations, and execution feedback to make and check changes \citep{yang2025unigen,yin2026autoue,huang2026guigames,li2026worldtestsuite}. Their methods differ in how they represent the project, specialize the model, and turn observed failures into repairs. Building a first prototype and revising an existing project expose different parts of this problem; maintenance additionally requires preserving behavior that a new request is intended to leave unchanged (\Cref{fig:build-directions}).

\draftmap{\begin{tikzpicture}[x=1cm,y=1cm]
\node[dm q] at (-1.65,2.35) {How the chapter moves: the horizon of the build lengthens, from a first run to a project someone else can continue.};
\node[dm node=3.3cm] (a) at (0,0) {\dmhead{6.1 Project construction}\\[1pt]{\scriptsize code, scene, and assets bound}};
\node[dm node=3.3cm] (b) at (4.3,0) {\dmhead{6.2 Tool-using agents}\\[1pt]{\scriptsize repository, editor, running game}};
\node[dm node=3.3cm] (c) at (8.6,0) {\dmhead{6.3 Execution and repair}\\[1pt]{\scriptsize symptom $\to$ stage $\to$ fix}};
\node[dm node=3.3cm] (d) at (12.9,0) {\dmhead{6.4 Maintenance and handoff}\\[1pt]{\scriptsize edits, regressions, handoff}};
\draw[dm arrow] (a) -- (b); \draw[dm arrow] (b) -- (c); \draw[dm arrow] (c) -- (d);
\node[dm bridge=2.7cm] at (2.15,0.8) {binding code to scene needs access to the project};
\node[dm bridge=2.7cm] at (6.45,0.8) {access shows the project; running it shows the failures};
\node[dm bridge=2.7cm] at (10.75,0.8) {one fix is not a maintained project};
\node[dm frame,fit=(current bounding box)] (F) {};\dmtag
\end{tikzpicture}}

\subsection{Code, Scenes, and Engine Projects}
\label{sec:build-projects}

\draftmap{\begin{tikzpicture}[x=1cm,y=1cm]
\node[dm q] at (-2.1,2.35) {What does ``runs in the engine'' require beyond correct code?};
\node[dm node=4.6cm] (a) at (0,0) {\dmhead{Scene construction and engine integration}\\[1pt]{\scriptsize UnrealLLM, UniGen, AutoUE, GameCraft-Bench}};
\node[dm node=4.6cm] (b) at (6.4,0) {\dmhead{Executable environment generation}\\[1pt]{\scriptsize Game Code World Models, Agent2World}};
\draw[dm arrow] (a) -- (b);
\node[dm bridge=3.6cm] at (3.2,0.8) {the same binding problem returns when the consumer is another agent};
\draw[dm axis] (-2.4,-1.15) -- (8.8,-1.15) node[midway,below=1pt,font=\scriptsize,text=GameSlate] {from a project a player runs to a project another agent runs};
\node[dm frame,fit=(current bounding box)] (F) {};\dmtag
\end{tikzpicture}}

Project construction requires agreement between code, scenes, assets, and engine configuration. A generator may work through structured graphs, component libraries, or source files, each exposing a different set of dependencies. Executable environments add a related requirement: observations, actions, rewards, and resets must implement the interface expected by a consuming agent.

\begin{figure}[!htb]
\centering
\setlength{\tilew}{0.47\linewidth}
\setlength{\tilelabelh}{14.0mm}
\blocktile{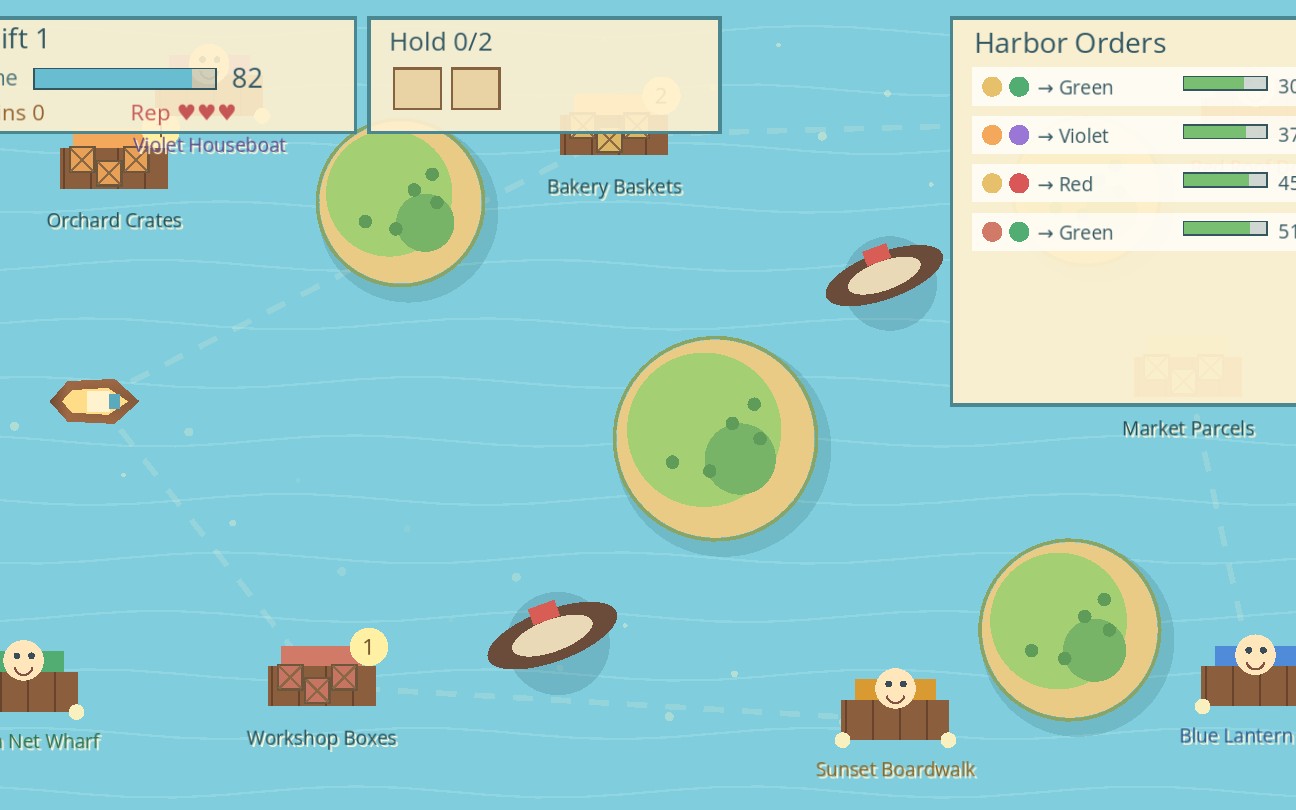}{(a) Scene construction and engine integration}{GameCraft-Bench, Cozy Harbor Delivery}{Agent-built Godot game running with live HUD}\hfill
\blocktile{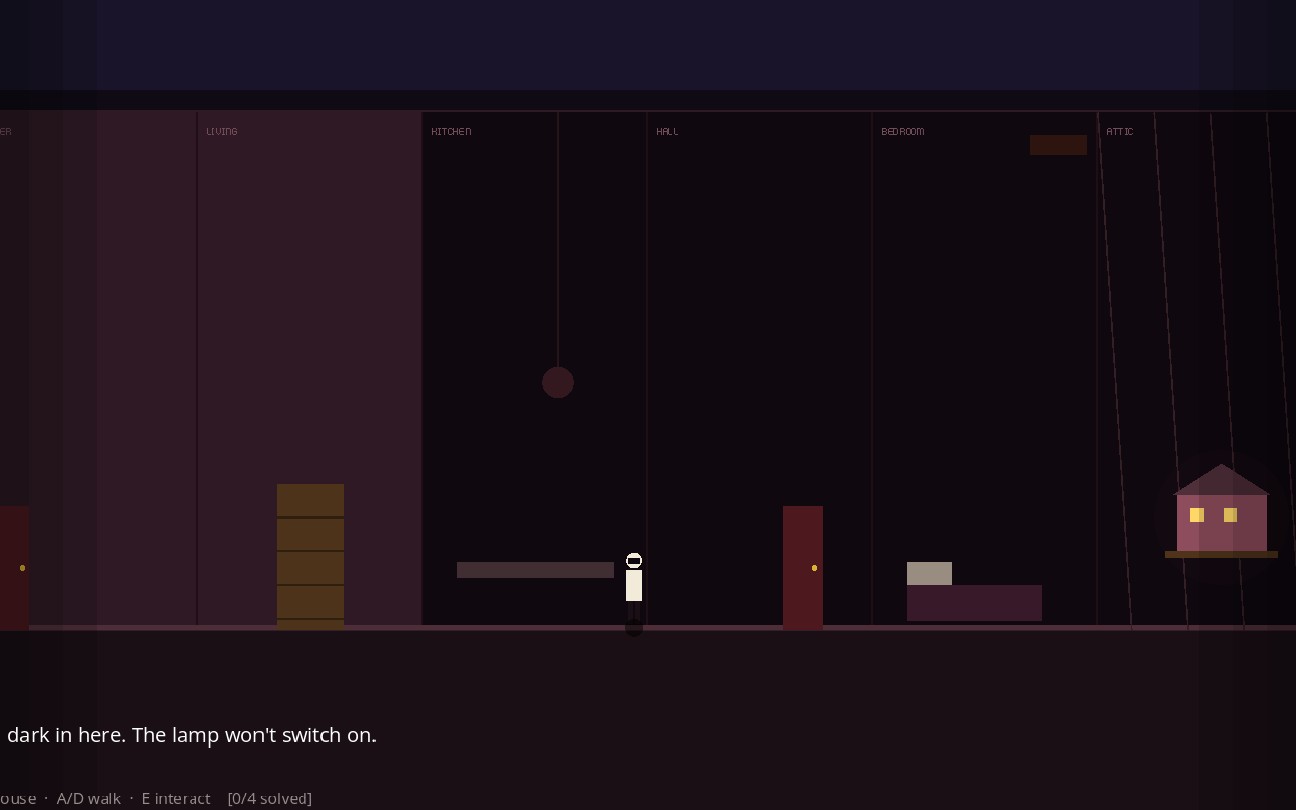}{(b) Scene construction and engine integration}{GameCraft-Bench, horror game}{A second agent-built game loaded and played}
\caption{Complete Godot projects constructed by coding agents and verified by running them: (a) GameCraft-Bench, Cozy Harbor Delivery \citep{luo2026gamecraft}; (b) GameCraft-Bench, horror game \citep{luo2026gamecraft}.}
\label{fig:strip-6-1}
\end{figure}

\runin{Scene construction and engine integration}\figmark{\ref{fig:strip-6-1}a--b} UnrealLLM translates natural-language scene descriptions into executable Unreal PCG graphs and supports basic in-engine interaction, connecting scene design directly to engine execution \citep{songtang-etal-2025-unrealllm}. UniGen coordinates planning, C\# generation, component binding, scene construction, and debugging in Unity. It evaluates three prototypes and, on one of them, reports a 91.4\% reduction in development time relative to a manual baseline implemented by one undergraduate developer with more than three years of experience \citep{yang2025unigen}. This is a small prototype comparison, not a population estimate of developer productivity. AutoUE combines model retrieval, scene generation, interaction code, and automated runtime playtesting for final evaluation in Unreal Engine \citep{yin2026autoue}. These workflows must coordinate assets with engine configuration and gameplay logic: generating the right script is insufficient when the scene does not bind or invoke it correctly.

These approaches expose different dependencies to the model. UnrealLLM's graph representation makes available operations and their connections explicit \citep{songtang-etal-2025-unrealllm}; UniGen must also generate C\# behavior and bind it to Unity components \citep{yang2025unigen}. Selecting and connecting an existing component restricts what can be built but supplies known interfaces. Generating a new component permits a wider range of mechanics while adding responsibility for initialization, callbacks, and references. The distinction is consequential even when both outputs look like a complete scene: an object may render correctly yet never receive input or invoke its script. Project context therefore includes more than nearby code. Object identifiers, asset paths, scene hierarchy, and lifecycle conventions can determine whether a generated change takes effect. A constrained intermediate representation can expose these requirements before code generation, but the translation into engine objects can itself fail. Mage's comparison of direct C\# generation and structured representations tests this trade-off: the representation that best preserves mechanic structure need not produce the most frequently running scene \citep{liu2026mage}.

GameCraft-Bench evaluates complete initial projects rather than isolated components. Its 140 Godot tasks across 15 game families use replayed interaction demonstrations and multimodal rubric judgments \citep{luo2026gamecraft}. The official leaderboard snapshot analyzed in \Cref{sec:evaluation-build} includes 16 configurations and a highest aggregate score of 75.86 out of 100 \citep{gamecraft2026results}. Separating mechanics, depth, visuals, and art reveals incomplete content or feedback within otherwise recognizable games. This differs from repository editing, where existing assets and working behavior already constrain and support the requested change.

\runin{Executable environment generation} Executable environment generation bridges modeling and software construction. Game Code World Models translate descriptions of rules, legal actions, observations, transitions, and rewards into Python environments. One study constructs a 30-game dataset, trains a 3B-parameter model on 23 games, and evaluates it on seven held-out games, combining supervised fine-tuning with execution-based reinforcement learning \citep{serapio2026gamecwm}. Structural properties are tested directly, whereas semantic checks use reference scenarios produced by a frontier language model, so their reliability also depends on the scenario generator. Agent2World assigns agents to research a domain, implement PDDL or code-based models, and test their behavior through unit tests and simulation \citep{hu2025agent2world}. Intended uses include planning and agent training, but the construction results principally measure whether the generated environments implement the specified behavior.

Generated environments need an operational contract with their consumers. Observation, action, transition, reward, and termination functions must agree, and resetting an episode must restore the assumed starting state. A structural check can verify that a function exists or returns the right type. Whether a legal action has the intended consequence is beyond it. Reference transitions, invariant checks, and simulated trajectories address that semantic gap. A generated engine can also supply state to a learned visual renderer instead of generating all observations itself. This separates implementation of rules from synthesis of appearance; \Cref{sec:engine-rendering} examines the emerging connection and the additional checks it requires.

\subsection{Tool-Using Development Agents}
\label{sec:build-tools}

\draftmap{\begin{tikzpicture}[x=1cm,y=1cm]
\node[dm q] at (-2.1,2.35) {What can a development agent observe and change in a project, and what helps it use that access?};
\node[dm node=4.0cm] (a) at (0,0) {\dmhead{Repository, editor, and runtime access}\\[1pt]{\scriptsize SWE-agent, Play2Code, AutoUE, UniGen, Unity AI tools}};
\node[dm node=4.0cm] (b) at (5.5,0) {\dmhead{Model specialization and reusable workflows}\\[1pt]{\scriptsize GameCoder, OpenGame procedures}};
\node[dm node=4.0cm] (c) at (11.0,0) {\dmhead{Multi-agent development coordination}\\[1pt]{\scriptsize RPGAgent, DreamGarden, Play2Code}};
\draw[dm arrow] (a) -- (b); \draw[dm arrow] (b) -- (c);
\node[dm bridge=3.2cm] at (2.75,0.8) {access exposes the project; specialization helps the agent use it};
\node[dm bridge=3.2cm] at (8.25,0.8) {several agents then need shared plans, artifacts, and traces};
\draw[dm axis] (-2.1,-1.15) -- (13.1,-1.15) node[midway,below=1pt,font=\scriptsize,text=GameSlate] {from project access to reusable procedures to coordinated work};
\node[dm frame,fit=(current bounding box)] (F) {};\dmtag
\end{tikzpicture}}

General coding knowledge becomes useful only when the agent can inspect and modify the particular project. Repository text, editor state, and running gameplay supply complementary context. Engine-specific training and reusable workflows offer two further forms of specialization, while multi-agent systems must coordinate the artifacts produced through these interfaces.

\begin{figure}[!htb]
\centering
\setlength{\tilew}{0.47\linewidth}
\setlength{\tilelabelh}{14.0mm}
\blocktile{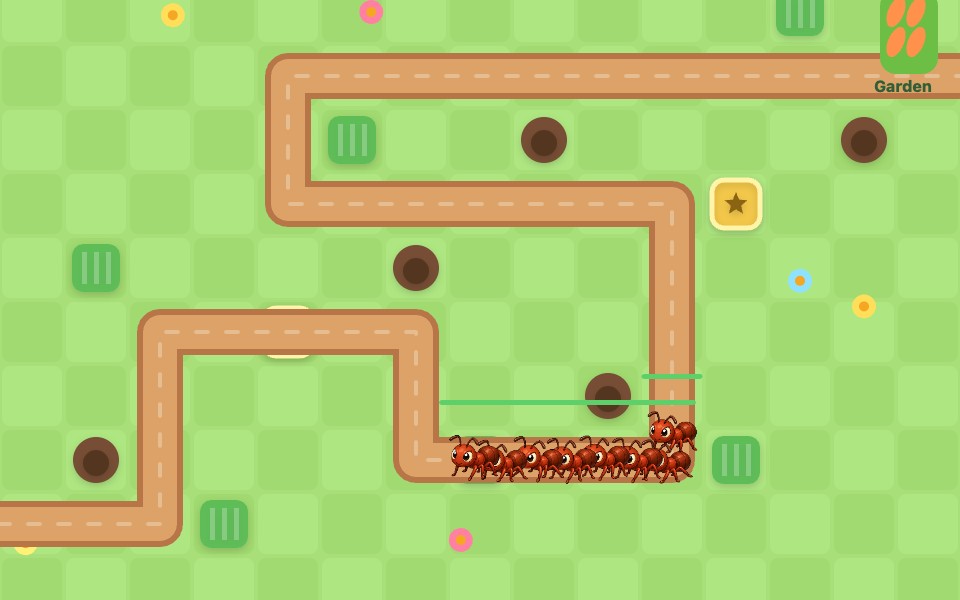}{(a) Repository, editor, and runtime access}{Play2Code, Garden Guard}{Agent testing the game through the running build}\hfill
\blocktile{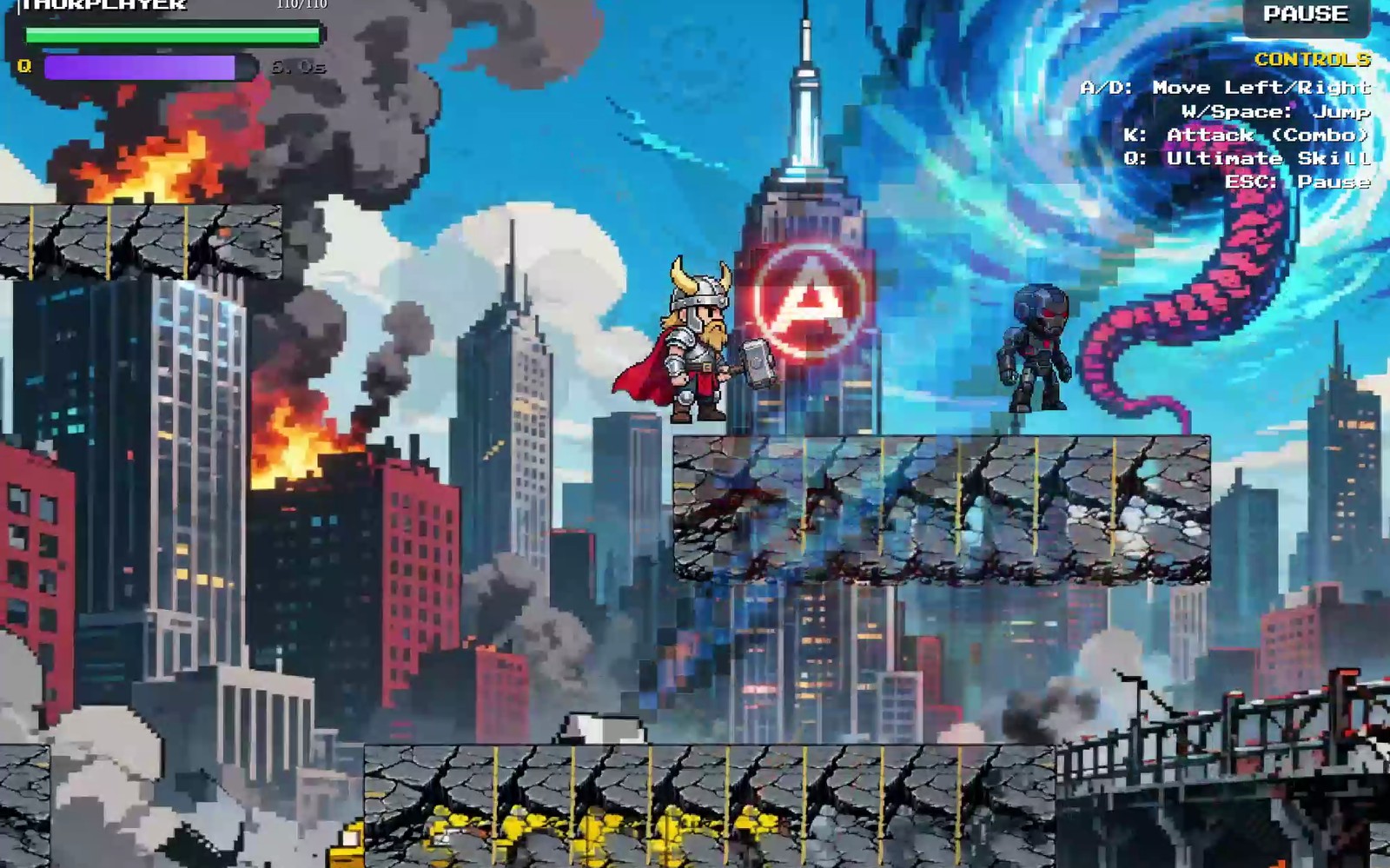}{(b) Model specialization and reusable workflows}{OpenGame, generated brawler}{Complete brawler produced by a specialised coder}
\caption{Games produced by development agents that observe the running build (a) or specialise a coding model (b): (a) Play2Code, Garden Guard \citep{huang2026guigames}; (b) OpenGame, generated brawler \citep{jiang2026opengame}.}
\label{fig:strip-6-2}
\end{figure}

\runin{Repository, editor, and runtime access}\figmark{\ref{fig:strip-6-2}a} General software-engineering benchmarks such as SWE-bench already test repository-level issue resolution \citep{jimenez2024swebench}, while systems such as SWE-agent study code editing, navigation, and executable feedback \citep{yang2024sweagent}. Game development adds engine state, scenes, assets, real-time behavior, and multimodal feedback. Development agents operate on whole projects through different interfaces. OpenGame combines reusable web-game templates with an evolving store of debugging procedures grounded in execution \citep{jiang2026opengame}. In Play2Code, a coding agent edits an HTML game while a separate GUI agent plays it in a browser and returns a play summary and a repair list through shared memory \citep{huang2026guigames}. AutoUE operates through Unreal-specific generation and runtime commands \citep{yin2026autoue}, whereas UniGen builds Unity scenes from generated plans and code and relies primarily on a developer's natural-language issue reports to prompt later fixes \citep{yang2025unigen}. Repository benchmarks instead ask agents to modify a specified part of an existing project.

Repository and editor access expose complementary dependencies. Textual search can locate a method, configuration file, or asset reference, while editor inspection reveals instantiated objects and their current properties. A development agent must also decide how much context to retrieve: a local change may require understanding a scene hierarchy, an inherited class, or a package version outside the edited file. SWE-agent's emphasis on agent--computer interaction is relevant here because tool granularity and feedback determine what the model can observe and change \citep{yang2024sweagent}. Game-specific adapters then supply operations that generic file editing cannot reliably reconstruct on its own.

Runtime access complements these project views. DreamGarden revises C++ actors using compiler errors, runtime logs, and vision-model feedback on screenshots \citep{earle2025dreamgarden}. Unity's official AI tools instead illustrate integration with a working editor: the assistant can use the scene graph, components, packages, and target platform, with plan review and reversible changes \citep{unity2026aitools}. These product features support inspection and correction; the documentation does not measure autonomous project completion. Playco's Playbot offers a further industrial example. An official case study describes generating three themed prototypes from a shared greybox and using a game-specific toolchain to iterate on them \citep{openai2026playbot}. The report concerns a studio workflow, not a controlled comparison across engines or project scales. Repository, editor, and runtime access are recorded alongside repair feedback in Appendix~\ref{app:resources}, \Cref{tab:development-systems}.

\runin{Model specialization and reusable workflows}\figmark{\ref{fig:strip-6-2}b} Engine knowledge can be supplied through training as well as through tools and retrieved instructions. OpenGame trains GameCoder with continued pretraining on game-development material, supervised fine-tuning, and execution-grounded reinforcement learning. Within the same OpenGame framework, the reported intent-alignment score rises from 49.8 for the base model to 54.1 for the fully trained variant \citep{jiang2026opengame}. A separate ablation, using a fixed Claude backbone, examines reusable development procedures such as hook-driven execution and structured project reading. These experiments concern different backbones and should not be combined into one estimate of the benefit of training versus workflow design. The two approaches address different sources of repeated work. Training can make familiar APIs and implementation patterns easier to produce; a retrieved procedure can specify how to inspect and test the current project. Neither removes version-dependent context. An API pattern learned during pretraining may be outdated, while a template may omit the mechanic requested by the designer. Their usefulness depends on whether the agent can check those assumptions against the engine and project it actually uses.

\runin{Multi-agent development coordination} Systems divide responsibilities in different ways. RPGAgent passes structured narrative, scene, mechanic, and code outputs between specialist agents \citep{zhang2026rpgagent}, while DreamGarden exposes a plan tree that a user can inspect and revise \citep{earle2025dreamgarden}. The former makes intermediate decisions available to the next generation step; the latter also provides an intervention point when a high-level decision needs changing. Decomposition is useful only if later outputs still implement those decisions.

Play2Code assigns a different role to its second agent: the browser player attempts interactions and returns a play summary and repairs to the coding agent \citep{huang2026guigames}. This separates implementing a mechanic from trying to exercise it. The returned summary can report that a jump command was issued but the character did not move, rather than simply judging the screenshot as unsatisfactory. Failure to reach the intended mechanic remains ambiguous, however: the control policy may have failed, or the implementation may be broken.

These comparisons make the shared artifact more important than the number of named agents. Scene identifiers and accepted requirements must remain consistent across outputs; a reported failure must refer to the version and initial state that produced it. Otherwise, one agent can repair an obsolete symptom or overwrite another's valid change. The surveyed systems provide mechanisms for exchanging plans, artifacts, and traces, but isolating the value of role specialization also requires holding the underlying models, tool access, and execution budget comparable.

\subsection{Execution, Debugging, and Repair}
\label{sec:build-repair}

\draftmap{\begin{tikzpicture}[x=1cm,y=1cm]
\node[dm q] at (-2.4,2.35) {What does running the project reveal, and how does a symptom become a fix?};
\node[dm node=4.6cm] (a) at (0,0) {\dmhead{Execution feedback}\\[1pt]{\scriptsize GameDevBench visual-feedback ablation}};
\node[dm node=4.6cm] (b) at (6.4,0) {\dmhead{Failure modes across execution stages}\\[1pt]{\scriptsize JAMER, Mage, GameEngineBench}};
\draw[dm arrow] (a) -- (b);
\node[dm bridge=3.6cm] at (3.2,0.8) {a symptom becomes a diagnosis only once it is placed at a stage};
\draw[dm axis] (-2.4,-1.15) -- (8.8,-1.15) node[midway,below=1pt,font=\scriptsize,text=GameSlate] {from symptom to stage: startup $\to$ behavior $\to$ scale};
\node[dm frame,fit=(current bounding box)] (F) {};\dmtag
\end{tikzpicture}}

Static analysis, compilation, startup checks, and interaction tests reveal different defects. Repair requires connecting an observed symptom to its cause and then checking that the change addresses the failure. Recent studies provide evidence about both the feedback available to an agent and the kinds of errors that persist after apparently successful execution.

\runin{Execution feedback} Executing a project exposes failures that static inspection often misses: broken scene wiring, absent feedback, unreachable objectives, or interactions that appear to satisfy a textual requirement but fail in play. Across 333 multimodal tasks in GameDevBench, the reported visual-feedback ablation raises GPT-5.4's pass rate from 41.1\% to 52.0\% with both screenshot and video feedback \citep{chi2026gamedevbench}. This benefit comes from a richer execution-and-inspection workflow, with additional tool and inference use. The editor and engine are part of the agent's working environment as well as the destination of generated code.

Compiler diagnostics and gameplay traces support different kinds of diagnosis. A missing symbol often identifies a local implementation problem; an unresponsive character could instead arise from input binding, scene activation, collision, or state logic. DreamGarden combines compile and visual feedback \citep{earle2025dreamgarden}, whereas Play2Code supplies a summary of attempted interactions from a separate player \citep{huang2026guigames}. The additional information is useful because it can narrow the cause, not merely because it adds another modality. For example, a screenshot cannot distinguish a failed action from an action that was never attempted. An input trace establishes the attempt; engine state or a targeted replay can then help locate the failure. Restricted-language systems such as ScriptDoctor make this feedback loop easier to automate through a fixed interpreter and search procedure (\Cref{sec:design-rules}). General engine projects additionally require diagnosis across code, scene bindings, and runtime state. In either setting, repair differs from generating a replacement project: a localized change must address the failure while preserving unrelated working behavior.

\runin{Failure modes and project scale} JAMER examines reconstruction in existing Godot projects. It selects 8,133 validated projects from more than 240,000 repositories: 7,833 form JamSet and 300 form JamBench. In Task~2a, directly prompted models fill removed function bodies while scene files remain available, and their mean runtime-pass rate drops from 80.4\% on small projects to 5.7\% on large ones \citep{sun2026jamer}. Here a pass means a 30-second headless run without player input. The result concerns successful reconstruction and startup as project size increases, rather than sustained interactive correctness. Larger projects raise the dependencies needed for either a successful build or a correct behavioral change. Mage tests generation from scratch: across 858 Unity scene-generation attempts with four open-weight models and 26 mini-game patterns, direct natural-language-to-C\# generation has the highest mean runtime-pass rate, 43\% \citep{liu2026mage}.

Other measures reveal defects within projects that run. In Mage, mechanism $F_1$ is near 0.12 for the configuration with the best runtime pass. This is a static measure extracted from event--condition--effect chains in generated code, not an observed gameplay success rate. The value 0.12 is its floor: three of the 26 patterns have empty reference win paths, and an empty extracted path counts as a perfect match there, giving 3/26. Structured intermediate representations improve adherence while reducing runtime success in the reported experiment. GameEngineBench tests native C++ changes in nine existing Unreal repositories through compilation, Play-in-Editor behavioral tests, and an LLM judge. The best configuration solves 55.5\% of 110 tasks, while 31 tasks defeat every tested configuration. Many failures compile successfully but violate engine lifecycle, replication, or cross-system behavior \citep{la2026gameenginebench}.

CEGame-Bench extends complete-game creation to five production engines (Defold, Godot, Roblox Studio, Unity, and Unreal Engine), with runs averaging up to 295 steps and a hidden evaluator that accepts a game only through real-input play. Under four scaffolds, agents earn most integration points but usually under half the gameplay points, and their games reach the ending in only 16\% of runs, whereas references built by engineers with coding agents score near the maximum; no two engines rank configurations alike (Kendall's $\tau$ 0.14--0.62) \citep{dong2026cegamebench}.

Four newer benchmarks separate these layers further. UnityBench-Mini3D scores 77 project-level Unity tasks with 1,178 in-engine milestone tests and no LLM judge: the best configuration reaches 72.49\%, and the average pass rate across 21 models and nine harnesses falls from 80.9\% on easy tasks to 35.2\% on hard ones \citep{zhou2026unitybench}. OpenGameEval runs agents in stateful Roblox Studio sessions and, depending on the task, uses edit-mode scene checks, scripted play-mode checks, or both: the best of 13 models passes 51.7\% of 84 core tasks on one attempt, and six tasks are never solved in 16 attempts \citep{turkel2026opengameeval}. 3D GameDev Arena asks agents to build a 3D browser game from a text specification and reference images in 108 tasks: two configurations with nearly equal functional scores (59.8 and 59.4) receive Bradley--Terry scores of 25.86 and 69.24 from blind pairwise human votes \citep{bai2026gamedevarena}. OCTAVE resolves 180 HTML5 game tasks into 1,800 requirements over eight capability dimensions and judges them by static review of the source code with an LLM. Across 67 generated games from eight mobile tasks, its verdicts match human reviewers on 149 of the 241 game--requirement pairs they could assess (61.8\%) \citep{li2026octave}.

STORY2GAME exposes the same separation between compilation and intended behavior in generated text-game actions. It derives executable preconditions and effects from story events, including updates to the represented world state. Of 90 dynamically generated actions, approximately 80\% compile and 60\% pass manual inspection of their semantic implementation \citep{zhou2025story2game}. The gap shows why checking generated code must include whether its state updates express the requested action; successful compilation establishes only part of that contract.

Read together, these results distinguish dependency failures, missing mechanic structure, and incorrect behavior under interaction. They also show that additional structure is not uniformly beneficial: it can improve specification following while introducing translation or integration errors. Comparisons are most informative when they separate these outcomes and identify whether the task constructs a new project, completes missing code, or revises existing behavior.

\subsection{Revision, Maintenance, and Handoff}
\label{sec:build-maintenance}

\draftmap{\begin{tikzpicture}[x=1cm,y=1cm]
\node[dm q] at (-2.1,2.35) {After the first success, what does the project need in order to keep living?};
\node[dm node=4.0cm] (a) at (0,0) {\dmhead{Regression-aware revision}\\[1pt]{\scriptsize GameXpert-Bench}};
\node[dm node=4.0cm] (b) at (5.5,0) {\dmhead{Update-aware testing}\\[1pt]{\scriptsize SAGE, SMART}};
\node[dm node=4.0cm] (c) at (11.0,0) {\dmhead{Developer handoff}\\[1pt]{\scriptsize DreamGarden plan, Unity reversible edits}};
\draw[dm arrow] (a) -- (b); \draw[dm arrow] (b) -- (c);
\node[dm bridge=3.2cm] at (2.75,0.8) {preserving behavior needs tests that know what changed};
\node[dm bridge=3.2cm] at (8.25,0.8) {finding regressions is not someone else continuing the work};
\draw[dm axis] (-2.1,-1.15) -- (13.1,-1.15) node[midway,below=1pt,font=\scriptsize,text=GameSlate] {the horizon lengthens: one edit $\to$ many edits $\to$ another developer};
\node[dm frame,fit=(current bounding box)] (F) {};\dmtag
\end{tikzpicture}}

Revision changes the target while retaining much of the existing project. A new mechanic may require intentional changes to movement, collision, or feedback without breaking unrelated behavior. The available evidence concerns bounded change sequences and update-aware testing more often than long-lived maintenance or transfer of a generated project to a new developer.

\runin{Regression-aware revision} GameXpert-Bench includes 97 greenfield generation tasks, 100 repair tasks, and 17 self-contained JavaScript games with six-turn optimization chains, totaling 102 cumulative requests \citep{chen2026gamexpert}. Its optimization track evaluates final artifacts against both requested changes and regression criteria. The reported results favor implementing explicit requests over discovering hidden defects and preserving functionality through accumulated edits. This setting tests a bounded sequence of revisions; dependency upgrades and collaboration in a long-lived repository impose additional demands. IGMBench moves revision into existing commercial games: 110 modding tasks, 57 in Minecraft and 53 in Terraria, are graded by in-game state and behavior checks at four depths of intervention. The best agent's strict success falls from 96.3\% for property edits to 48.1\% for system-level ones, and most failed submissions pass compilation and loading and fail behavioral validation \citep{ku2026worldediting}.

Testing must also change with the specification, as shown by the update-aware testing component in \Cref{fig:build-directions}. SAGE uses update information to prioritize and maintain tests \citep{cai2025sage}, while SMART relates code changes to functional testing goals \citep{mu2025smart}. These systems identify behavior to re-exercise after a revision; \Cref{sec:test-verification} details their test-selection and coverage mechanisms. A movement update, for example, may intentionally change jump trajectories while leaving inventory behavior intact. Replaying every earlier output as an immutable target would misclassify intended changes as regressions. Linking tests to accepted requirements makes it possible to distinguish preservation, deliberate revision, and accidental breakage.

\runin{Developer handoff} Handoff asks whether someone other than the generating agent can continue the work. The recipient needs dependencies, a build procedure, editable assets, and reproducible failures as well as an executable artifact. DreamGarden's visible plan \citep{earle2025dreamgarden} and Unity's reviewable, reversible editor operations \citep{unity2026aitools} offer mechanisms for inspecting and correcting generated changes. The design and tool interfaces are discussed in \Cref{sec:design-narrative,sec:build-tools}. Their relevance to handoff is that developers can inspect decisions and revise changes; evidence about a new developer taking over a large, evolving project remains limited. Evidence for handoff would need to show whether a developer can understand the generated structure, locate a defect, and implement a new requirement without reconstructing the original interaction with the model. This is different from automatically replaying a fixed change sequence. Bounded revision benchmarks provide a starting point, but maintainability also depends on what explanations, tests, and project conventions survive after the generating agent's session ends.

\begin{gameaiinsight}[Builder]{Game development couples code, scenes, and play}
\begin{insightpoints}
\item \textbf{Feedback exposes different parts of a project.} Repository access reveals dependencies; editor access reveals scene bindings; gameplay reveals their effects. Play2Code and GameDevBench show why execution observations can improve repair: general coding knowledge alone does not reveal what this particular project does \citep{huang2026guigames,chi2026gamedevbench}.
\item \textbf{Implementation quality has competing failure modes.} Mage's structured representations improve mechanic adherence while reducing runtime success in the reported comparison \citep{liu2026mage}. GameEngineBench likewise finds compilable but behaviorally incorrect edits \citep{la2026gameenginebench}. Better specification following can coexist with worse integration, so one aggregate build score can hide the trade-off.
\item \textbf{Maintenance requires deciding what should change.} GameXpert-Bench's cumulative requests \citep{chen2026gamexpert} and SAGE's update-aware tests \citep{cai2025sage} expose a problem absent from one-shot construction: preserving earlier behavior is correct only where the new requirement leaves it intact. Useful project memory must retain accepted requirements and reproducible tests as well as dialogue history.
\end{insightpoints}
\end{gameaiinsight}

\gameaisectionaccent{Resident}
\renewcommand{\dmcolor}{Resident}
\pretitlemark{section}{AI That Generates and Adapts at Runtime}
\section{AI That Generates and Adapts at Runtime}
\label{sec:adapt}

\begin{figure}[H]
  \centering
  \includegraphics[width=\linewidth]{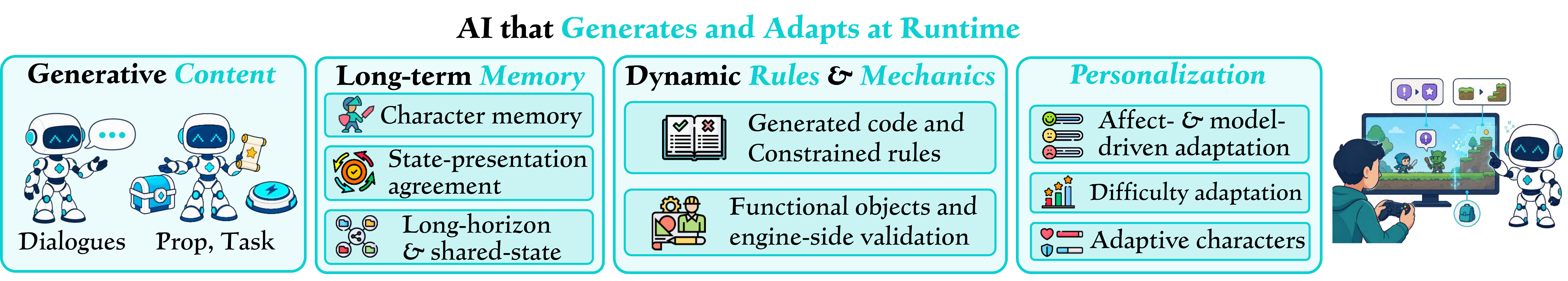}
  \caption{Research directions for AI that generates and adapts at runtime: generative content for dialogues, props, and tasks; long-term memory across characters, state agreement, and shared sessions; dynamic rules and mechanics with engine-side validation; and personalization through affect models, difficulty adaptation, and adaptive characters.}
  \label{fig:runtime-persistence}
\end{figure}

Design and development mostly finish before a player arrives. Runtime generation continues during play, making player input and the current session part of the content-production process. A dialogue, quest, rule, or difficulty adjustment must fit what has already happened and arrive in time to affect play. Earlier narrative managers and adaptive generators addressed these requirements with authored structures and selected parameters; language and multimodal models expand what can be proposed during interaction \citep{mateas2005facade,yannakakis2011edpcg,kumaran2024narrativegenie}. This creates practical choices about generation, validation, and memory. A fluent response may be purely presentational, trigger a state change, or introduce new executable behavior. Player-experience claims therefore require integrated-system evaluation (\Cref{fig:runtime-persistence}).

\draftmap{\begin{tikzpicture}[x=1cm,y=1cm]
\node[dm q] at (-1.65,2.35) {How the chapter moves: what is generated live grows in consequence, then must hold over the session, then must prove its value to the player.};
\node[dm node=3.3cm] (a) at (0,0) {\dmhead{7.1 Generative content}\\[1pt]{\scriptsize dialogue, hints, items, beats}};
\node[dm node=3.3cm] (b) at (4.3,0) {\dmhead{7.2 Runtime rules}\\[1pt]{\scriptsize code, rule languages, schemas}};
\node[dm node=3.3cm] (c) at (8.6,0) {\dmhead{7.3 State consistency}\\[1pt]{\scriptsize memory, variables, plot, sessions}};
\node[dm node=3.3cm] (d) at (12.9,0) {\dmhead{7.4 Personalization and deployment}\\[1pt]{\scriptsize content, character, deployment}};
\draw[dm arrow] (a) -- (b); \draw[dm arrow] (b) -- (c); \draw[dm arrow] (c) -- (d);
\node[dm bridge=2.7cm] at (2.15,0.8) {a line can be shown; a rule must run};
\node[dm bridge=2.7cm] at (6.45,0.8) {everything generated so far has to stay true};
\node[dm bridge=2.7cm] at (10.75,0.8) {consistency is necessary; value is the claim};
\node[dm frame,fit=(current bounding box)] (F) {};\dmtag
\end{tikzpicture}}

\subsection{Generative Characters, Narrative, and Content}
\label{sec:runtime-content}

\draftmap{\begin{tikzpicture}[x=1cm,y=1cm]
\node[dm q] at (-2.4,2.35) {How does a generated line, item, or story beat enter a running game?};
\node[dm node=4.6cm] (a) at (0,0) {\dmhead{Character dialogue and player-interpreted output}\\[1pt]{\scriptsize Craft an Iron Sword, CALYPSO}};
\node[dm node=4.6cm] (b) at (6.4,0) {\dmhead{Narrative progression and generated content}\\[1pt]{\scriptsize NarrativeGenie, PANGeA, 1001 Nights, Drama Llama}};
\draw[dm arrow] (a) -- (b);
\node[dm bridge=3.6cm] at (3.2,0.8) {a line can be shown; an item or a trigger must be represented and executed};
\draw[dm axis] (-2.4,-1.15) -- (8.8,-1.15) node[midway,below=1pt,font=\scriptsize,text=GameSlate] {how deeply the output enters the game: interpreted by a person, presented, executed, gating progression};
\node[dm frame,fit=(current bounding box)] (F) {};\dmtag
\end{tikzpicture}}

Interactive narrative combines plot structure, character intention, and game state. Fa\c{c}ade uses autonomous characters and drama management \citep{mateas2005facade}, while narrative planning coordinates events with intentional action \citep{riedl2010narrative,riedl2013interactive}. Foundation models broaden the dialogue and content available within these arrangements. The integration differs according to who interprets a proposal, which operations the game exposes, and how generated details influence subsequent play.

\begin{figure}[!htb]
\centering
\setlength{\tilew}{0.47\linewidth}
\setlength{\tilelabelh}{14.0mm}
\blocktile{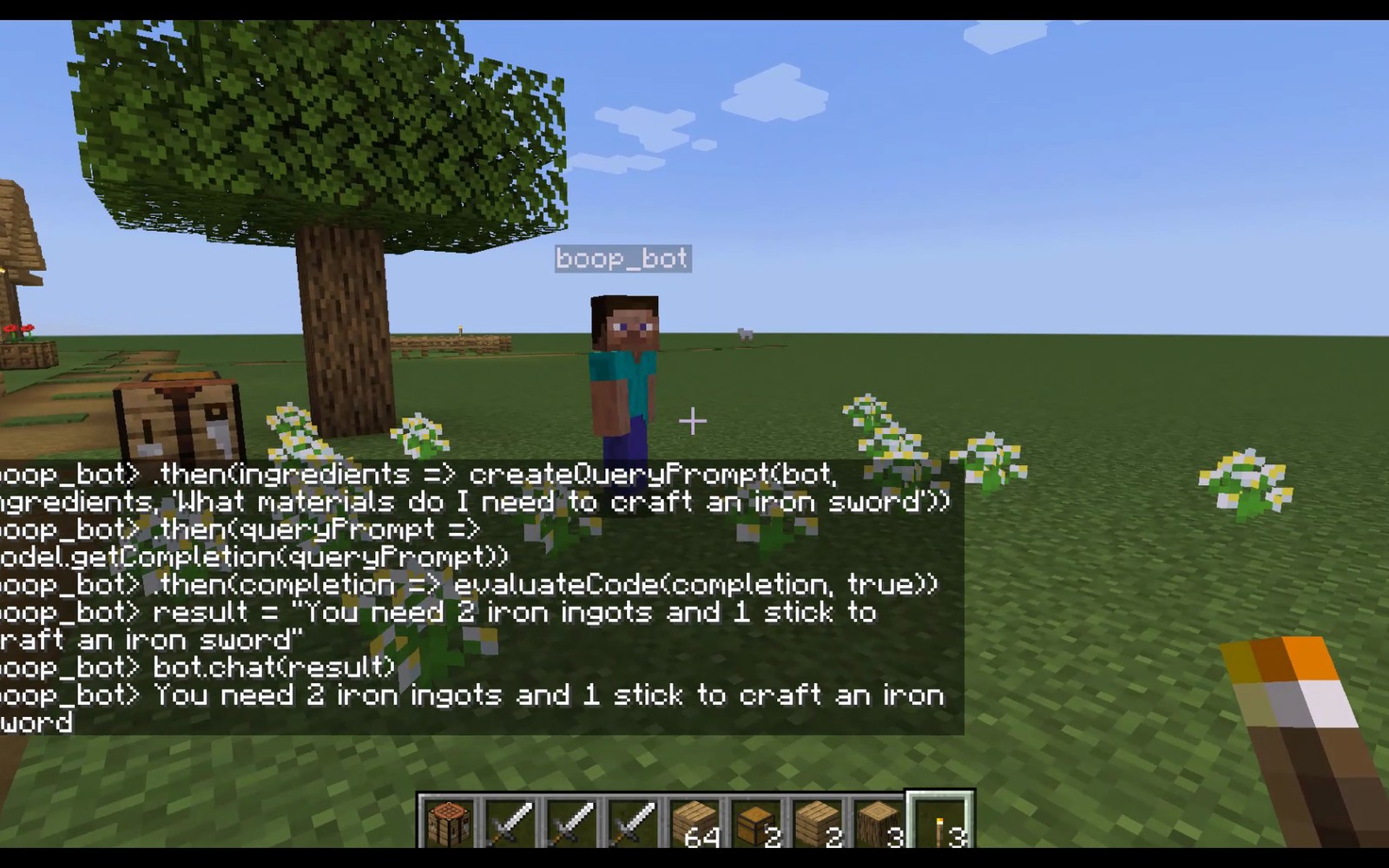}{(a) Character dialogue}{Craft an Iron Sword, Minecraft}{Recipe advice in a code-enabled NPC's chat}\hfill
\blocktile{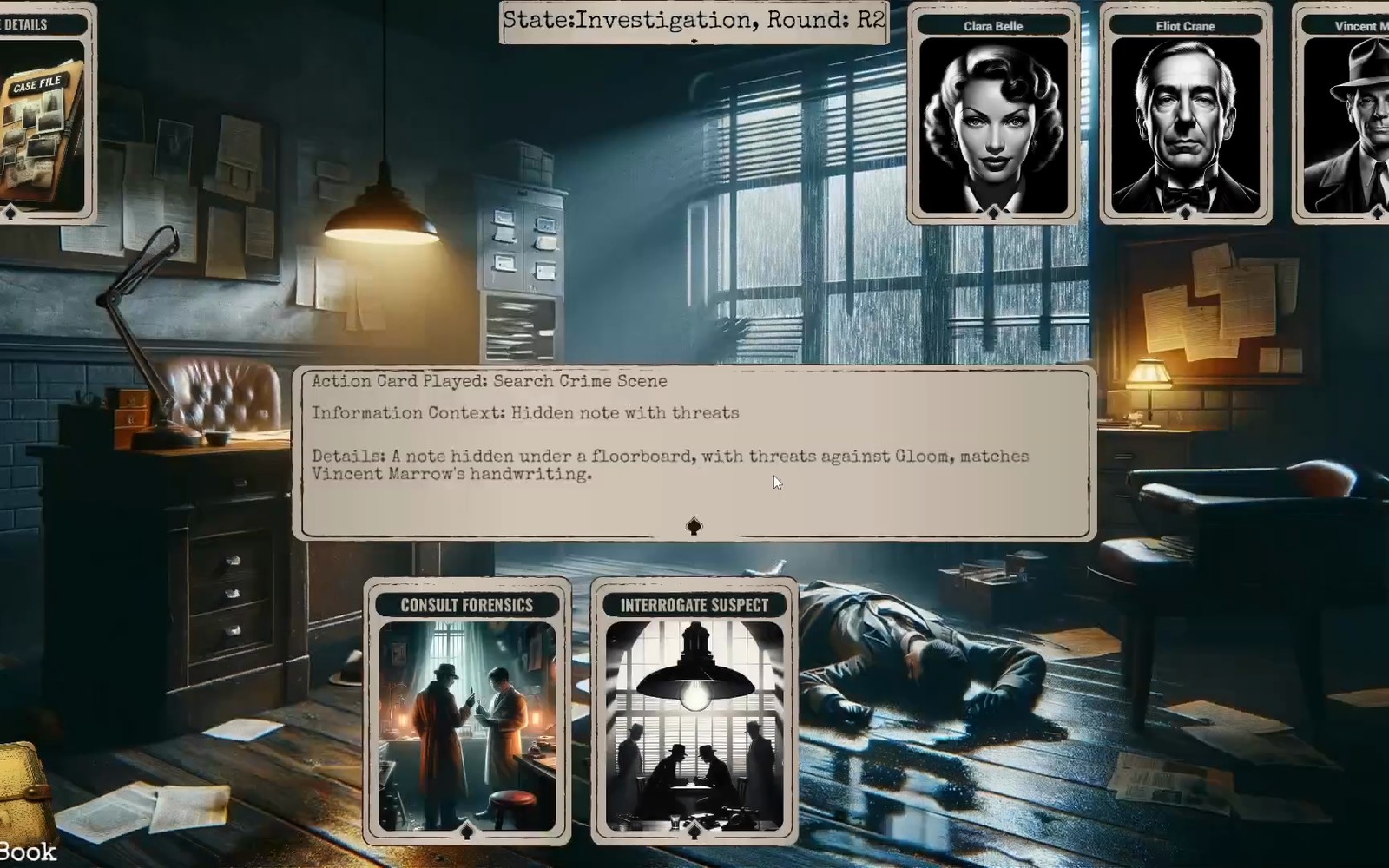}{(b) Narrative progression and generated content}{PANGeA, Dark Shadows}{Generated clue with designer-defined action cards}
\caption{Runtime generation in two settings: a question-answering exchange with a code-enabled NPC (a) and generated narrative content integrated into game progression (b): (a) Craft an Iron Sword, Minecraft \citep{volum2022craft}; (b) PANGeA, Dark Shadows \citep{buongiorno2024pangea}.}
\label{fig:strip-7-1}
\end{figure}

\runin{Character dialogue and player-interpreted output}\figmark{\ref{fig:strip-7-1}a} LIGHT connected dialogue, emotes, and actions in a text-based fantasy environment, grounding response prediction in characters, objects, and locations \citep{urbanek2019light}. Craft an Iron Sword later used a code-trained language model to generate free-form NPC dialogue and game actions in Minecraft \citep{volum2022craft}. Other systems preserve a human intermediary. CALYPSO assists Dungeon Masters with encounter summaries and brainstorming within an established world. Its field study involved 71 players and DMs; DMs selected generated material, presenting some text directly and developing other ideas further \citep{zhu2023calypso}. These setups distribute interpretation differently: an environment supplies available actions, generated code invokes operations, or a person decides how prose affects play.

Concordia makes the Game Master itself a configurable model-driven entity. Actors propose actions, and GM components determine consequences and the observations returned to participants; memory and other components can be reused across scenarios \citep{vezhnevets2025concordia}. This differs from CALYPSO's human adjudication and from a character acting through fixed engine operations. Component reuse makes scenarios easier to construct, but a generative GM's account of an action is still an interpretation of its consequences. The framework illustrates a division of computation, while player outcomes and consistency depend on the scenario and its components.

\runin{Spoken dialogue and character voices} A voiced NPC couples language generation to speech input, synthesis, and turn-taking. In \emph{The Interview}, Azure speech recognition transcribes the player's input, GPT-4o generates a response, and OpenAI TTS voices it. Completed sentence fragments enter a playback queue before the full response is ready \citep{figueiredo2025scaffolded}. This makes streaming an interaction design choice: speech can begin sooner, but material already spoken cannot be silently revised. Listening indicators and captions help players detect recognition failures during play.

The Mecha BREAK technology showcase connects on-device Whisper recognition and Nemotron dialogue to cloud-based ElevenLabs speech, with Audio2Face animating the character \citep{nvidia2024mechabreak}. These components solve distinct problems: TTS renders a line audibly; speaker conditioning or voice cloning aims for a consistent vocal identity; facial animation aligns visible performance with audio. Fortnite's Darth Vader integration uses a recognizable voice with the estate's permission \citep{epic2025vader}. It demonstrates an authorized character-voice application, without disclosing a reproducible cloning-training pipeline. Vocal similarity should therefore be distinguished from intelligibility, expressive delivery, and whether the spoken statement agrees with the current game state.

Speech also introduces timing failures absent from a static text box. A queued hint may arrive after the player leaves the scene, overlapping speakers can disrupt turn detection, and interrupting an NPC requires cancelling obsolete audio as well as future text. The engine must distinguish spoken promises from actions it has actually accepted. Relevant tests include end-of-turn to first-audio delay, interruption recovery, recognition errors under game audio, and dialogue--action agreement. \Cref{sec:runtime-players} discusses the player evidence for these systems.

\runin{Narrative progression and generated content}\figmark{\ref{fig:strip-7-1}b} Systems integrated directly into digital games require more explicit links between language and gameplay. NarrativeGenie's runtime Adaptive Dialogue Manager derives hints and summaries from current state and play history. Its evaluation combines automated checks on 30 simulated histories with a 19-participant study in one mystery episode \citep{kumaran2024narrativegenie}. PANGeA generates settings, items, characters, and dialogue for turn-based role-playing scenarios, using memory and a language-model validator to keep free-form input within designer-specified rules \citep{buongiorno2024pangea}. In 1001 Nights, words elicited through co-created storytelling become usable battle equipment, while narrated scenes are visualized during play \citep{sun2023language}. The difference is what the game does with the response: a hint helps interpret existing state, whereas an item or action must also be represented and executed by the game.

Drama Llama provides a further hybrid: authors specify natural-language storylet triggers, and a language-model drama manager checks when to inject the corresponding stage directions into an unfolding text interaction \citep{sun2025dramallama}. Its preliminary study with six authors examines authoring and the creative quality of the resulting stories. Compared with a fixed branching script, triggers can respond to a wider range of phrasing. Compared with unconstrained narration, they preserve explicit points of author control. Its responsiveness depends on the learned trigger checker's interpretation.

Personalized quest generation makes the intermediate structure explicit. Ashby et al. match player input to a game-world knowledge graph, use graph paths and grammars to construct quests, and condition a fine-tuned GPT-2 model on those quests to generate titles and NPC dialogue \citep{ashby2023quests}. Compared with NarrativeGenie's hints within an episode, this pipeline generates the task as well as its presentation. Personalization follows the player's expressed request, rather than a profile inferred from past behavior. The graph constrains task construction, but generated dialogue can still mention absent objects or locations, so structural consistency and text grounding remain separate checks in evaluation.

This comparison identifies a practical source of narrative failure. A generated clue may sound appropriate yet refer to an inaccessible location; an item may be described without any applicable action; an improvised branch may have no supported continuation. Checking prose quality cannot resolve these failures alone. The system must also know which generated details later play can use. Authoring constraints, retrieval, and executable bindings address different aspects of this problem.

\subsection{Runtime Rules, Mechanics, and Worlds}
\label{sec:runtime-rules}

\draftmap{\begin{tikzpicture}[x=1cm,y=1cm]
\node[dm q] at (-2.4,2.35) {When the generated output is a rule, how much expressiveness is traded for validation?};
\node[dm node=4.6cm] (a) at (0,0) {\dmhead{Generated code and constrained rule languages}\\[1pt]{\scriptsize GROMIT, Real-Time World Crafting, IF:CARGO}};
\node[dm node=4.6cm] (b) at (6.4,0) {\dmhead{Functional objects and engine-side validation}\\[1pt]{\scriptsize Roblox Cube}};
\draw[dm arrow] (a) -- (b);
\node[dm bridge=3.6cm] at (3.2,0.8) {a schema constrains what is generated and exposes more behavior to validation};
\draw[dm axis] (-2.4,-1.15) -- (8.8,-1.15) node[midway,below=1pt,font=\scriptsize,text=GameSlate] {more constrained interfaces expose more behavior to validation};
\node[dm frame,fit=(current bounding box)] (F) {};\dmtag
\end{tikzpicture}}

Runtime mechanics make player requests into executable changes. Systems use generated code, constrained rule languages, or schemas linked to existing behaviors. These choices differ in what the model is allowed to propose and what the engine can check. A narrower language can simplify some checks, but valid rules can interact in ways that players struggle to predict.

\begin{figure}[!htb]
\centering
\setlength{\tilew}{0.32\linewidth}
\setlength{\tilelabelh}{21.0mm}
\blocktile{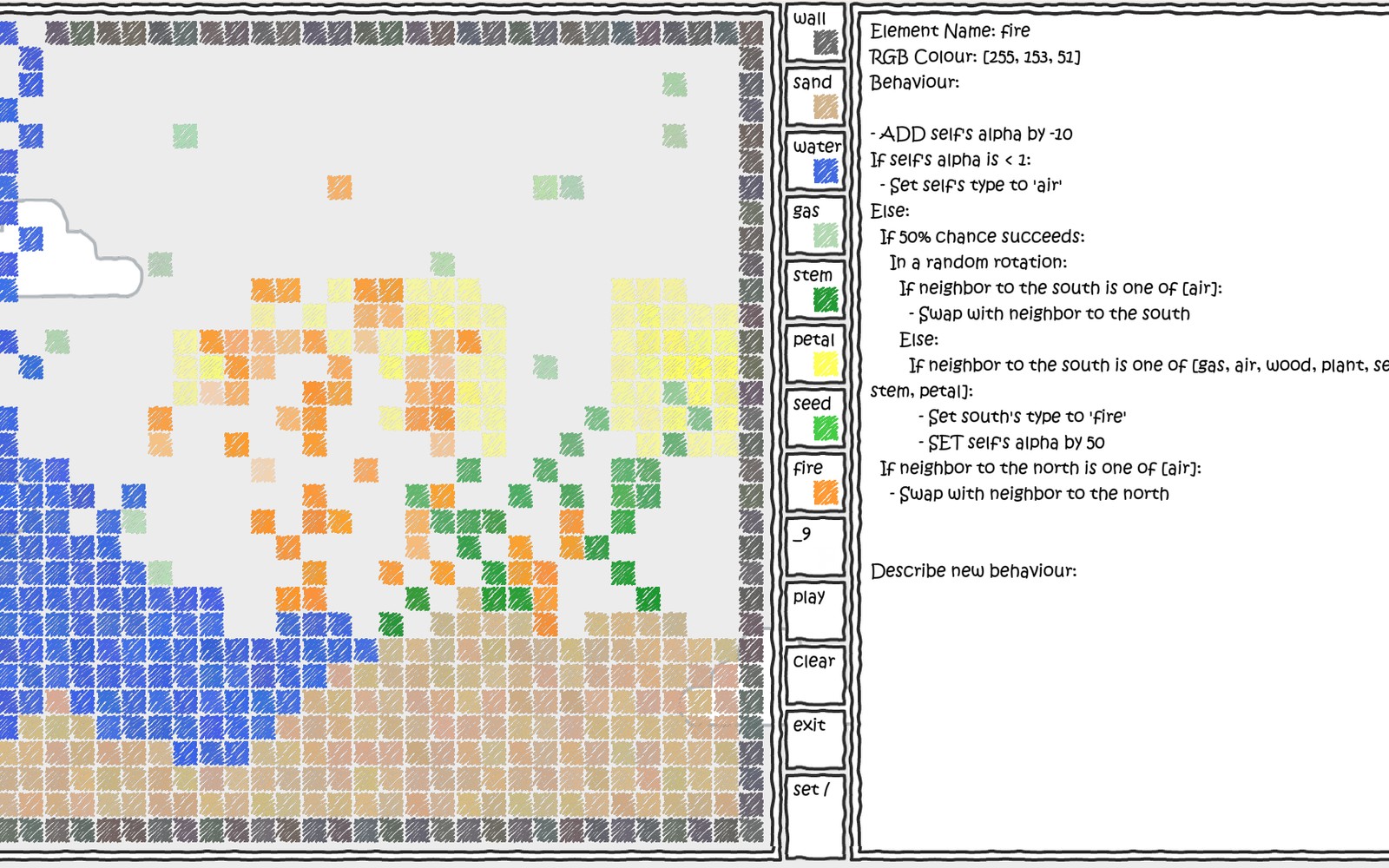}{(a) Generated code and\\constrained rule languages}{Real-Time World Crafting, Latent Space}{Generated rule text beside the world it governs}\hfill
\blocktile{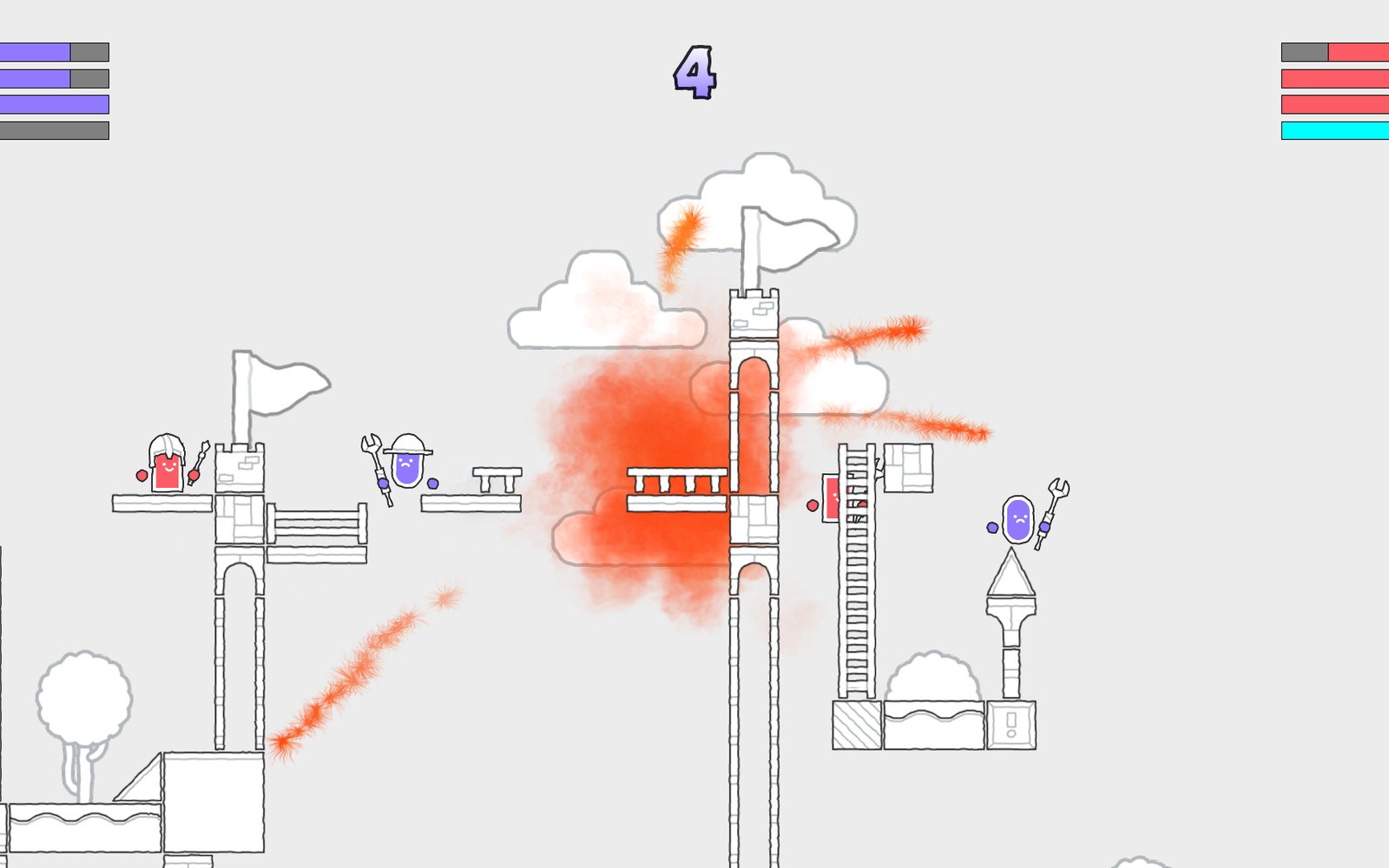}{(b) Generated code and\\constrained rule languages}{Real-Time World Crafting, Latent Space}{The compiled spell firing in battle}\hfill
\blocktile{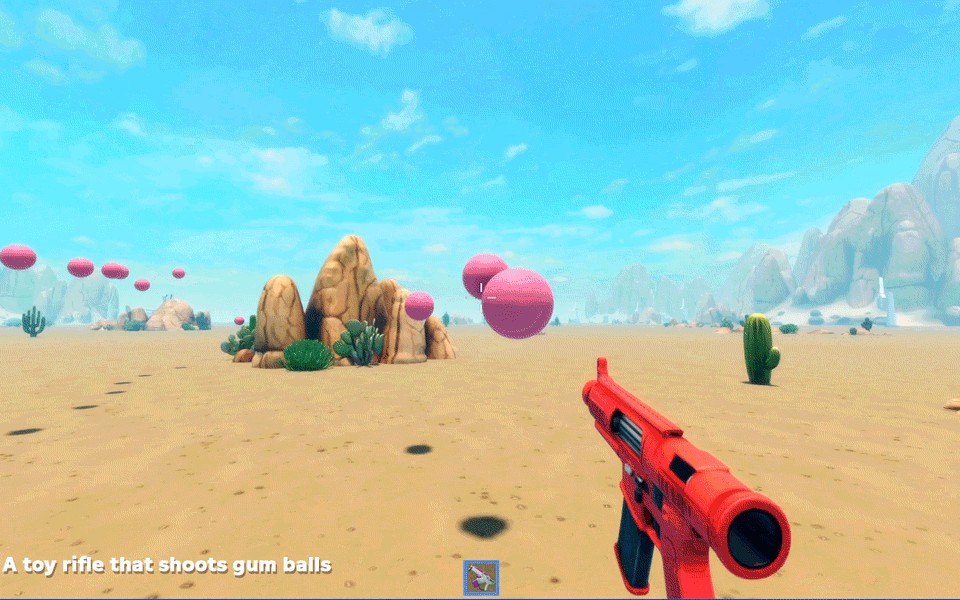}{(c) Functional objects and\\engine-side validation}{Roblox Cube}{A just-generated toy rifle, immediately usable}
\caption{Rules and objects generated during play and executed by the engine: a domain-specific rule language (a--b) and schema-constrained objects (c): (a--b) Real-Time World Crafting, Latent Space \citep{drake2025worldcrafting}; (c) Roblox Cube \citep{singh2026robloxcube}.}
\label{fig:strip-7-2}
\end{figure}

\runin{Generated code and constrained rule languages}\figmark{\ref{fig:strip-7-2}a--b} Some runtime systems generate executable behavior or rules. GROMIT generates and compiles Unity behaviors during play, allowing model-generated code to produce gameplay effects \citep{jennings2024gromit}. Its evaluation combines three demonstration scenarios with interviews of 13 game developers, so the evidence primarily concerns runtime behavior generation and development-workflow risks, not player outcomes. Real-Time World Crafting translates player commands into a constrained domain-specific language that configures an entity--component system in a spell-crafting prototype \citep{drake2025worldcrafting}. IF:CARGO similarly maps player-authored rules to commands that the engine validates and executes deterministically. In its 24-participant, eight-level playtest, players used execution feedback to revise rules, while periodic commands, multi-robot coordination, and priority rules increased diagnostic demands \citep{hsu2026ifcargo}. Direct code allows new implementations; constrained languages expose a smaller set of operations with explicit validation rules. Their reliability still depends on the generated program and the checks available.

The representation also determines what a player can debug. Direct code can express behavior outside an existing command set, but a compilation error may be remote from the player's intended mechanic. A constrained language limits the proposal space and can make rule priority, targets, and parameters visible in engine feedback. IF:CARGO's multi-robot and periodic-command cases show that even deterministic execution can remain difficult to reason about when several valid rules interact. Validating each command separately therefore differs from explaining the behavior of the combined rule set.

\runin{Functional objects and engine-side validation}\figmark{\ref{fig:strip-7-2}c} Industrial object generation illustrates another form of constraint. Roblox's Cube-powered 4D generation beta lets players create functional objects inside creator-enabled experiences. Its initial Car-5 and Body-1 schemas prescribe the parts to generate, after which behavior scripts are retargeted to the generated geometry \citep{singh2026robloxcube}. Here the reusable component is a generative shape model, and functionality comes from a schema and script integration, with no unrestricted invention of behavior. The official report documents deployed functionality, not a controlled comparison of player outcomes.

Dialogue can often be presented directly, whereas generated quests, behaviors, or rules may require parsing, compilation, constraint checks, or engine-side validation. A weak line of dialogue may briefly disrupt characterization, a malformed quest condition can block progression, and an incorrect rule can alter later strategy. Safeguards must therefore match both the type of runtime content and the route by which it enters the engine.

\subsection{State Consistency and Long-Term Memory}
\label{sec:runtime-state}

\draftmap{\begin{tikzpicture}[x=1cm,y=1cm]
\node[dm q] at (-2.1,2.35) {What must be retained as the session continues, and what test shows it survived?};
\node[dm node=4.0cm] (a) at (0,0) {\dmhead{Character and interaction memory}\\[1pt]{\scriptsize Generative Agents, PANGeA}};
\node[dm node=4.0cm] (b) at (5.5,0) {\dmhead{State--presentation agreement}\\[1pt]{\scriptsize Unbounded, AnimeGamer}};
\node[dm node=4.0cm] (c) at (11.0,0) {\dmhead{Long-horizon and shared-state consistency}\\[1pt]{\scriptsize NCP-Bench, MultiGen}};
\draw[dm arrow] (a) -- (b); \draw[dm arrow] (b) -- (c);
\node[dm bridge=3.2cm] at (2.75,0.8) {recalling a line is not preserving an inventory change};
\node[dm bridge=3.2cm] at (8.25,0.8) {scene-level consistency does not establish plot-level consistency};
\draw[dm axis] (-2.1,-1.15) -- (13.1,-1.15) node[midway,below=1pt,font=\scriptsize,text=GameSlate] {the record grows in scope: utterance $\to$ variable $\to$ plot $\to$ session and shared state; the evidence thins as it grows};
\node[dm frame,fit=(current bounding box)] (F) {};\dmtag
\end{tikzpicture}}

Quests, relationships, inventories, and discovered facts connect earlier choices to later play \citep{mateas2001poetics,riedl2010narrative}. Runtime systems use event records, structured state, retrieved memories, and generative presentation to preserve different aspects of that history. Recalling a conversation, respecting a quest condition, and synchronizing a shared event require different information and update procedures.

\runin{Character and interaction memory} Generative Agents ranks stored observations by relevance, recency, and importance, then synthesizes higher-level reflections that inform later plans \citep{park2023generativeagents}. Retrieval decides which past events remain available, and reflection can connect several events into a character's belief or intention. Their failure modes differ: missing retrieval can erase a relevant encounter, while an incorrect reflection can propagate a false interpretation into later dialogue. In the reported sandbox evaluations, retrieval failures and invented embellishments remain observable even when behavior is judged believable. GARP runs this architecture in real time inside Godot with a local 3B-parameter model and about 20 agents. Full responses still take 10--15 seconds, and believability was not tested with human raters \citep{molino2026garp}. PANGeA, introduced with runtime content in \Cref{sec:runtime-content}, combines retrieved narrative context with validation of out-of-scope requests. Its ten-scenario ablation uses synthetic player inputs and a GPT-4 judge whose modal answer over 30 runs matched that of 30 experts on 79 of 80 examples. The reported 98--99\% accuracy concerns handling those inputs \citep{buongiorno2024pangea}. This tests a different property from the character memories in Generative Agents: constraining scenario requests rather than retrieving and reflecting on social encounters.

Retrieval and state updates also occur at different points in an interaction. A remembered statement may describe an intended action, a mistaken belief, or an event later corrected by the player. Treating each as an accomplished game event would introduce errors even with perfect retrieval. The useful distinction is between what a character can recall and what the running game has actually recorded. Where systems maintain both, evaluation can test whether dialogue is regenerated appropriately after a state correction, rather than checking recollection in isolation. State-Grounded Conditioning applies this distinction to an in-game conversational coaching agent that guides players through consecutive competitive matches: moving state-dependent control into rule kernels over structured state raised human-annotated session-level grounded accuracy on 200 logged sessions from 20.0\% (prompting) and 26.5\% (a tool-using agent) to 83.5\%, while holding mean first-token latency at 1.5~s versus 6.1~s for the tool-using agent \citep{liu2026stategrounded}.

\runin{State--presentation agreement} Generative life simulations also maintain visual context. Unbounded updates character-life variables alongside language-routed narrative and images \citep{li2025unbounded}. AnimeGamer instead predicts character variables and animation-shot representations from multimodal history, then decodes the latter with video diffusion. Its improvements concern visual continuity and motion as well as instruction following, while character-state accuracy remains similar to its language-router baseline \citep{cheng2025animegamer}. Evaluation combines generated instruction sequences with 20 raters viewing prerecorded outputs, without an interactive player trial. Jointly generating images and state therefore addresses representational consistency without yet demonstrating sustained play. State and presentation can disagree in either direction. A character may describe an item that was never awarded, or the engine may record a completed event that the generated dialogue ignores. Structured state constrains the set of valid references, but the generation interface must expose those constraints and update them after accepted actions. Unbounded and AnimeGamer make some character variables explicit, whereas PANGeA emphasizes retrieved context and rule checks. Their different state stores explain why memory quality, rule adherence, and visual continuity need separate tests.

\runin{Long-horizon and shared-state consistency} Narrative continuity requires later responses to respect established facts and plot constraints, including when a player attempts to disrupt the story. NCP-Bench operationalizes this requirement through narrator interactions checked against an explicit narrative specification \citep{ma2026ncpbench}. Its auditing procedure and results are discussed in \Cref{sec:test-verification}; here it specifies what runtime memory must preserve.

Longer deployments and shared play widen the scope further. Returning players must encounter recognizable affordances, character memories must survive gaps between sessions and changes in the underlying model, and saved sessions must remain compatible with later versions of the game. Multiplayer play adds synchronization and fairness requirements. Participants need a consistent account of events that affect later play, even when narrative presentation or personalization differs. A generated reward, rule, or piece of strategic information can change competitive balance, while latency affects the order in which actions enter shared state. MultiGen's external map and synchronized generated viewpoints illustrate a concrete shared-state mechanism \citep{po2026multigen}. Existing learned-world studies provide more evidence about model-level synchronization than about fairness or persistence in live multiplayer populations (\Cref{sec:model-generative}).

Cross-session testing can make these requirements concrete: resume a saved quest after a gap, correct an earlier event, or update the character model while keeping the saved game fixed. The expected outcome differs in each case. Dialogue may change after a correction, whereas completed objectives and shared rewards should remain consistent with the revised record. Such tests examine the interaction between persistence and generation rather than transcript storage alone.

\subsection{Personalization, Adaptation, and Deployment}
\label{sec:runtime-players}

\draftmap{\begin{tikzpicture}[x=1cm,y=1cm]
\node[dm q] at (-1.5,2.35) {Does adapting to the player make play better, and by whose evidence?};
\node[dm node=2.8cm] (a) at (0,0) {\dmhead{Affect- and model-driven adaptation}\\[1pt]{\scriptsize physiological DDA, Affectively}};
\node[dm node=2.8cm] (b) at (3.6,0) {\dmhead{Difficulty adaptation and balancing}\\[1pt]{\scriptsize Hamlet, EOMM, Replicants}};
\node[dm node=2.8cm] (c) at (7.2,0) {\dmhead{Content and narrative personalization}\\[1pt]{\scriptsize EDPCG, match-three, interactive drama}};
\node[dm node=2.8cm] (d) at (10.8,0) {\dmhead{Adaptive characters and companions}\\[1pt]{\scriptsize Campus Culture Week, LeagueBot}};
\node[dm node=2.8cm] (e) at (14.4,0) {\dmhead{Evaluation, transparency, and deployment}\\[1pt]{\scriptsize CHI PLAY studies, PUBG Ally}};
\draw[dm arrow] (a) -- (b); \draw[dm arrow] (b) -- (c); \draw[dm arrow] (c) -- (d); \draw[dm arrow] (d) -- (e);
\node[dm bridge=2.4cm] at (1.8,0.8) {the signal is only useful once something is changed};
\node[dm bridge=2.4cm] at (5.4,0.8) {a parameter is the smallest thing to change; content is next};
\node[dm bridge=2.4cm] at (9.0,0.8) {adapting a character changes whom the player talks to};
\node[dm bridge=2.4cm] at (12.6,0.8) {player evidence is needed to settle the experience claim};
\draw[dm axis] (-1.5,-1.15) -- (15.9,-1.15) node[midway,below=1pt,font=\scriptsize,text=GameSlate] {what drives adaptation $\to$ what is adapted (parameter, content, character) $\to$ how the claim is tested};
\node[dm frame,fit=(current bounding box)] (F) {};\dmtag
\end{tikzpicture}}

Personalization uses observed performance, inferred preferences, or explicit requests to choose what a player receives. Difficulty, content, and character behavior are different intervention targets, with different effects on control and experience \citep{mortazavi2024dynamic,zohaib2018dynamic,paraschos2023game}. Evaluation must separate the quality of the inferred player model from the effect of the change made using it.

\runin{Affect- and model-driven adaptation} Performance, inferred preferences, affect, and explicit player choices can jointly drive an adaptation. Affective game computing connects sensing and prediction to experience adjustment \citep{yannakakis2023affective,chanel2011emotion}. A review of 23 empirical studies finds complete sensing-to-adaptation loops comparatively uncommon in the literature it retrieved \citep{lopes2025closing}. The Affectively Framework makes predicted affect available in Gym environments for agent training \citep{barthet2024affectively}, while language-model studies test engagement inference from gameplay footage \citep{melhart2025large}. Training against an affect model is useful for developing a policy, but also permits the policy to exploit errors in that model. Model accuracy and intervention benefits need separate checks (\Cref{sec:model-players}).

\runin{Difficulty adaptation and balancing} Hunicke's Hamlet system adjusts resources using player performance, with attention to intelligible feedback and intended experience \citep{hunicke2005dda}. Later approaches optimize different objectives: an Electronic Arts deployment reports engagement gains from difficulty adjustment \citep{xue2017dynamic}, while Dungeons \& Replicants uses learned individual behavior models to balance content offline \citep{pfau2020dungeons}. Engagement, challenge, and perceived fairness are distinct targets; improvement in one cannot be assumed to improve the others. Timing is another substantive choice. Within-session adjustment responds quickly but can make the current challenge feel unstable. Between-level generation allows more time to validate content and present a consistent challenge. Longer-term profiles can incorporate repeated play but may become stale as a player learns. Foundation models expand the information used to choose an adjustment, without resolving the choice of objective or intervention timescale.

Controlled player studies of difficulty adaptation form a line of their own. Challenge adjustment has been tested for its effect on player performance and experience \citep{denisova2015adaptation}, different adjustment systems have been compared on game experience \citep{ang2017comparing}, player-oriented systems that expose adjustment as a choice have been studied for how that choice is represented and how often it is offered \citep{ang2019representation}, and adjustment has been shown to affect players' confidence \citep{constant2019dynamic}. A player-centered framework of game personalization collects the resulting open problems, among them modeling accuracy, controllability, and players' mental models of the personalization itself \citep{zhu2021player}. Language-model agents can also serve the evaluation: although they play below the level of an average human, their results can still indicate relative difficulty. In Wordle, agent guess counts correlated with players' average guess counts. In Slay the Spire, agents' remaining health after boss fights correlated with players' win rates against those bosses. The strength of these associations varied with the model and prompting method \citep{xiao2024llms}.

\begin{figure}[!htb]
\centering
\setlength{\tilew}{0.47\linewidth}
\setlength{\tilelabelh}{14.0mm}
\blocktile{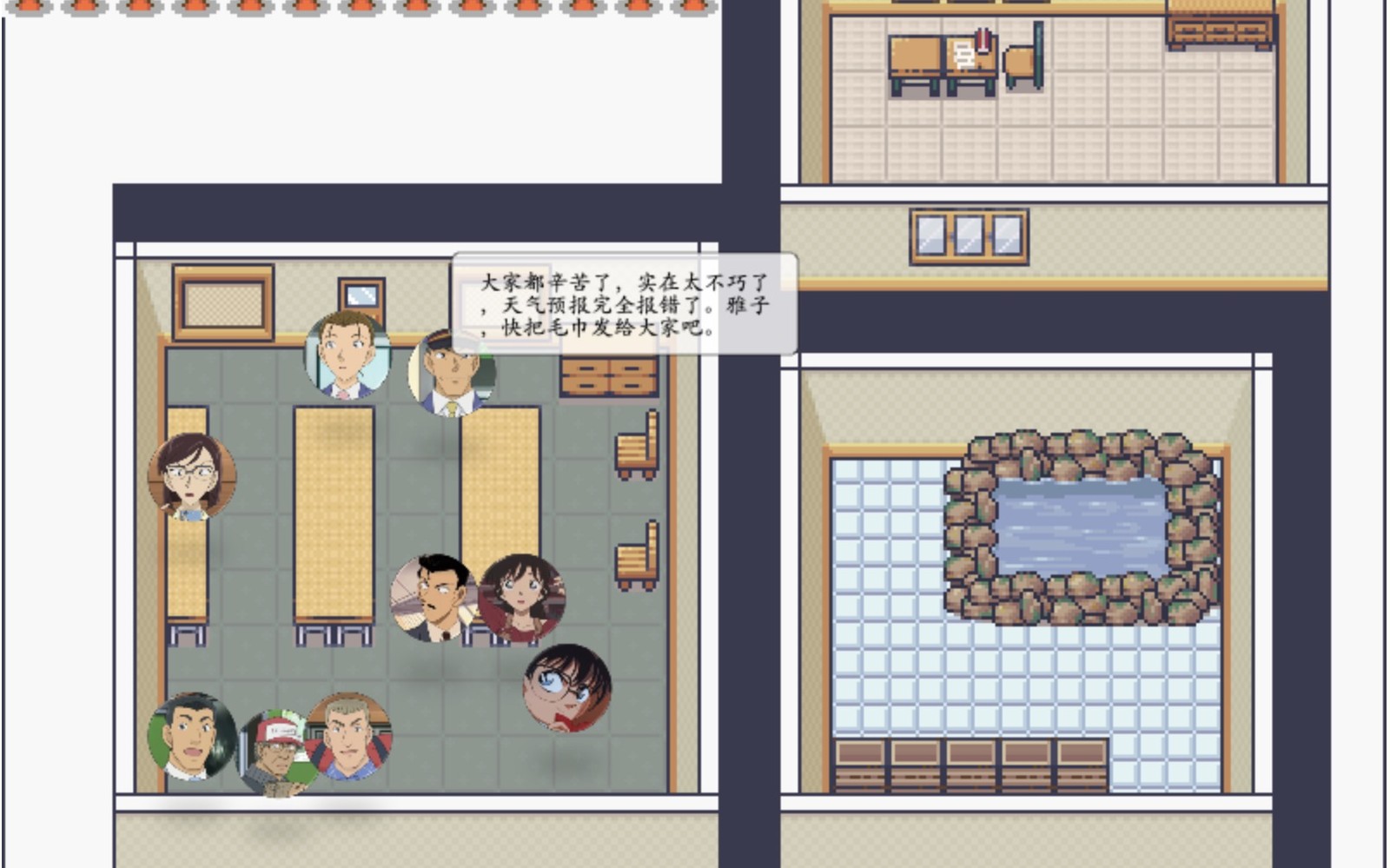}{(a) Content and narrative personalization}{Interactive drama (Wu et al.), waiting-room scene}{Model-played characters improvising around the player}\hfill
\blocktile{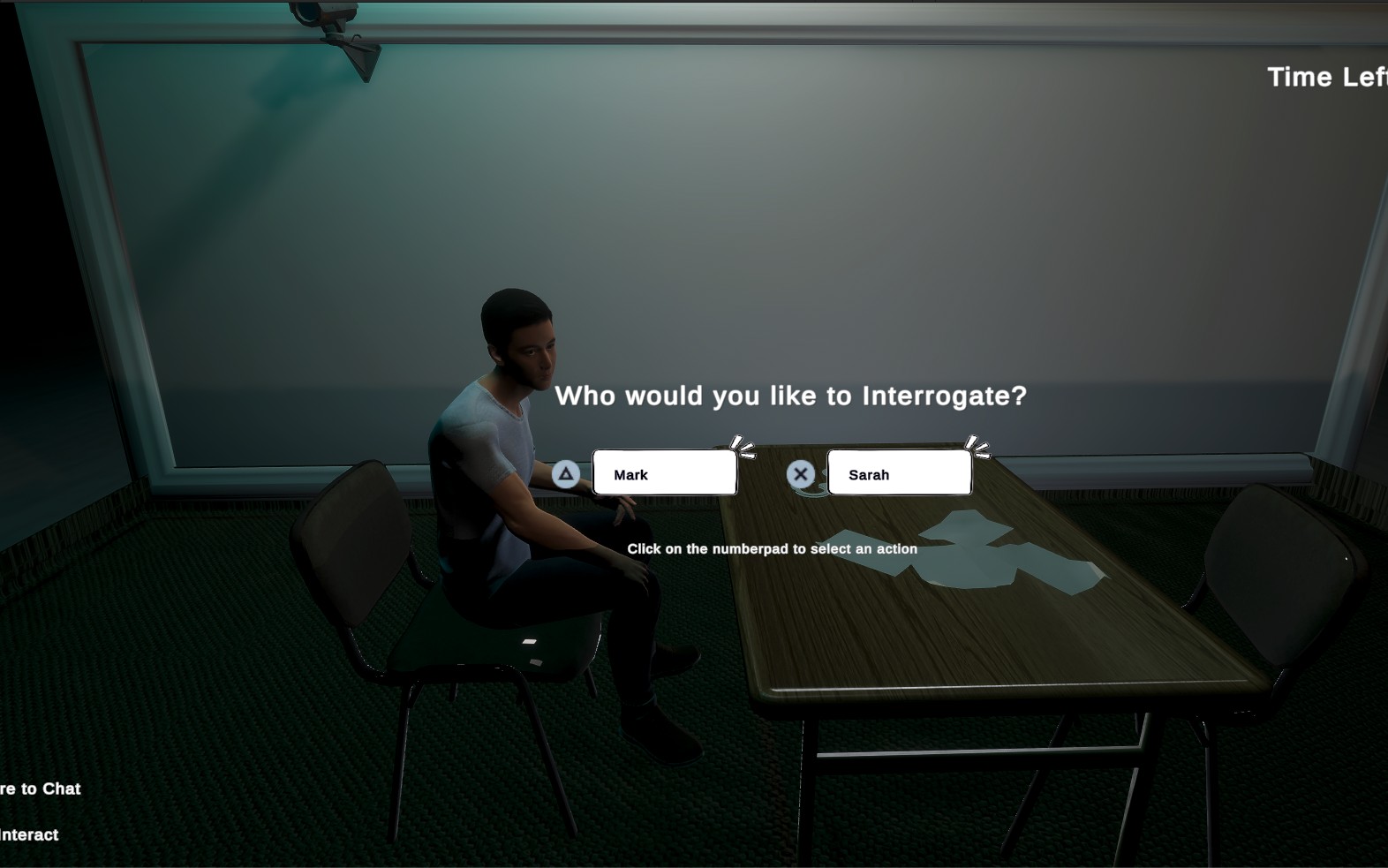}{(b) Adaptive characters and companions}{Figueiredo et al., detective prototype}{Free-form dialogue with an adaptive suspect}\\[3pt]
\blocktile{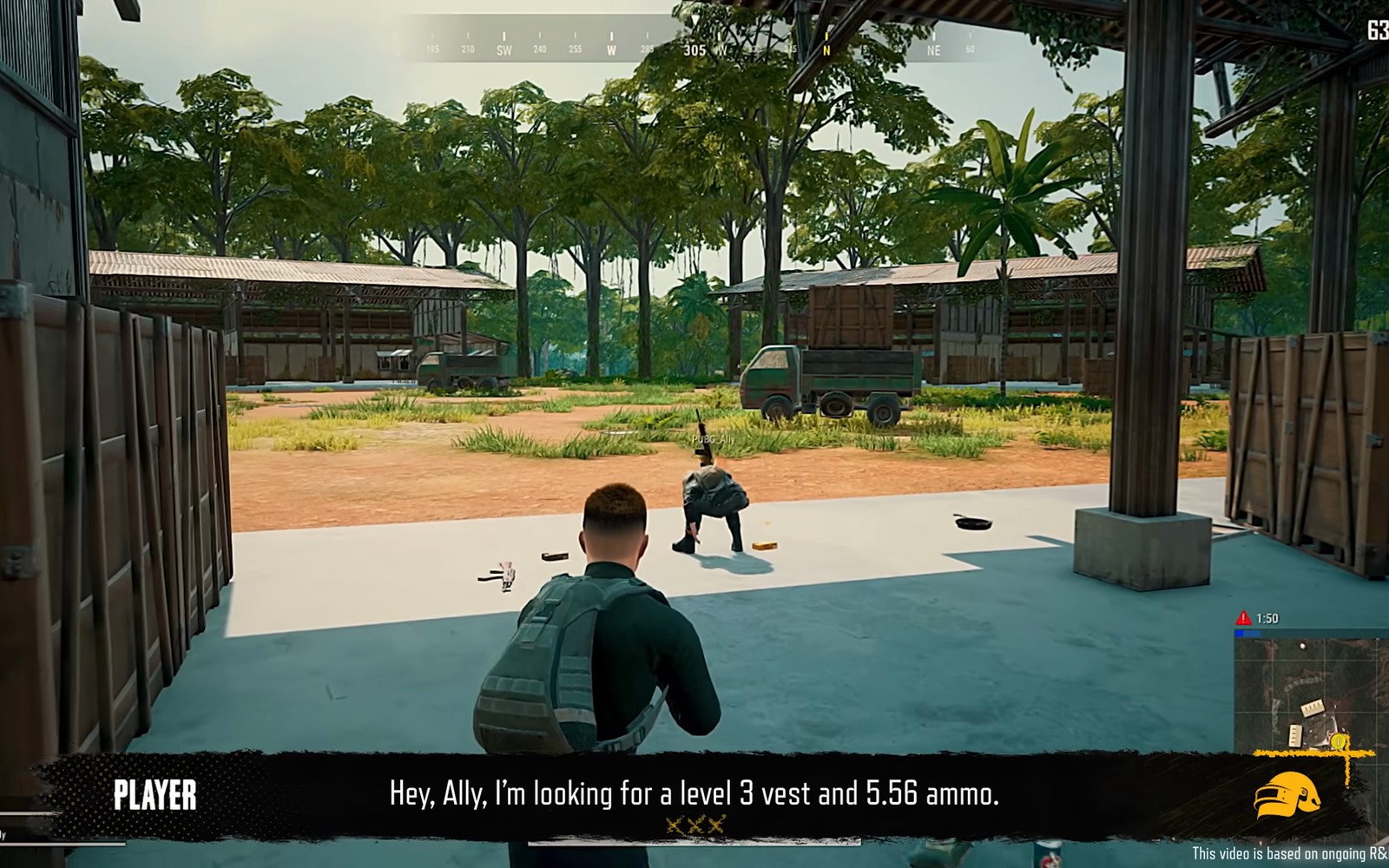}{(c) Evaluation, transparency, and deployment}{PUBG Ally}{AI teammate answering a spoken request}\hfill
\blocktile{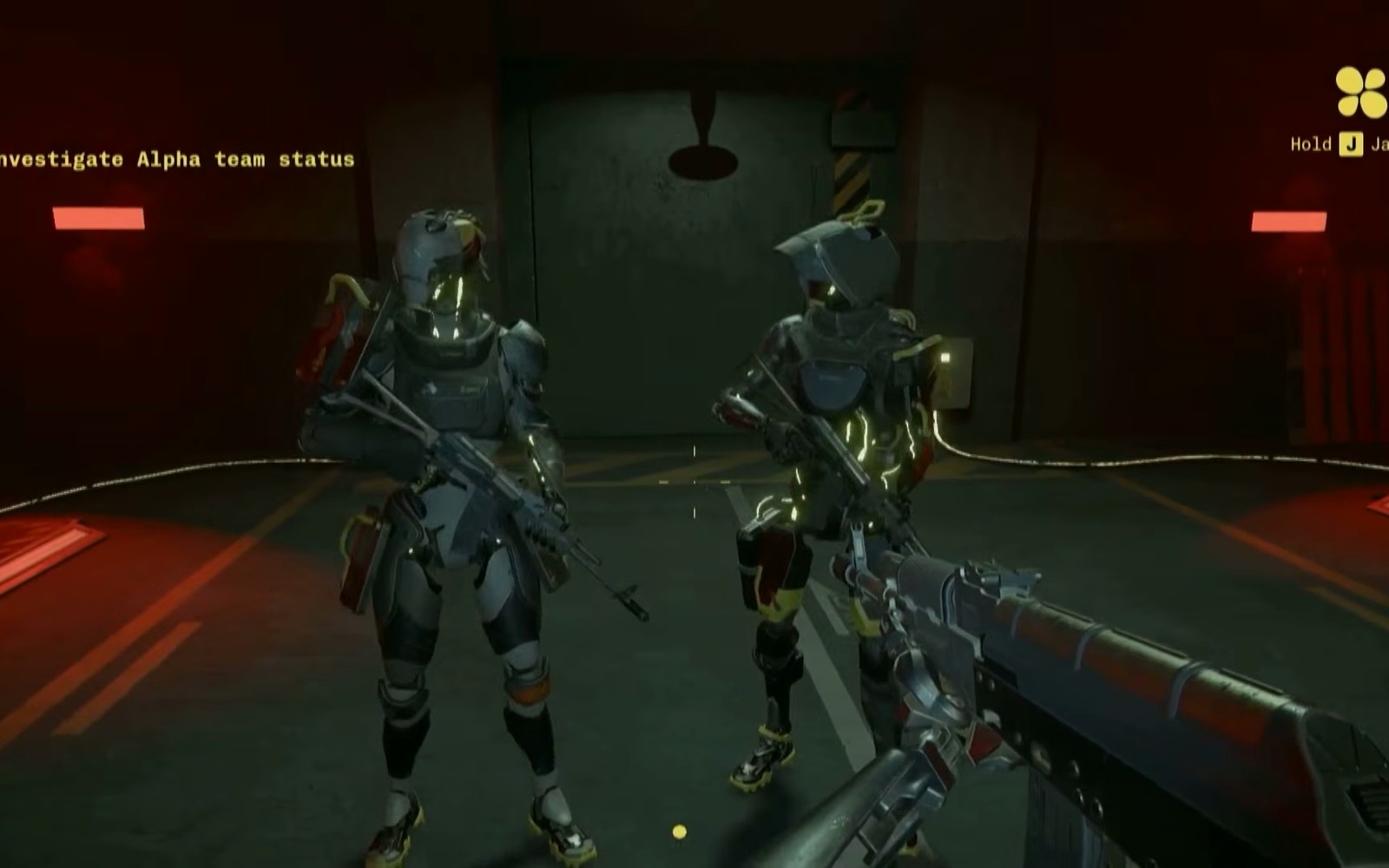}{(d) Evaluation, transparency, and deployment}{Ubisoft Teammates, first-person prototype}{Voice-addressable AI squadmates in a lab prototype}
\caption{Runtime personalization and companion systems: personalised narrative (a), adaptive characters (b), an AI teammate in a limited-time public beta (c), and AI squadmates in a closed-playtest prototype (d): (a) Interactive drama (Wu et al.), waiting-room scene \citep{wu2025enhanced}; (b) Figueiredo et al., detective prototype \citep{figueiredo2025scaffolded}; (c) PUBG Ally \citep{krafton2026allyduo}; (d) Ubisoft Teammates, first-person prototype \citep{ubisoft2025teammates}.}
\label{fig:strip-7-4}
\end{figure}

\runin{Content and narrative personalization}\figmark{\ref{fig:strip-7-4}a} Adaptation can change a difficulty parameter, select an encounter, or generate content. Experience-driven PCG \citep{yannakakis2011edpcg} and PaSSAGE \citep{thue2007passage} extend the parameter adjustments illustrated by Hamlet to content generation and narrative selection. Language models broaden how a system interprets behavior and proposes changes, while the available action or content space remains game-specific.

Profile-conditioned generation and direct changes to level structure expose different parts of the adaptation pipeline. Beyond Asking compares inferred, ground-truth, and mismatched profiles primarily using simulated players, followed by an exploratory 12-person pilot with real players \citep{lu2026personalized}. A separate approach trains a classifier on agent trajectories at three skill levels together with clustered human trajectories, collected from the authors deliberately mimicking those levels, and is designed to make targeted changes to level content from inferred skill during play. Its edits are evaluated offline, by applying skill-labeled prompts to 15 Super Mario Bros.\ levels \citep{elshamy2026adaptive}. The former tests the profile supplied to the generator, and the latter pairs skill classification with an explicit content intervention.

Deployment adds variation in starting conditions and participation. A match-three study deployed GPT-4-conditioned PCG through Google Play during a 26-day recruitment window. Among 102 players who completed at least one level, it recorded 928 level starts and 422 completions. Overall completion was 55\% under the LLM condition versus 35\% with random parameter selection, but first-level completion also differed before personal gameplay history was available. Generation quality, initial difficulty, and personalization therefore contribute to the comparison together \citep{hafnar2025zeroshot}. The system buffered three generated levels to absorb network delays, making scheduling part of the deployed adaptive experience. Separating player-model evaluation from the content policy, as discussed in \Cref{sec:model-players}, helps locate the source of improvement.

Narrative personalization changes characters' responses and story progression, with immersion and agency as evaluation targets. In LLM-based interactive drama, a player acts among characters played by language-model agents; playwriting-guided generation supports story structure, and plot-based reflection aligns the agents' reactions with the player's intentions \citep{wu2025enhanced}. The storylet approach of Drama Llama is discussed in \Cref{sec:runtime-content}.

\runin{Adaptive characters and companions}\figmark{\ref{fig:strip-7-4}b} Research on game design and agency connects meaningful choice to available actions and intelligible feedback \citep{salen2003rules,wardripfruin2009agency,ryan2006motivation}. Runtime generation can broaden those possibilities when players understand how their input is interpreted and can respond to the result. In a study of 28 players, Player-Driven Emergence identified narrative events beyond the authored premise by converting dialogue logs into graphs with GPT-4 and inspecting the resulting events. Improvised events may not be actionable and may lead to dead ends, and players with a preference for discovery and experimentation contributed more of them \citep{peng2024emergence}. In a separate Minecraft study, 28 players collaborated with two GPT-4-driven characters during 30 minutes of play. Players sometimes compensated verbally for missing visual or state information, but only 25\% completed the full quest \citep{rao2024collaborative}. Open dialogue can therefore create new opportunities for cooperation while leaving progress constrained by the character's ability to perceive and act in the game.

Persistent characters and companions raise questions that short dialogue tests cannot answer. Game-companion studies emphasize context-sensitive behavior, personality, and integration with the surrounding story \citep{emmerich2018companions}, while relational-agent research examines continuity and trust across repeated encounters \citep{bickmore2005relationships}. Proact-VL evaluates when a companion should comment or assist during streaming gameplay, using a benchmark of response timing and quality \citep{yan2026proact}.

Player studies examine how character and companion designs affect the immediate interaction. In a 10-participant voice-based detective-game study, high- and low-constraint prompts produced no reliable differences in first-play ratings. A separate synthetic experiment with an LLM judge found role-dependent effects of a JSON-and-retrieval redesign, and those results should not be read as player-experience effects \citep{figueiredo2025scaffolded}.

Larger player studies show benefits and costs under different companion roles. In a randomized between-subjects experiment with 130 players in the \emph{Campus Culture Week} prototype, players who entered free text and received LLM-generated replies reported higher perceived autonomy and cognitive load but lower usability and trust than players who selected preset options with scripted replies, with no significant difference in overall game experience \citep{hsu2026doubleedged}. LeagueBot, a voice companion providing informational and emotional support during live League of Legends matches, reduced cognitive challenge, performative challenge, and perceived tension in a within-subjects experiment with 33 novice players \citep{lee2026leaguebot}. These studies use different games, populations, and outcomes, so their contrast cannot be attributed to the companion architecture alone. A VR puzzle study shows what a within-game comparison can reveal: across four modes, from no assistance to autonomous agent play, assistance reduced workload for 24 participants, but autonomous play, despite the lowest workload, reduced challenge, autonomy, immersion, and enjoyment \citep{zhang2026vrcompanions}. Together they motivate controlled comparisons of task-focused assistance and open-ended dialogue within the same game.

Players also form models of the AI they face. A qualitative study of ten player pairs in the drawing game iNNk traced how mental models of an adversarial AI player develop along dimensions of focus and style \citep{villareale2022i}. The benefit of dialogue freedom depends on whether it helps the player accomplish and understand the current activity.

More targeted experiments isolate how generation enters play. In GenFlora, a 72-participant, $2\times2$ within-subject study varies whether generated items have dynamic game functionality and whether NPC dialogue responds to the player and generated content. Both factors improve reported presence, autonomy, and enjoyment, with no significant interaction between them \citep{yin2026contextualized}. Unlike a comparison of complete NPC architectures, this tests two integration choices within one farming-game prototype. Its short tasks support those local effects rather than a general advantage for all generative characters.

Response timing can also be tested independently of model choice. Roso et al. vary time to first token and time per output token in a 34-participant simulated RPG conversation study. Sustained token delay has more pronounced negative effects than the initial wait, and character differences also influence ratings \citep{roso2026latency}. Total response time therefore hides an important design choice: waiting before an utterance and slowing its delivery produce different conversational rhythms. This complements control-rate measurements for acting agents, where delay instead changes whether an action arrives in time.

\runin{Evaluation, transparency, and deployment}\figmark{\ref{fig:strip-7-4}c--d} Ubisoft's Teammates, following NEO NPC, combines authored characters and an FPS scenario with voice-driven dialogue and actions. Ubisoft reports a closed playtest with a few hundred players but no controlled outcome study \citep{ubisoft2025teammates}. KRAFTON's PUBG Ally ran as a public Arcade beta from 17 June to 1 July 2026 \citep{krafton2026allyduo}. Its on-device speech--language pipeline consumes textual engine observations, while a behavior tree executes actions and time-critical reactions \citep{nvidia2026pubgally}. The system paper describes a language-model agent that steers a faster control layer and is trained iteratively on records from nearly 39k sessions with real players; among surveyed players whose play was confirmed in game records, positive responses on recommending Ally exceeded negative ones by 25.1 percentage points \citep{pubg2026ally}. Teammates and PUBG Ally reuse speech and language technology but retain game-specific observations, action execution, authored character constraints, and hardware budgets. Public availability and survey responses demonstrate integration and acceptance under those conditions, but sustained cooperation or retention benefits still require controlled evidence about player outcomes.

AI Dungeon and Fortnite show two other product integrations. AI Dungeon uses player text to continue an open-ended narrative \citep{latitude2026aidungeon}; Fortnite's Darth Vader feature lets squad members address a recruitable character through voice, using Gemini 2.0 Flash for conversation and ElevenLabs Flash v2.5 for speech \citep{epic2025vader}. In one, generated text carries the evolving adventure; in the other, conversation is embedded in an existing multiplayer game. The official descriptions establish deployed interaction features, not autonomous tactical control or measured long-term player benefit. More general HCI research distinguishes the qualities sustaining use from those driving initial novelty \citep{karapanos2009uxovertime}; repeated encounters remain important even for publicly available systems. Appendix~\ref{app:resources}, \Cref{tab:runtime-systems}, indexes generated elements, execution, study populations, exposure, and outcomes.

\begin{gameaiinsight}[Resident]{More open interaction is harder to execute and to attribute}
\begin{insightpoints}
\item \textbf{Players can propose more in dialogue than the game can execute.} Player-Driven Emergence reports improvised events that can end without executable continuations \citep{peng2024emergence}; Minecraft collaborators sometimes compensate for NPC perception gaps \citep{rao2024collaborative}. Meaningful agency depends on whether the game can act on a player's contribution.
\item \textbf{Constraints can support player control.} GROMIT compiles generated code directly into the running game \citep{jennings2024gromit}, while IF:CARGO uses a constrained language with execution feedback \citep{hsu2026ifcargo}. The narrower representation makes some errors easier to explain, although interacting valid rules remain hard to diagnose. Expressiveness and intelligibility are coupled design choices.
\item \textbf{Personalized content shapes the behavior observed next.} Later behavior reflects the content already selected for the player. The match-three deployment's first-level difference illustrates why a stronger generator, easier initial content, and better profile inference cannot be credited to adaptation as one effect \citep{hafnar2025zeroshot,lu2026personalized}.
\end{insightpoints}
\end{gameaiinsight}

\gameaisectionaccent{Evaluator}
\renewcommand{\dmcolor}{Evaluator}
\pretitlemark{section}{AI That Tests and Evaluates Games}
\section{AI That Tests and Evaluates Games}
\label{sec:testing}

\begin{figure}[H]
  \centering
  \includegraphics[width=\linewidth]{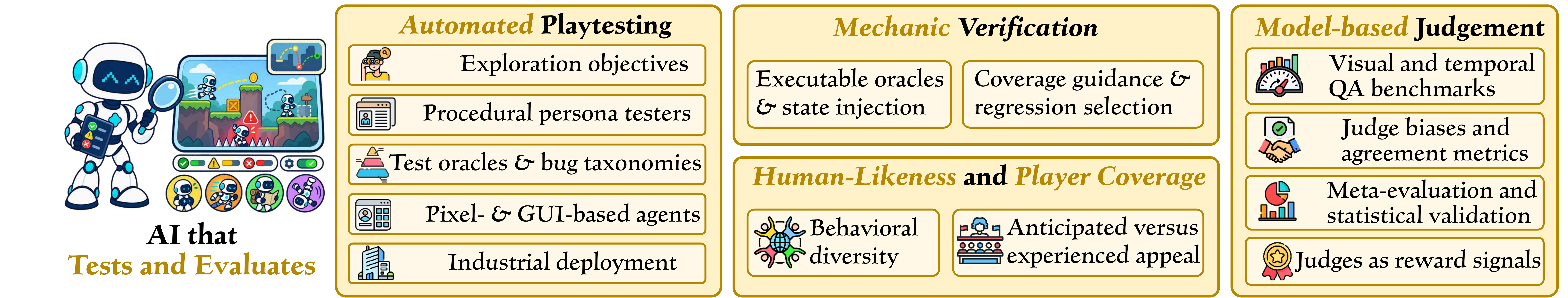}
  \caption{Research directions for AI that tests and evaluates games: automated playtesting through exploration objectives, procedural personas, and pixel- or GUI-based agents; mechanic verification with executable oracles and coverage-guided regression selection; human-likeness and player coverage; and model-based judgement with QA benchmarks, agreement metrics, and meta-evaluation.}
  \label{fig:testing-evidence}
\end{figure}

Every role above produces something that can be wrong: an action, a prediction, a design, a build, or a live response. An automated tester must both encounter relevant behavior and recognize when it is wrong. These are different capabilities: a skilled player can miss a defect, while a reliable local check may bypass the sequence that a person would need to reach it. Foundation-model systems broaden testing through language-based requirements, visual inspection, and tool use, alongside established search, reinforcement learning, and executable oracles \citep{zheng2019wuji,cai2025sage,jia2026gamegenverifier}. The useful comparisons concern how a test is directed, what observations support its verdict, and how the resulting trace or diagnosis helps development. Claims that a test predicts player experience require a human reference beyond software correctness (\Cref{fig:testing-evidence}).

\draftmap{\begin{tikzpicture}[x=1cm,y=1cm]
\node[dm q] at (-1.65,2.35) {How the chapter moves: a test finds a state, checks it against an oracle, judges it when no oracle exists, and must stand in for the right player.};
\node[dm node=3.3cm] (a) at (0,0) {\dmhead{8.1 Automated playtesting}\\[1pt]{\scriptsize objective, access, trace}};
\node[dm node=3.3cm] (b) at (4.3,0) {\dmhead{8.2 Mechanic verification}\\[1pt]{\scriptsize oracles and guidance}};
\node[dm node=3.3cm] (c) at (8.6,0) {\dmhead{8.3 Model-based judges}\\[1pt]{\scriptsize model judges, calibration}};
\node[dm node=3.3cm] (d) at (12.9,0) {\dmhead{8.4 Player representativeness}\\[1pt]{\scriptsize diversity, human reference}};
\draw[dm arrow] (a) -- (b); \draw[dm arrow] (b) -- (c); \draw[dm arrow] (c) -- (d);
\node[dm bridge=2.7cm] at (2.15,0.8) {reaching a state is not enough; an oracle checks whether behavior violates expectations};
\node[dm bridge=2.7cm] at (6.45,0.8) {without an executable oracle, a model judges};
\node[dm bridge=2.7cm] at (10.75,0.8) {a correct verdict about the wrong player is still the wrong test};
\node[dm frame,fit=(current bounding box)] (F) {};\dmtag
\end{tikzpicture}}

\subsection{Automated Playtesting}
\label{sec:test-playtesting}

\draftmap{\begin{tikzpicture}[x=1cm,y=1cm]
\node[dm q] at (-1.5,2.35) {What lets a tester reach states a player would not, and how does it know that a state is a fault?};
\node[dm node=2.8cm] (a) at (0,0) {\dmhead{Exploration objectives and coverage}\\[1pt]{\scriptsize Wuji, curiosity agents}};
\node[dm node=2.8cm] (b) at (3.6,0) {\dmhead{Procedural personas and human-like testers}\\[1pt]{\scriptsize personas, APF}};
\node[dm node=2.8cm] (c) at (7.2,0) {\dmhead{Test oracles and bug taxonomies}\\[1pt]{\scriptsize GLIB, bug taxonomy}};
\node[dm node=2.8cm] (d) at (10.8,0) {\dmhead{Pixel- and GUI-based agents}\\[1pt]{\scriptsize Inspector, Play2Code, GBQA}};
\node[dm node=2.8cm] (e) at (14.4,0) {\dmhead{Industrial deployment and evaluation practice}\\[1pt]{\scriptsize EA, iv4XR, coverage metrics}};
\draw[dm arrow] (a) -- (b); \draw[dm arrow] (b) -- (c); \draw[dm arrow] (c) -- (d); \draw[dm arrow] (d) -- (e);
\node[dm bridge=2.4cm] at (1.8,0.8) {coverage says where the tester went, not whom it resembles};
\node[dm bridge=2.4cm] at (5.4,0.8) {reaching behavior is not enough; an oracle checks whether it violates expectations};
\node[dm bridge=2.4cm] at (9.0,0.8) {oracles need observations; pixels change what is observed};
\node[dm bridge=2.4cm] at (12.6,0.8) {what runs in a studio decides what counts};
\draw[dm axis] (-1.5,-1.15) -- (15.9,-1.15) node[midway,below=1pt,font=\scriptsize,text=GameSlate] {objective $\to$ behavior $\to$ verdict $\to$ observation interface $\to$ practice};
\node[dm frame,fit=(current bounding box)] (F) {};\dmtag
\end{tikzpicture}}

Automated playtesting uses interaction to examine a game, with objectives that may include progress, state coverage, anomaly discovery, performance stress, or representative behavior. The exploration policy and the test oracle need not be learned together. Surveys distinguish approaches by their access, objectives, and validation mechanisms, while studies of practice document the continuing role of manual testing \citep{albaghajati2020assessment,politowski2021survey}. Target access, exploration mechanisms, and test oracles are compared system by system in Appendix~\ref{app:resources}, \Cref{tab:testing-systems}.

\begin{figure}[!htb]
\centering
\setlength{\tilew}{0.32\linewidth}
\setlength{\tilelabelh}{17.5mm}
\blocktile{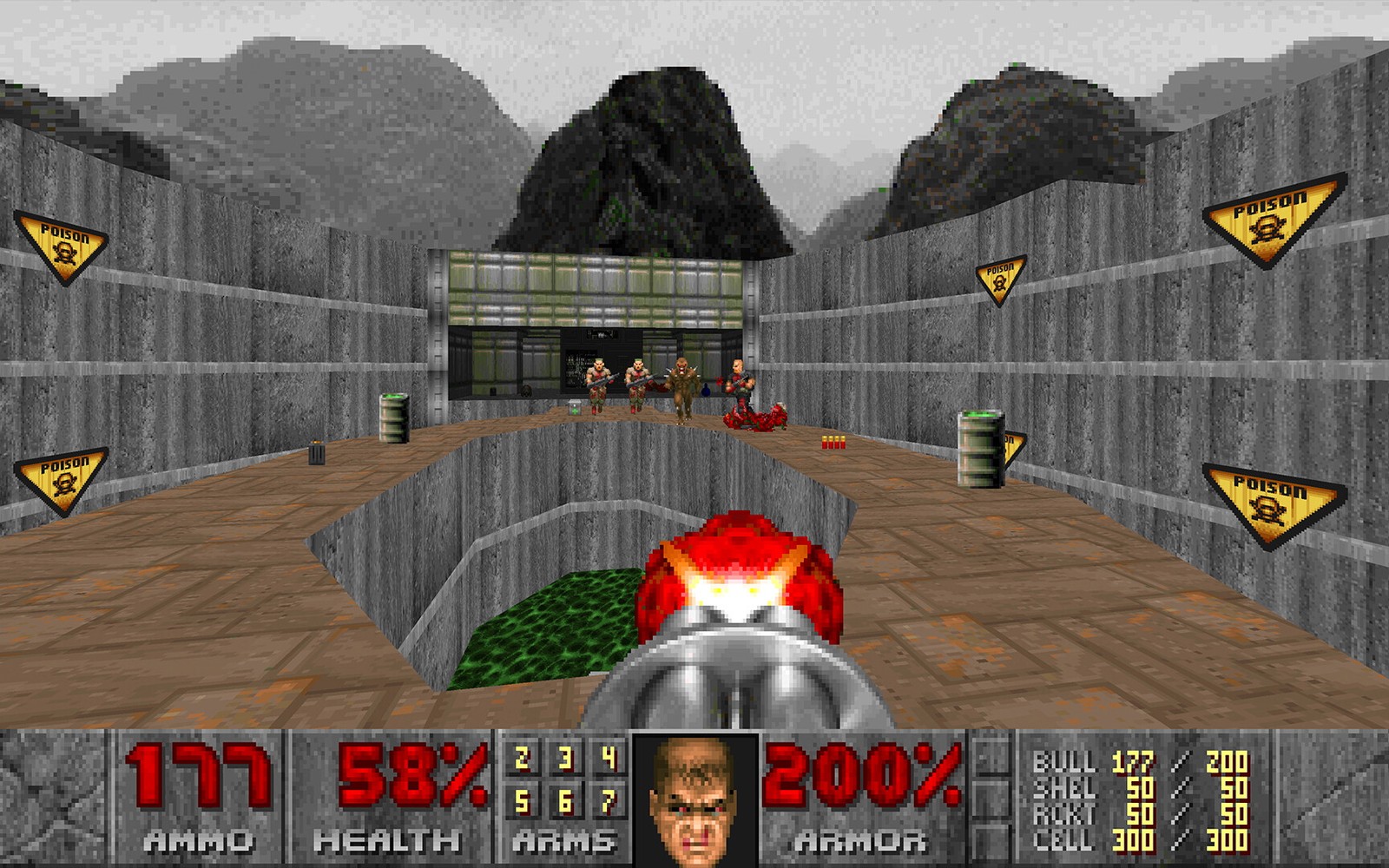}{(a) Procedural personas\\and human-like testers}{Developing personas, DOOM}{DOOM gameplay (illustrative screenshot)}\hfill
\blocktile{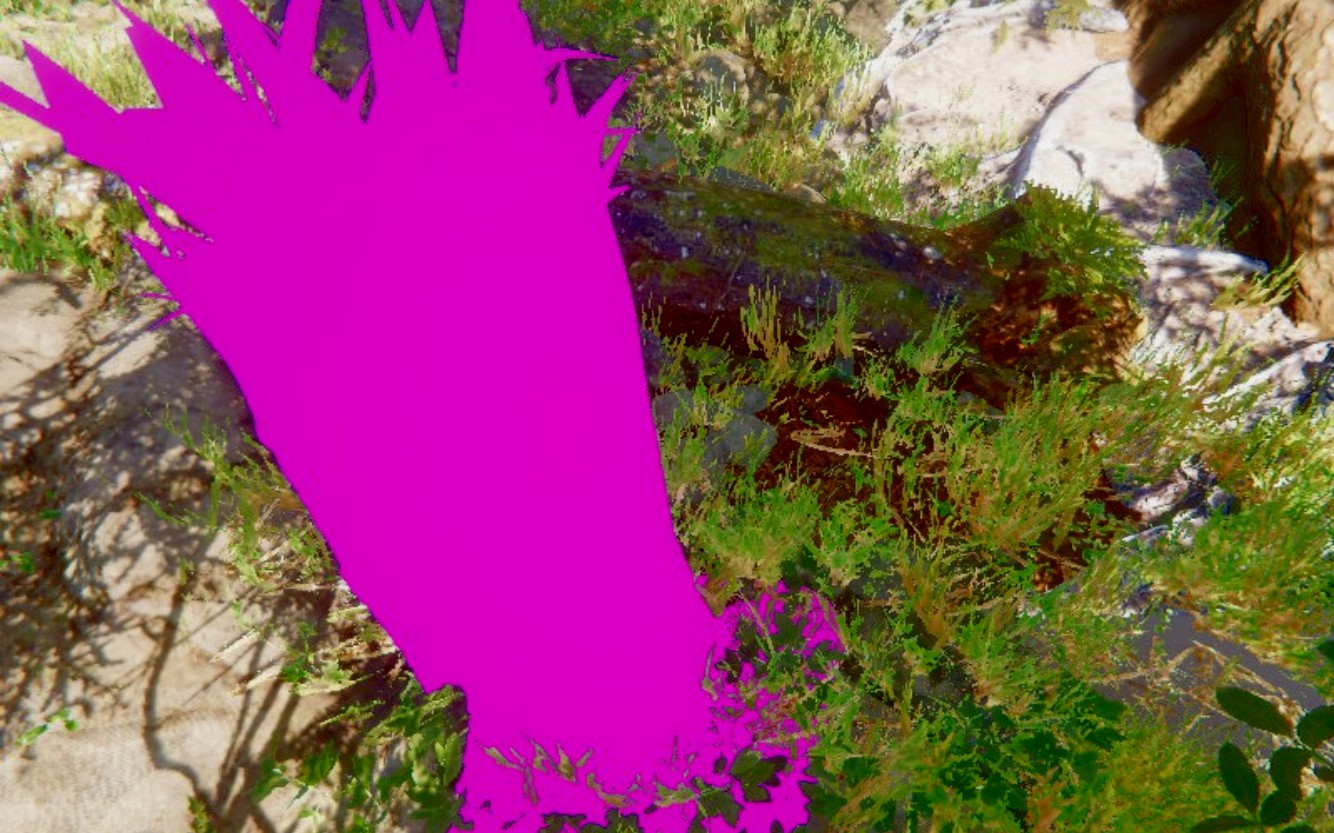}{(b) Test oracles and bug taxonomies}{Ling et al., glitch sample}{Missing texture that an oracle must flag}\hfill
\blocktile{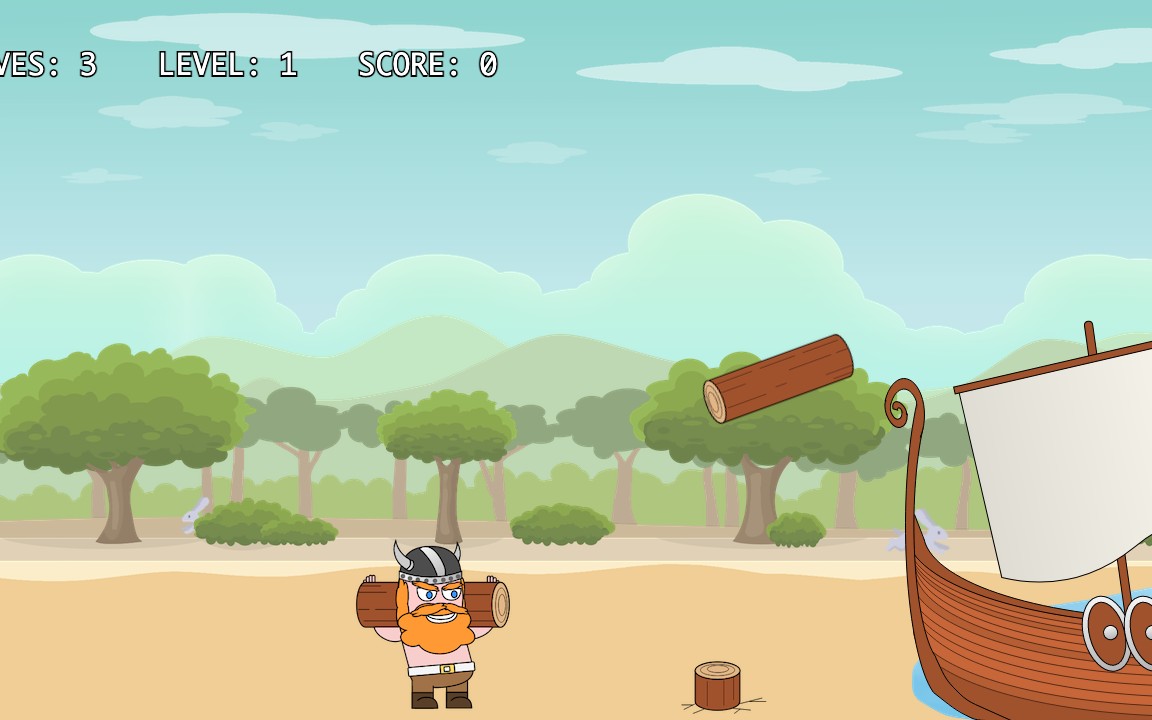}{(c) Test oracles and bug taxonomies}{Macklon et al., HTML5 canvas game}{Thrown log in the canvas test game}\\[3pt]
\blocktile{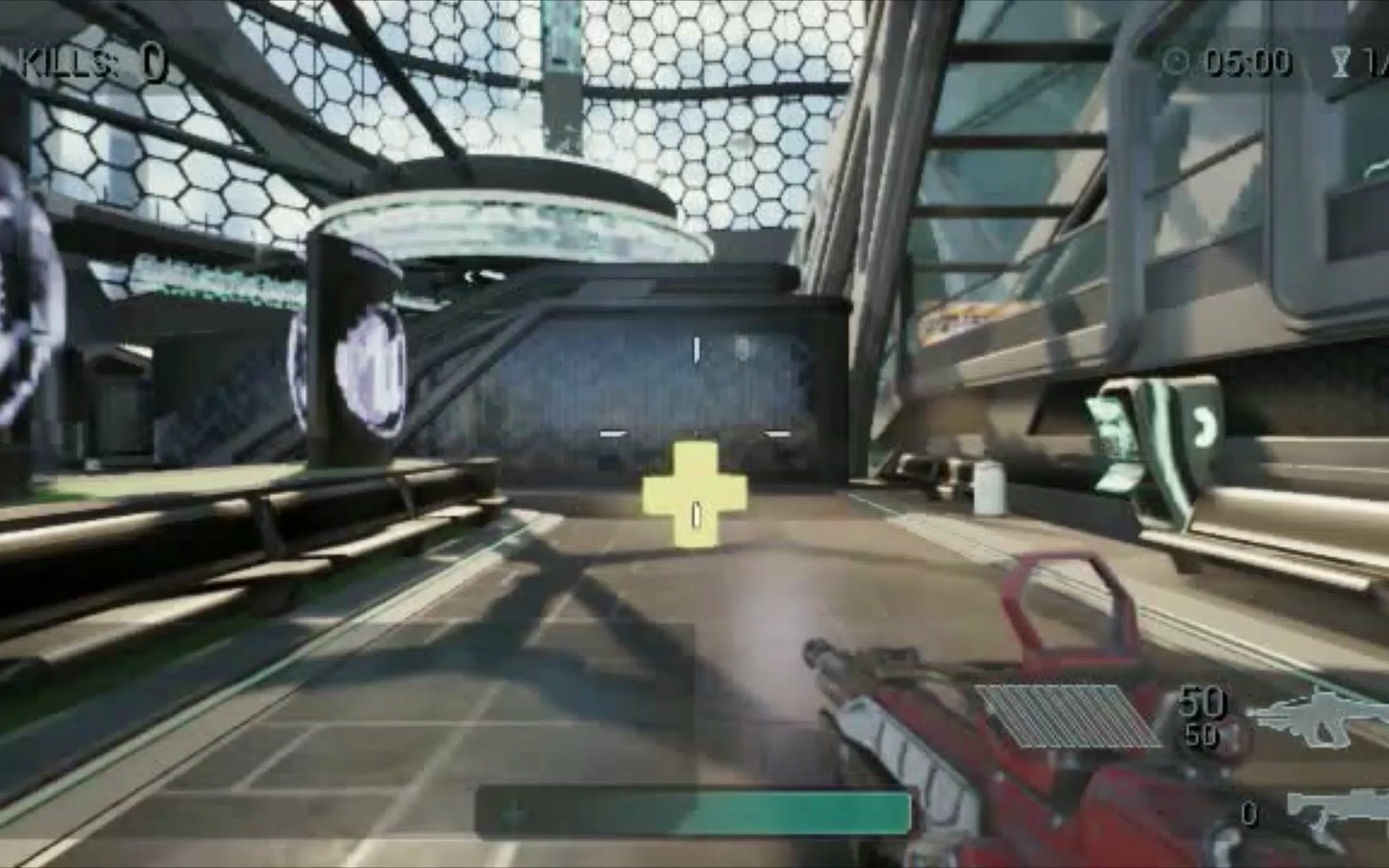}{(d) Pixel- and GUI-based agents}{Inspector, FPS test arena}{Tester observing only the raw screen}\hfill
\blocktile{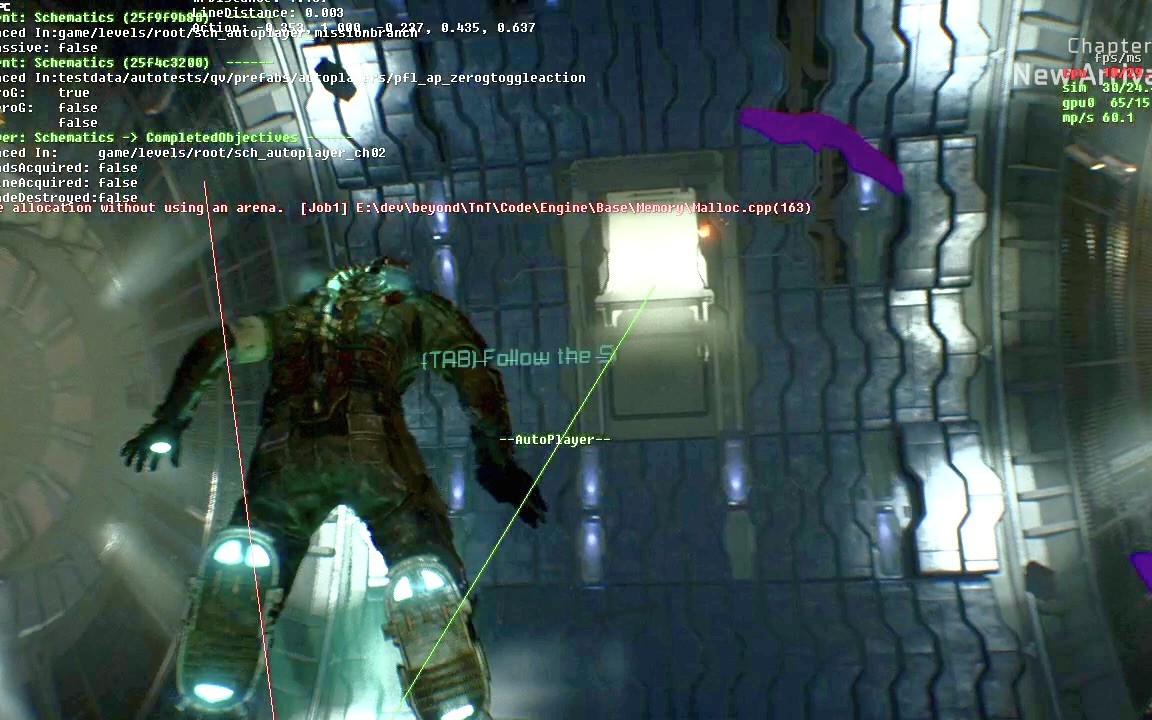}{(e) Industrial deployment\\and evaluation practice}{EA SEED AutoPlayer, Dead Space}{Zero-gravity navigation in a production build}\hfill
\blocktile{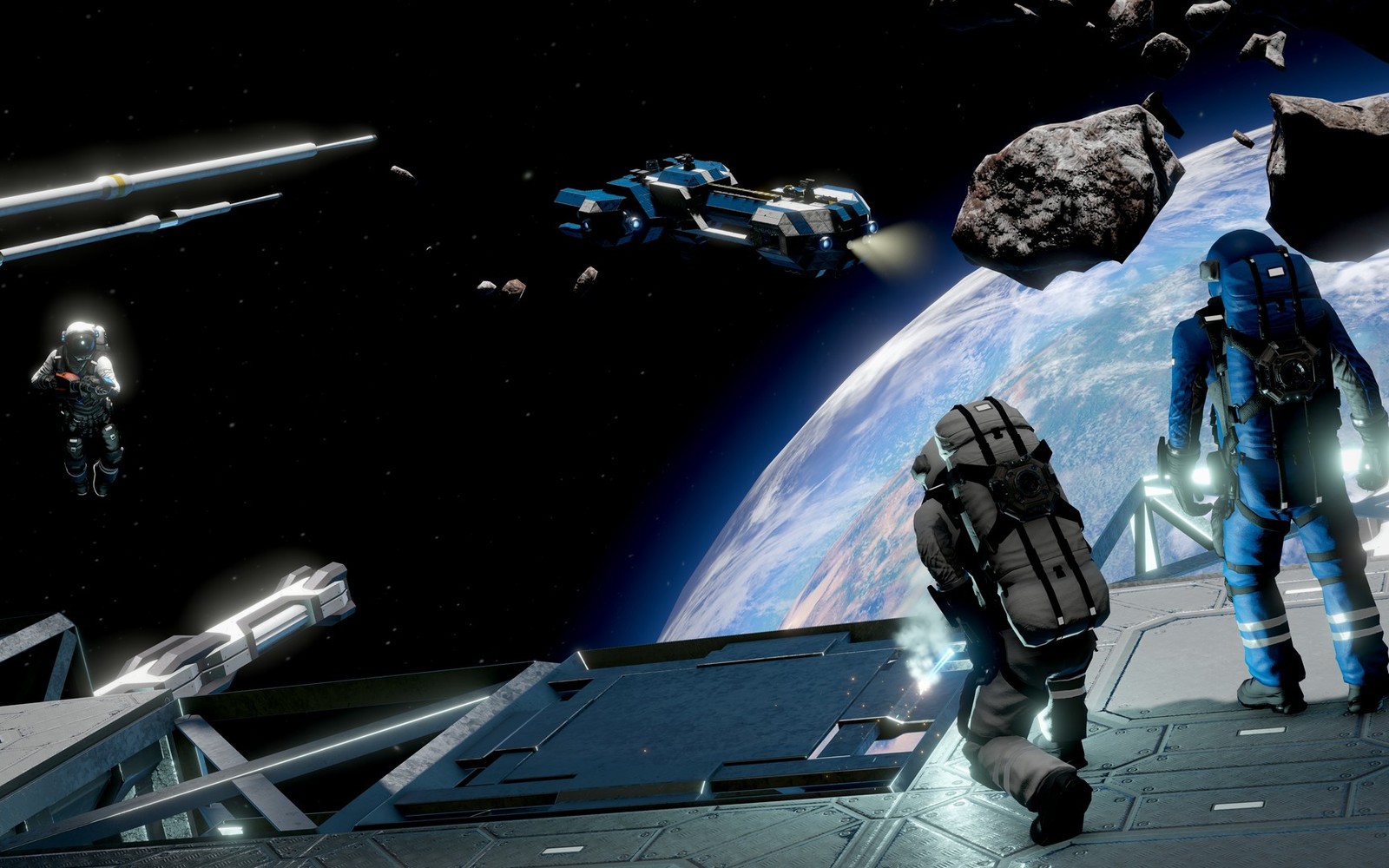}{(f) Industrial deployment\\and evaluation practice}{iv4XR, Space Engineers}{Official screenshot of the tested game}
\caption{Automated playtesting, from persona-driven exploration (a) to learned and asset-based oracles for visual defects (b--c), screen-only testers (d), production test integration at EA (e), and exploratory testing in Space Engineers (f): (a) Developing personas, DOOM \citep{ariyurek2021playtesting}; (b) Ling et al., glitch sample \citep{ling2024using}; (c) Macklon et al., HTML5 canvas game \citep{macklon2022automatically}; (d) Inspector, FPS test arena \citep{liu2022inspector}; (e) EA SEED AutoPlayer, Dead Space zero-gravity navigation \citep{gillberg2023productiontesting}; (f) iv4XR, Space Engineers \citep{prasetya2022agent}.}
\label{fig:strip-8-1}
\end{figure}

\runin{Exploration objectives and coverage} An objective of reaching states shapes the policy differently from an objective of winning. A score-maximizing policy may repeatedly follow a safe route and never exercise optional mechanics, failure recovery, or unusual action sequences. Wuji combines evolutionary search and deep reinforcement learning to balance progress with exploration, and its evaluation includes two commercial combat games and three newly discovered, developer-confirmed bugs, using four oracles derived from 1,349 real bugs, not learned verdicts \citep{zheng2019wuji}. A line of work at Electronic Arts made coverage the testing objective. Deep reinforcement learning with a user-defined reward increased test coverage, found exploits, and tested map difficulty in first-person-shooter scenarios \citep{bergdahl2021augmenting}, and curiosity-driven agents rewarded for the novelty of their actions raised game-state coverage on a complex 3D scenario where earlier exploration techniques performed poorly \citep{gordillo2021improving}. Multi-agent curiosity extends the same objective: cMarlTest deploys two cooperating agents and reports higher coverage than a single-agent variant under three coverage criteria on levels of a 3D game \citep{ferdous2025curiosity}. Code-aware guidance, which directs exploration toward target functions or changed code, is covered with verification in \Cref{sec:test-verification}. The balance is practical: exploration without progress misses later content, but progress without variation can repeatedly certify the same narrow path.

\runin{Procedural personas and human-like testers}\figmark{\ref{fig:strip-8-1}a} Procedural personas enact different play styles through search, using Monte Carlo tree search with evolved heuristics to play as distinct types of player \citep{holmgard2019playtesting}. Synthetic and humanlike tester agents make the contrast measurable: one study seeds 45 bugs across three GVG-AI games, checks test runs with a separate execution oracle, and compares the fraction each tester group detects, while 427 human trajectories provide a distinct reference for behavioral similarity \citep{ariyurek2021testing}. Two limits of the original personas have been addressed since. Developing personas progress through different goals during play instead of holding one fixed goal, and the Alternative Path Finder trains on previously tested paths and modulates the reward so that a reinforcement-learning tester is encouraged to explore alternatives to previously tested paths \citep{ariyurek2021playtesting}. RIDGE blends four persona rewards in one PPO agent with state-conditioned weights. In Crafter, its single agent reaches both the crafting and the combat branch, each at lower success than the matching fixed-persona specialist, in an evaluation with three seeds and one environment \citep{shifan2026ridge}. Personas broaden the behaviors a test exercises. Whether that breadth represents real players is the subject of \Cref{sec:test-representativeness}.

\runin{Test oracles and bug taxonomies}\figmark{\ref{fig:strip-8-1}b--c} Test oracles decide whether an observed behavior is a fault; a bug taxonomy specifies the kinds of faults to look for. A multivocal review that selected 189 of 436 sources derived a taxonomy of 63 game bug categories, validated with industry practitioners, so that testers and researchers can name what a test is meant to find \citep{butt2023deriving}. Learned oracles extend detection to patterns that are difficult to encode as fixed rules. GLIB targets graphical glitches, which a study of NetEase bug reports found to be frequent and to escape crash-only testing, and provides an automated oracle for graphically rich applications \citep{chen2021glib}. For HTML5 canvas games, whose contents are absent from the DOM and which snapshot oracles handle poorly, an asset-based oracle instead compares rendered objects with the game's own graphics assets \citep{macklon2022automatically}. Supervised classifiers detect rendered texture glitches at 86.8\% accuracy with a 6.7\% false-positive rate on generated data \citep{ling2024using}, and co-finetuning on labeled and unlabeled data from several games reduces the dependence on labeled bugs from the target game \citep{yi2025hybrid}. Oracles also cover performance: an RL agent can be trained to reach demanding 3D scenes, where frame rate is then measured as the test property \citep{tufano2022using}.

\runin{Pixel- and GUI-based agents}\figmark{\ref{fig:strip-8-1}d} Early testing agents relied on game-internal state, which requires deep integration with each game. Inspector uses only screenshots, combining a game-space explorer, a key-object detector, and a human-like object investigator, so it needs no game-internal state. All three components still learn from the game under test, the detector and investigator from labeled screenshots and human demonstrations \citep{liu2022inspector}. A preference-conditioned pixel agent lets test engineers steer exploration toward a preferred style, such as the golden path, while keeping the state representation pixel-based \citep{abdelfattah2023preference}. Language-model testers inherit the same constraint from the other direction: Lap targets non-text games that lack APIs, where the model cannot naively read game state \citep{zhao2025llm}. Multimodal agents now extend testing to rendered interfaces and generated games. PlaytestArena uses a GUI agent to play browser games and judge observed behavior against task-specific rubrics \citep{huang2026guigames}. Play2Code instead withholds those scoring rubrics from its playtester: the GUI agent returns observations and suggested fixes to a coding agent for the next software revision \citep{huang2026guigames}. In PlayWorld, an agent adapts human-annotated reference actions to pursue comparable objectives across learned environments, while a separate visual-question-answering verifier scores the rollouts \citep{ding2026playworld}. GBQA measures the whole pipeline, with 30 games and 124 human-verified injected bugs across three difficulty levels and a ReAct-plus-memory baseline agent \citep{jiang2026gbqa}. TITAN, which couples long-horizon task execution with a language-model bug oracle in commercial MMORPGs, is discussed with verification in \Cref{sec:test-verification}.

Rendered-interface testing reduces dependence on internal instrumentation but makes exploration and diagnosis less direct. The agent must discover what is interactive, distinguish a missed input from a broken mechanic, and remember the sequence leading to a symptom. The resulting trace is useful beyond the final verdict: a coding agent or developer can replay it to localize a defect. This explains why Play2Code can use playtesting for repair even though its playtester does not receive the benchmark's scoring rubric.

\runin{Industrial deployment and evaluation practice}\figmark{\ref{fig:strip-8-1}e--f} Practice surveys document a substantial role for manual testing alongside growing automation \citep{politowski2021survey,roque2025literature}. In Battlefield 2042 and Dead Space, EA used reinforcement learning for difficult navigation while retaining scripted bots for other behavior. Structured observations reduced rendering dependencies, but the learned policies still required training and integration with changing builds \citep{gillberg2023productiontesting}. The iv4XR framework takes an agent-programming approach, combining goal-directed agents with conventional testing algorithms in applications including Space Engineers \citep{prasetya2022agent}. EA's deployments and iv4XR's integration approach make task selection and instrumentation central to the testing workflow. Hybrid testers divide the work accordingly. Scripted actions can reliably initialize a scenario or perform a known interaction, and a learned navigation policy can reach a location that would be expensive to script across changing geometry. Instrumented observations simplify perception and permit faster execution, while rendering can be enabled for failures that require visual inspection. Reusing this arrangement across builds still requires checking that observations, controls, and target conditions retain their meanings. Production integration is therefore a separate achievement from high return in a fixed training environment.

\subsection{Software and Mechanic Verification}
\label{sec:test-verification}

\draftmap{\begin{tikzpicture}[x=1cm,y=1cm]
\node[dm q] at (-2.4,2.35) {How is a specific mechanic checked, and what brings the tester to it?};
\node[dm node=4.6cm] (a) at (0,0) {\dmhead{Executable oracles and state injection}\\[1pt]{\scriptsize GameGen-Verifier, GameEngineBench, GameWorld, Agent2World}};
\node[dm node=4.6cm] (b) at (6.4,0) {\dmhead{Coverage guidance and regression selection}\\[1pt]{\scriptsize CA2, SAGE, TITAN, SMART}};
\draw[dm arrow] (a) -- (b);
\node[dm bridge=3.6cm] at (3.2,0.8) {an oracle checks the observed behavior; guidance decides which states are reached};
\draw[dm axis] (-2.4,-1.15) -- (8.8,-1.15) node[midway,below=1pt,font=\scriptsize,text=GameSlate] {what judges, then what guides};
\node[dm frame,fit=(current bounding box)] (F) {};\dmtag
\end{tikzpicture}}

Mechanic verification connects an expected behavior to an observable result. An executable assertion may inspect state directly; a visual judge may infer the result from rendered interaction. Either approach also needs a way to reach or construct the relevant test condition. Code structure and update history can guide that selection.

\runin{Executable oracles and state injection} Software tests can inspect states that ordinary play may not reach. GameGen-Verifier injects target states and runs bounded checks around specified mechanics. Its verifier combines a vision-language judge with programmatic assertions where available. Across 100 generated web games, the best configuration reports 92.2\% specification-label agreement with expert annotations under Acc@5, computed over five repeated runs, versus 58.8\% for a coverage-enforced baseline, with up to 16.6$\times$ lower wall-clock time \citep{jia2026gamegenverifier}. This is agreement on mechanic checks, not the fraction of fully correct games. State injection also changes what is tested: it can isolate a transition while bypassing the progression needed to reach it. Injected states must respect the game's invariants, or be identified as robustness tests, and successful local checks still leave ordinary reachability to be tested.

Deterministic checks provide evaluation infrastructure. GameEngineBench uses hidden Play-in-Editor tests, reviewed by an LLM judge, to score development agents \citep{la2026gameenginebench}. GameWorld evaluates 170 tasks across 34 browser games using state-verifiable task metrics, with both keyboard/mouse and semantic-action interfaces \citep{ouyang2026gameworld}. Agent2World instead generates unit-test and simulation evidence while executable world models are being constructed \citep{hu2025agent2world}. Handwritten, hidden, and model-generated tests differ in independence from the implementation.

Narrative constraints also support targeted verification. NCP-Bench checks narrator responses against an explicit fact ledger and fixed plot commitments across 100 movie-derived environments. Prompt-fixed LLM auditors detect contradictions, update facts, and track commitment satisfaction during adversarial player interventions \citep{ma2026ncpbench}. The best reported narrator remains conflict-free in 42\% of environments after 20 turns under this protocol \citep{ma2026ncpbench}. The test compares generated interaction with a stated narrative specification, with verdicts that depend on the auditors' interpretation; it measures continuity under simulated interventions rather than longitudinal player experience. FAIRGAMER supplies a complementary behavioral diagnostic by testing social bias in NPC decisions \citep{shi2026fairgamer}; its human-relevance implications are discussed in \Cref{sec:test-representativeness}.

\runin{Coverage guidance and regression selection} Code-aware and regression testing use different sources of guidance. CA2 uses call-stack traces to guide reinforcement-learning testers toward target functions in state- and image-based environments \citep{adaikkappan2026ca2}. SAGE combines LLM-guided exploration, compact test-suite construction, and update-log-based prioritization. Its Overcooked Plus and Minecraft experiments use predefined bug-trigger predicates: language interpretation selects tests, while these predicates determine detection \citep{cai2025sage}. TITAN couples long-horizon task execution with an LLM bug oracle. In two commercial MMORPGs, its authors report 95\% success across 100 runs (20 tasks, five runs per task) and four previously unknown bugs, and separately report adoption in eight QA pipelines \citep{wang2025titan}. Target reachability, regression selection, and bug classification therefore improve different parts of the testing process.

SMART makes a further connection between code changes and testing intent. It uses structural code differences to identify changed behavior, translates that context into functional goals, and combines coverage with semantic guidance in reinforcement-learning rewards \citep{mu2025smart}. Its custom Overcooked and Minecraft experiments study reaching update-relevant behavior with instrumented access. CA2 targets functions, SAGE manages regression effort, and SMART uses changes to shape exploration; each grounds testing in game or code structure.

The usefulness of code-aware guidance depends on how closely coverage matches the intended behavior. Reaching a target function establishes execution of that code, but not necessarily a rare branch, a particular parameter combination, or the player-visible effect of interest. Update logs can suggest such targets even when the tester lacks source access; structural differences can locate changed implementation when source is available. Compare guidance by faults and behaviors exposed at a matched testing budget, not target coverage alone.

A useful regression record links the build version and initial state to an action sequence, an expected result, and the observed failure. This supports both reproduction and diagnosis. After an update, the expected result may need revision: an old trace can become invalid because the requirement changed, not because a defect was introduced. SAGE \citep{cai2025sage} selects update-relevant tests and SMART \citep{mu2025smart} steers exploration toward changed code, while checked oracles decide whether the new outcomes are faults. Retaining a reproducer connects detection to repair without assuming that the testing agent can perform the repair itself.

\subsection{Model-Based Judges}
\label{sec:test-judges}

\draftmap{\begin{tikzpicture}[x=1cm,y=1cm]
\node[dm q] at (-1.65,2.35) {Without an executable oracle, what can a model judge, and when can its verdict be trusted?};
\node[dm node=3.3cm] (a) at (0,0) {\dmhead{Visual and temporal QA benchmarks}\\[1pt]{\scriptsize GlitchBench, VideoGameQA-Bench, VideoGlitchBench}};
\node[dm node=3.3cm] (b) at (4.3,0) {\dmhead{Judge biases and agreement metrics}\\[1pt]{\scriptsize self-preference, CALM, MLLM-as-a-Judge}};
\node[dm node=3.3cm] (c) at (8.6,0) {\dmhead{Meta-evaluation and statistical validation}\\[1pt]{\scriptsize JudgeBench, alt-test}};
\node[dm node=3.3cm] (d) at (12.9,0) {\dmhead{Judges as reward signals}\\[1pt]{\scriptsize Agent-as-a-Judge, reward hacking}};
\draw[dm arrow] (a) -- (b); \draw[dm arrow] (b) -- (c); \draw[dm arrow] (c) -- (d);
\node[dm bridge=2.7cm] at (2.15,0.8) {a benchmark score hides which errors the judge makes};
\node[dm bridge=2.7cm] at (6.45,0.8) {a known bias still needs a statistical bar to clear};
\node[dm bridge=2.7cm] at (10.75,0.8) {a judge that trains the generator can be gamed};
\draw[dm axis] (-1.65,-1.15) -- (14.55,-1.15) node[midway,below=1pt,font=\scriptsize,text=GameSlate] {what the judge can do $\to$ how it fails $\to$ how failure is measured $\to$ what happens when its verdict is optimized against};
\node[dm frame,fit=(current bounding box)] (F) {};\dmtag
\end{tikzpicture}}

Model-based judges are useful where correctness cannot be fully encoded as an executable oracle: interpreting a visual glitch, matching an interaction to a requirement, or assessing an open-ended design. Their flexibility makes validation important. Game-specific benchmarks test visual and temporal recognition, while broader judge research helps explain bias and the risks of using verdicts as optimization feedback \citep{gu2024survey,li2024generation}.

\begin{figure}[!htb]
\centering
\setlength{\tilew}{0.47\linewidth}
\setlength{\tilelabelh}{14.0mm}
\blocktile{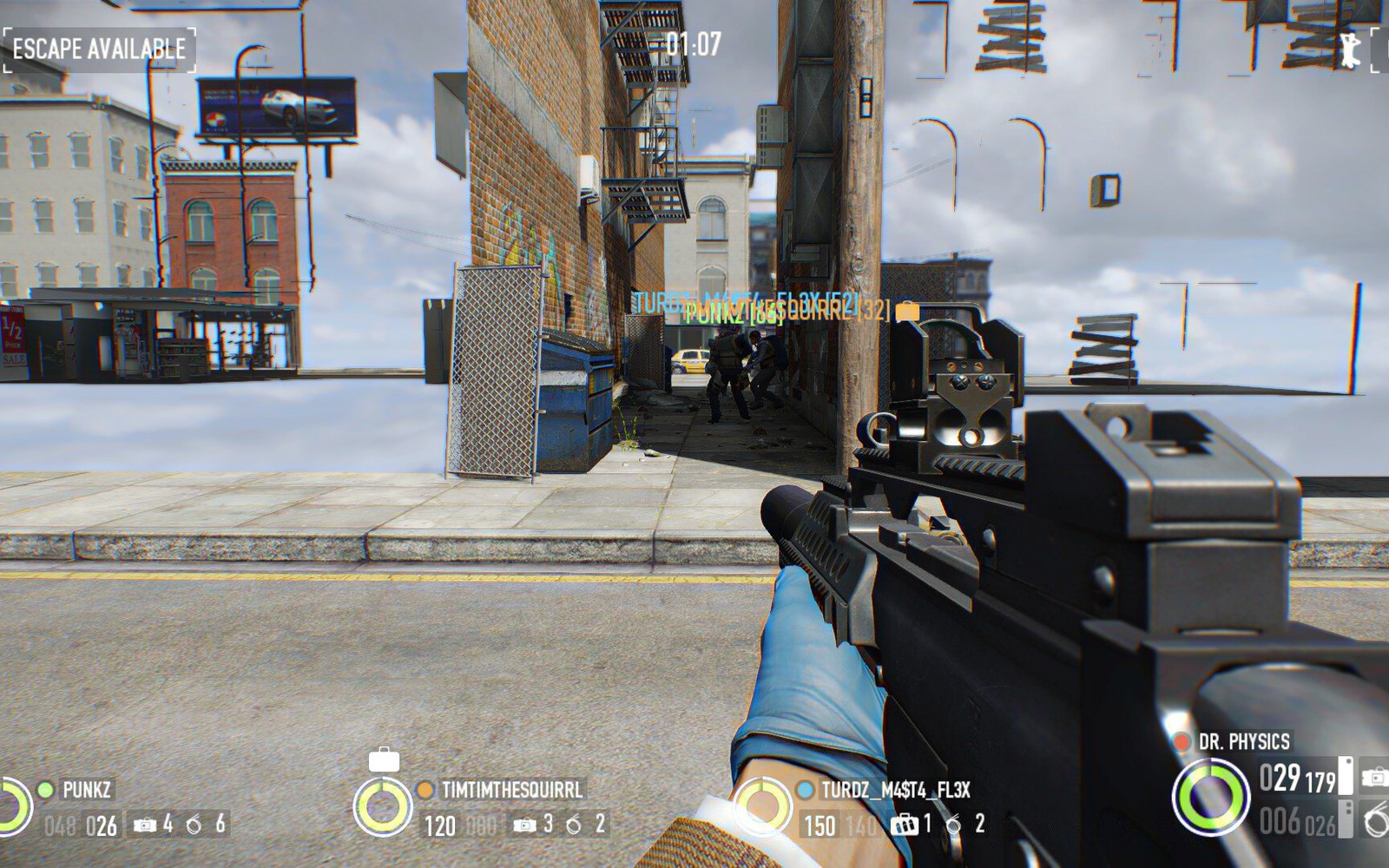}{(a) Visual and temporal QA benchmarks}{VideoGameQA-Bench, co-op FPS heist}{Render corruption a model must detect}\hfill
\blocktile{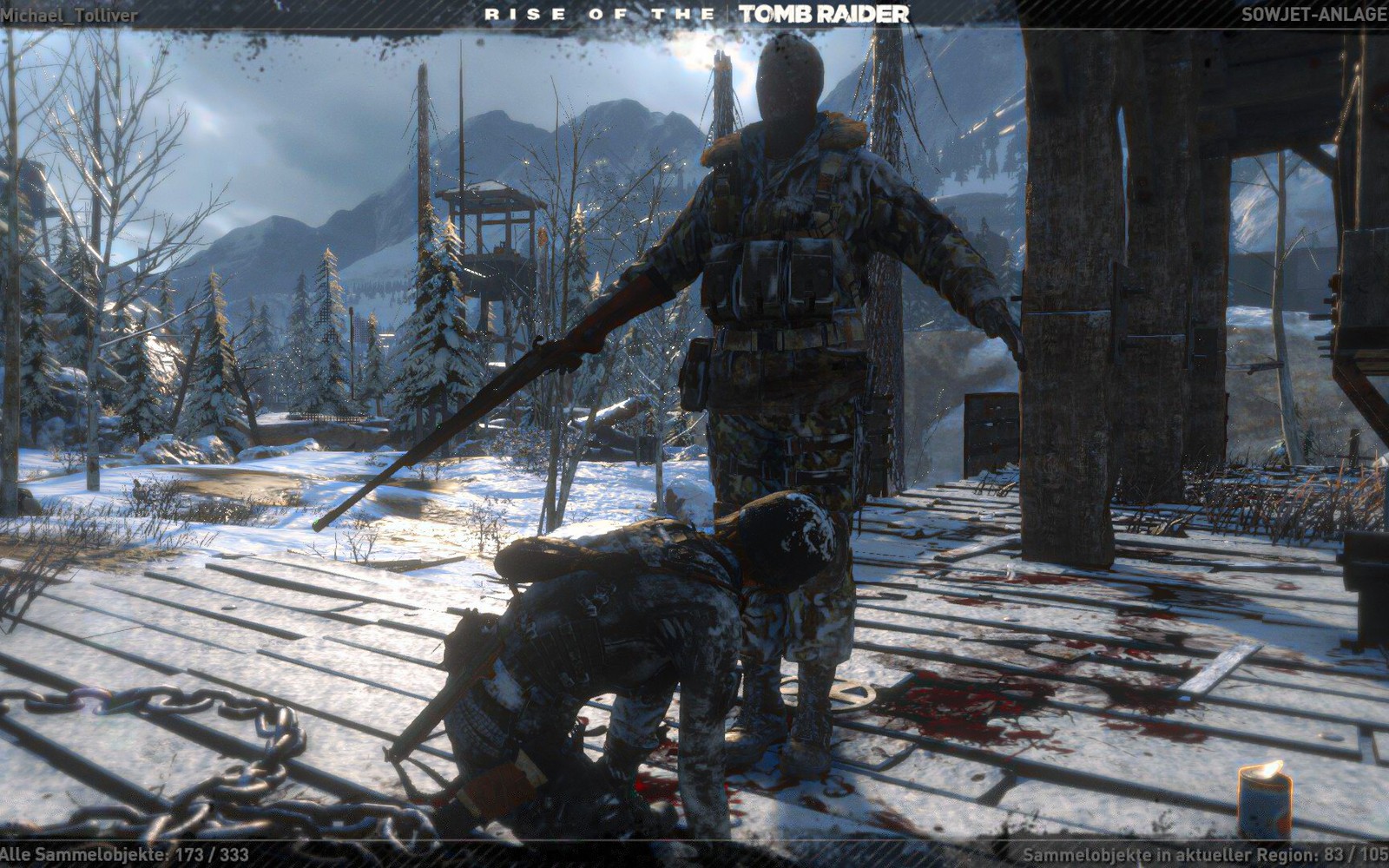}{(b) Visual and temporal QA benchmarks}{VideoGameQA-Bench, Rise of the Tomb Raider}{Frozen T-pose versus normal posture}\\[3pt]
\blocktile{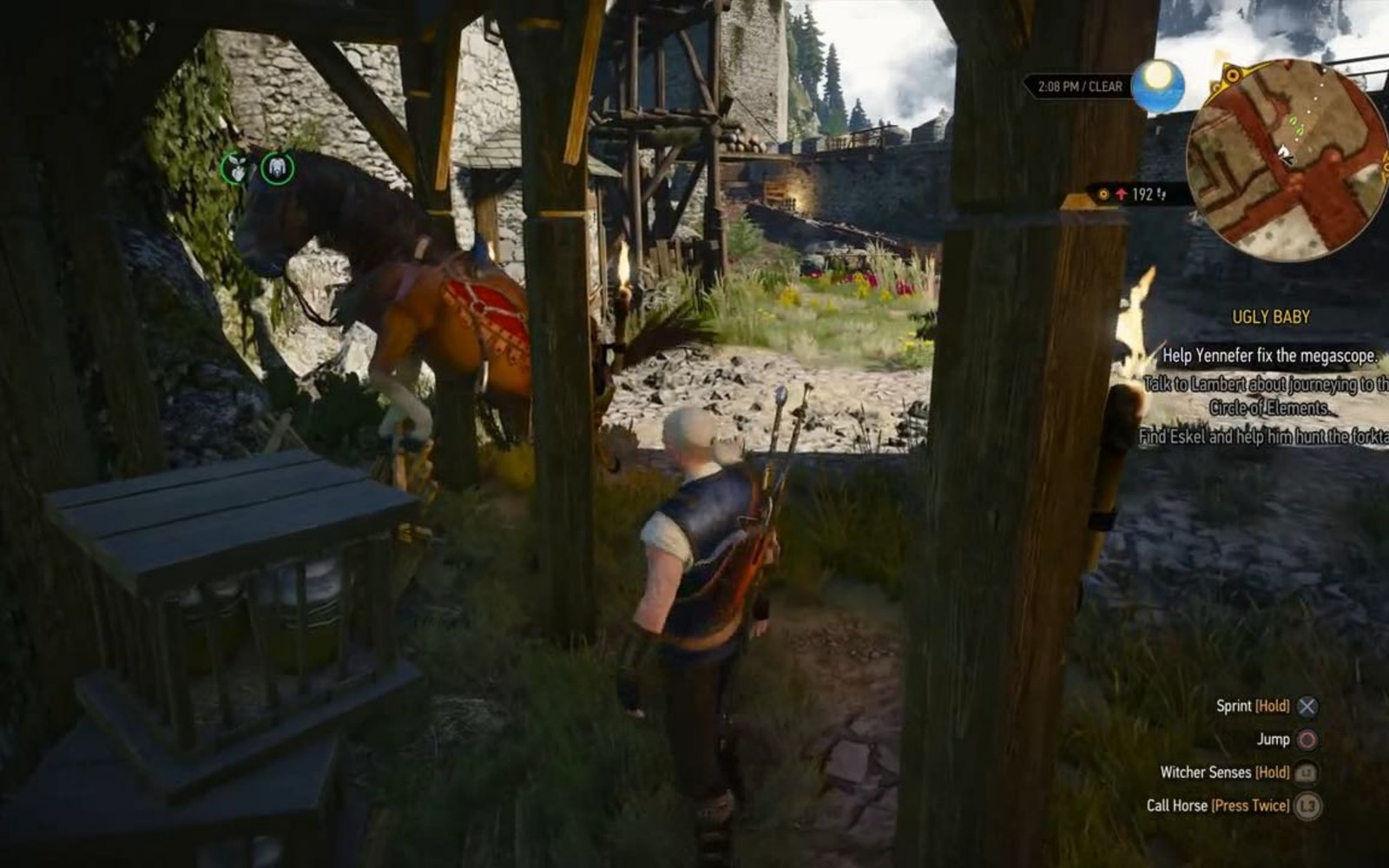}{(c) Visual and temporal QA benchmarks}{GlitchBench, The Witcher 3}{Horse clipped into scenery}\hfill
\blocktile{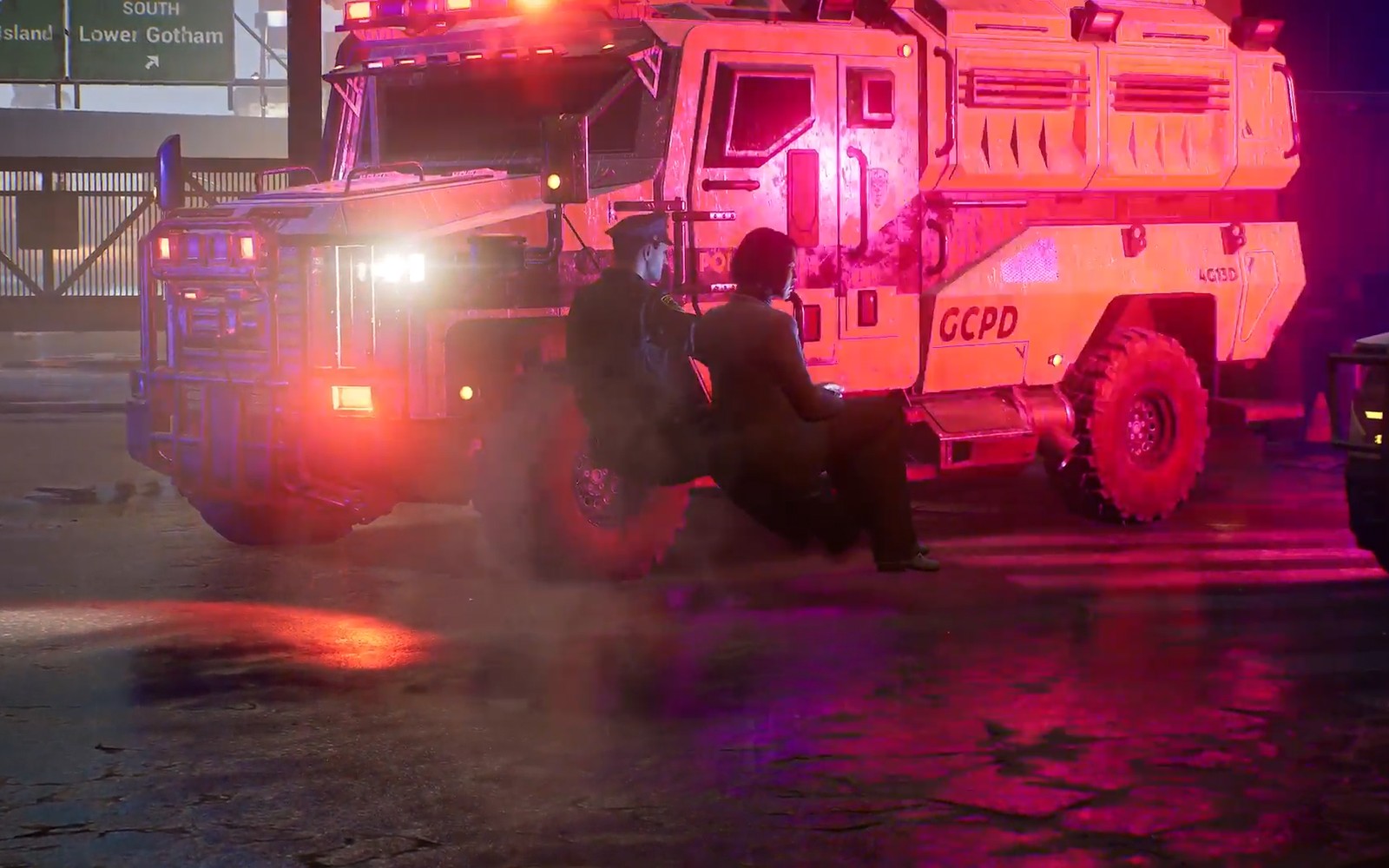}{(d) Visual and temporal QA benchmarks}{GlitchBench, Gotham Knights}{Civilian floating in mid-air}
\caption{Benchmark cases in which a model must recognize a visual defect in a screenshot: (a) VideoGameQA-Bench, co-op FPS heist \citep{taesiri2025videogameqa}; (b) VideoGameQA-Bench, Rise of the Tomb Raider \citep{taesiri2025videogameqa}; (c) GlitchBench, The Witcher 3 \citep{taesiri2023glitchbench}; (d) GlitchBench, Gotham Knights \citep{taesiri2023glitchbench}. For GlitchBench (c--d), the evaluated model describes the screenshot, and a separate text judge compares its response with the reference description.}
\label{fig:strip-8-3}
\end{figure}

\runin{Visual and temporal QA benchmarks}\figmark{\ref{fig:strip-8-3}a--d} Visual QA benchmarks test whether models can recognize an anomaly, locate it in time, and explain the fault. GlitchBench uses unusual and glitched game scenarios to test whether large multimodal models detect and interpret out-of-the-ordinary events, reporting that the task remains challenging for state-of-the-art models \citep{taesiri2023glitchbench}. VideoGameQA-Bench evaluates nine visual QA tasks, including UI checks, visual regression, glitch detection, temporal localization, and bug reporting. Its best reported visual-regression accuracy is 45.2\%, despite stronger glitch-recognition results \citep{taesiri2025videogameqa}. VideoGlitchBench examines open-ended detection over continuous video, with 5,238 gameplay videos from 120 games annotated with glitch descriptions and temporal spans. The evaluated model must distinguish true glitches from unusual but valid in-game events \citep{zheng2026open}.

Industrial footage tests visual inspection under a different distribution of scenes and faults. Across 100 videos totaling 41 hours and 19,738 keyframes, a single-prompt vision--language baseline reaches a mean per-video precision of 0.50 and accuracy of 0.72 \citep{lu2026far}. Reference-guided detection compares each test frame with an earlier frame of the same video. On real gameplay, automatically selected references improve frame-level detection modestly, and unreliable references remain a source of error \citep{yu2026resp}. This uses temporal context to address scene variation. These studies use different datasets and evaluation protocols, so their scores describe performance within each setting. Vision--language models have also been explored for canvas applications with procedurally generated graphics, where no asset-based ground truth exists for direct comparison \citep{macklon2025exploring}.

An acting tester must also reach the behavior that a passive visual judge inspects. For interactive testing, PlaytestArena reports 87.9\% raw criterion agreement with human judgments across its 200 games, with Cohen's $\kappa=0.66$ \citep{huang2026guigames}. This evaluates judgments grounded in play traces; passive visual QA alone leaves the exploration requirement untested. Auditing interactive 3D worlds adds navigation and triggering to recognition. Six fine-tuned 4B explorers with an offline frontier reviewer succeed on 55.9\% of 213 auditing tasks, matching the 54.9\% of the reviewer model exploring alone at 4.6 times lower cost, against 83.4\% for human auditors \citep{yang2026worldauditor}.

Temporal evidence is especially important for distinguishing symptoms from causes. A screenshot can reveal a missing object or visual artifact, but a failed interaction may require comparing the state before input, the action actually executed, and the later response. A useful report identifies the triggering sequence and violated requirement, not just an unusual image.

\runin{Narrative and logical bug detection} Text-based generative games introduce faults that need not appear in graphics or crash a program. Jin et al. analyze DejaBoom! gameplay logs for errors caused by hallucination, forgetting, and misinterpretation, comparing LLM-based detection with bugs reported by players \citep{jin2024bugdetection}. This complements the same game's study of emergent player narratives \citep{peng2024emergence}: an improvised event may be a useful addition or a contradiction of the game's premise. The judge needs both prior interaction and the relevant design constraints to distinguish them. Unlike an executable assertion, its verdict interprets whether the narrative remains consistent; unlike a preference rating, it targets an identifiable deviation. Dialogue-log inspection extends coverage but still needs state checks.

\runin{Judge biases and agreement metrics} A generator and evaluator can share errors, and language-model evaluators may prefer their own outputs \citep{panickssery2024selfpreference}. Broader judge studies document task-dependent biases and differences between pairwise comparison, absolute scoring, and ranking \citep{ye2024justice,chen2024mllm}, and evaluation criteria also affect judge reliability \citep{yamauchi2025empirical}. In game evaluation, this means that agreement on attractive screenshots cannot be transferred automatically to judgments of mechanics or interaction traces. Clear criteria help isolate the target, while engine checks and human annotations can reveal errors shared by generator and judge.

\runin{Meta-evaluation and statistical validation} Raw agreement depends on the prevalence of faults: a judge that usually approves output may appear accurate when defects are rare. Per-category detection, false positives, and chance-adjusted agreement reveal different weaknesses. Repeated runs measure variability, not independence. JudgeBench distinguishes objective correctness from preference \citep{tan2024judgebench}, and the alternative annotator test evaluates whether a model can substitute for human annotation on a specified task \citep{calderon2025alternative}. These are methodological precedents; a game judge still needs validation against the relevant game evidence. SWE-Game supplies such a validation for game construction: on human-labeled behaviors from 100 agent-built games, executable checks achieve 92.59\% balanced accuracy compared with 78.41\% for a video-based VLM judge, while rubric-based visual scores reach a Spearman correlation of 0.829 with human ratings of 200 gameplay clips \citep{chen2026swegame}. The result supports assigning mechanic correctness to executable checks and presentation to calibrated model judges. Disagreement can itself be informative. An assertion may confirm a state update while the rendered feedback incorrectly suggests failure. Conversely, an animation may look successful without changing inventory. Retaining the requirement, trace, and oracle output makes such cases diagnosable rather than reducing them to conflicting scores. Human judgments are needed for experience claims, but need not replace executable checks of a well-defined rule.

\runin{Judges as reward signals} A verdict used for repeated revision creates a different risk from a one-off score. GameGen-Verifier, Play2Code, and PlayWorld place automated assessment within construction, repair, or simulation workflows. More general agentic evaluation likewise supplies intermediate feedback during development \citep{zhuge2024agent}. Research on imperfect verifiers \citep{helff2026llms} and tool-use reward hacking \citep{thaman2026reward} shows how optimization can exploit scoring shortcuts. These findings motivate a specific test for game-generation loops: after optimization against one judge, does the improvement survive withheld interaction scenarios, independent engine checks, and human review? Reusing the optimization judge as the only final evaluator leaves that validation question unanswered.

\subsection{Player Representativeness and Human Relevance}
\label{sec:test-representativeness}

\draftmap{\begin{tikzpicture}[x=1cm,y=1cm]
\node[dm q] at (-2.4,2.35) {Whose behavior does the test stand in for, and against which human reference is a judgment checked?};
\node[dm node=4.6cm] (a) at (0,0) {\dmhead{Behavioral diversity}\\[1pt]{\scriptsize procedural personas, Ariyurek et al., MIMIC-Py}};
\node[dm node=4.6cm] (b) at (6.4,0) {\dmhead{Human-reference evaluation}\\[1pt]{\scriptsize Collins et al., FAIRGAMER}};
\draw[dm arrow] (a) -- (b);
\node[dm bridge=3.6cm] at (3.2,0.8) {diverse testers are not a human reference; the reference must match the decision};
\draw[dm axis] (-2.4,-1.15) -- (8.8,-1.15) node[midway,below=1pt,font=\scriptsize,text=GameSlate] {from who plays to who is the reference};
\node[dm frame,fit=(current bounding box)] (F) {};\dmtag
\end{tikzpicture}}

Behavioral diversity helps discover failures, but it need not reproduce the distribution of human play. A deliberately unusual test can be valuable for robustness and a poor predictor of usability. Claims about player relevance therefore require a reference appropriate to the decision: trajectories for behavioral similarity, interaction for usability, or repeated encounters for an ongoing experience.

\runin{Behavioral diversity} A tester's competence limits the conclusions of any play-based test. Failure to reach a mechanic can be mistaken for its absence, whereas a highly skilled policy may never reveal novice confusion. On a 20-game testbed, GUI playtesters reach a Pass@10 of 0.68 to 0.76 against 0.92 for humans, failing mostly on real-time timing and small interface elements \citep{huang2026guigames}. The procedural personas discussed in \Cref{sec:test-playtesting} broaden coverage through different objectives or play styles \citep{holmgard2019playtesting}. Their representativeness depends on how those behaviors compare with human strategies. Ariyurek et al. explicitly compare synthetic and humanlike testers with human trajectories, showing that bug-finding effectiveness and behavioral similarity can be measured separately instead of being assumed from one another \citep{ariyurek2021testing}.

MIMIC-Py makes personality-conditioned exploration a reusable testing interface. It separates planning, execution, and memory, with configurable personalities and game adapters demonstrated for Dungeon Adventures, Shattered Pixel Dungeon, and Minecraft \citep{chen2026mimicpy}. The tool supports structured action calls and generated interaction code, and adopting a new game still requires game-side integration. Such profiles can broaden test behavior, but a personality prompt is not itself a validated model of a player population. Human trajectory comparisons remain necessary when the claim is representativeness and not diversity alone.

\runin{Human-reference evaluation} Predictions of player response need a different reference from rule correctness. The distinction between anticipated and experienced appeal in \Cref{fig:testing-evidence} concerns when the human judgment is elicited. In the study by Collins et al., whose prediction results are discussed in \Cref{sec:model-players}, participants judged games before playing; the fun ratings therefore measure anticipated appeal \citep{collins2025gameevaluations}. Such judgments can inform concept selection, while claims about enjoyment during play require evidence from interaction. Human relevance also includes fairness: FAIRGAMER tests social bias in NPC decisions across transaction, cooperation, and competition, covering four bias types over 12 tasks \citep{shi2026fairgamer}. Such diagnostics complement studies of actual play without replacing them.

The human reference must match the decision for which the evaluator is used. Selecting a promising concept can reasonably use anticipated appeal, whereas assessing usability requires interaction and assessing a persistent companion requires repeated encounters. Similarly, testing whether NPC choices change with a demographic cue is a targeted bias diagnostic, not a complete account of fairness during multiplayer play. Automated judgment is most useful when its target and reference are specified narrowly enough that errors can be interpreted and corrected.

\begin{gameaiinsight}[Evaluator]{Playing well and testing well are different skills}
\begin{insightpoints}
\item \textbf{Reaching a state and judging it are separate steps.} Play2Code can miss mechanics its tester cannot trigger \citep{huang2026guigames}; GameGen-Verifier injects states to inspect specified requirements \citep{jia2026gamegenverifier}. Injection reduces exploration cost but bypasses the path to the state. The two approaches are complementary.
\item \textbf{Varied testers may still play unlike real players.} Procedural personas deliberately explore different goals \citep{holmgard2019playtesting}, while human-trajectory comparisons test whether those goals resemble actual play \citep{ariyurek2021testing}. A useful bug-finding portfolio may include atypical behavior; it should not be used as a population model without a separate human reference.
\item \textbf{Repair loops raise the cost of judge errors.} When a judge guides repeated revision, its blind spots can be reinforced rather than revealed \citep{panickssery2024selfpreference}. Independent execution checks and withheld defects then matter beyond ordinary judge agreement: they test whether apparent repair survives outside the feedback that produced it \citep{helff2026llms}.
\end{insightpoints}
\end{gameaiinsight}

\gameaisectionaccent{GameInk}
\section{Benchmarks and Evidence Across the Six Roles}
\label{sec:evaluation}

Sections~3--8 surveyed what AI systems have been shown to accomplish within each of the six roles. This section asks what their evaluations establish: what counts as success, which conditions make results comparable, and where interpretation should stop. It is distinct from \emph{Test and Evaluate} in \cref{sec:testing}, where an AI system itself produces test evidence, a verdict, or a diagnosis. A player study, engine test, or benchmark can evaluate any role; its target may be action quality, predictive fidelity, design quality, software behavior, runtime adaptation, or testing reliability.

\Cref{tab:role-evaluation-guide} provides a reading guide to these targets. The quantitative examples below illustrate distinct interpretation problems: progress versus completion, model versus harness, trajectory validity versus world fidelity, and scoped development versus complete delivery. For design and player-facing systems, study conditions and human outcomes are more informative than imposing the same score format.

\begin{table}[H]
\centering
\caption{Evaluation targets and interpretation boundaries across the six roles.}
\label{tab:role-evaluation-guide}
\begingroup
\small
\setlength{\tabcolsep}{4pt}
\renewcommand{\arraystretch}{1.10}
\begin{tabularx}{\linewidth}{@{}>{\raggedright\arraybackslash}p{0.19\linewidth}>{\raggedright\arraybackslash}X>{\raggedright\arraybackslash}X@{}}
\toprule
\textbf{Role} & \textbf{Targets and example measures} & \textbf{Interpretation boundary} \\
\midrule
\textcolor{Player}{\textbf{Play and Act}} & Completion; progress; action efficiency & Protocol-specific; transfer across rules, controls, or timing remains untested \\
\addlinespace[2pt]
\textcolor{Simulator}{\textbf{Model Players\newline and Games}} & World-state fidelity; reference-game policy return; player prediction or trait recovery & Short-horizon fidelity does not establish persistence; synthetic traits do not establish real-player validity \\
\addlinespace[2pt]
\textcolor{Creator}{\textbf{Design}} & Validity; diversity; constraint satisfaction; designer control & Does not establish authoring efficiency or intended player experience \\
\addlinespace[2pt]
\textcolor{Builder}{\textbf{Build and Maintain}} & Launchability; requirement satisfaction; repair; regression & Does not establish long-term maintenance or developer handoff \\
\addlinespace[2pt]
\textcolor{Resident}{\textbf{Generate and Adapt at Runtime}} & Latency; consistency; controllability; player outcomes & Single-session evidence; adaptation effects require suitable controls \\
\addlinespace[2pt]
\textcolor{Evaluator}{\textbf{Test and Evaluate}} & Coverage; validated defects; verdict accuracy; reproduction cost & Coverage is not correctness; sampled behavior is not population validity \\
\bottomrule
\end{tabularx}
\endgroup
\end{table}

\subsection{Role-Specific Evaluation Targets and Metrics}
\label{sec:evaluation-targets}

\subsubsection{Play and Act: Task Performance and Generalization}

Game-agent benchmarks have a long tradition of standardized outcomes. General Game Playing evaluates decisions
under unseen formal rules \citep{genesereth2005ggp}; ALE fixes an emulator and action set across Atari games \citep{bellemare2013ale}; MineRL adds human
demonstrations and native Minecraft control \citep{guss2019minerl}, while BASALT poses hard-to-specify tasks without reward functions and scores them through human comparisons \citep{shah2022basalt}. Recent suites extend coverage
to language-mediated play, multimodal real-time games, and common evaluation harnesses
\citep{paglieri2024balrog,li2025textatari,zhang2025videogamebench,ouyang2026gameworld}. Task success measures
end-to-end behavior only under the specified observation, action, timing, and adaptation conditions.

Text-game environments expose different kinds of generalization. TextWorld generates games with controlled
vocabularies, quests, and layouts \citep{cote2018textworld}, whereas Jericho exposes human-authored interactive fiction whose puzzles and
language are not generated by one common grammar \citep{hausknecht2020jericho}. TextArena
broadens language-mediated evaluation to a collection of single- and multi-player games
\citep{guertler2025textarena}. These interfaces simplify motor control but retain action selection, partial
observability, and interpretation of rules. In Minecraft, Malmo established configurable missions and
observation/action interfaces that later data and agent platforms build on \citep{johnson2016malmo}.

Longer tasks also require more informative progress measures. The NetHack Learning Environment emphasizes
exploration and skill acquisition in a procedurally varied roguelike \citep{kuttler2020nle}; Crafter scores semantically defined
achievements in a visual survival game \citep{hafner2022crafter}. FlashAdventure instead tests full
story arcs in graphical adventures, where clues and inventory must support later actions
\citep{ahn2025flashadventure}. OmniGameArena adds solo, competitive, and cooperative UE5 games and evaluates
improvement through repeated skill-prompt revision \citep{lin2026omnigamearena}. Achievement coverage, story
progress, and improvement after reflection therefore answer different questions from a first-attempt win rate.

PuzzleJAX offers a complementary route to task variety: it compiles PuzzleScript game descriptions into GPU-executable environments and provides a benchmark of over 2,500 game environments. Validation replays action sequences found in the original engine on 3,012 games, and 87.0\% of their 20,308 levels pass. Search, reinforcement learning, and language-model players are compared on 12 games from its 15-game Select set \citep{earle2025puzzlejax}. Here the varied object is executable game logic, rather
than another level generated under a fixed rule set. Scores still need to specify search depth, training budget, and rule access.

GameWorld makes the distinction between partial progress and task completion visible across a larger model
comparison (\Cref{tab:visual-agent-results}). Agents often advance toward the objective without reaching it, even
when inference pauses the game. The ordering also depends on the interface: semantic action parsing helps some
model configurations, but does not consistently outperform direct computer use. Both interfaces score below the two human
reference participants in this evaluation. Because each experience condition contains only one participant
\citep{ouyang2026gameworld}, these are illustrative reference cases, not population-level estimates of novice and
expert performance.

\begin{table}[!htbp]
\centering
\caption{GameWorld results for 18 model--interface configurations and two human reference participants across 170 tasks in 34 games \citep{ouyang2026gameworld}.}
\label{tab:visual-agent-results}
\small
\renewcommand{\arraystretch}{1.08}
\begin{tabular*}{0.70\linewidth}{@{\extracolsep{\fill}}lrr@{}}
\toprule
\textbf{Model / player} & \textbf{Progress (\%) $\uparrow$} & \textbf{Success (\%) $\uparrow$} \\
\midrule
\multicolumn{3}{@{}l}{\textbf{Computer use: keyboard and mouse commands}} \\
Seed-1.8 & 39.8 & 20.0 \\
Claude Sonnet 4.6 & 38.3 & 19.4 \\
Gemini 2.5 Computer Use & 36.1 & 16.5 \\
OpenAI Computer Use & 35.8 & 16.5 \\
Qwen3-VL-Plus & 33.6 & 15.9 \\
Qwen3-VL-235B-A22B & 31.4 & 14.1 \\
UI-TARS-1.5-7B & 31.1 & 12.4 \\
Qwen3-VL-30B-A3B & 30.8 & 12.9 \\
\midrule
\multicolumn{3}{@{}l}{\textbf{Semantic action parsing}} \\
Gemini 3 Flash Preview & 41.9 & 21.2 \\
GPT-5.2 & 40.6 & 20.6 \\
Claude Sonnet 4.6 & 39.3 & 20.6 \\
Seed-1.8 & 39.0 & 20.0 \\
Kimi K2.5 & 37.4 & 18.8 \\
Grok 4.1 Fast Reasoning & 36.0 & 16.5 \\
Qwen3-VL-Plus & 35.4 & 16.5 \\
GLM-4.6V & 30.8 & 14.1 \\
Qwen3-VL-235B-A22B & 30.8 & 13.5 \\
Qwen3-VL-30B-A3B & 30.6 & 12.4 \\
\midrule
\multicolumn{3}{@{}l}{\textbf{Human reference}} \\
Novice human & 64.1 & 55.3 \\
Expert human & 82.6 & 77.1 \\
\bottomrule
\end{tabular*}
\par\smallskip
\begin{minipage}{\linewidth}
\footnotesize
For agents, the game is paused during model inference; each task permits at most 100 actions. Success measures target attainment, whereas progress awards partial credit. Each human label refers to one participant ($n=1$ per condition); the expert reference had studied the rules and practiced the controls beforehand.
\end{minipage}
\end{table}

Learning unfamiliar rules is tested more directly by ARC-AGI-3, whose abstract turn-based environments provide no
natural-language instructions \citep{arcprize2026benchmark}. The results in
\Cref{tab:interactive-reasoning-results} illustrate sensitivity to the agent harness. At the reported high
reasoning-effort setting, GPT-6 Astra scores 54.82 with the Standard harness and 99.95 with the Provider Adapter
\citep{arcprize2026astra}, and Claude Fable 5.2 scores 79.26 and 97.72 \citep{arcprize2026fable}. The configurations differ in how they carry reasoning state and manage conversation
history; the effort label alone does not establish matched computational cost.
On the public games, scores are also near the ceiling. Under one frozen Claude Opus~5 configuration, the Kepler harness,
which represents hypotheses as executable world models and checks them against recorded transitions, obtained a
server-verified 100.00 on all 25 public games, and 48 of 50 game--model cells across Claude Opus~5 and GPT-5.6 Sol
reached 100. These are final replays after development on the public games, not runs under the official first-exposure protocol and its per-level action budget. Its author also reports a source-code leak that produced an invalid perfect run and recommends
first-attempt, cost-conditioned, and verification-aware reporting \citep{wu2026kepler}.

\begin{table}[H]
\centering
\caption{ARC-AGI-3 results on the Semi-Private split \citep{arcprize2026astra,arcprize2026benchmark,arcprize2026carbon,arcprize2026fable,arcprize2026gpt56,arcprize2026gpt62,arcprize2026grok45,arcprize2026grok46,arcprize2026opus}. The first panel reports the highest published Standard-harness score for each listed model at the reasoning-effort setting shown; the second compares harnesses for GPT-6 Astra and Claude Fable 5.2.}
\label{tab:interactive-reasoning-results}
\small
\renewcommand{\arraystretch}{1.08}
\begin{tabular*}{0.70\linewidth}{@{\extracolsep{\fill}}llr@{}}
\toprule
\multicolumn{3}{@{}l}{\textbf{(a) Standard harness}} \\
\midrule
\textbf{Model} & \textbf{Reasoning effort} & \textbf{Score (\%) $\uparrow$} \\
\midrule
Claude Fable 5.2 & XHigh & 81.17 \\
GPT-6.2 Astra & Max & 70.31 \\
Gemini 4 Carbon & Low & 65.12 \\
GPT-6 Astra & Max & 62.71 \\
Claude Opus 5 & High & 30.16 \\
GPT-5.6 Sol & Max & 7.78 \\
Grok 4.6 & XHigh & 2.11 \\
GPT-5.6 Terra & Max & 0.80 \\
Grok 4.5 & Medium & 0.32 \\
GPT-5.6 Luna & Max & 0.18 \\
\bottomrule
\end{tabular*}
\par\medskip
\begin{tabular*}{0.70\linewidth}{@{\extracolsep{\fill}}lrr@{}}
\toprule
\multicolumn{3}{@{}l}{\textbf{(b) Harness comparison}} \\
\midrule
\textbf{Reasoning effort} & \textbf{Standard (\%) $\uparrow$} & \textbf{Provider Adapter (\%) $\uparrow$} \\
\midrule
\multicolumn{3}{@{}l}{GPT-6 Astra} \\
\quad High & 54.82 & 99.95 \\
\quad Max & 62.71 & 98.55 \\
\midrule
\multicolumn{3}{@{}l}{Claude Fable 5.2} \\
\quad High & 79.26 & 97.72 \\
\quad Max & 65.66 & 94.90 \\
\bottomrule
\end{tabular*}
\par\smallskip
\begin{minipage}{\linewidth}
\footnotesize
Scores measure human-relative action efficiency, not game completion rate. Reasoning-effort labels do not imply matched computational budgets across models or harnesses. The Claude Fable 5.2 result at XHigh comes from an earlier checkpoint whose tuning at that setting differs from the released model, while its High and Max settings are unchanged. ARC Prize omits the Standard-harness result for Gemini 4 Carbon at High because that run completed only 30 of 55 games.
\end{minipage}
\end{table}

\subsubsection{Model Players and Games: World-Model Fidelity and Player-Model Validity}
\label{sec:evaluation-models}

World-model fidelity has at least three distinct targets: perceptual quality, mechanics correctness, and persistent state (\Cref{tab:world-fidelity-tests}). Perceptual measures describe the generated sequence, including its temporal appearance. FVD compares distributions of spatiotemporal video features \citep{unterthiner2018fvd}; it can respond to visually apparent failures but neither tests a specified rule nor verifies a hidden variable. A convincing animation can accompany an incorrect health decrement, and a correct immediate transition can later be forgotten. Action following cuts across these targets: responding to a movement command does not establish that a collision or inventory rule is respected \citep{fang2026iworldbench,yang2026worldexam,ding2026playworld}.

\begin{table}[htbp]
\centering
\caption{Complementary tests of world-model fidelity. State-based checks require an independent specification or reference; visible agreement must also be checked when state drives a learned renderer.}
\label{tab:world-fidelity-tests}
\begingroup
\footnotesize
\setlength{\tabcolsep}{4pt}
\renewcommand{\arraystretch}{1.12}
\begin{tabularx}{\linewidth}{@{}L{2.65cm}Y L{5.0cm}@{}}
\toprule
\textbf{Target} & \textbf{What to check} & \textbf{Suitable evidence} \\
\midrule
Perceptual quality & Appearance; motion continuity; visual diversity & FVD; perceptual scores; human~ratings \\
\addlinespace[3pt]
Mechanics correctness & Legal transitions; health/resource changes; contact and terminal events & Reference-state error; rule assertions; controlled replay \\
\addlinespace[3pt]
Persistent state & Identity and consequences after occlusion, revisit, or delayed events & Revisit/mutation probes; save--restore; synchronized views \\
\bottomrule
\end{tabularx}
\endgroup
\end{table}

When a reference engine or suitable state instrumentation is available, evaluation can compare predicted health, resources, object identities, and terminal events against recorded transitions. Collision and friction tests should use the game's specified geometry, contact rules, and tolerances, rather than assume realistic physical behavior is always intended. For stochastic games, controlled random seeds or repeated trials distinguish model error from valid alternative outcomes. The reference state must also be independent of the model under test: a variable generated by the same model is an output to check, not ground truth. Agreement between a valid state and its rendered observation is a further requirement; a blocked path must not appear traversable.

Recent benchmarks make parts of this separation measurable. WildBench combines video quality, camera control, action following, and pose-based state alignment \citep{li2026wildworld}; pose alignment alone does not verify hidden resources. WorldMark v2 supplies common scenes and action adapters for ten models across 500 cases, separating per-axis control accuracy, purity, latency, and stability from visual quality and memory \citep{xu2026worldmark}. These controlled comparisons are more informative about response behavior than a gallery of diverse worlds. CombatStateBench tests a different property: visible agreement with an engine-maintained scenario. Programmable World Model reports 94\% count accuracy and 98\% death-state accuracy on 50 clips, using a VLM to inspect generated frames. The latter check accepts any visible dead character in sampled post-event frames, not necessarily the correct individual \citep{huang2026programmable}.

Training-oriented world
models add another target: whether policies learned in the model retain performance in the reference environment
\citep{kaiser2020simple,alonso2024diamond,hafner2025dreamer4}. A planning model can be validated through decision
quality; a training environment through return after transfer to the reference game and tests for model
exploitation; and a player-facing simulator through sustained, controllable interaction.

Player models require a
different target: prediction of later behavior, preference, skill, or response against independent observations.
Maia4All tests individual behavior prediction on held-out human chess moves \citep{tang2025maia4all}.
Beyond Asking instead evaluates recovery of controlled synthetic traits, tests profile-conditioned adaptation,
and includes an exploratory human study \citep{lu2026personalized}. These are complementary targets, not
interchangeable validations: recovering a bot parameter does not establish a real player's latent traits, and
improving generated content does not by itself establish profile accuracy. The mobile PCG study by
\citet{hafnar2025zeroshot} evaluates downstream play outcomes.

PlayWorld's nine-model comparison reports trajectory validity alongside world quality assessed by rubrics
(\Cref{tab:world-model-results}). SANA-WM has the second-highest trajectory-validity rate among the listed models,
yet its two evolution scores remain near the lower end of the rubric scale. LingBot-World2 scores higher than
LingBot-World on Insight Evolution (visible evolution), but lower on Out-of-sight Evolution. The rubric scores also
incorporate an instruction-following check: applicable failures receive a score of 1 \citep{ding2026playworld}.
They should therefore be read together with validity rather than as fidelity measured only on successful trajectories.
For other uses of a world model, these measures should be complemented by outcomes matched to that use:
reference-game policy return for training-oriented models \citep{kaiser2020simple,alonso2024diamond}, or numerical
state error when reference state is available \citep{lin2026stateplaystateawaregameworld}.

\begin{table}[!htbp]
\centering
\caption{PlayWorld evaluation of nine interactive world models \citep{ding2026playworld}. Rubric scores range from 1 to 5; trajectory validity is a percentage. Higher values are better in every column.}
\label{tab:world-model-results}
\small
\setlength{\tabcolsep}{4pt}
\renewcommand{\arraystretch}{1.08}
\begin{tabular*}{0.72\linewidth}{@{\extracolsep{\fill}}lrrrrrr@{}}
\toprule
\textbf{Model} & \textbf{GC} & \textbf{IF} & \textbf{IE} & \textbf{OE} & \textbf{Overall} & \textbf{Validity (\%)} \\
\midrule
Genie 3 & 2.74 & 2.40 & 1.51 & 1.81 & 2.12 & 87.1 \\
HappyOyster & 2.54 & 2.15 & 1.47 & 1.54 & 1.92 & 79.6 \\
LingBot-World2 & 2.04 & 2.13 & 1.95 & 1.16 & 1.82 & 78.8 \\
LingBot-World & 2.11 & 2.23 & 1.33 & 1.43 & 1.78 & 72.7 \\
HY-World2 & 2.14 & 2.06 & 1.13 & 1.09 & 1.61 & 51.7 \\
SANA-WM & 1.72 & 1.89 & 1.13 & 1.16 & 1.48 & 80.4 \\
Hunyuan-GameCraft & 1.62 & 1.52 & 1.21 & 1.31 & 1.42 & 61.5 \\
HY-WorldPlay & 1.12 & 1.63 & 1.01 & 1.08 & 1.21 & 41.6 \\
Matrix-Game-3.0 & 1.30 & 1.25 & 1.00 & 1.00 & 1.14 & 68.2 \\
\bottomrule
\end{tabular*}
\par\smallskip
\begin{minipage}{\linewidth}
\footnotesize
GC: geometry consistency; IF: interaction fidelity; IE: Insight Evolution (the source term for continuously visible evolution); OE: out-of-sight evolution. Overall
averages the four dimension scores; validity averages the three movement-dependent dimensions. Applicable
instruction-following failures receive a score of 1. Gemini 3.1 Pro judges sample-specific rubrics.
\end{minipage}
\end{table}

\subsubsection{Design: Design Quality and Designer Control}

PCG and automated-design research measure validity, diversity, controllability, difficulty, novelty, and the
behavior enabled by generated rules or levels \citep{togelius2011searchpcg,yannakakis2011edpcg}. Co-creative systems additionally require evidence about steering,
transparency, designer control, iteration, and creator experience
\citep{lai2022mixedinitiative,liapis2013sketchbook,lanzi2023llmco,earle2025dreamgarden}.
Human evaluation is needed when a claim extends from formal validity to player comprehension, agency, pacing,
enjoyment, or creator control. The checks must match the generated object. ChatGPT4PCG evaluates whether Science
Birds structures remain standing and resemble requested letters, with later editions adding diversity \citep{taveekitworachai2023chatgpt4pcg,taveekitworachai2024chatgpt4pcg2}; a Sokoban
solver instead checks whether a puzzle admits a solution within its search budget
\citep{todd2023level}. ScriptDoctor further
separates compilation from finding sufficiently nontrivial solutions for every generated level
\citep{earle2025scriptdoctor}. These tests reveal different failures.

A design comparison should therefore specify the brief, accepted constraints, candidate budget, and editing tools
available to the creator. Validity and diversity concern the generated artifacts; correction effort and control
concern the authoring process; enjoyment and comprehension concern the resulting player experience. Reporting
these separately makes it possible to identify whether a method improves the generator, the creative workflow,
or both.

\subsubsection{Build and Maintain: Executable Artifacts and Revision}
\label{sec:evaluation-build}

Development benchmarks examine different stages of implementation. GameDevBench evaluates multimodal Godot
development tasks \citep{chi2026gamedevbench}; GameCraft-Bench scores complete greenfield Godot artifacts against a rubric \citep{luo2026gamecraft}; and GameEngineBench
tests scoped native-C++ changes through hidden runtime tests reviewed by an LLM judge
\citep{la2026gameenginebench}. PlaytestArena provides interaction-grounded
evaluation for generated browser games \citep{huang2026guigames}. Mage combines compilation and runtime checks with
static structural and mechanic analysis: the latter tests what the code appears to implement, not whether the
mechanic behaves correctly in play \citep{liu2026mage}. JAMER evaluates theme-driven generation and multi-granularity
code completion with compilation, structural, and behavioral measures \citep{sun2026jamer}; GameXpert-Bench adds
repair and multi-turn regression tests \citep{chen2026gamexpert}. These establish different levels of execution
evidence. A compilation rate, fraction of projects passing a test, and average rubric score are different quantities
even when all are reported as percentages.

Repair benchmarks additionally condition success on the starting artifact. Restoring a removed function in a
supplied project tests reconstruction with scene context intact; fixing a known defect tests diagnosis and
modification; satisfying a new request tests implementation while preserving earlier behavior. A benchmark that
supplies the failing location removes part of the diagnostic burden. Reports should state this initial information alongside the test outcome.

The results in \Cref{tab:development-results,tab:gamecraft-results} distinguish scoped development from complete
game construction. The GameDevBench results in \Cref{tab:development-results} cover 21 model--harness
configurations with uncertainty intervals; small score differences should be interpreted alongside those intervals
and the configuration differences. The GameCraft-Bench results in \Cref{tab:gamecraft-results} cover 16
configurations on complete artifacts. Across its reported rows, mechanics scores exceed content-depth scores: under
this benchmark's rubric, generated projects more consistently realize recognizable rules than substantial
progression and content.

\begin{table}[!htbp]
\centering
\caption{GameDevBench results for 21 model--harness configurations evaluated on 333 Godot tasks \citep{chi2026gamedevbench,chi2026gamedevbenchresults}. Values are Pass@1 estimates with 95\% confidence intervals.}
\label{tab:development-results}
\small
\renewcommand{\arraystretch}{1.08}
\begin{tabular*}{0.72\linewidth}{@{\extracolsep{\fill}}llr@{}}
\toprule
\textbf{Model (effort, where specified)} & \textbf{Harness} & \textbf{Pass@1 (\%) $\uparrow$} \\
\midrule
Claude Opus 5.5 (xhigh) & Claude Code & $70.6 \pm 4.9$ \\
GPT-6.1 Sol (xhigh) & Codex & $70.0 \pm 4.9$ \\
GPT-6 Astra (high) & Codex & $68.8 \pm 5.0$ \\
Claude Opus 5.5 (high) & Claude Code & $68.2 \pm 5.0$ \\
Claude Fable 5 (xhigh) & Claude Code & $67.3 \pm 5.0$ \\
GPT-5.6 Sol (xhigh) & Codex & $63.7 \pm 5.2$ \\
GPT-5.6 Sol (high) & Codex & $63.1 \pm 5.2$ \\
Muse Spark 1.2 (high) & Muse Code & $61.0 \pm 5.2$ \\
GPT-5.6 Sol (medium) & Codex & $58.6 \pm 5.3$ \\
Kimi K3 & Kimi Code & $58.0 \pm 5.3$ \\
Claude Opus 4.8 & Claude Code & $55.9 \pm 5.3$ \\
GPT-5.5 & Codex & $54.7 \pm 5.3$ \\
Gemini 3 Pro Preview & Gemini CLI & $53.8 \pm 5.4$ \\
GPT-5.4 & Codex & $52.0 \pm 5.4$ \\
Gemini 3 Flash Preview & Gemini CLI & $46.8 \pm 5.4$ \\
GPT-5.4 Mini & Codex & $43.2 \pm 5.3$ \\
GLM-5.2 & OpenCode & $38.4 \pm 5.2$ \\
Claude Sonnet 4.5 & Claude Code & $34.8 \pm 5.1$ \\
Kimi K2.5 & OpenHands & $20.7 \pm 4.4$ \\
Claude Haiku 4.5 & Claude Code & $18.6 \pm 4.2$ \\
Qwen3.5-397B & OpenHands & $5.4 \pm 2.4$ \\
\bottomrule
\end{tabular*}
\end{table}

These leaderboard results complement controlled component comparisons. GameDevBench's visual-feedback ablation
holds a model--harness pair fixed while varying visual tools, and the gains vary across models \citep{chi2026gamedevbench}.

GameCraft-Bench's overall score combines rubric dimensions with unequal weights and a validity gate for
submissions; it is not the arithmetic mean of the displayed category scores. Failure to launch a project or supply
a parseable interaction trace yields a zero score, so delivery failures must be distinguished from lower rubric
quality in an artifact that enters the benchmark's replay-and-scoring pipeline \citep{luo2026gamecraft}.

\begin{table}[!htbp]
\centering
\caption{GameCraft-Bench results for 16 model--harness configurations across 140 tasks in 15 game families \citep{luo2026gamecraft,gamecraft2026results}. Scores are normalized to a 0--100 scale, not project pass rates; higher values are better.}
\label{tab:gamecraft-results}
\small
\setlength{\tabcolsep}{3pt}
\renewcommand{\arraystretch}{1.12}
\begin{tabularx}{\linewidth}{@{}>{\raggedright\arraybackslash}Xlrrrrr@{}}
\toprule
\textbf{Model (effort)} & \textbf{Harness} &
\shortstack[r]{\textbf{Core}\\\textbf{mechanics}} &
\shortstack[r]{\textbf{Content}\\\textbf{depth}} &
\shortstack[r]{\textbf{Functional}\\\textbf{visuals}} &
\shortstack[r]{\textbf{Art and}\\\textbf{presentation}} &
\textbf{Overall} \\
\midrule
GPT-6 Astra (xhigh) & Codex & 84.99 & 68.74 & 77.83 & 78.07 & 75.86 \\
GPT-6 Sol (xhigh) & Codex & 81.43 & 63.83 & 71.54 & 67.69 & 69.00 \\
Claude Opus 5 (xhigh) & Claude Code & 76.75 & 63.31 & 68.04 & 70.11 & 68.44 \\
Claude Fable 5 (high) & Claude Code & 76.52 & 58.61 & 66.93 & 67.68 & 65.72 \\
GPT-5.6 Sol (high) & Codex & 74.50 & 56.10 & 64.80 & 57.00 & 60.50 \\
Kimi K3 & Kimi Code & 73.03 & 53.38 & 56.90 & 53.59 & 56.96 \\
Seele02-pro & SeeleAgent & 68.42 & 48.76 & 52.74 & 44.17 & 50.70 \\
Claude Opus 4.7 (high) & Claude Code & 55.34 & 39.48 & 42.78 & 36.86 & 41.46 \\
DeepSeek V4 Flash-0731 & Claude Code & 52.11 & 37.14 & 43.70 & 37.72 & 40.61 \\
GPT-5.5 (high) & Codex & 54.36 & 38.61 & 41.84 & 32.94 & 39.49 \\
GLM-5.2 & Claude Code & 50.09 & 36.40 & 41.06 & 36.29 & 39.12 \\
Kimi K2.6 & Kimi Code & 39.76 & 28.07 & 33.66 & 27.99 & 30.65 \\
MiMo-V2.5-Pro & Claude Code & 32.33 & 22.59 & 27.45 & 20.65 & 24.10 \\
GLM-5.1 & Code Buddy & 25.23 & 17.80 & 21.14 & 14.59 & 18.29 \\
MiniMax-M2.7 & Code Buddy & 14.27 & 9.92 & 14.92 & 8.85 & 10.95 \\
DeepSeek V4 Pro & Codex & 2.25 & 1.69 & 1.97 & 2.63 & 2.15 \\
\bottomrule
\end{tabularx}
\par\smallskip
\begin{minipage}{\linewidth}
\footnotesize
Scores are assigned from replayed gameplay using hidden rubrics. Overall follows the benchmark's weighted, validity-gated aggregation rather than an arithmetic mean of the displayed category scores.
\end{minipage}
\end{table}

\subsubsection{Generate and Adapt at Runtime: System Behavior and Player Outcomes}

Runtime studies assess narrative coherence, character behavior, rule interpretation, latency, controllability, and
player response. The player studies considered here examine bounded encounters
\citep{peng2024emergence,hsu2026doubleedged,hsu2026ifcargo}, while field studies and live deployments such as CALYPSO
and mobile personalized PCG extend evaluation beyond such sessions
\citep{zhu2023calypso,hafnar2025zeroshot}. Study design determines which change can explain an outcome. GenFlora
independently varies item functionality and NPC dialogue \citep{yin2026contextualized}, whereas
\citet{roso2026latency} vary response onset and token delivery during otherwise comparable conversations. Such
component-level comparisons complement full-system studies: an apparent dialogue-quality effect may partly arise
from responsiveness, and a content-generation benefit may not require personalization. Repeated-session studies
additionally need to distinguish sustained benefits from familiarity and novelty.

For spoken characters, text-token latency and time to audible response are different measurements. Recognition, dialogue generation, synthesis, and playback each add delay. Evaluation should separate speech intelligibility and vocal identity from turn-taking, interruption recovery, and whether utterances remain appropriate to the current game state. The queued speech pipeline in The Interview illustrates these dependencies \citep{figueiredo2025scaffolded}.

\subsubsection{Test and Evaluate: Coverage and Verdict Reliability}

An AI tester is evaluated through state or behavior coverage, defect discovery, verdict accuracy against reference
oracles, diagnostic usefulness, and efficiency. Reliability also depends on the evaluation setup: whether the
oracle or judge is sufficiently independent of the system under test, and whether the tester's behavior covers the
relevant player population. A tester that reaches many states but misjudges them differs from a verifier that
checks mechanics accurately but cannot represent human play. Human agreement, target-state injection, deterministic
engine tests, and independent evaluator models establish different forms of reliability (\Cref{sec:testing}).

Efficiency should be measured at the same unit as reliability. Wall-clock time per checked mechanic, interactions
needed to reach a target, and developer time to reproduce a bug are different costs. A faster verifier can still
depend on an expensive state injector or a manually authored specification. Separating setup cost from repeated
execution helps identify reusable testing components.

\subsection{A Cross-Role Framework for Interpreting Evidence}
\label{sec:evaluation-framework}

\subsubsection{Dimensions for Comparing Evaluations}

The six roles require different task-specific measures, but their evidence can still be compared without collapsing unlike outcomes into a single maturity score. We use five cross-role dimensions that apply across empirical targets:

\textbf{Standardization.} Are tasks, protocols, and measures shared across studies, or does each system define its own demonstration and harness?

\textbf{Execution grounding.} Is the claim checked against an engine, reference environment, deterministic test, or other executable behavior rather than surface quality alone?

\textbf{Scope and transfer.} Does the evidence span games, genres, engines, interfaces, or player populations, and which of these actually changes at test time?

\textbf{Horizon and revision.} Does evaluation cover sustained interaction, revisits, repeated edits, regression, or handoff rather than one clip, build, or session?

\textbf{External validation.} Two questions matter: is the reference sufficiently independent of the system being evaluated, and does it represent the intended task or player population? Engine checks, held-out behavior, human judgments, and deployment outcomes address different parts of these questions. Human participation does not by itself make a verdict independent, and an independent verdict need not establish real-world relevance.

These dimensions explain why evidence may be well standardized yet narrow, grounded in execution yet short-term, or
human-centered yet small and system-specific. Within any of these dimensions,
evidential strength also depends on statistical support: repeated runs or seeds, uncertainty reporting, sample
size, and replication determine how much weight should be placed on an observed difference.

\subsubsection{Statistical Support and Repeated Evaluation}

Variance matters when benchmarks contain few games, expensive model calls, or stochastic generations.
\citet{agarwal2021statistical} show that point estimates across small numbers of reinforcement-learning runs can
support unstable conclusions; they develop interval estimates, performance profiles, and robust aggregates such as
the interquartile mean. Reporting an interval cannot make two incompatible success measures
comparable. The unit of resampling should follow the unit of variation. Repeated outputs for the same task estimate
sampling variability, whereas new games, projects, or participants test broader variation. For generation and
repair, results should also distinguish a typical attempt from a best-of-budget selection. A paired comparison on
the same tasks helps isolate a system change, while confidence intervals and per-task results expose how much the
aggregate depends on a small subset. Human studies require particular care when several observations come from the
same person.

The GameDevBench rows in \Cref{tab:development-results} include confidence intervals \citep{chi2026gamedevbenchresults}, whereas the GameCraft-Bench
rows in \Cref{tab:gamecraft-results} do not report row-wise intervals
\citep{gamecraft2026results}. Repeated scoring of a fixed interaction trace measures
evaluator variability, which is distinct from variation across independently generated artifacts.

\subsection{Limits of Interpretation and Generalization}
\label{sec:evaluation-limits}

\subsubsection{Interfaces, Adaptation Budgets, and Held-Out Settings}
\label{sec:evaluation-generalization}

Interfaces and test harnesses strongly affect reported performance. An agent using symbolic state or semantic
actions faces a different problem from one using pixels, a keyboard, and a mouse under real-time constraints.
Gym-Anything-RT isolates timing by running the same 100 generated minigames, at their highest difficulty, in real time or paused: GPT-6 Astra
(xHigh) passes 9.0\% in real time and 75.0\% when the environment is paused, and when allowed to write code,
frontier agents write programs that bypass real-time perception and action \citep{aggarwal2026gymanythingrt}.
Two further benchmarks pause the game during inference and so measure perception and decision without the clock.
AtariBench runs vision--language models on 20 Atari 2600 games in 30-second windows, where PPO agents still rank first on 14 of the 16 games on which they are compared \citep{huo2026ataribench}. PlaySuite covers 5,734 games from itch.io and PyWeek and scores progress from gameplay video with a model judge on a 0--4 scale: without interaction history, the best of fourteen open models averages 0.84 on the itch.io games \citep{varghese2026playsuite}.
Latency, context length, tool access, and online adaptation change the task. Retry and selection protocols also
matter: success on one attempt, success after repair, and the best result selected from several runs impose
different costs and support different reliability claims. A benchmark score describes the configured system, not
just its base model.

Generalization results depend on both the held-out setting and the capability measured.
\Cref{tab:generalization-evidence} separates changes of map, mode, and game. These labels are not a difficulty ranking:
a mode can change rules substantially, while two different titles may share familiar controls and objectives.
Engine, controller, and partner changes introduce software, control, and coordination shifts.

\begin{table}[htbp]
\centering
\caption{Map, mode, and game changes test different generalization settings. The measured capability and permitted target adaptation must be specified in each case.}
\label{tab:generalization-evidence}
\begingroup
\setcitestyle{numbers,square,comma}
\footnotesize
\setlength{\tabcolsep}{4pt}
\renewcommand{\arraystretch}{1.12}
\begin{tabularx}{\linewidth}{@{}L{2.1cm}L{3.65cm}Y L{4.05cm}@{}}
\toprule
\textbf{Held out} & \textbf{What changes} & \textbf{Claim supported by testing} & \textbf{Example / test} \\
\midrule
Map / level & Layout, placement, visual~setting & Robustness within specified rules & Procgen: unseen levels \citep{cobbe2020procgen} \\
\addlinespace[3pt]
Mode & Objectives, opponents, difficulty, or rules & Adaptation to named rule/task variants & Atari mode transfer \citep{farebrother2018generalization} \\
\addlinespace[3pt]
Game & Title; potentially rules, assets, controls & Transfer of the evaluated behavior or model function & SIMA: task success \citep{sima2024} \\
\bottomrule
\end{tabularx}
\endgroup
\end{table}

Three kinds of transfer should be identified explicitly. \emph{Visual representation} transfer concerns recognition
or generation under changed appearance. \emph{Action-control} transfer concerns whether commands retain their
intended effect and timing. \emph{Rule reasoning} concerns selecting or explaining valid consequences when
objectives, legal actions, or transitions change. SIMA evaluates instructed behavior in games held out from behavioral cloning, but its pretrained encoders were fine-tuned on visual data from some of those games. Goat Simulator~3 was excluded from that fine-tuning, which makes its transfer result unconfounded \citep{sima2024}. SCOPE and GameFactory test action-conditioned generation in unseen visual scenes \citep{tong2026scope,yu2025gamefactory}, which supports appearance and control transfer and leaves each game's original mechanics untested. Tests isolating appearance, control mappings, and rules can
attribute a gain to one component; joint-change tests examine whether those components still work together.

A generated image can establish a starting scene or illustrate a hypothesis. Evidence for transfer comes from
what happens after actions: held-out tasks, rule-sensitive interventions, reference transitions, or sustained
play under stated budgets. Even a generated rollout should be identified as a selected demonstration or a
systematically scored test. Zero-shot use, retrieval of target demonstrations, and target-game fine-tuning
test different forms of transfer. Public gameplay, code,
walkthroughs, or benchmark repositories may overlap with pretraining data.
SpireBench shows how such overlap can be probed: in the complete commercial deck-builder Slay the Spire, a
comprehensive, self-consistent renaming of every game term shows that part of model performance rests on prior
knowledge of the game, and frontier models that match the average human player fall behind experts as runs
lengthen \citep{lowe2026spirebench}.

\subsubsection{Measurement Validity and Evaluator Dependence}

The source of a verdict matters as much as its label. The fidelity targets in
\Cref{tab:world-fidelity-tests} can be assessed with instruments of different strength. StatePlay separately measures
numerical state error and uses VLM judgments for mechanics fidelity \citep{lin2026stateplaystateawaregameworld}, while WorldMind's reported NPC preferences are
also model-judged \citep{deng2026worldminddecoupledgameworld}. GameGen-Verifier
combines VLM judgments with programmatic assertions and compares repeated-run verdicts with human labels
\citep{jia2026gamegenverifier}. Agreement across related judges is not
a substitute for an engine check, and aggregate accuracy can hide missed defects when most test cases pass.
NCP-Bench similarly interprets narrative consistency through structured LLM audits, with checks of auditor
sensitivity and a human review of a subset of outcomes \citep{ma2026ncpbench}.

\subsubsection{Human-Study Validity}

Human-centered results require equally careful comparisons. Repeated level starts by the same players are not
independent participants, and a deployment's calendar duration does not establish repeated-session exposure for
each person. In personalized PCG, groups already differ before individual gameplay history is available because
their initial levels use different generation policies; later differences cannot be attributed to personalization
alone \citep{hafnar2025zeroshot}. Profile inference, content generation, and adaptation can be separated
experimentally by keeping the generator fixed while varying the profile or adaptation policy.

\subsubsection{Reproducibility and Data Provenance}

A reproducible evaluation needs the model and engine versions, prompts, observation and action interfaces,
tool permissions, randomization settings, and retry or selection budgets. For changing APIs and leaderboards,
the specific model or benchmark version should accompany the score so that later model or benchmark updates can be distinguished.

Gameplay data, retrieved assets, source code, and private project traces also need an identifiable origin and
clear access conditions. Reports should specify which inputs can be redistributed, which depend on private
resources, and which player records are retained or updated. When a component cannot be released, documenting
its interface and preserving representative inputs, outputs, and evaluation traces still helps others assess
what can be reproduced.

The evidence map in \Cref{tab:benchmark-overview} summarizes the measured targets, evidence sources, and
interpretation limits of the reviewed evaluation settings.

\begingroup
\footnotesize
\setcitestyle{comma}
\setlength{\tabcolsep}{4pt}
\renewcommand{\arraystretch}{1.0}
\setlength{\LTleft}{0pt plus 1fill}
\setlength{\LTright}{0pt plus 1fill}
\setlength{\LTcapwidth}{\linewidth}
\begin{longtable}{@{}
>{\raggedright\arraybackslash}p{\dimexpr0.23\linewidth-1.5\tabcolsep\relax}
>{\raggedright\arraybackslash}p{\dimexpr0.32\linewidth-1.5\tabcolsep\relax}
>{\raggedright\arraybackslash}p{\dimexpr0.16\linewidth-1.5\tabcolsep\relax}
>{\raggedright\arraybackslash}p{\dimexpr0.29\linewidth-1.5\tabcolsep\relax}
@{}}
\caption{Benchmarks and evaluation settings across the six roles: measured targets, principal evidence sources, and limits of interpretation.}
\label{tab:benchmark-overview} \\
\toprule
\textbf{Benchmark / setting} & \textbf{Task and measure} & \textbf{Evidence} & \textbf{Comparison limits} \\\midrule
\endfirsthead
\multicolumn{4}{@{}l}{\tablename~\thetable\ continued} \\
\toprule
\textbf{Benchmark / setting} & \textbf{Task and measure} & \textbf{Evidence} & \textbf{Comparison limits} \\\midrule
\endhead
\midrule
\multicolumn{4}{r@{}}{\emph{Continued on the next page}} \\
\endfoot
\bottomrule
\endlastfoot
\addlinespace[4pt]
\multicolumn{4}{@{}l}{\textcolor{Player}{\textbf{Play and Act}}} \\*

TextWorld / Jericho / TextArena~[\citenum{cote2018textworld,hausknecht2020jericho,guertler2025textarena}]
& Generated quests, authored adventures, and single-/multi-player text games; task-specific outcomes
& Play; executable checks
& Command vocabulary, rule access, partners, and scoring conventions differ \\
\addlinespace[1pt]

NLE / Crafter~[\citenum{kuttler2020nle,hafner2022crafter}]
& Procedural exploration and survival; game scores and achievement coverage
& Play; executable checks
& Progress within a game, not arbitrary-rule transfer \\
\addlinespace[1pt]

BALROG~[\citenum{paglieri2024balrog}]
& Six game environments; text/visual inputs; normalized progress
& Play; executable checks
& Abstracted actions; native motor timing excluded \\
\addlinespace[1pt]

VideoGameBench~[\citenum{zhang2025videogamebench}]
& Full: 10 test games played through real-time interaction; Lite: three development and three test games, with the game paused during model inference; checkpoint progress
& Gameplay traces
& Different game sets; progress is not completion \\
\addlinespace[1pt]

GameWorld~[\citenum{ouyang2026gameworld}]
& 170 tasks / 34 browser games; state-verifiable success and partial progress
& Interaction; state checks
& Native vs. semantic actions; game paused during inference \\
\addlinespace[1pt]

ARC-AGI-3~[\citenum{arcprize2026benchmark,arcprize2026astra}]
& Unfamiliar abstract environments; human-relative action efficiency
& Interaction; human reference
& Harness, reasoning-state handling, and effort settings affect scores \\
\addlinespace[1pt]

Gym-Anything-RT~[\citenum{aggarwal2026gymanythingrt}]
& 100 generated real-time minigames; pass rate in real time vs.\ paused
& Interaction; paired paused condition
& Agent-generated minigames, not commercial games \\
\addlinespace[1pt]

RealtimeBench~[\citenum{pan2026realtimebench}]
& 180 tasks in six game types; progress score with the game running during inference
& Interaction; paired paused condition
& Structured text state, not pixels \\
\addlinespace[1pt]

AtariBench / PlaySuite~[\citenum{huo2026ataribench,varghese2026playsuite}]
& 20 Atari games; 5,734 games from itch.io and PyWeek; game score or judged progress
& Interaction with the game paused; model judge in PlaySuite
& No real-time reaction; PlaySuite progress rests on a model judge \\
\addlinespace[1pt]

SpireBench~[\citenum{lowe2026spirebench}]
& Complete Slay the Spire runs at the highest difficulty; average-human and expert references
& Play; renamed-term control for prior knowledge
& One deck-builder; harness differences affect comparison \\
\addlinespace[1pt]

MC-EvoBench~[\citenum{song2026mcevobench}]
& Minecraft evolution chains; source-to-target transfer through one declared artifact
& Stage-level environment verifiers; matched budgets and seeds
& Early bottlenecks can leave later capabilities unobserved \\
\addlinespace[1pt]

FlashAdventure~[\citenum{ahn2025flashadventure}]
& 34 GUI adventures; full-story progress
& Interaction; model-assisted verification
& Clue/inventory persistence; verification is not an executable rule check \\
\addlinespace[1pt]

OmniGameArena~[\citenum{lin2026omnigamearena}]
& 12 UE5 games; solo/competitive/cooperative play; improvement after reflection
& Play; executable outcomes
& Skill-prompt revision; not weight learning or first-attempt performance \\
\addlinespace[1pt]

PuzzleJAX~[\citenum{earle2025puzzlejax}]
& Human-designed executable puzzles; outcomes for search, RL, and language agents
& Play; executable checks
& Rule diversity and exposure; search/training budgets matter \\
\addlinespace[1pt]

\addlinespace[4pt]
\multicolumn{4}{@{}l}{\textcolor{Simulator}{\textbf{Model Players and Games}}} \\*

Atari policy transfer~[\citenum{kaiser2020simple,alonso2024diamond}]
& Train in learned dynamics; policy return in reference ALE
& Reference-game execution
& Decision utility, not player-facing simulation or cross-game transfer \\
\addlinespace[1pt]

iWorld-Bench / WorldExam~[\citenum{fang2026iworldbench,yang2026worldexam}]
& Controlled interaction clips; action following, memory, and reactivity
& Interaction-based probes
& Control adapters and task definitions differ \\
\addlinespace[1pt]

PlayWorld~[\citenum{ding2026playworld}]
& 171 objectives; geometry consistency; interaction fidelity; Insight Evolution; out-of-sight evolution
& Agent rollouts; model judgments
& Scores rest on one player policy and one verifier, and instruction failures affect the rubric \\
\addlinespace[1pt]

WorldMark v2~[\citenum{xu2026worldmark}]
& 500 cases; control accuracy, purity, latency, stability
& Shared action adapters
& Visual/memory scores separate from control; no full rule oracle \\
\addlinespace[1pt]

WildBench~[\citenum{li2026wildworld}]
& Video quality; camera/action control; pose alignment
& Instrumented game data
& One ARPG; pose alignment is not resource correctness \\
\addlinespace[1pt]

MultiWorldBench~[\citenum{xu2026multiworldbench}]
& 495 multiplayer cases; ten capabilities including shared state, persistence, and revisit
& Engine reference; human pairwise check
& Minecraft only; two multiplayer systems evaluated \\
\addlinespace[1pt]

PROWBench~[\citenum{huang2026prowbench}]
& 170 program-built episodes; 600 proxy videos; interaction success and logic adherence
& Engine event record; model judge
& Judge not calibrated against human annotation \\
\addlinespace[1pt]

CombatStateBench~[\citenum{huang2026programmable}]
& Visible counts and post-event deaths
& Engine scenario; VLM frame checks
& Sampled frames; death check does not resolve identity \\
\addlinespace[1pt]

WorldOlympiad~[\citenum{zhao2026worldolympiad}]
& Physical, geometric, and interaction diagnostics
& Model-based diagnostics
& Diagnostic probes, not player-experience evidence \\
\addlinespace[1pt]

State / NPC studies~[\citenum{lin2026stateplaystateawaregameworld,deng2026worldminddecoupledgameworld,wang2026reactivegwmsteeringnpcreactive,zhu2026incantationnaturallanguageaction}]
& State error; mechanics fidelity; NPC reactivity; entity transfer
& Reference checks; model and human judgments
& Selected variables/domains; model judgments differ from numerical state error \\
\addlinespace[1pt]

WorldRoamBench~[\citenum{xu2026worldroambenchopenworldbenchmarklonghorizon}]
& 1000+ cases; action, vision, physics, and memory; 10--60 s
& Interaction-based probes
& Fixed probes and finite horizons \\
\addlinespace[1pt]

Maia4All~[\citenum{tang2025maia4all}]
& Held-out move prediction for individual chess players
& Human move records
& Chess-specific; move accuracy does not validate preference \\
\addlinespace[1pt]

Beyond Asking~[\citenum{lu2026personalized}]
& Synthetic-trait recovery; profile-conditioned adaptation; exploratory human study
& Controlled bot parameters; human study
& Synthetic recovery, real-player validity, and adaptation benefit are distinct \\
\addlinespace[1pt]

\addlinespace[4pt]
\multicolumn{4}{@{}l}{\textcolor{Creator}{\textbf{Design}}} \\*

Levels and rules~[\citenum{sudhakaran2023mariogpt,todd2024gavel,nasir2025mortar}]
& Solvability, constraints, prompt adherence, and simulated play
& Solvers; simulated play
& Validity and agent preference differ from player value \\
\addlinespace[1pt]

Science Birds competitions~[\citenum{taveekitworachai2023chatgpt4pcg,taveekitworachai2024chatgpt4pcg2,llms4pcg2025}]
& Structure stability, letter resemblance, and diversity
& Physics and shape checks
& Formal criteria, not human enjoyment \\
\addlinespace[1pt]

Co-creative authoring~[\citenum{liapis2013sketchbook,earle2025dreamgarden}]
& Designer edits and plan inspection; creator control and usability
& Creator studies
& Small, task-specific studies \\
\addlinespace[1pt]

\addlinespace[4pt]
\multicolumn{4}{@{}l}{\textcolor{Builder}{\textbf{Build and Maintain}}} \\*

GameDevBench / GameEngineBench~[\citenum{chi2026gamedevbench,la2026gameenginebench}]
& 333 Godot tasks / 110 Unreal tasks; executable success
& Runtime and acceptance tests, plus an LLM judge for the Unreal tasks
& Different engines, tasks, and acceptance criteria \\
\addlinespace[1pt]

Mage / JAMER~[\citenum{liu2026mage,sun2026jamer}]
& Mage: scene/runtime/static-mechanic checks; JAMER: generation/code completion
& Compilation, runtime, static/behavioral checks
& Bounded checks do not establish long-term maintenance \\
\addlinespace[1pt]

PlaytestArena / GameCraft-Bench~[\citenum{huang2026guigames,luo2026gamecraft}]
& Browser / Godot artifacts; interaction-grounded rubrics
& Agent play or replay; model judgments
& Coverage and judge reliability; weighted scores are not pass rates \\
\addlinespace[1pt]

WebGameBench / GameXpert-Bench~[\citenum{zhang2026webgamebench,chen2026gamexpert}]
& Requirement delivery; repair and six-turn cumulative revision
& Play; executable tests; criteria judging
& Bounded requests; long-term handoff untested \\
\addlinespace[1pt]

CEGame-Bench~[\citenum{dong2026cegamebench}]
& Complete games from design documents in five production engines
& Hidden real-input play to the ending
& Engine-dependent rankings; long, costly runs \\
\addlinespace[1pt]

UnityBench-Mini3D / OpenGameEval~[\citenum{zhou2026unitybench,turkel2026opengameeval}]
& 77 Unity tasks with 1,178 milestone tests; 84 core Roblox Studio tasks
& In-engine executable tests; scene and/or scripted play checks
& One engine each; scripted rather than human play \\
\addlinespace[1pt]

3D GameDev Arena / OCTAVE~[\citenum{bai2026gamedevarena,li2026octave}]
& 108 from-scratch 3D browser games; 180 HTML5 tasks with 1,800 requirements
& Code-level tests, agent playtests, human pairwise votes; static LLM review in OCTAVE
& Per-game adapted tests; OCTAVE judge matches humans on 149 of 241 human-assessable game--requirement pairs (61.8\%) \\
\addlinespace[1pt]

IGMBench~[\citenum{ku2026worldediting}]
& 110 modding tasks in Minecraft and Terraria at four intervention depths
& In-game state and behavior checks; regression checks
& Two games; one run per model and task \\
\addlinespace[1pt]

SWE-Game~[\citenum{chen2026swegame}]
& 247 tasks from 41 Godot reference games: construction, repair, and porting
& Engine-state checks; reference-input replay; VLM presentation rubrics
& Godot-centred; porting tested only to Unity \\
\addlinespace[1pt]

\addlinespace[4pt]
\multicolumn{4}{@{}l}{\textcolor{Resident}{\textbf{Generate and Adapt at Runtime}}} \\*

PANGeA / IF:CARGO~[\citenum{buongiorno2024pangea,hsu2026ifcargo}]
& PANGeA: input validation; IF:CARGO: executable player-authored rules
& Model validation; engine execution
& Validator accuracy and rule execution concern different outputs \\
\addlinespace[1pt]

NCP-Bench~[\citenum{ma2026ncpbench}]
& 100 narrative environments; fact and plot consistency
& Structured model audits
& Adversarial simulated players and LLM auditors \\
\addlinespace[1pt]

Player studies / deployments~[\citenum{zhu2023calypso,peng2024emergence,hsu2026doubleedged,hafnar2025zeroshot}]
& Agency, workload, trust, completion, and experience
& Human studies; deployment records
& Protocols/exposure differ; deployment duration is not longitudinal exposure \\
\addlinespace[1pt]

Generation / latency ablations~[\citenum{yin2026contextualized,roso2026latency}]
& Item/dialogue factors; onset vs. streaming delay
& Controlled human studies
& Within-system comparisons; bounded encounters \\
\addlinespace[1pt]

\addlinespace[4pt]
\multicolumn{4}{@{}l}{\textcolor{Evaluator}{\textbf{Test and Evaluate}}} \\*

GameGen-Verifier~[\citenum{jia2026gamegenverifier}]
& 100 web games; injected-state checks; expert-label agreement
& Assertions; model verdicts; expert labels
& Acc@5 averages five runs; injection bypasses ordinary reachability \\
\addlinespace[1pt]

PlaytestArena / PlayWorld~[\citenum{huang2026guigames,ding2026playworld}]
& Agent-generated traces; rubric or visual-question-answering verdicts
& Interaction; model verdicts
& Reaching a condition and judging it are separate tasks \\
\addlinespace[1pt]

CA2 / SAGE / SMART~[\citenum{adaikkappan2026ca2,cai2025sage,mu2025smart}]
& Code coverage; regression selection; update-relevant testing
& Instrumented execution; bug predicates
& Coverage is not correctness; SAGE uses predefined bug triggers \\
\addlinespace[1pt]

TITAN~[\citenum{wang2025titan}]
& Long-horizon task execution and commercial-game bug discovery
& Interaction; LLM bug oracle
& Task completion, bug discovery, and deployment claims are distinct \\
\addlinespace[1pt]

Human relevance~[\citenum{ariyurek2021testing}]
& Human-trajectory similarity; seeded-defect detection
& Human traces; executable bug oracles
& Sampled behavior; not a population guarantee \\
\addlinespace[1pt]

FAIRGAMER~[\citenum{shi2026fairgamer}]
& NPC decision disparities under demographic cues
& Controlled decision comparisons
& Targeted bias diagnostics; not live multiplayer fairness \\
\addlinespace[1pt]

VideoGameQA-Bench~[\citenum{taesiri2025videogameqa}]
& Screenshot/video diagnosis, localization, and reporting
& VLM diagnoses evaluated against references
& The VLM is the tested diagnostic system; no active exploration \\
\addlinespace[1pt]

Text-game bug detection~[\citenum{jin2024bugdetection}]
& Narrative/logical defects in logs; recall of player-reported bugs
& Model verdicts, player survey
& Recorded-interaction detection; no active coverage \\
\addlinespace[1pt]
\end{longtable}
\endgroup

Taken together, these evaluations support claims within specified tasks and protocols. The next section discusses
how outputs connect roles, what information those connections must preserve, and which downstream claims require
additional evidence.

\gameaisectionaccent{GameInk}
\section{Discussion: Cross-Role Connections}
\label{sec:synthesis}

Sections~3--8 examined individual roles, and \cref{sec:evaluation} considered how their evidence should be interpreted. We now discuss what emerges when these roles are connected: how broader model interfaces interact with game-specific structure, what information passes between components, and whether those exchanges improve downstream workflows.

\subsection{Broader Interfaces and Game-Specific Structure}

Language, multimodal, and code interfaces let AI participate in several stages of a game workflow, but the connection to executable behavior differs across systems. DreamGarden turns a high-level prompt into an editable hierarchical plan whose leaves call implementation modules in Unreal Engine \citep{earle2025dreamgarden}. GameWorld maps semantic actions to deterministic keyboard and mouse operations \citep{ouyang2026gameworld}. In these cases, a flexible model-facing interface connects to a more constrained plan, tool, or action representation. What the model can express and what the game can execute remain distinct.

Learned simulation offers a contrasting case. GameNGen learns action-conditioned visual transitions from gameplay and supports interaction through neural simulation rather than executing the original game's rules at each generated step \citep{valevski2024gamengen}. It nevertheless has a defined action interface and must be assessed for the consistency of the behavior it produces. Game-specific action semantics and state requirements therefore remain relevant whether dynamics are implemented explicitly or approximated by a learned model.

The resulting design question is where a workflow needs precise constraints and where learned approximation is useful. A generated plan needs enough structure for implementation; a control command needs a defined effect; and a reused observation must preserve the information required for later decisions. Broader interfaces change which parts of a workflow can be learned, generated, or connected, while reliability depends on the receiving component's requirements.

\subsection{Artifact Reuse and Capability Transfer}

As defined in \cref{sec:taxonomy}, \emph{artifact reuse} concerns an identifiable output consumed by another task or role, whereas \emph{capability transfer} concerns competence under a changed game, engine, interface, player population, or task. In the representative pipelines reviewed here, reuse is visible in implemented component connections. Artifact reuse can support capability transfer, but does not by itself establish it. A transfer claim requires specifying what changes and testing whether the relevant competence survives.

\Cref{tab:cross-role-flows} organizes implemented exchanges around the information they carry: interaction traces, imagined experience, specifications, test feedback, and player profiles. Its final column states outcomes to validate, not a claim that every listed system has completed all of those checks. For example, ScriptDoctor uses compiler errors and search feedback to revise PuzzleScript games \citep{earle2025scriptdoctor}; Play2Code returns a GUI agent's play summary and fix list to a coding agent \citep{huang2026guigames}; and Beyond Asking uses an inferred profile to condition adaptation \citep{lu2026personalized}. Iterative repair with a fixed model is not the same claim as learning a generally reusable repair skill. MC-EvoBench tests the distinction in Minecraft: only a declared agent artifact may cross from a source world to a reset target world, and under matched environment-step budgets and five seeds, related experience yields no stable transfer, with all intervals including zero; early bottlenecks also leave later capabilities unobserved \citep{song2026mcevobench}.

\begin{table}[H]
\centering
\caption{Representative implemented cross-role exchanges and downstream outcomes to validate.}
\label{tab:cross-role-flows}
\begingroup
\setcitestyle{numbers,square,comma}
\footnotesize
\setlength{\tabcolsep}{3.5pt}
\renewcommand{\arraystretch}{1.10}
\begin{tabularx}{\linewidth}{@{}L{2.60cm}L{2.70cm}L{3.15cm}Y@{}}
\toprule
\textbf{Flow} & \textbf{Artifact} & \textbf{Example system} & \textbf{Outcome to validate} \\
\midrule
\textcolor{Player}{\textbf{Play}} $\rightarrow$ \textcolor{Simulator}{\textbf{Model}} & Action-linked traces & GameNGen \citep{valevski2024gamengen} & Prediction fidelity; interactive consistency \\
\addlinespace[2pt]
\textcolor{Simulator}{\textbf{Model}} $\rightarrow$ \textcolor{Player}{\textbf{Play}} & Imagined trajectories & Dreamer~4 \citep{hafner2025dreamer4} & Reference-game policy performance \\
\addlinespace[2pt]
\textcolor{Creator}{\textbf{Design}} $\rightarrow$ \textcolor{Builder}{\textbf{Build}} & Editable plan & DreamGarden \citep{earle2025dreamgarden} & Executable output satisfying requirements \\
\addlinespace[2pt]
\textcolor{Builder}{\textbf{Build}} $\rightarrow$ \textcolor{Simulator}{\textbf{Model}} & Executable state; rendering controls & Programmable World Model \citep{huang2026programmable} & Generated views agreeing with engine events \\
\addlinespace[2pt]
\textcolor{Evaluator}{\textbf{Test}} $\rightarrow$ \textcolor{Creator}{\textbf{Design}} & Compiler/search feedback & ScriptDoctor \citep{earle2025scriptdoctor} & Compilation; solvability within search budget \\
\addlinespace[2pt]
\textcolor{Creator}{\textbf{Design}} $\rightarrow$ \textcolor{Resident}{\textbf{Runtime}} & Narrative structure & NarrativeGenie \citep{kumaran2024narrativegenie} & Narrative continuity; intended player experience \\
\addlinespace[2pt]
\textcolor{Evaluator}{\textbf{Test}} $\rightarrow$ \textcolor{Builder}{\textbf{Build}} & Play/repair feedback & Play2Code \citep{huang2026guigames} & Corrected failures; unaffected behavior preserved \\
\addlinespace[2pt]
\textcolor{Evaluator}{\textbf{Test}} $\rightarrow$ \textcolor{Simulator}{\textbf{Model}} & Tests/simulation feedback & Agent2World \citep{hu2025agent2world} & Dynamics matching target behavior \\
\addlinespace[2pt]
\textcolor{Simulator}{\textbf{Model}} $\rightarrow$ \textcolor{Resident}{\textbf{Runtime}} & Player profile & Beyond Asking \citep{lu2026personalized} & Adaptation benefit; separate from profile accuracy \\
\bottomrule
\end{tabularx}
\par\smallskip
\begin{minipage}{\linewidth}
\footnotesize
Rows denote implemented exchanges; the final column lists downstream outcomes to validate, not outcomes established by every cited study.
\end{minipage}
\endgroup
\end{table}

Reuse also requires preserving the meaning of the exchanged information. An action trace needs the corresponding observations and control conventions; a design plan needs its accepted requirements; and a player profile needs the behavioral context in which it was inferred. A recipient may otherwise parse an output correctly but use it incorrectly. For instance, a successful search trace certifies a solution only under the associated rules and search conditions, not after arbitrary edits to the game. This is a connection-level concern: it cannot be resolved by evaluating the producing and receiving components in isolation.

Exposing an intermediate representation is a useful step toward reuse, but it is not itself an implemented downstream application. StatePlay exposes selected numerical variables alongside frames \citep{lin2026stateplaystateawaregameworld}, while WorldMind incorporates reconstructed state into NPC behavior \citep{deng2026worldminddecoupledgameworld}. \citet{zhou2026trajectory} use game-development trajectories for executable scene construction and report downstream embodied-policy experiments; those results concern world construction and policy utility rather than action-conditioned video-game dynamics.

\runin{Executable grayboxes and generative rendering}
\label{sec:engine-rendering}
A concrete emerging connection separates game construction, state execution, and visual synthesis. Programmable World Model uses a coding agent to create entity programs, a lightweight engine to execute them, and state-augmented 3D boxes compiled into controls for a video renderer \citep{huang2026programmable}. In another implementation, Generative World Renderer conditions appearance on engine G-buffers, and its AlayaRenderer-Flash follow-up brings that rendering path to a live game engine \citep{huang2026worldrenderer,lin2026rendererflash}. Unlike Programmable World Model, these G-buffer renderers do not generate the engine's executable rules. Matrix-Game 4.0 combines both directions in real time: a lightweight game engine maintains world state and executes interaction rules, a multimodal coding agent translates player ideas into game code and scene captions, and a distilled autoregressive video renderer turns engine-rendered low-poly proxies into detailed visuals at 480p and 60~FPS with 112~ms key-press-to-display latency on a single RTX~5090, demonstrated in shooter and driving games with engine-enforced mechanics \citep{zhang2026matrixgame4}.

A corresponding UE/Unity workflow could start with a code agent translating intent into an executable \emph{graybox}: simple geometry, controls, collision, objectives, and tests. Retrieved or generated assets would replace its proxies while retaining object identities and behavior bindings, building on the construction workflows of UniGen and AutoUE \citep{yang2025unigen,yin2026autoue}. During play, the engine would update positions, health, inventory, and events. A state-conditioned video model would then use a low-cost render, depth, masks, motion, and other selected buffers to synthesize the displayed appearance. Input would return to the engine, so generated pixels would not silently redefine collision or score. This full production-engine combination remains a research direction.

The benefit is that visual richness need not carry the whole burden of rule execution. The difficulty moves to agreement between the executable scene and what players see. Asset replacement can alter collision scale; a video-to-video pass can paint a door where the graybox has a wall; and delayed rendering can show an obsolete state. Comparing a graybox, an asset-enriched engine render, and a state-conditioned neural render under the same action traces would isolate these effects. Tests should combine engine invariants, visible object/event alignment, input-to-display latency, and player control errors. Changing the intended layout must revise the executable scene before that new layout becomes a player-facing promise.

\begin{gameaiinsight}[GameInk]{What must cross a role boundary?}
\begin{insightpoints}
\item \textbf{The usable record.} GameNGen consumes action-linked trajectories \citep{valevski2024gamengen}; Play2Code consumes a summary of attempted interactions and a fix list \citep{huang2026guigames}. Both reuse gameplay, but the receiving computation needs different information about actions, state, and requirements.
\item \textbf{The conditions behind a verdict.} An output can retain a verified property under unchanged assumptions while still needing a new evaluation for a different use. A playable simulation, for example, is not yet evidence that training inside it improves a policy when evaluated in the reference game.
\end{insightpoints}
\end{gameaiinsight}

\subsection{Validating Downstream Benefits}
\label{sec:downstream-validation}

Three questions help assess a cross-role connection and its downstream effects. \textbf{Compatibility:} does the output preserve the format and meaning required by the receiving component? \textbf{Use:} does that component actually consume the output during the evaluated workflow? \textbf{Benefit:} does the connection improve the intended downstream outcome relative to a suitable baseline, at a reported cost? Compatibility does not establish actual use, and use alone does not establish benefit.

The downstream outcome depends on the connection. For a model reused for policy training, it is behavior in the reference game; for test feedback, it is a repaired project that also preserves unaffected requirements; for a player profile, it is the experience produced by the resulting adaptation. Where the goal is to isolate the value of the exchanged information, comparisons should hold the receiving component and total budget as constant as possible while removing, replacing, or varying that information. GameNGen's comparison of training on agent-generated and random-policy trajectories \citep{valevski2024gamengen} illustrates this approach: the experiments test how the data source affects subsequent predictions.

Feedback loops also create characteristic blind spots. GameGen-Verifier injects states to check local mechanics \citep{jia2026gamegenverifier}, whereas Play2Code supplies feedback from played trajectories \citep{huang2026guigames}. The first route can inspect a condition without demonstrating an ordinary path to it; the second can reveal such a path but is limited by what the tester visits. A repair optimized only against either feedback source could leave other failures unchanged. Preserving legal reproduction traces, retesting unaffected requirements, and checking held-out behaviors would test whether the connection improves the project rather than only the feedback score.

A similar feedback issue arises when a player profile changes the content that generates the next behavioral record. Avoiding an item after adaptation may reflect the opportunities offered by the new level rather than a change in preference. This motivates retaining content and intervention context alongside player observations, consistent with the opportunity-aware representation examined in Beyond Asking \citep{lu2026personalized}. Upstream prediction accuracy and downstream experience should then be tracked separately.

Claims of improvement require validation in the intended setting, especially when rules, project versions, or player contexts change. The research agenda in the next section therefore concerns both stronger components and connections that preserve their useful information over time.

\gameaisectionaccent{GameInk}
\section{Open Challenges and Research Directions}
\label{sec:challenges}

The evidence reviewed in \cref{sec:evaluation} and the discussion of cross-role connections in \cref{sec:synthesis} motivate six challenges. \Cref{tab:capability-frontier} summarizes the setting-dependent findings and the gaps addressed below. Each subsection connects a technical problem to a candidate approach and a test of progress. The role label identifies where the problem first arises, not an exclusive boundary: a change to a model, design, or test can affect several downstream components.

\begin{table}[H]
\centering
\caption{Representative evidence and research gaps across the six roles. Findings are setting-dependent, not a ranking of role maturity.}
\label{tab:capability-frontier}
\begingroup
\setcitestyle{numbers,square,comma}
\footnotesize
\setlength{\tabcolsep}{3.5pt}
\renewcommand{\arraystretch}{1.10}
\begin{tabularx}{\linewidth}{@{}L{1.55cm}Y L{3.10cm}L{3.65cm}@{}}
\toprule
\textbf{Role} & \textbf{Reported evidence} & \textbf{Evaluation conditions} & \textbf{Research gap} \\
\midrule
\textcolor{Player}{\textbf{Play}} & Cross-game skills \citep{sima22025,magne2026nitrogen}; rule inference \citep{skoutnev2026twin} & Selected games; specified adaptation budgets & Joint transfer across rules, controls, and timing \\
\addlinespace[2pt]
\textcolor{Simulator}{\textbf{Model}} & State prediction \citep{lin2026stateplaystateawaregameworld}; NPC behavior \citep{deng2026worldminddecoupledgameworld}; player prediction \citep{tang2025maia4all} & Domain-specific state and behavior data & Persistent world state; reliable player-model updates and transfer \\
\addlinespace[2pt]
\textcolor{Creator}{\textbf{Design}} & Solvable levels \citep{sudhakaran2023mariogpt}; rule search \citep{todd2024gavel}; designer steering \citep{earle2025dreamgarden} & Fixed representations; solver/creator feedback & Intended player experience; sustained creative control \\
\addlinespace[2pt]
\textcolor{Builder}{\textbf{Build}} & Complete projects \citep{luo2026gamecraft}; scoped edits \citep{la2026gameenginebench}; short revision chains \citep{chen2026gamexpert} & Specified engines, requests, tests & Evolving requirements; developer handoff \\
\addlinespace[2pt]
\textcolor{Resident}{\textbf{Runtime}} & Constrained rule execution \citep{hsu2026ifcargo}; player-outcome studies \citep{hsu2026doubleedged,hafnar2025zeroshot} & Bounded encounters or field deployments & Cross-session consistency, experience, and shared-state fairness \\
\addlinespace[2pt]
\textcolor{Evaluator}{\textbf{Test}} & Mechanic verification \citep{jia2026gamegenverifier}; human-reference defect testing \citep{ariyurek2021testing}; fairness diagnostics \citep{shi2026fairgamer} & Specified access, oracles, sampled behaviors & Independent verification; broader behavioral coverage \\
\bottomrule
\end{tabularx}
\endgroup
\end{table}

\subsection{Play and Act: Transfer Across Interfaces and Rules}

Generalist policies benefit from recurring visual affordances, language goals, and shared control representations, but an unfamiliar game can change appearance, action timing, and hidden rules at once. SIMA~2 \citep{sima22025} and NitroGen \citep{magne2026nitrogen} evaluate different forms of transfer, while Cradle \citep{tan2025cradle} and Orak \citep{park2025orak} illustrate contrasting assumptions about native and structured interfaces. The next problem is deciding which prior skill applies, which interface mapping must be relearned, and which apparent failure reflects an unknown rule rather than poor control.

Connecting executable rule acquisition to real-time control offers one route forward. Twin constructs an executable model from simulation and interaction \citep{skoutnev2026twin}, whereas Code World Models uses supplied natural-language rules and game trajectories to synthesize an executable model \citep{lehrach2025cwm}. A hybrid agent could use uncertainty over candidate rules to choose informative actions, then hand routine movement back to a fast controller. The unresolved issue is when the cost of another experiment is justified during play. Target-game demonstrations, semantic queries, model calls, and exploratory actions all consume resources; comparing only the final score would hide that cost.

The map/mode/game distinctions in \Cref{tab:generalization-evidence} suggest factorial tests: retain rules while changing layout or appearance, remap controls without changing the task, then alter legal actions or objectives while retaining the visual scene. These tests could identify which component transfers. Joint changes under a fixed real-time and interaction budget would then test whether the components work together. Social games add another held-out factor: partner conventions. Evaluations with unfamiliar teammates should distinguish successful execution from compatible coordination, following the motivation of Other-Play \citep{hu2020otherplay} and Fictitious Co-Play \citep{strouse2021fcp}.

\subsection{Model Players and Games: Persistent World State and Reliable Player Models}
\label{sec:challenge-models}

Visual history, spatial memory, and explicit variables preserve different information. WorldMem retrieves relevant memory frames using stored states \citep{xiao2025worldmem}; ReWorld retrieves pose-indexed landmarks from a bounded memory bank \citep{chen2026reworld}; PERSIST maintains a latent 3D scene \citep{garcin2026persist}; StatePlay predicts selected numerical state \citep{lin2026stateplaystateawaregameworld}; WorldMind uses reconstructed state to choose NPC behavior \citep{deng2026worldminddecoupledgameworld}. These approaches raise a common unresolved question: which information should be remembered, recomputed, or invalidated after an action? Returning an old view is wrong if an object has moved, just as regenerating a plausible reward is wrong if it has already been collected.

A candidate approach is to couple generative observations with selectively explicit, updatable state. It requires identifying variables whose errors alter later decisions, defining how actions update them, and resolving disagreement between stored state and generated observations. A rule change may also invalidate an earlier state update. Tests could revisit a location after moving an object, spending a resource, changing a rule, or triggering a delayed event. Save--restore and synchronized-view tests would expose contradictions that uninterrupted, single-view video metrics overlook. MultiWorldBench provides an initial synchronized-view and delayed-revisit test for multiplayer worlds \citep{xu2026multiworldbench}. The engine--renderer route in \Cref{sec:engine-rendering} offers one testable way to divide this work. Progress would require visual improvements without increasing rule violations, state disagreement, or player control errors, including after replacing assets or revising a mechanic. Memory cost, action latency, and correction frequency should be measured together.

WorldRoamBench's diagnostic probes \citep{xu2026worldroambenchopenworldbenchmarklonghorizon} and PlayWorld's active objectives \citep{ding2026playworld} offer starting points. BadWorld additionally shows that visually subtle adversarial perturbations can destabilize later rollouts, suggesting robustness tests beyond ordinary sampled play \citep{shen2026badworld}. Active evaluation needs both a competent probing policy and an independently checked account of what the world should do under the tested actions.

Player models face a distinct persistence problem: an observed change in behavior may reflect learning, a temporary goal, or different opportunities to act. Maia4All's individual move prediction \citep{tang2025maia4all} and Beyond Asking's controlled synthetic trait recovery \citep{lu2026personalized} provide different starting points. The technical challenge is to update a useful model without treating every behavioral change as a permanent preference. A context-aware model could retain the conditions of each observation, distinguish stable tendencies from session-specific state, and retain uncertainty under sparse observations.

Progress should be tested on later, held-out behavior, with separate evaluations of adaptation to genuine change and resistance to misleading observations. Useful comparisons include a fixed profile, a recent-history predictor, and an updated model under the same observation budget. Cross-game claims additionally require testing which player information remains meaningful when actions and opportunities change. Player inspection and correction, as proposed by open and player-centered modeling, provide a further interface for contesting an inferred profile \citep{zhu2021open,zhu2021player}. Whether using that profile improves experience is the separate runtime question in \cref{sec:challenge-runtime}.

\subsection{Design: Designer Intent, Diversity, and Control}

Broader generation does not remove the difficulty of specifying a good game. GAVEL evolves games and mechanics in an executable rule language \citep{todd2024gavel}, while Mortar evolves mechanics and evaluates them in complete games assembled through tree search \citep{nasir2025mortar}. Both optimize particular properties of simulated play. A design may score well under an evaluator that does not represent all intended play styles. Equally, a designer may only discover the intended experience after inspecting several proposals. Treating the initial prompt as a complete, fixed specification misses this iterative design process.

Supporting this process requires separating hard constraints from revisable preferences, preserving diverse playable alternatives, and exposing the effects of proposed edits before replacing accepted content. Tanagra's constraint-preserving edits \citep{smith2011tanagra}, Sentient Sketchbook's alternatives \citep{liapis2013sketchbook}, and DreamGarden's editable plans \citep{earle2025dreamgarden} supply concrete precedents at different scales. The challenge is to extend these forms of control to language- and code-based generators while retaining dependencies between mechanics, levels, and narrative branches. A local request should not silently erase an unrelated design decision.

Evaluation should follow actual revision work: whether designers can locate the relevant choice, reach an intended result, retain alternatives, and recover from an unwanted edit. Artifact tests can measure validity and variation; creator studies can measure correction effort and control; player studies can test the resulting experience. Comparing these outcomes would reveal whether a richer generator expands the practical design space or merely produces more candidates to inspect.

\subsection{Build and Maintain: Reliable Revision and Developer Handoff}

Current benchmarks reveal failures that span code, scenes, assets, and engine conventions. GameEngineBench tests scoped changes in existing projects \citep{la2026gameenginebench}; JAMER reports a sharp performance drop as project scale increases \citep{sun2026jamer}; GameXpert-Bench adds cumulative requests and regression criteria \citep{chen2026gamexpert}. Maintaining a game requires linking these problems over time. A visible failure may originate outside the edited file, while a later requirement may intentionally invalidate an earlier test.

Version-aware project context could retain accepted requirements, dependency links, reproducible traces, and tests as editable project artifacts rather than only as conversation history. SAGE \citep{cai2025sage} and SMART \citep{mu2025smart} connect update information to test selection or exploration, but deciding which assertion should change is a separate problem. A repair agent could propose both a patch and its expected behavioral effect, with independent checks for unchanged requirements. Giving the same agent unrestricted control over implementation and acceptance tests would reduce the independence of that evidence.

Multi-version benchmarks could combine feature additions, dependency changes, scene edits, and bug reports while retaining tests for unaffected behavior. Human handoff adds a different endpoint: a developer who did not generate the project must reproduce a failure and implement a new request. Completion time, erroneous edits, and the amount of reconstruction needed would test maintainability more directly than readability or the generating agent's success.

\subsection{Generate and Adapt at Runtime: Cross-Session Consistency and Player Experience}
\label{sec:challenge-runtime}

The runtime studies discussed here report mixed effects on agency, cognitive load, trust, and enjoyment within bounded encounters \citep{peng2024emergence,hsu2026doubleedged,lee2026leaguebot}. Cross-session play introduces changes in the player as well as the system: skill develops, preferences vary by context, character memories accumulate, and models are updated. Preserving a transcript is insufficient if a saved quest, an inferred preference, or a character's recollection no longer matches the current game.

Building on the player-model questions in \cref{sec:challenge-models}, the runtime challenge is to turn uncertain predictions into appropriate, reversible interventions. A system could retain which quest facts are confirmed, which preferences are inferred, and which adaptations the player has accepted or rejected. Conservative fallbacks and player-visible controls would then limit the consequences of an uncertain profile or incompatible model update. The technical trade-off is between responsive personalization and continuity: changing content to match a new prediction should not silently invalidate an established event, difficulty choice, or shared game state.

To isolate the contribution of adaptation, comparisons should hold the generator and starting conditions fixed while varying the adaptation policy or its access to player information. Otherwise, easier initial content or stronger generation can be mistaken for successful personalization. Evaluation should include both the benefits of correct interventions and the cost of inappropriate changes, including the player's effort to correct or reverse them.

Repeated-session trials can examine continuity, correction, and agency alongside enjoyment. Planned model updates and resumed saves would test whether established events survive system changes. Non-adaptive controls, usage logs, and interviews can help distinguish novelty from sustained benefit \citep{karapanos2009uxovertime}. For multiplayer systems, these studies also need to observe whether personalized information, response delays, or generated rewards create unequal opportunities within shared play.

\subsection{Test and Evaluate: Independent Verification and Behavioral Coverage}

Automated testing faces three separable limitations: the policy may miss a relevant state, the oracle may misjudge observed behavior, and the sampled behavior may fail to represent the player population relevant to a player-facing claim. GameGen-Verifier makes local mechanic checks cheaper through state injection \citep{jia2026gamegenverifier}; CA2 directs exploration using code \citep{adaikkappan2026ca2}; procedural personas broaden the goals a tester pursues \citep{holmgard2019playtesting}. A different risk appears when the generator learns from the evaluator: model judges can exhibit self-preference \citep{panickssery2024selfpreference}, and optimization against an imperfect verifier can reward shortcuts \citep{helff2026llms}. Repeating a judge or changing its prompt does not by itself establish independent verification.

Combining complementary exploration policies with independent oracles could improve both coverage and verification, while reserving expensive visual or human inspection for uncertain and consequential cases. Injected states would still need invariant checks and, when ordinary reachability is part of the claim, a legal reproduction trace. Withheld defect families, execution-based checks, and periodic human annotations could test whether repairs generalize beyond the feedback used to produce them. Judge disagreement is useful diagnostic information: it can indicate hidden state, misleading presentation, or an incorrect assertion.

For defect-discovery systems, a useful outcome is the number of additional validated defects found within a fixed testing budget, with false positives and reproduction cost reported alongside coverage. Human relevance needs its own sampling: novice and expert trajectories, different strategies \citep{ariyurek2021testing}, and targeted fairness cases \citep{shi2026fairgamer} answer different questions. Keeping these references separate would let automated testing support both software reliability and player-facing decisions.

\medskip
\noindent Across all six research directions, deployment depends on how outputs are controlled and used in a live workflow. Research prototypes should make tool permissions, data retention, and version changes explicit; player-facing systems should support correction or withdrawal of inferred profiles, consistent with the goals of open player modeling \citep{zhu2021open}. End-to-end evaluations should report compute and latency alongside task outcomes. They should also exercise rollback or fallback behavior after a failed update and test whether personalized actions remain compatible with shared multiplayer state.

\section{Conclusion}
\label{sec:conclusion}

Foundation models are expanding the parts of a game that AI can interpret, generate, predict, and revise, while their use across the game lifecycle continues to depend on explicit game structure, including controls, rules, state representations, engine interfaces, and human decisions. Across the six roles surveyed here, the most concrete cross-role connections arise through the reuse of interaction traces, learned environments or imagined experience, specifications and rules, execution and test evidence, and player models or profiles. These exchanges enable new workflows, but artifact reuse does not by itself establish capability transfer: performance remains conditioned by the game, engine, interface, player population, and task in which a system is evaluated. Evidence thins as outputs move from bounded play toward sustained use: a match can be scored when it ends, whereas an evolving software project, a persistent world, or a repeated player encounter has to be observed over time. Progress in AI for games will depend on carrying the information and constraints that make an output useful as it moves between roles, without losing developer control, game consistency, or relevance to the players these systems are intended to serve.

\clearpage
\pagestyle{gameairefs}
\phantomsection
\addcontentsline{toc}{section}{References}
\makeatletter
\renewcommand{\@biblabel}[1]{[#1]}
\let\@bibsetup\NAT@bibsetnum
\makeatother
\bibliographystyle{plainnat}
\bibliography{references}

\clearpage
\pagestyle{gameai}
\appendix
\pretitlemark*{section}{System Index}
\section{System Index}
\thispagestyle{gameairesources}
\label{app:resources}

\noindent\small
The index compares interfaces, mechanisms, and evaluation settings behind the
main-text discussion. The tables follow the six roles, with game models and
player models separated for lookup. References link each row to its source.
\Cref{app:game-examples} provides an illustrated guide to selected game demonstrations and their public resources.

\smallskip
\noindent
\indexrole{P} Play and Act \quad
\indexrole{M} Model Players and Games \quad
\indexrole{D} Design \quad
\indexrole{B} Build and Maintain \quad
\indexrole{R} Runtime \quad
\indexrole{T} Test and Evaluate

\smallskip
\noindent
The first code identifies the indexed system's role; parentheses mark additional
functions. Named test components are identified separately from their parent
systems. A dash denotes a benchmark, dataset, or framework rather than an AI
system. Industry rows identify official workflow reports separately from measured research outcomes.
Each table carries four columns; the two wide columns hold two sub-fields apiece, and a bold
inline lead-in names the sub-field that follows, as the column heading lists them.

\begingroup
\setcitestyle{numbers,square,comma}
\fontsize{7.8}{9.15}\selectfont
\setlength{\tabcolsep}{3pt}
\renewcommand{\arraystretch}{1.13}
\hyphenpenalty=7000
\exhyphenpenalty=7000
\begin{longtable}{@{}L{2.90cm}L{1.00cm}J{5.70cm}J{6.10cm}@{}}
\caption{Game-playing systems and their transfer conditions. Grouped entries share a technical purpose, not a common score scale.}\label{tab:agent-systems}\\
\toprule
\textbf{System / source} & \textbf{Role} & \textbf{Environment and mechanism} & \textbf{Data and evaluation}\\
\midrule
\endfirsthead
\multicolumn{4}{@{}l}{\footnotesize\itshape Table~\thetable\ continued}\\
\toprule
\textbf{System / source} & \textbf{Role} & \textbf{Environment and mechanism} & \textbf{Data and evaluation}\\
\midrule
\endhead
\midrule
\multicolumn{4}{r}{\footnotesize\itshape Continued on the next page}\\
\endfoot
\bottomrule
\endlastfoot
\indexgroup{Player}{4}{Shared policies and learned controllers}
\textbf{VPT} \citep{baker2022vpt}
 & \indexrole{P} & \textbf{Env:}~Minecraft; pixels; keyboard/mouse\newline \textbf{Mech:}~Inverse-dynamics labels; BC, then~RL & \textbf{Data:}~About 70K h of video; smaller labeled~set\newline \textbf{Eval:}~Native control; Minecraft-specific pretraining \\
\textbf{Gato} \citep{reed2022gato}
 & \indexrole{P} & \textbf{Env:}~Atari plus language/robotics~tasks\newline \textbf{Mech:}~One autoregressive, multimodal~policy & \textbf{Data:}~Mixed offline~demonstrations\newline \textbf{Eval:}~Shared weights across tasks; adaptation evaluated~separately \\
\textbf{Multi-Game DT}~\citep{lee2022multigamedt}
 & \indexrole{P} & \textbf{Env:}~Atari; pixels and discrete~actions\newline \textbf{Mech:}~Offline return-conditioned Transformer & \textbf{Data:}~41 training / 5 held-out~games\newline \textbf{Eval:}~Shared training-game policy; new titles require~fine-tuning \\
\textbf{DreamerV3} \citep{hafner2025dreamerv3}
 & \indexrole{P} (\indexrole{M}) & \textbf{Env:}~Visual or vector~observations\newline \textbf{Mech:}~Latent world model; imagination~RL & \textbf{Data:}~150+ tasks; one~configuration\newline \textbf{Eval:}~Algorithm reuse with per-task~learning \\
\textbf{STEVE-1} \citep{lifshitz2023steve1}
 & \indexrole{P} & \textbf{Env:}~Minecraft; language goals; native~actions\newline \textbf{Mech:}~VPT policy; latent goals; hindsight & \textbf{Data:}~Relabeled Minecraft~behavior\newline \textbf{Eval:}~Text-to-behavior transfer within one~game \\
\textbf{SIMA / SIMA 2}~\citep{sima2024,sima22025}
 & \indexrole{P} & \textbf{Env:}~3D worlds; pixels; keyboard/mouse\newline \textbf{Mech:}~Shared instruction following; demonstrations & \textbf{Data:}~Multi-world data; held-out~worlds\newline \textbf{Eval:}~Bounded tasks; SIMA 2 adds generated~feedback \\
\end{longtable}
\endgroup

\Needspace{180pt}
\begingroup
\setcitestyle{numbers,square,comma}
\fontsize{7.8}{9.15}\selectfont
\setlength{\tabcolsep}{3pt}
\renewcommand{\arraystretch}{1.13}
\hyphenpenalty=7000
\exhyphenpenalty=7000
\begin{longtable}{@{}L{2.90cm}L{1.00cm}J{5.70cm}J{6.10cm}@{}}
\caption{Learned game models: what they represent, how they are controlled, and what is evaluated. Reported speed and horizon use different hardware and protocols.}\label{tab:world-model-systems}\\
\toprule
\textbf{System / source} & \textbf{Role} & \textbf{Representation and control} & \textbf{Data and evidence}\\
\midrule
\endfirsthead
\multicolumn{4}{@{}l}{\footnotesize\itshape Table~\thetable\ continued}\\
\toprule
\textbf{System / source} & \textbf{Role} & \textbf{Representation and control} & \textbf{Data and evidence}\\
\midrule
\endhead
\midrule
\multicolumn{4}{r}{\footnotesize\itshape Continued on the next page}\\
\endfoot
\bottomrule
\endlastfoot
\indexgroup{Simulator}{4}{Planning and policy-learning models}
\textbf{World Models}~\citep{ha2018worldmodels}
 & \indexrole{M} (\indexrole{P}) & \textbf{Repr:}~VAE + recurrent latent~dynamics\newline \textbf{Control:}~Continuous control; visual~observations & \textbf{Data:}~CarRacing and~VizDoom\newline \textbf{Evidence:}~Controller return; transfer from imagined to real~rollouts \\
\textbf{MuZero} \citep{schrittwieser2020muzero}
 & \indexrole{P} (\indexrole{M}) & \textbf{Repr:}~Value-, reward-, and policy-trained latent~state\newline \textbf{Control:}~Discrete actions; tree~search & \textbf{Data:}~Atari, Go, chess, shogi\newline \textbf{Evidence:}~Decision quality; no image reconstruction requirement \\
\textbf{Dreamer / DreamerV3}~\citep{hafner2020dreamer,hafner2025dreamerv3}
 & \indexrole{P} (\indexrole{M}) & \textbf{Repr:}~Recurrent latent dynamics; imagined~trajectories\newline \textbf{Control:}~Actions and~rewards & \textbf{Data:}~Visual and vector control~tasks\newline \textbf{Evidence:}~Policy return in reference~environments \\
\textbf{I2A} \citep{racaniere2017i2a}
 & \indexrole{P} (\indexrole{M}) & \textbf{Repr:}~Encoded rollouts from an imperfect~model\newline \textbf{Control:}~Discrete~actions & \textbf{Data:}~Sokoban; MiniPacman\newline \textbf{Evidence:}~Sokoban performance with accurate and imperfect environment~models \\
\textbf{Action-conditional prediction}~\citep{oh2015actionconditional}
 & \indexrole{M} & \textbf{Repr:}~Feedforward / recurrent image~prediction\newline \textbf{Control:}~Recorded Atari~actions & \textbf{Data:}~Five Atari~games\newline \textbf{Evidence:}~Prediction error; selected control~tests \\
\textbf{Recurrent simulators}~\citep{chiappa2017res}
 & \indexrole{M} & \textbf{Repr:}~Recurrent latent propagation; image~decoder\newline \textbf{Control:}~Environment action~labels & \textbf{Data:}~Atari, racing, 3D~mazes\newline \textbf{Evidence:}~Hundreds-step visual~rollouts \\
\end{longtable}
\endgroup

\Needspace{180pt}
\begingroup
\setcitestyle{numbers,square,comma}
\fontsize{7.8}{9.15}\selectfont
\setlength{\tabcolsep}{3pt}
\renewcommand{\arraystretch}{1.13}
\hyphenpenalty=7000
\exhyphenpenalty=7000
\begin{longtable}{@{}L{2.90cm}L{1.00cm}J{5.70cm}J{6.10cm}@{}}
\caption{Player modeling by prediction target. Input, inferred construct, and intervention outcome occupy different columns; datasets and frameworks are marked with a dash.}\label{tab:player-model-systems}\\
\toprule
\textbf{System / source} & \textbf{Role} & \textbf{Input and construct} & \textbf{Validation and boundary}\\
\midrule
\endfirsthead
\multicolumn{4}{@{}l}{\footnotesize\itshape Table~\thetable\ continued}\\
\toprule
\textbf{System / source} & \textbf{Role} & \textbf{Input and construct} & \textbf{Validation and boundary}\\
\midrule
\endhead
\midrule
\multicolumn{4}{r}{\footnotesize\itshape Continued on the next page}\\
\endfoot
\bottomrule
\endlastfoot
\indexgroup{Simulator}{4}{Behavior, skill, and individual action prediction}
\textbf{player2vec} \citep{wang2024player2vec}
 & \indexrole{M} & \textbf{Input:}~Mobile-game event~sequences\newline \textbf{Model:}~Masked-token~Longformer & \textbf{Data:}~Masked-event recovery; embedding~structure\newline \textbf{Boundary:}~No independent preference~labels \\
\textbf{Behavior Structformer}~\citep{smirnov2024behaviorstructformer}
 & \indexrole{M} & \textbf{Input:}~Structured tracking~events\newline \textbf{Model:}~Transformer over typed event~fields & \textbf{Data:}~Constructed session-count~targets\newline \textbf{Boundary:}~Activity proxy rather than~experience \\
\textbf{TrueSkill} \citep{herbrich2006trueskill}
 & \indexrole{M} & \textbf{Input:}~Match~outcomes\newline \textbf{Model:}~Bayesian skill and~uncertainty & \textbf{Data:}~Predictive performance of~ratings\newline \textbf{Boundary:}~Historical skill model; not preference~inference \\
\textbf{Maia / individual Maia}~\citep{mcilroyyoung2020aligning,mcilroyyoung2020learning}
 & \indexrole{M} & \textbf{Input:}~Chess positions and player~histories\newline \textbf{Model:}~Skill-group / individual behavior~models & \textbf{Data:}~Held-out human~moves\newline \textbf{Boundary:}~Behavior imitation, not strongest~play \\
\textbf{Maia-2} \citep{tang2024maia2}
 & \indexrole{M} & \textbf{Input:}~Position and skill~level\newline \textbf{Model:}~Shared model; skill-aware~attention & \textbf{Data:}~Human moves across rating~groups\newline \textbf{Boundary:}~Population~conditioning \\
\textbf{Maia4All} \citep{tang2025maia4all}
 & \indexrole{M} & \textbf{Input:}~Individual chess~histories\newline \textbf{Model:}~Prototype-enriched player~embedding & \textbf{Data:}~800 positions: 53.2\% move~accuracy\newline \textbf{Boundary:}~Individual prediction within~chess \\
\end{longtable}
\endgroup

\Needspace{180pt}
\begingroup
\setcitestyle{numbers,square,comma}
\fontsize{7.8}{9.15}\selectfont
\setlength{\tabcolsep}{3pt}
\renewcommand{\arraystretch}{1.13}
\hyphenpenalty=7000
\exhyphenpenalty=7000
\begin{longtable}{@{}L{2.90cm}L{1.00cm}J{5.70cm}J{6.10cm}@{}}
\caption{Design and development: proposed artifacts, operating mechanisms, feedback, and evaluation settings. Benchmark rows have no AI-role assignment.}\label{tab:development-systems}\\
\toprule
\textbf{System / source} & \textbf{Role} & \textbf{Artifact and mechanism} & \textbf{Feedback and evaluation}\\
\midrule
\endfirsthead
\multicolumn{4}{@{}l}{\footnotesize\itshape Table~\thetable\ continued}\\
\toprule
\textbf{System / source} & \textbf{Role} & \textbf{Artifact and mechanism} & \textbf{Feedback and evaluation}\\
\midrule
\endhead
\midrule
\multicolumn{4}{r}{\footnotesize\itshape Continued on the next page}\\
\endfoot
\bottomrule
\endlastfoot
\indexgroup{Creator}{4}{Content, rules, and mixed-initiative design}
\textbf{Tanagra / Sentient Sketchbook}~\citep{smith2011tanagra,liapis2013sketchbook}
 & \indexrole{D} & \textbf{Artifact:}~Platformer levels / map~sketches\newline \textbf{Mech:}~Constraint solving / novelty~search & \textbf{Feedback:}~Playability checks; designer~edits\newline \textbf{Eval:}~Historical mixed-initiative authoring \\
\textbf{ANGELINA / Mechanic Miner}~\citep{cook2017angelina1,cook2013mechanicminer}
 & \indexrole{D} & \textbf{Artifact:}~Games and~mechanics\newline \textbf{Mech:}~Automated design; reflective mechanism~search & \textbf{Feedback:}~Simulated play and design~criteria\newline \textbf{Eval:}~Historical automated-game-design systems \\
\textbf{CubePart} \citep{zhu2026cubepart}
 & \indexrole{D} & \textbf{Artifact:}~Part-controllable 3D~assets\newline \textbf{Mech:}~Semantic schemas; separate~meshes & \textbf{Feedback:}~Part identity and~geometry\newline \textbf{Eval:}~Part control; external behavior~scripts \\
\textbf{WorldSculpt} \citep{niu2026worldsculpt}
 & \indexrole{D} & \textbf{Artifact:}~Compositional mesh~scenes\newline \textbf{Mech:}~Posed views; object mesh~prior & \textbf{Feedback:}~Mesh/scene~fidelity\newline \textbf{Eval:}~Unreal reconstruction~tests \\
\textbf{CoBG} \citep{li2026autobg}
 & \indexrole{D} (\indexrole{T}) & \textbf{Artifact:}~Board-game ideas and~rulebooks\newline \textbf{Mech:}~Rule iteration; persona~feedback & \textbf{Feedback:}~Designer dialogue; critic-gated~revision\newline \textbf{Eval:}~207 held-out games; 30-creator study; paper-prototype~playtests \\
\textbf{MarioGPT} \citep{sudhakaran2023mariogpt}
 & \indexrole{D} & \textbf{Artifact:}~Mario tile~levels\newline \textbf{Mech:}~Fine-tuned GPT-2; novelty~search & \textbf{Feedback:}~A* solver; prompt-feature~checks\newline \textbf{Eval:}~250 levels; 88.4\% solved within five~attempts \\
\end{longtable}
\endgroup

\Needspace{180pt}
\begingroup
\setcitestyle{numbers,square,comma}
\fontsize{7.8}{9.15}\selectfont
\setlength{\tabcolsep}{3pt}
\renewcommand{\arraystretch}{1.13}
\hyphenpenalty=7000
\exhyphenpenalty=7000
\begin{longtable}{@{}L{2.90cm}L{1.00cm}J{5.70cm}J{6.10cm}@{}}
\caption{Runtime systems: generated elements, execution context, and human evidence. Raters, authors, and interactive players are distinguished; n.s. means no statistically significant difference was reported.}\label{tab:runtime-systems}\\
\toprule
\textbf{System / source} & \textbf{Role} & \textbf{Generated element and execution} & \textbf{Human evidence and finding}\\
\midrule
\endfirsthead
\multicolumn{4}{@{}l}{\footnotesize\itshape Table~\thetable\ continued}\\
\toprule
\textbf{System / source} & \textbf{Role} & \textbf{Generated element and execution} & \textbf{Human evidence and finding}\\
\midrule
\endhead
\midrule
\multicolumn{4}{r}{\footnotesize\itshape Continued on the next page}\\
\endfoot
\bottomrule
\endlastfoot
\indexgroup{Resident}{4}{Narrative and content systems}
\textbf{CALYPSO} \citep{zhu2023calypso}
 & \indexrole{R} (\indexrole{D}) & \textbf{Element:}~Encounter prose; summaries; ideas\newline \textbf{Execution:}~Human-mediated D\&D; current~scenario & \textbf{Exposure:}~71 players and DMs; field~deployment\newline \textbf{Finding:}~51 chat threads; direct use plus DM~revision \\
\textbf{Concordia} \citep{vezhnevets2025concordia}
 & \indexrole{R} (\indexrole{P}) & \textbf{Element:}~Character actions and narrated~consequences\newline \textbf{Execution:}~Actor / game-master components; memory & \textbf{Exposure:}~Framework demonstrations\newline \textbf{Finding:}~Configurable social simulation; no common player-outcome~trial \\
\textbf{KG-conditioned quests}~\citep{ashby2023quests}
 & \indexrole{R} (\indexrole{D}) & \textbf{Element:}~Personalized quests and~dialogue\newline \textbf{Execution:}~Player input; graph paths; grammar; GPT-2 & \textbf{Exposure:}~Text-rating study; generated-quest analysis\newline \textbf{Finding:}~Intent grounding; unsupported entities remain~possible \\
\textbf{NarrativeGenie} \citep{kumaran2024narrativegenie}
 & \indexrole{D} (\indexrole{R}) & \textbf{Element:}~Story graphs; hints; summaries\newline \textbf{Execution:}~Graph/state dialogue~manager & \textbf{Exposure:}~19 players; one mystery~episode\newline \textbf{Finding:}~Ratings plus 30 simulated-history checks \\
\textbf{Drama Llama}~\citep{sun2025dramallama}
 & \indexrole{R} (\indexrole{D}) & \textbf{Element:}~Responsive text~narrative\newline \textbf{Execution:}~Storylet triggers checked by an~LLM & \textbf{Exposure:}~Six authors; authoring~study\newline \textbf{Finding:}~Authorable responses; not a player-outcome~trial \\
\end{longtable}
\endgroup

\Needspace{180pt}
\begingroup
\setcitestyle{numbers,square,comma}
\fontsize{7.8}{9.15}\selectfont
\setlength{\tabcolsep}{3pt}
\renewcommand{\arraystretch}{1.03}
\hyphenpenalty=7000
\exhyphenpenalty=7000
\begin{longtable}{@{}L{2.90cm}L{1.00cm}J{5.70cm}J{6.10cm}@{}}
\caption{AI testing systems: exploration, oracles, and evaluation. Named testing components are indexed separately.}\label{tab:testing-systems}\\
\toprule
\textbf{System / source} & \textbf{Role} & \textbf{Target and tester} & \textbf{Oracle and evidence}\\
\midrule
\endfirsthead
\multicolumn{4}{@{}l}{\footnotesize\itshape Table~\thetable\ continued}\\
\toprule
\textbf{System / source} & \textbf{Role} & \textbf{Target and tester} & \textbf{Oracle and evidence}\\
\midrule
\endhead
\midrule
\multicolumn{4}{r}{\footnotesize\itshape Continued on the next page}\\
\endfoot
\bottomrule
\endlastfoot
\indexgroup{Evaluator}{4}{Exploration, coverage, and regression}
\textbf{Procedural Personas}~\citep{holmgard2019playtesting}
 & \indexrole{T} & \textbf{Target:}~Executable game~states\newline \textbf{Tester:}~MCTS; evolved goal~heuristics & \textbf{Oracle:}~Game-specific~outcomes\newline \textbf{Evidence:}~Deliberate style diversity; not population~sampling \\
\textbf{Wuji} \citep{zheng2019wuji}
 & \indexrole{T} & \textbf{Target:}~Instrumented combat~games\newline \textbf{Tester:}~Multiobjective evolutionary~RL & \textbf{Oracle:}~Bug predicates; developer~confirmation\newline \textbf{Evidence:}~Two commercial games; three newly confirmed~bugs \\
\textbf{Synthetic / humanlike agents}~\citep{ariyurek2021testing}
 & \indexrole{T} & \textbf{Target:}~Three GVG-AI~games\newline \textbf{Tester:}~Agents with different behavior~objectives & \textbf{Oracle:}~Separate execution oracle; human~trajectories\newline \textbf{Evidence:}~45 seeded bugs; 427 human~traces \\
\textbf{EA navigation AutoPlayers}~\citep{gillberg2023productiontesting}
 & \indexrole{T} & \textbf{Target:}~Battlefield 2042 / Dead Space; structured~state\newline \textbf{Tester:}~RL navigation within scripted test~bots & \textbf{Oracle:}~Existing production QA~infrastructure\newline \textbf{Evidence:}~Navigation deployment, not end-to-end learned~QA \\
\textbf{CA2} \citep{adaikkappan2026ca2}
 & \indexrole{T} & \textbf{Target:}~State/images and call~stacks\newline \textbf{Tester:}~RL guided toward code~targets & \textbf{Oracle:}~Function-entry instrumentation\newline \textbf{Evidence:}~Improved target reachability; diagnosis is~separate \\
\textbf{SAGE} \citep{cai2025sage}
 & \indexrole{T} & \textbf{Target:}~Overcooked Plus / Minecraft; update~logs\newline \textbf{Tester:}~LLM-guided RL; test~prioritization & \textbf{Oracle:}~Predefined bug-trigger~predicates\newline \textbf{Evidence:}~Detection/cost and update adaptation; two custom~environments \\
\end{longtable}
\endgroup

\clearpage
\pagestyle{gameaicases}
\gameaisectionaccent{GameInk}
\pretitlemark*{section}{Game Examples and Interactive Demonstrations}
\section{Game Examples and Interactive Demonstrations}
\label{app:game-examples}

\Cref{tab:game-examples-index} collects 23 examples across game genres and AI roles, from familiar commercial titles to generated prototypes. Twelve illustrated cases follow. Specialist systems provide historical context alongside foundation-model applications; the links distinguish research demonstrations, industry reports, and product features.

\begingroup
\setcitestyle{numbers,square,comma}
\begingroup
\fontsize{9.3}{10.8}\selectfont
\gameaitabletext
\setlength{\tabcolsep}{4pt}
\renewcommand{\arraystretch}{1.06}
\begin{longtable}{@{}L{3.8cm}L{2.25cm}L{1.05cm}L{3.95cm}L{4.1cm}@{}}
\caption{Game examples and public materials. P: Play; M: Model; D: Design; B: Build; R: Runtime; T: Test. $\dagger$: specialist or pre-foundation-model lineage.}\label{tab:game-examples-index}\\
\toprule
\textbf{Game / system} & \textbf{Family} & \textbf{Role} & \textbf{AI contribution} & \textbf{Resource / status} \\
\midrule
\endfirsthead
\multicolumn{5}{@{}l}{\footnotesize\itshape Table~\thetable\ continued}\\
\toprule
\textbf{Game / system} & \textbf{Family} & \textbf{Role} & \textbf{AI contribution} & \textbf{Resource / status} \\
\midrule
\endhead
\midrule
\multicolumn{5}{r}{\footnotesize\itshape Continued on the next page}\\
\endfoot
\bottomrule
\endlastfoot
\indexgroup{Player}{5}{Players, opponents, and teammates}
Minecraft / Voyager~\citep{wang2023voyager} & Sandbox & \indexrole{P} & Code-skill acquisition & \href{https://voyager.minedojo.org/}{Research: code and videos} \\
GT7 / Sophy$^\dagger$~\citep{wurman2022sophy,polyphony2025sophy21} & Racing & \indexrole{P} & Learned rival driving & \href{https://www.gran-turismo.com/us/news/00_4957003.html}{PS5 custom-race feature} \\
PUBG / Ally~\citep{krafton2026allyduo,nvidia2026pubgally} & Battle royale & \indexrole{P},\indexrole{R} & Voice-guided cooperation & \href{https://pubg.com/en/news/10179}{Limited beta: June 2026} \\
StarCraft II / AlphaStar$^\dagger$~\citep{vinyals2019alphastar} & RT strategy & \indexrole{P} & Strategy and unit control & \href{https://deepmind.google/discover/blog/alphastar-mastering-the-real-time-strategy-game-starcraft-ii/}{Research match recordings} \\
Dota 2 / OpenAI~Five$^\dagger$~\citep{berner2019dota} & Team MOBA & \indexrole{P} & Team coordination & \href{https://openai.com/index/openai-five/}{Historical match videos} \\
Diplomacy / CICERO~\citep{fair2022cicero} & Negotiation & \indexrole{P},\indexrole{M} & Strategic dialogue & \href{https://ai.meta.com/research/cicero/}{Research: code and report} \\
\indexgroup{Simulator}{5}{Game dynamics and human behavior}
DOOM / GameNGen~\citep{valevski2024gamengen} & Shooter & \indexrole{M} & Visual simulation & \href{https://gamengen.github.io/}{Research rollout videos} \\
Street Fighter / ReactiveGWM~\citep{wang2026reactivegwmsteeringnpcreactive} & Fighting & \indexrole{M},\indexrole{P},\indexrole{R} & Player and NPC control & \href{https://inv-wzq.github.io/ReactiveGWM/}{Research demos + models} \\
Chess / Maia4All~\citep{tang2025maia4all} & Board game & \indexrole{M} & Individual move forecasts & \href{https://arxiv.org/abs/2507.21488}{Behavior-modeling study} \\
Oasis~\citep{decart2024oasis} & Sandbox & \indexrole{M} & Interactive visual world & \href{https://oasis-model.github.io/}{Demo, code, and weights} \\
\indexgroup{Creator}{5}{Content and co-creative design}
MarioGPT~\citep{sudhakaran2023mariogpt} & Platformer & \indexrole{D} & Text-to-level layouts & \href{https://github.com/shyamsn97/mario-gpt}{Code and level samples} \\
Science Birds / LLMs4PCG~\citep{llms4pcg2025} & Physics puzzle & \indexrole{D} & Letter-shaped structures & \href{https://chatgpt4pcg.github.io/2025-llms4pcg/}{Competition and tests} \\
DreamGarden~\citep{earle2025dreamgarden} & 3D prototypes & \indexrole{D},\indexrole{B} & Design-to-code iteration & Creator study and demos \\
\indexgroup{Builder}{5}{Executable game construction}
GameCraft-Bench~\citep{luo2026gamecraft} & 15 families & \indexrole{B} & Complete Godot projects & \href{https://tongxuluo.github.io/gamecraft-bench-website/}{Playable/video gallery} \\
Play2Code~\citep{huang2026guigames} & Browser games & \indexrole{B},\indexrole{T} & GUI-feedback repair & \href{https://continual-game-generation.vercel.app/}{Eight browser demos} \\
Playco / Playbot~\citep{openai2026playbot} & Prototypes & \indexrole{B} & Greybox-to-themed games & \href{https://openai.com/index/playco-game-prototyping-with-astra/}{Industry case: GPT-6 Astra} \\
\indexgroup{Resident}{5}{Live dialogue, objects, and narrative}
Fortnite / Darth Vader~\citep{epic2025vader} & Battle royale & \indexrole{R} & Voiced NPC dialogue & \href{https://www.fortnite.com/news/this-will-be-a-day-long-remembered-speak-with-darth-vader-in-fortnite}{Seasonal feature: 2025} \\
1001 Nights~\citep{sun2023language,adaeden2026nights} & Adventure & \indexrole{R} & Stories become equipment & \href{https://store.steampowered.com/app/2542850/1001_Nights/}{Demo; release forthcoming} \\
Roblox / Cube 4D~\citep{singh2026robloxcube} & UGC platform & \indexrole{D},\indexrole{R} & Functional objects & \href{https://about.roblox.com/newsroom/2026/02/accelerating-creation-powered-roblox-cube-foundation-model}{Creator-configured beta} \\
AI Dungeon~\citep{latitude2026aidungeon} & Text RPG & \indexrole{R} & Player-directed narrative & \href{https://play.aidungeon.com/}{Live product} \\
\indexgroup{Evaluator}{5}{Testing authored and generated worlds}
Battlefield 2042$^\dagger$~\citep{gillberg2023productiontesting} & Shooter & \indexrole{T} & Helicopter control for QA & \href{https://www.ea.com/seed/news/cog23-challenges-deploying-rl-agents-game-testing}{Production testing report} \\
Dead Space$^\dagger$~\citep{gillberg2023productiontesting} & Survival horror & \indexrole{T} & Zero-gravity QA control & \href{https://arxiv.org/abs/2307.11105}{Production QA study} \\
PlayWorld~\citep{ding2026playworld} & Virtual worlds & \indexrole{T} & Interaction-driven probing & \href{https://kxding.github.io/project/PlayWorld/}{Benchmark videos and code} \\
\end{longtable}
\endgroup

\newcommand{\gamecase}[7]{\begin{tcolorbox}[enhanced,colback=white,colframe=GameLine,
    boxrule=.55pt,arc=1.7mm,left=3mm,right=3mm,top=2.5mm,bottom=2.5mm,
    before skip=7pt,after skip=7pt]
    {\sffamily\bfseries\color{#1}#2}\hfill{\footnotesize\color{GameSlate}#3}\par\smallskip
    \noindent\begin{minipage}[c]{.39\linewidth}
      \centering\includegraphics[width=\linewidth,height=39mm,keepaspectratio]{figures/case_#4.png}
    \end{minipage}\hfill
    \begin{minipage}[c]{.58\linewidth}
      \fontsize{10}{12}\selectfont\gameaitabletext
      \textbf{AI contribution.} #5\par\smallskip
      \textbf{Game context.} #6
    \end{minipage}\par\smallskip
    {\footnotesize\color{GameSlate}#7}
  \end{tcolorbox}}
\newcommand{\casemediacredit}{\par\smallskip{\footnotesize\color{GameSlate}Original media from the linked project teams and game rights holders; no images are reconstructed or generated for this survey. Research demos, product illustrations, and deployed features are identified separately.}}

\clearpage
\subsection{Racing, Team Play, and Learned Simulation}

\gamecase{Player}{Gran Turismo 7 / Sophy}{Learned Opponent}{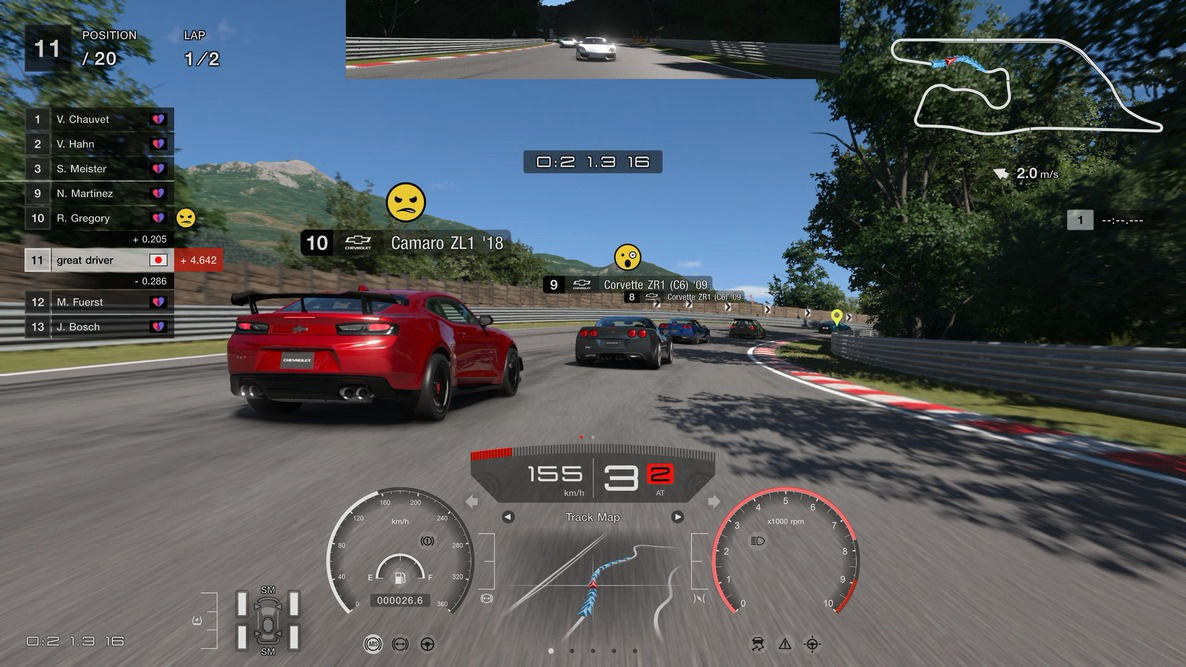}
{A reinforcement-learning driving policy races against people. Sophy 2.1 extends this role to selected custom-race configurations.}
{Tracks, vehicles, physics, and race rules remain supplied by Gran Turismo 7. This is a specialist-policy deployment, not a general language agent.}
{Sources: the Sophy research and official PS5 feature announcement \citep{wurman2022sophy,polyphony2025sophy21}. Shown: official race screenshot. \href{https://www.gran-turismo.com/us/news/00_4957003.html}{Feature and supported settings}.}

\gamecase{Player}{PUBG / Ally Duo}{Co-Playable Character}{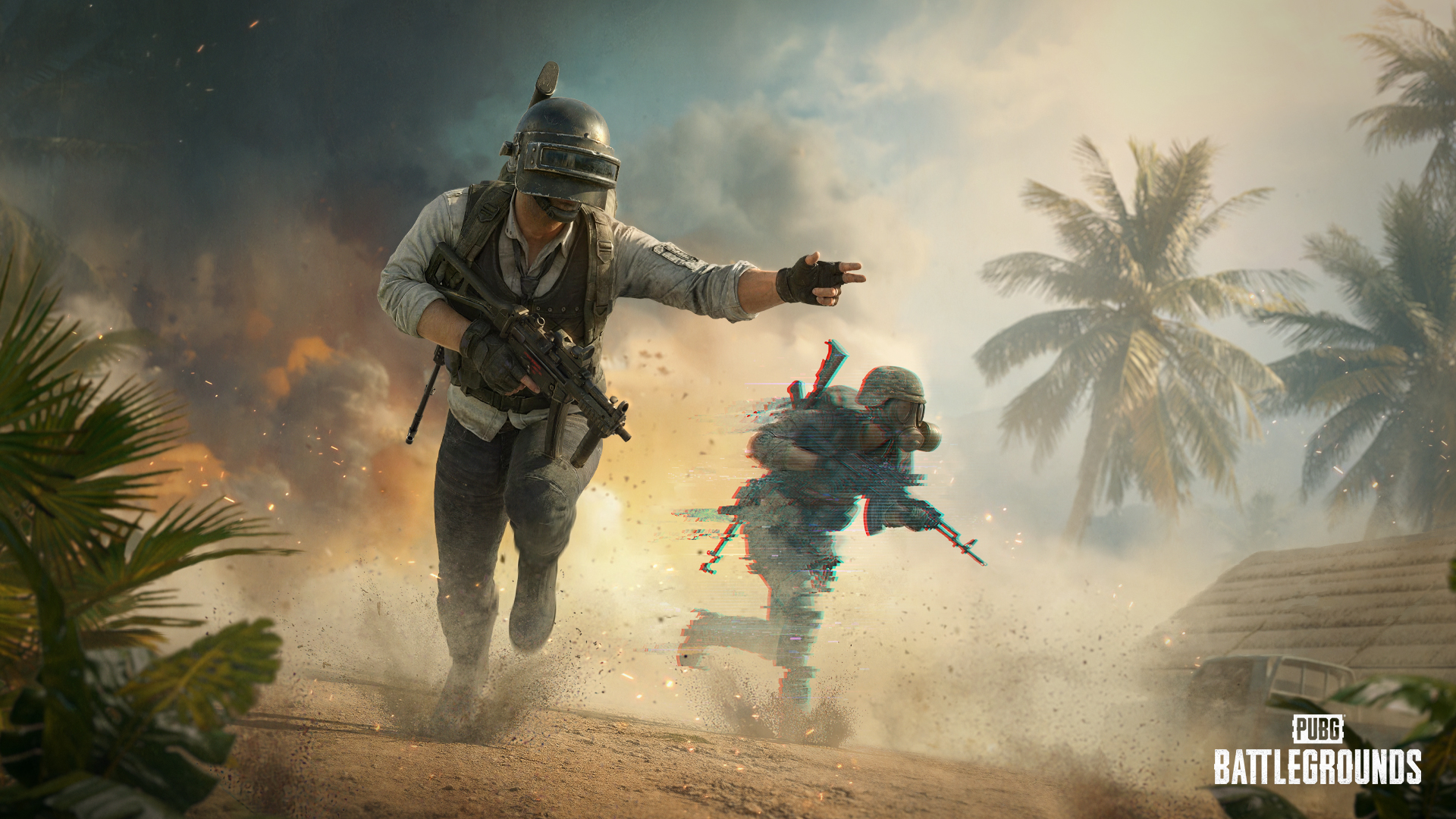}
{Ella combines language-based communication with grounded teammate behavior, allowing spoken requests during a battle-royale match.}
{PUBG supplies the Sanhok map, combat mechanics, controls, and action affordances. The PC beta ran from 17 June to 1 July 2026.}
{Sources: KRAFTON and NVIDIA \citep{krafton2026allyduo,nvidia2026pubgally}. Shown: official mode illustration, not a generated gameplay frame. \href{https://pubg.com/en/news/10179}{Beta details} \enspace\textbullet\enspace \href{https://developer.nvidia.com/blog/how-krafton-built-pubg-ally-a-co-playable-character-powered-by-nvidia-ace/}{Technical account}.}

\gamecase{Simulator}{DOOM / GameNGen}{Learned Environment}{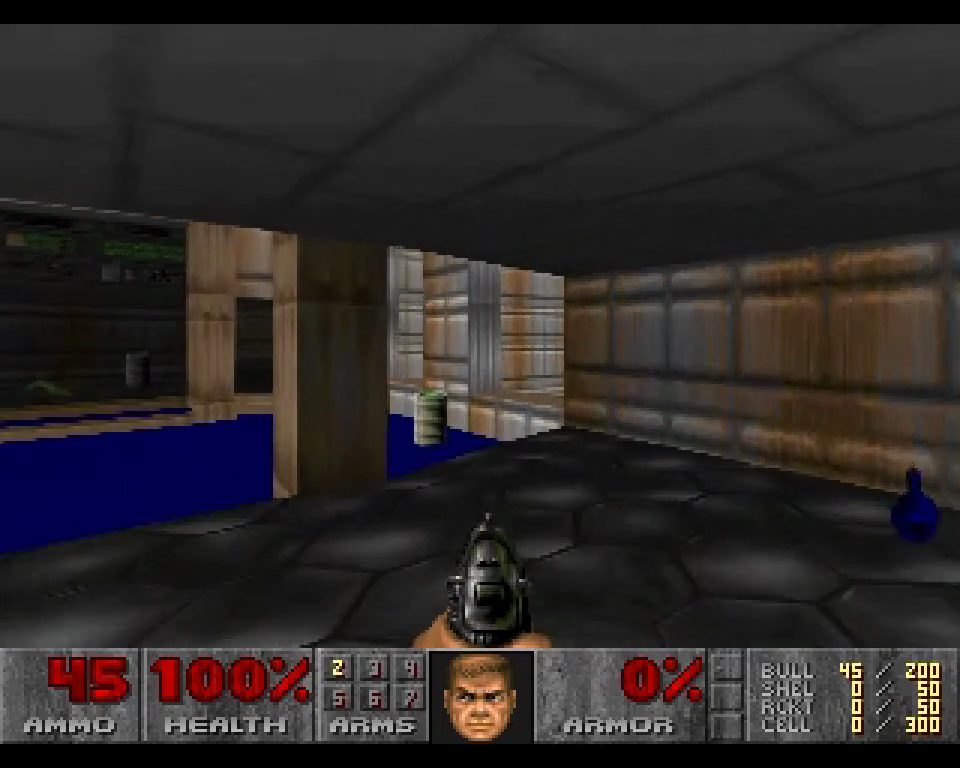}
{An action-conditioned diffusion model produces subsequent game frames. An RL player's trajectories provide training examples from the original game.}
{DOOM supplies the rules and states represented in the training data. The demonstrated learned simulator generates the visual continuation.}
{Source: GameNGen \citep{valevski2024gamengen}. Shown: frame from a research rollout. \href{https://gamengen.github.io/}{Recorded demonstrations}.}

\clearpage
\subsection{Fighting, Platforming, and Physics Puzzles}

\gamecase{Simulator}{Street Fighter / ReactiveGWM}{NPC-Aware Simulation}{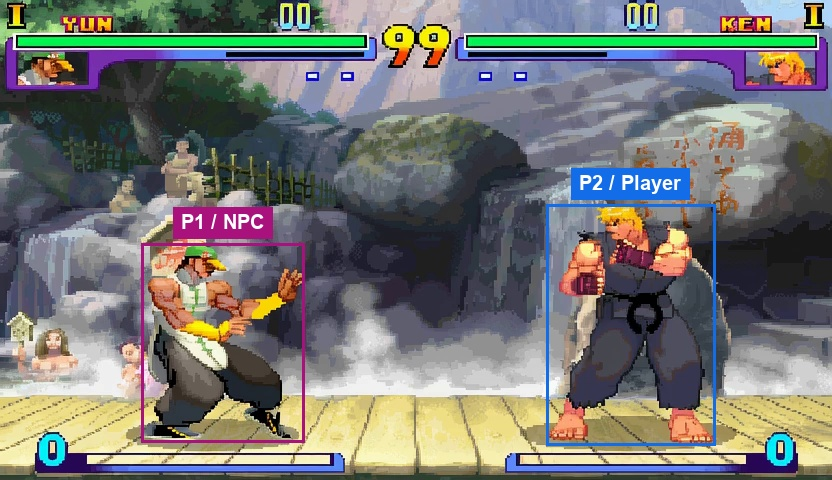}
{Users choose which fighters follow player actions and which act as NPCs, and natural-language if--then rules set how each NPC reacts in the generated fight.}
{Experiments train separate models for Street Fighter III and the open-source fighting game Hikou no Mizu. This is a research simulator, not a released Capcom game feature.}
{Source: ReactiveGWM \citep{wang2026reactivegwmsteeringnpcreactive}. Shown: Street Fighter III scene with player and NPC role labels. \href{https://inv-wzq.github.io/ReactiveGWM/}{Demos, code, and models}.}

\gamecase{Creator}{Mario Levels / MarioGPT}{Level Design}{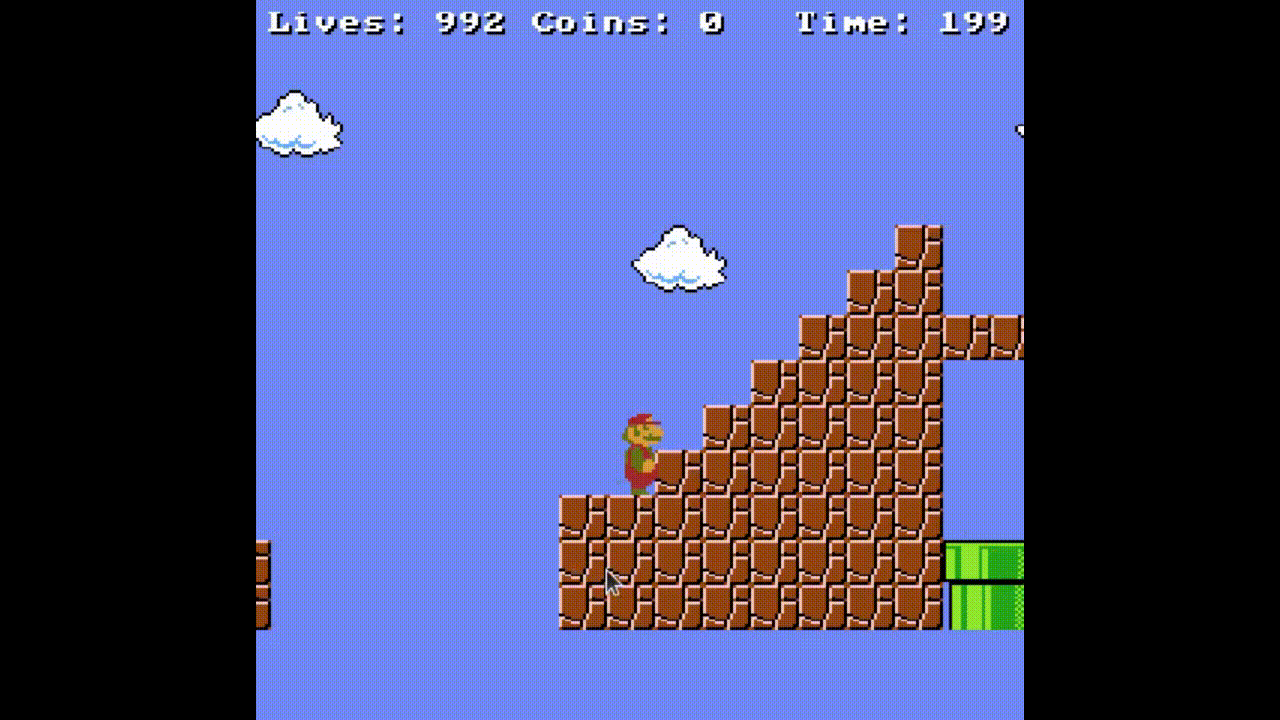}
{A fine-tuned language model proposes tile layouts from requests about pipes, enemies, blocks, and elevation. Levels can be played or checked with A*.}
{An existing Mario-style environment supplies the tile vocabulary, movement rules, renderer, and controller. Generation changes the level, not the engine.}
{Source: MarioGPT \citep{sudhakaran2023mariogpt}. Shown: a level from the authors' interactive example. \href{https://github.com/shyamsn97/mario-gpt}{Code and local player} \enspace\textbullet\enspace \href{https://huggingface.co/spaces/multimodalart/mariogpt}{Hosted demo}.}

\gamecase{Creator}{Science Birds / LLMs4PCG}{Physics-Constrained Design}{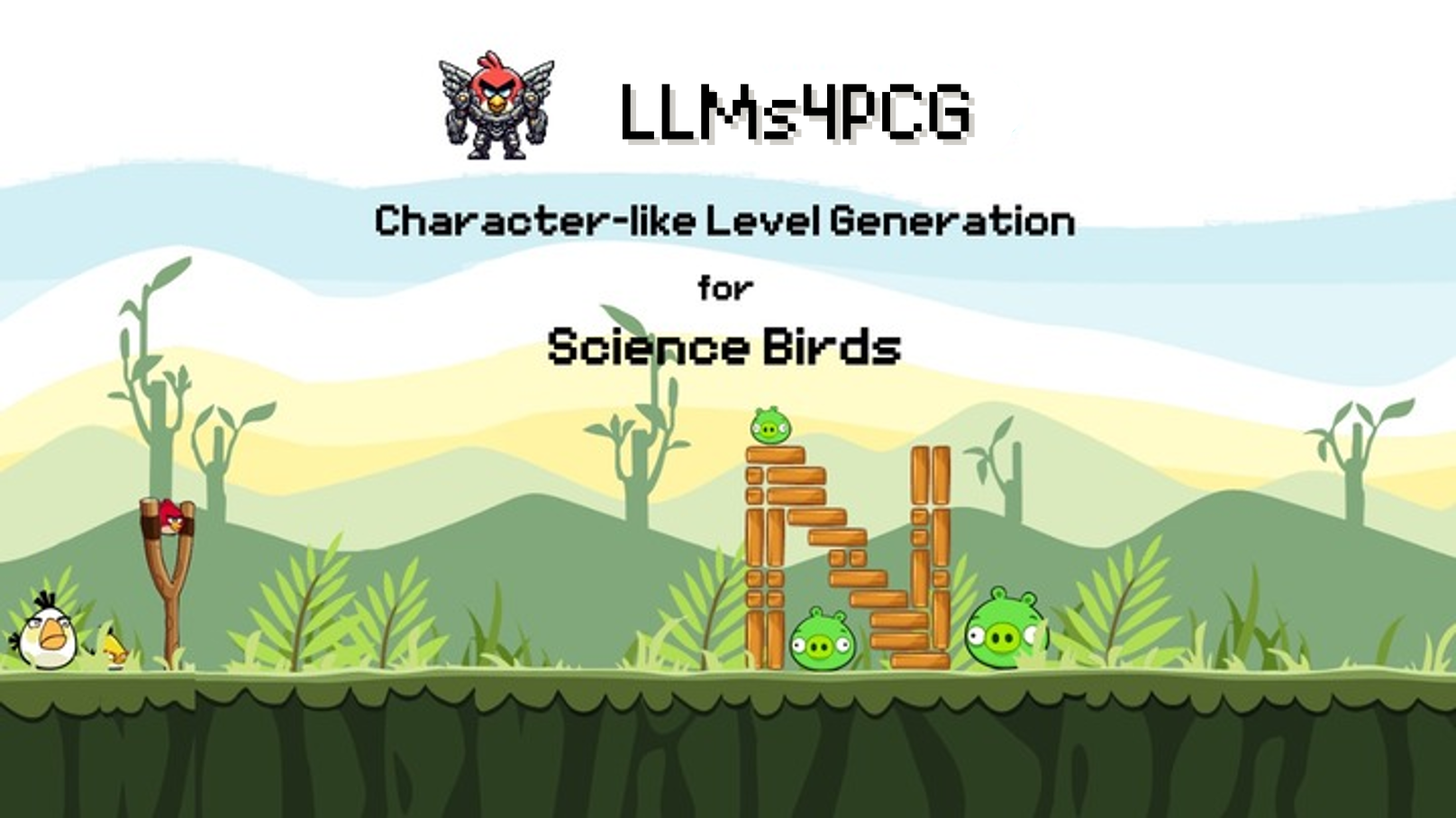}
{Prompted language models arrange blocks into letter-shaped structures. The competition checks stability, shape resemblance, and diversity.}
{Science Birds supplies executable physics and construction elements. This is an Angry Birds-style research platform, not an official Angry Birds release.}
{Source: LLMs4PCG \citep{llms4pcg2025}. Shown: official competition illustration, rather than a scored submission. \href{https://chatgpt4pcg.github.io/2025-llms4pcg/}{Rules, results, and platform}.}

\clearpage
\subsection{Building Games and Generating Live Content}

\gamecase{Builder}{GameCraft-Bench / Godot Games}{Project Construction}{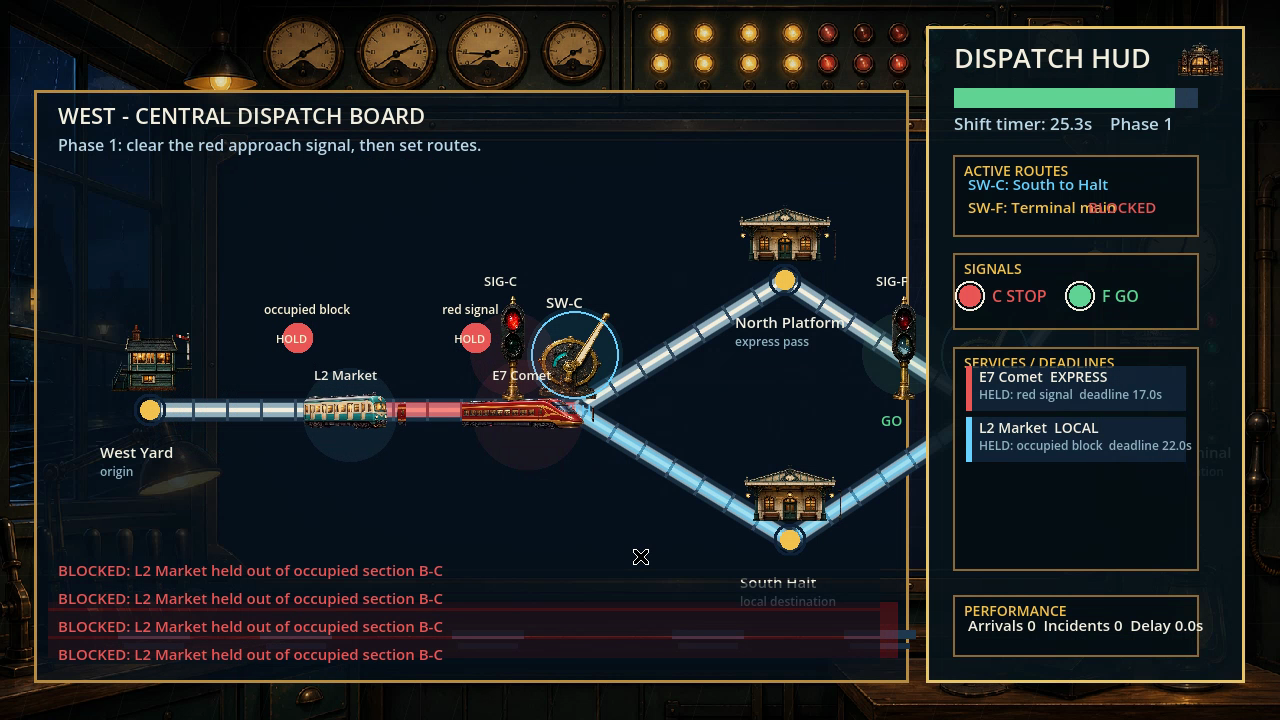}
{Coding agents implement complete game projects across 15 families. The gallery includes racing, roguelikes, puzzles, sports, and visual novels.}
{Godot, task specifications, tools, and available assets are supplied. Gallery selections illustrate outputs; the full-task comparison is in Table~\ref{tab:gamecraft-results}.}
{Source: GameCraft-Bench \citep{luo2026gamecraft,gamecraft2026results}. Shown: Seele02-pro's \emph{Signal Rail Dispatcher}. \href{https://tongxuluo.github.io/gamecraft-bench-website/}{Recorded and playable examples}.}

\gamecase{Resident}{Book of Infinity: 1001 Nights}{Interactive Narrative}{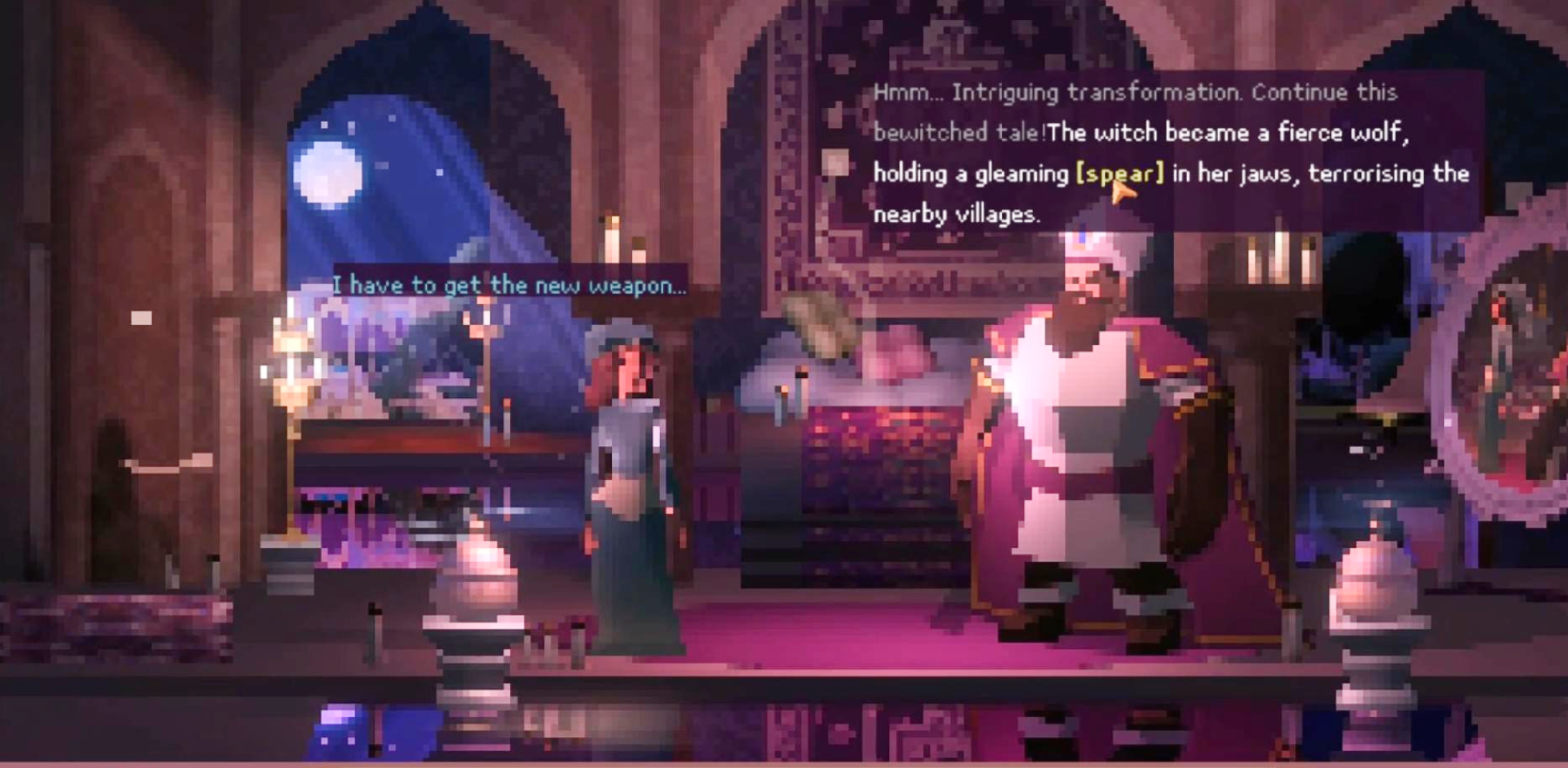}
{Generated story continuations respond to player-authored tales, while words can become equipment through the game's narrative mechanic.}
{The game supplies the premise and conversion rules. Its current product disclosure identifies the core art assets as human-made.}
{Sources: the research system and Ada Eden \citep{sun2023language,adaeden2026nights}. Shown: official demo media. \href{https://store.steampowered.com/app/2542850/1001_Nights/}{Steam demo}; full release listed as forthcoming.}

\gamecase{Resident}{Roblox / Cube 4D}{Functional Object Generation}{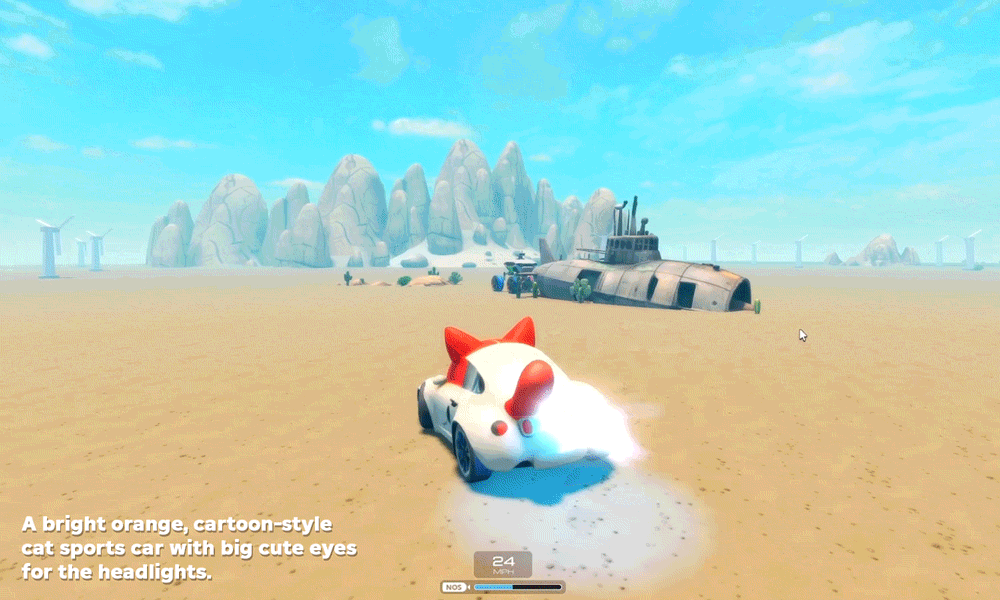}
{Text prompts produce objects with distinct functional parts, such as a car body and wheels. Behavior scripts are retargeted to the generated geometry.}
{The beta provides object schemas and scripts; Roblox supplies physics and the surrounding creator-configured experience.}
{Source: Roblox \citep{singh2026robloxcube}. Shown: official demonstration frame. \href{https://about.roblox.com/newsroom/2026/02/accelerating-creation-powered-roblox-cube-foundation-model}{Beta examples and technical account}.}

\clearpage
\subsection{Character Dialogue and Automated Testing}

\gamecase{Resident}{Fortnite / Darth Vader}{Voiced NPC Dialogue}{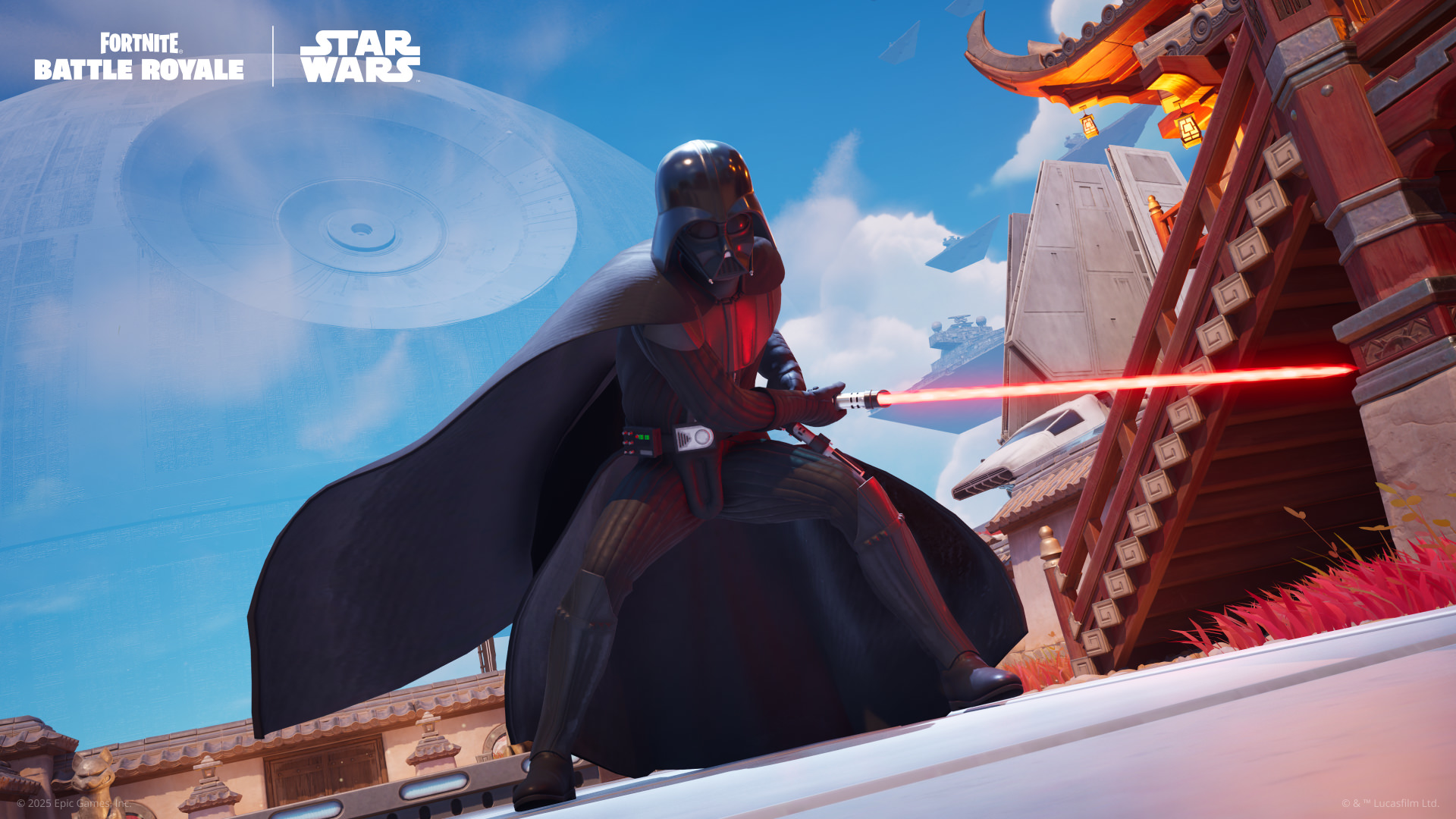}
{A language model generates conversational responses, which a voice model delivers in character. Dialogue is embedded in an otherwise authored battle-royale encounter.}
{Epic supplies the character, combat behavior, game rules, and safety controls. The 2025 seasonal feature is a deployment example, not a current availability promise.}
{Source: Epic Games \citep{epic2025vader}. Shown: official announcement artwork. \href{https://www.fortnite.com/news/this-will-be-a-day-long-remembered-speak-with-darth-vader-in-fortnite}{Feature and safeguards}.}

\gamecase{Evaluator}{Battlefield 2042 / EA SEED}{Production QA}{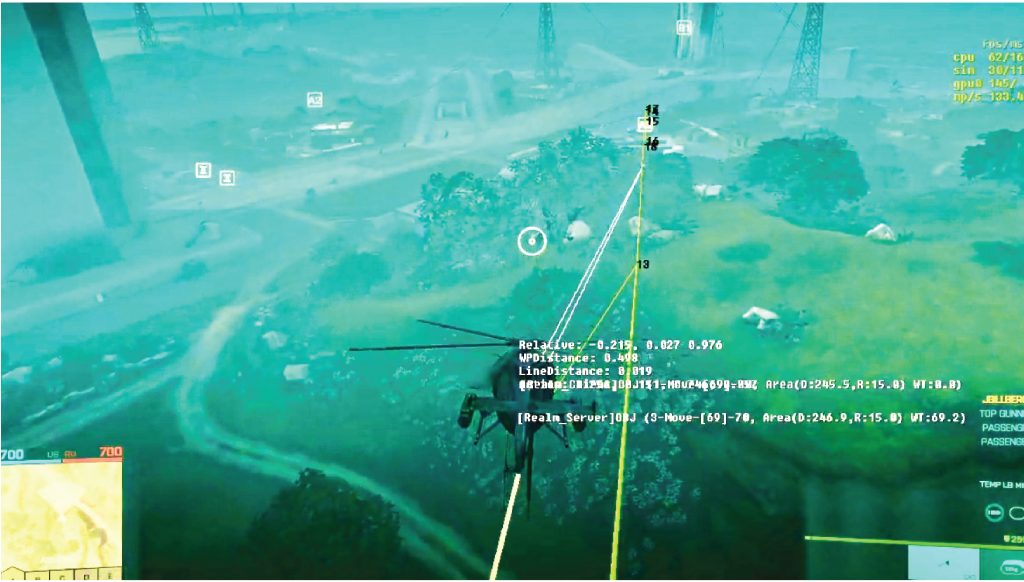}
{A learned helicopter controller executes testing tasks. The same engineering study integrates zero-gravity navigation into Dead Space tests.}
{Scripted test frameworks specify objectives and check outcomes. RL supplies difficult control skills inside the QA workflow.}
{Source: EA SEED \citep{gillberg2023productiontesting}. Shown: original production-test image, paper Fig.~3(c). \href{https://www.ea.com/seed/news/cog23-challenges-deploying-rl-agents-game-testing}{Technical report}; specialist RL, not an LLM tester.}

\gamecase{Evaluator}{PlayWorld / Interactive Worlds}{Agent-Based Probing}{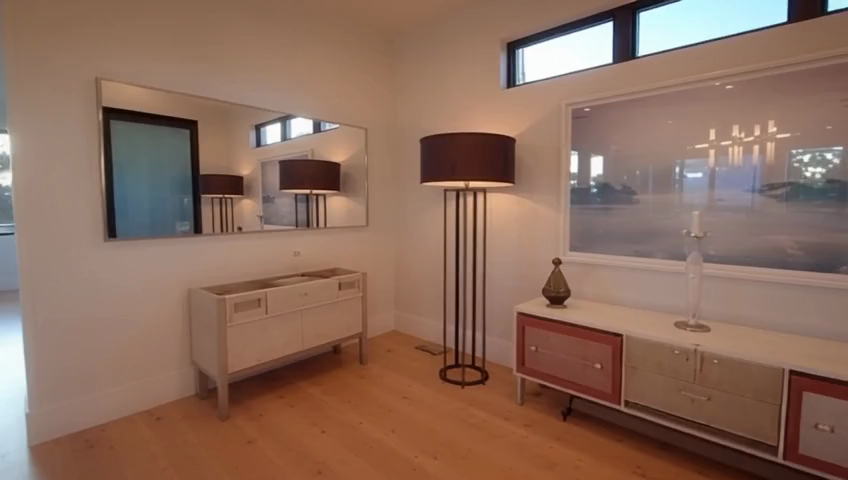}
{A visual agent acts to expose geometric and temporal inconsistencies, then a judge scores the resulting trajectories against test rubrics.}
{Scenario goals, the action interface, and scoring rules are supplied. These are probes of generated environments, rather than a commercial game.}
{Source: PlayWorld \citep{ding2026playworld}. Shown: a HappyOyster geometry-test rollout. \href{https://kxding.github.io/project/PlayWorld/}{Demonstrations and code}; full results in Table~\ref{tab:world-model-results}.}

\endgroup

\end{document}